\documentclass[dvipsnames]{article}

\PassOptionsToPackage{numbers, compress}{natbib}

\usepackage[main,final]{neurips_2026}

\usepackage[utf8]{inputenc} 
\usepackage[T1]{fontenc}    
\usepackage{hyperref}       
\usepackage{url}            
\usepackage{booktabs}       
\usepackage{amsfonts}       
\usepackage{nicefrac}       
\usepackage{microtype}      
\usepackage{xcolor}         

\usepackage{subfigure} 
\usepackage{multirow}
\usepackage{makecell}
\usepackage{seqsplit}
\usepackage{minibox}
\usepackage{booktabs}       
\usepackage{amsfonts}       
\usepackage{nicefrac}       
\usepackage{amssymb}
\usepackage{enumitem} 
\usepackage{wrapfig}
\usepackage{adjustbox}      
\usepackage{algorithm} 
\usepackage[noend]{algpseudocode} 
\usepackage{listings} 
\usepackage{tikz}
\usetikzlibrary{shapes,backgrounds,arrows,fit,calc,patterns}

\newcommand{\ie}{\emph{i.e.,}~}
\newcommand{\eg}{\emph{e.g.,}~}

\usepackage{pifont}

\newcommand{\para}[1]{\textbf{#1}}
\usepackage{amsmath} 
\usepackage{bm} 
\usepackage{bbm} 
\usepackage{amsthm}
\usepackage{mathtools}

\newcommand{\ep}{\varepsilon}
\renewcommand{\epsilon}{\ep}

\makeatletter
\@ifundefined{proposition}{%
    
}{}
\@ifundefined{definition}{%
}{}
\@ifundefined{lemma}{%
    \newtheorem{lemma}{Lemma}
}{}
\@ifundefined{corollary}{%
    \newtheorem{corollary}{Corollary}
}{}
\@ifundefined{theorem}{%
    \newtheorem{theorem}{Theorem}
}{}
\@ifundefined{corollary}{%
    \newtheorem{corollary}{Corollary}
}{}
\@ifundefined{assumption}{%
    
}{}
\@ifundefined{problem}{%
    
}{}
\@ifundefined{problem*}{%
    \newtheorem*{problem*}{Problem}
}{}
\@ifundefined{claim}{%
    
}{}
\@ifundefined{example}{%
    
}{}
\@ifundefined{implication}{%
    
}{}
\@ifundefined{remark*}{%
    \newtheorem*{remark*}{Remark}
}{}
\makeatother

\newcommand{\argmin}{\operatorname*{arg\,min}}

\newtheorem{innercustomgeneric}{\customgenericname}
\providecommand{\customgenericname}{}
\newcommand{\newcustomtheorem}[2]{%
  \newenvironment{#1}[1]
  {%
   \renewcommand\customgenericname{#2}%
   \renewcommand\theinnercustomgeneric{##1}%
   \innercustomgeneric
  }
  {\endinnercustomgeneric}
}
\newcustomtheorem{customclaim}{Claim}

\providecommand{\realnum}					{\mathbb{R}}

\providecommand{\naturalnum}				{\mathbb{N}}

\renewcommand{\(}						{\left(}
\renewcommand{\)}						{\right)}
\renewcommand{\[}						{\left[}
\renewcommand{\]}						{\right]}

\providecommand{\Prob}{\mathbbm{P}}

\providecommand{\Exp}{\mathbbm{E}}

\def\x{\mathbf{x}}
\def\y{\mathbf{y}}

\def\sh{\hat{{s}}}

\def\Sh{\hat{{S}}}

\def\Es{\mathcal{{E}}}

\def\Hs{\mathcal{{H}}}

\def\Os{\mathcal{{O}}}
\def\Ps{\mathcal{{P}}}

\def\Vs{\mathcal{{V}}}
\def\Ws{\mathcal{{W}}}
\def\Xs{\mathcal{{X}}}
\def\Ys{\mathcal{{Y}}}

 \newcommand{\Reg}{\textbf{Reg}}

\newcommand{\SR}{\textbf{SR}}
\newcommand{\Ine}{\textbf{Ineff}}
\newcommand{\ewns}{\texttt{EW}}
\newcommand{\ew}{\ewns~}
\newcommand{\eeens}{\texttt{Exp3}}
\newcommand{\eee}{\eeens~}

\newcommand{\eeeixns}{\texttt{Exp3-IX}}
\newcommand{\eeeix}{\eeeixns~}

\newcommand{\oursns}{\texttt{ExAUL}}
\newcommand{\ours}{\oursns~}

\newcommand{\hl}{black}

\usepackage[normalem]{ulem} 

\title{Online Conformal Abstention Under Adversarial Bandit Feedback}

\author{%
  \hspace{-2ex}
  Minjae Lee$^*$
  \\
  GSAI
  \\
  POSTECH
  \\
  \texttt{minjae.lee@postech.ac.kr}
  \And
  Yoonjae Jung$^*$
  \\
  GSAI
  \\
  POSTECH
  \\
  \texttt{jyjllll1025@postech.ac.kr}
  \And
  Sangdon Park
  \\
  GSAI \& CSE
  \\
  POSTECH
  \\
  \texttt{sangdon@postech.ac.kr} 
  \hspace{-2ex}
}

\begin{document}

\begin{NoHyper}
\def\thefootnote{*}\footnotetext{Equal contribution}
\end{NoHyper}

\maketitle

\begin{abstract}
As interactive generative systems are increasingly deployed in real-world applications, their tendency to generate unreliable or false responses raises serious concerns.
{\color{\hl}\emph{Conformal abstention}} mitigates this risk by ensuring that the system answers only when confident.
However, real-world deployments typically provide only partial user feedback (\eg thumbs up/down) on the selected response and often operate in non-stationary or adversarial environments, for which effective learning methods are largely missing. 
To bridge this gap, we propose \oursns, a novel online learning framework for {\color{\hl}conformal abstention} with adversarial and partial feedback.
Technically, we introduce
(i) a novel \emph{conversion lemma} that translates the regret of any bandit algorithm into a selective risk (SR) bound, and
(ii) \emph{feedback unlocking}, a strategy that exploits the structure of {\color{\hl}conformal abstention} to extract additional learning signals from partial feedback.
We prove that \ours achieves a regret bound of $\Os(\sqrt{T \ln |\Hs|})$, 
{\color{\hl}which translates into an $\Os(\sqrt{T})$ bound on SR violation control, matching the controllability of full-information settings despite receiving only partial feedback.}
While applicable to general generative tasks, we demonstrate the efficacy of \ours for ensuring the reliability of Large Language Models (LLMs) through empirical validation on question-answering tasks across diverse non-stationary and adversarial settings.
Our results demonstrate that \ours robustly controls the SR while maintaining competitive answering coverage. Code and datasets are
provided at \url{https://github.com/trustml-lab/oca}.
\end{abstract}

\section{Introduction}
\label{sec:introduction}


As large language models \citep{openai2026gpt54,meta2024llama31-8binstruct} surpass average human performance, their  
deployment in real-world human-facing applications has become widespread.
However, their tendency to generate unintended or incorrect information 
raises serious concerns about reliability and safety in high-stakes settings.
To address such reliability issues, a common practice is to utilize heuristic uncertainty estimates---such as self-consistency \citep{wang2022self, manakul2023selfcheckgpt} or entropy-based measures \citep{kadavath2022language, kamath2020selective, kuhn2023semantic}---and abstain when the uncertainty exceeds a heuristic threshold.
While these methods show strong empirical performance, they lack formal guarantees on reliability,
where such certification is required for the high-stakes situations in ensuring the trustworthiness of large language models.

A promising approach for providing such guarantees is \emph{selective prediction}, also known as {\color{\hl}\emph{conformal  abstention}}, \citep{geifman2017selective,goren2024hierarchical,mohri2023learning,lee2024selective,angelopoulos2021learn,yadkori2024mitigatingllmhallucinationsconformal}, which aims to control an error rate (\eg the false discovery rate) in a certified manner by abstaining from uncertain predictions.
Ideally, {\color{\hl}the abstainer} ensures that whenever it provides an answer, the answer is mostly correct under a specified risk level.
However, existing methods \citep{geifman2017selective,goren2024hierarchical,mohri2023learning,lee2024selective,angelopoulos2021learn,yadkori2024mitigatingllmhallucinationsconformal} that provide such theoretical guarantees are developed under limited \emph{stochastic assumptions}, \ie data are independently drawn from a fixed distribution.
This assumption undermines their applicability to dynamic real-world environments involving adversarial distribution shifts.
Moreover, these methods require \emph{full feedback} (\ie ground truth answers), whereas real-world systems typically receive only \emph{partial} or \emph{bandit feedback} (\eg thumbs-up/down to a generated answer), which is far more practical to obtain.
{\color{\hl}
Motivated by these challenges, we ask the following question:
``\emph{how can we adaptively learn a certified abstainer with adversarial bandit feedback?}''}

\begin{figure*}[tb]
  \centering
  \includegraphics[width=0.9\linewidth,clip,trim=0 8.5cm 0 0]{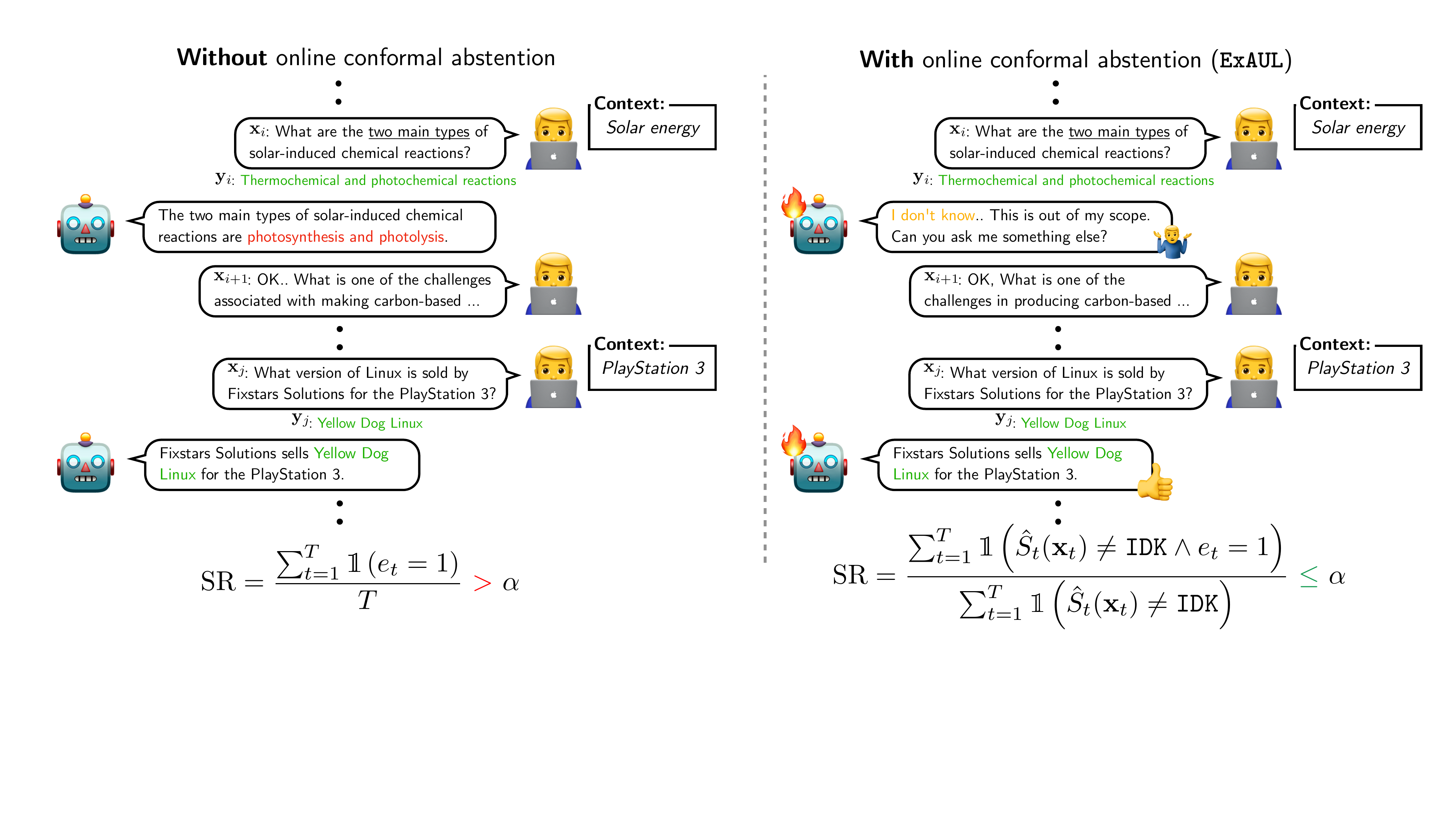}
  \vspace{-1ex}
  \caption{
    Qualitative examples from an interactive dialog simulation.
    This 
    demonstrates that our proposed method \ours effectively controls the rate of hallucination in the SR by abstaining from answering
    under a practical online setup with partial feedback.
    See Section \ref{sec:exp:interactive} for details.
    }
  \label{fig:method_overview}
  \vspace{-3ex}
\end{figure*}

{\color{\hl}
To mitigate these limitations, we introduce
an \emph{online learning} setup for conformal abstention under \emph{partial} and potentially \emph{adversarial} feedback as learning supervision.
We address this problem by  
proposing a novel online {\color{\hl}conformal abstention} algorithm to control a selective risk (SR), while maximizing selection efficiency (\ie the ratio of non-abstaining cases).}
In particular, we first reduce online {\color{\hl}conformal abstention} to bandits for leveraging any regret minimization algorithms, \eg \texttt{Exp3-IX} \citep{neu2015explore}, for leaning, and then we introduce a \emph{Regret-to-SR conversion lemma} to interpret any regret bounds into SR bounds for guarantees on conformal abstention. 
%
Second, we design an algorithm tailored for conformal abstention which fully exploits feedback information. 
Although partial feedback is practical, it provides an insufficient amount of information compared to full feedback, 
often requiring more samples for SR guarantees.
To address this challenge, we exploit the unique structure of conformal abstention to \emph{unlock} additional information from partial feedback, a technique we call \emph{partial feedback unlocking}. 
Leveraging this idea, we extend the \texttt{\underline{Ex}p3-IX} for adversarial bandits to online conformal \underline{\texttt{A}}bstention with adversarially partial feedback \underline{\texttt{U}}n\underline{\texttt{L}}ocking (\oursns).
Moreover, we provide the $\Os(\sqrt{T\ln |\Hs|})$ expected regret bound of \ours---which enables tighter control of the SR guarantee over time---better than a learner with partial feedback, \ie $\Os(\sqrt{T |\Hs| \ln |\Hs|})$, and compatible to that with full feedback, \ie $\Os(\sqrt{T \ln |\Hs|})$, where $T$ is a time horizon and $\Hs$ is a set of hypotheses of abstainers.
This provides our SR guarantee by a desired level $\alpha$ at the rate of $\Os(\sqrt{\ln |\Hs| / T})$ due to our conversion lemma.
%
Finally, we empirically evaluate the efficacy of \oursns. 
We consider two tasks (\ie QA over TriviaQA \citep{joshi2017triviaqa} and Natural Question \citep{kwiatkowski2019natural}; and dialog conversations via two dialog agents), four learning environments (\ie stochastic, distribution-shifted, interactive, and adversarial), and two language models (\ie GPT-5.4 \citep{openai2026gpt54} and LLaMA3.1 \citep{meta2024llama31-8binstruct}),
demonstrating that our method 
(1) controls a desired SR and 
(2) maintains reasonably low selection inefficiency. 
See Figure \ref{fig:method_overview} for qualitative results controlling a SR in the interactive environment.

\subsection{Related Work}\label{sec:rel}


Here, we introduce closely related literature, including selective prediction, conformal abstention, online learning, and bandit problems. See Appendix \ref{sec:additionalrel} for additional related work.

\para{Heuristic-Based Abstention.}
Recent approaches utilize uncertainty metrics, such as self-consistency \citep{wang2022self,manakul2023selfcheckgpt} and likelihood-/entropy-based scores \citep{kadavath2022language,kamath2020selective,kuhn2023semantic}, to detect hallucinations.
While these metrics serve as effective signals for abstention, the decision rule typically relies on empirical thresholds lacking theoretical justification,
thereby failing to provide formal guarantees on error rates (\eg SR).

\para{Selective Prediction and Conformal Abstention.}
In high-stakes domains such as medical diagnosis, rejecting an answer rather than making a risky prediction is critical for system reliability \citep{cortes2016learning}.
Selective prediction addresses this by abstaining from uncertain predictions to control the error rate in a certified manner.
\citet{geifman2017selective} propose selective classification, which learns a threshold for a scoring function to abstain from prediction and controls an SR.
Recent advances in conformal prediction provide interpretation of selective prediction in the lens of conformal prediction.
\citet{angelopoulos2021learn} proposes a general risk control framework for conformal prediction, including the SR of a selective classifier as the risk definition.
\citet{yadkori2024mitigatingllmhallucinationsconformal} consider an un-normalized surrogate of the SR as the risk definition, where it coined the term of \emph{conformal abstention}. 
\citet{jin2023selection} develops a rigorous method to control the SR of a given selected subset from a large candidate pool (also called screening).
Extensions of these ideas include exploiting hierarchical labels in selective classification \citep{goren2024hierarchical} or adapting to generation tasks \citep{mohri2023learning,yadkori2024mitigatingllmhallucinationsconformal,lee2024selective};
\citet{lee2024selective} generalizes selective prediction to open-ended generation by introducing entailment sets with SR guarantees.
However, these prior works predominantly focus on \emph{batch learning} (requiring a calibration set) under \emph{stochastic assumptions} (\ie i.i.d. data), which limits their applicability in real-world scenarios with distribution shifts.
In contrast, we consider a non-stochastic setup that allows for arbitrary distribution shifts and adversarial feedback, proposing a robust online conformal abstention method.


\para{Online Learning and Bandit Problems.}
Sequential prediction \citep{cesa2006prediction,mohri2012foundations,sankararaman2016semi,foster2023foundations} designs learners that adapt to sequentially arriving data. 
Unlike stochastic learning, online learning methods, \eg exponential weighting \citep{littlestone1994weighted}, make minimal distributional assumptions and analyze worst-case, possibly adaptive sequences, with guarantees in terms of regret.
In a full feedback online learning setting, the learner observes the loss of every prediction each round.
In contrast, bandit problems, such as multi-armed bandits \citep{agrawal1995sample} and adversarial bandits \citep{auer2002nonstochastic, bubeck2012regret}, consider a partial-feedback setting, where the learner only observes feedback of a chosen arm each round.
A well-known algorithm is \eee \citep{auer2002nonstochastic}, which leverages exponential weighting \citep{littlestone1994weighted} with an importance-weighted  unbiased loss estimator.
However, \eee suffers from a dishearteningly large variance of the regret, leading to looser concentration.
To obtain high-probability bounds, variants introduce exploration explicitly or implicitly, \ie
\texttt{Exp3.P} \citep{auer2002nonstochastic,bubeck2012regret} explicitly mixes the uniform distribution
and \eeeix \citep{neu2015explore} implements implicit exploration by adding an exploration parameter to the loss estimator.
To efficiently leverage partial feedback, structured bandits \citep{russo2013eluder} exploit the functional structure of arms-to-loss functions and semi-bandits \citep{sankararaman2016semi} allows to choose a set of arms with fixed size for better learning efficiency.
Here, we exploit adversarial bandits to mitigate stochastic assumptions in traditional conformal abstention
and to consider learning under partial feedback. 
Moreover, we leverage a unique structure between arms and loss in conformal abstention to achieve better learning efficiency.

\section{Online Risk-Controlled Abstention Under Adversarial Bandit Feedback}
\label{sec:problem}


\begin{figure*}[tb]
  \centering
  \includegraphics[width=0.9\linewidth]{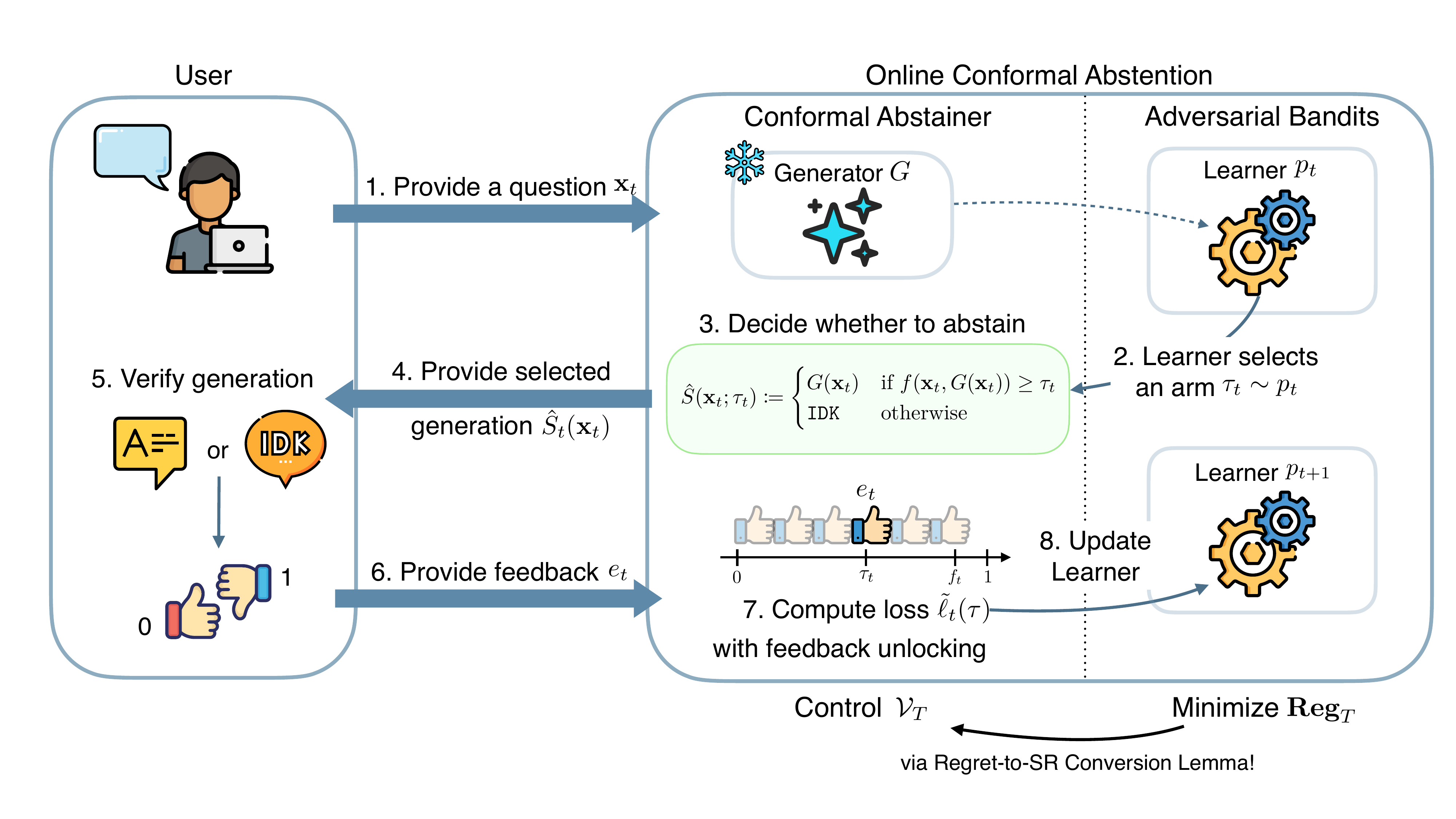}
  \vspace{-3ex}
  \caption{
    An example of our proposed framework for online conformal abstention. At each round $t$, (1) the user provides an input $\x_t$, (2) the learner selects an arm $\tau_t \sim p_t$, (3-4) generates $\Sh_t(\x_t; \tau_t)$, (5-6) the user provides partial feedback $e_t$, (7) the loss $\ell_t(\tau_t)$ is computed from $e_t$, and (8) the learner update $p_t$ by the loss estimator $\tilde\ell_t$. Note that the user can be modeled as an adversary.
    }
  \label{fig:method_pipeline}
\end{figure*}

We consider online learning of an abstainer under adversarial bandit feedback for language models (see Appendix \ref{sec:bg} for its preliminary). 
Let $\Ws$ be a set of tokens,
$\Xs= \Ys \coloneqq \cup_{i=0}^\infty \Ws^i$ be a set of (token) sequences,
and
$t$ be a round index.
Then,
let $\Hs$ be a finite class of \emph{abstainers} $\Sh_t: \Xs \to \Ys \cup \{\texttt{IDK}\}$.
Here, given a generator $G: \Xs \to \Ys$, each abstainer $\Sh_t \in \Hs$ is determined by a selection function $\sh_t: \Xs \times \Ys \to \{0, 1\}$ and abstains from answering when a generated answer $G(\x_t)$ of a question $\x_t$ is uncertain, \ie
$
    \Sh_t(\x_t) \coloneqq
    \begin{cases}
        G(\x_t) &\text{if~} \sh_t(\x_t, G(\x_t)) = 1
        \\
        \texttt{IDK} &\text{otherwise}
    \end{cases},
$
where $\texttt{IDK}$ represents ``I don't know''.
At each round $t$, the learner selects an abstainer $\Sh_t \in \Hs$.
To formalize supervision signals in online learning,
we define a feedback function $\Es_t: \Ys \cup\{\texttt{IDK}\} \rightarrow {\color{\hl}[0,1]}$,
which is induced from $(\x_t,\y_t)$ chosen by an adversary at round $t$,
and outputs feedback for an abstainer $\Sh_t$,
\eg
$\Es_t(\Sh_t(\x_t)) \coloneqq \mathbbm{1}(\Sh_t(\x_t) \neq_E \y_t)$.
We assume that
a learner does not observe a ground-truth $\y_t$ but instead receives \emph{partial feedback} $\Es_t(\Sh_t(\x_t))$,
which we simply denote as $e_t$.
Here, $A \neq_E B$ means that $A$ and $B$ are different in terms of a given correctness relation $E$, \eg textual-entailment \citep{bowman2015large}; thus, $e_t=0 $ if the generation is correct and $e_t=1$ if it is incorrect or $\hat{S}_t(\x_t)=\texttt{IDK}$ by definition. This setup includes real-world user feedback as a special case, where a user typically provides $e_t=0$ or $1$, \eg thumbs-up or down, instead of $\y_t$.
{\color{\hl}Note that we consider binary feedback for real world user feedback but $e_t$ can be a real value in $[0, 1]$.}


To learn an online {\color{\hl} abstainer} with this partial feedback,
we consider the following online learning protocol with a \emph{non-anticipating adaptive adversary}:
each round $t$ until a time horizon $T$, 
(1) an adversary chooses $\x_t \in \Xs$ and
$\y_t \in \Ys$, which determine the feedback function $\Es_t$,
(2) a learner observes $\x_t \in \Xs$ and predicts $\Sh_t(\x_t)$, where $\Sh_t$ is drawn from a learned distribution $p_t$ over {\color{\hl} abstainers},
and
(3) the leaner observes partial feedback $e_t$ and update $p_t$ by using it.
Here, we consider an \emph{adaptive adversary}, \ie $(\x_t, \y_t)$ is drawn from an arbitrary distribution that depends on
the interaction history with the learner, \ie $(\x_1, \y_1), \dots (\x_t, \y_t)$ and $\Sh_1(\x_1), \dots, \Sh_{t-1}(\x_{t-1})$ but crucially not on the current prediction $\Sh_t(\x_t)$, hence \emph{non-anticipating}.

\para{Goal.}
Under this learning setup, 
our primary goal is to control risk, where we specifically consider
the \emph{selective} risk (SR), of an {\color{\hl} abstainer} at a desired level $\alpha \in [0, 1]$ up to a time horizon $T$.
To this end, we first define 
the \emph{selective risk violation}
as follows:
\begin{equation}
  \Vs_T \coloneqq \sum_{t=1}^T \Bigl[ \mathbbm{1}( \Sh_t(\x_t) \neq  \texttt{IDK} ) {\color{\hl}\cdot e_t}
    - \alpha\mathbbm{1}( \Sh_t(\x_t) \neq \texttt{IDK}) \Bigr].
  \label{eq:fdrrisk}
\end{equation}

Ideally, we would ensure $\Vs_T \le 0$, equivalent to controlling the following SR by $\alpha$: 
\begin{equation}
    \SR_T \coloneqq \frac{\sum_{t=1}^T \mathbbm{1} ( \Sh_t(\x_t) \neq \texttt{IDK}  ) {\color{\hl}\cdot e_t}} {\sum_{t=1}^T \mathbbm{1} ( \Sh_t(\x_t) \neq \texttt{IDK} )} \le \alpha,
    \label{eq:empiricalfdr}
\end{equation}
where $\SR_T = \alpha$ if 
$\Sh_t(\x_t) = \texttt{IDK}$ for all $t$ by convention.
In the binary-feedback case where $e_t \in \{0,1\}$, $\SR_T$ is the fraction of incorrect generations among all non-\texttt{IDK} generations, \ie the false discovery rate (FDR).

As strict control of $\Vs_T$, \ie $\Vs_T \le 0$, cannot generally be guaranteed,
the goal is to learn a distribution $p_t$ over abstainers
such that a drawn abstainer $\Sh_t \sim p_t$  controls $\Vs_T$
to be sub-linear in $T$, \ie
$
     \Vs_T
     \le \ep(T),
$
where $\ep(T) = o(T)$.
Accordingly, the learner controls $\Vs_T$ while the adaptive adversary tries to maximize it.
%
Additionally, SR controllability is the primary objective, but we also wish to minimize selection inefficiency 
$
\Ine_T \coloneqq \frac{1}{T}\sum_{t=1}^T \mathbbm{1}( \Sh_t(\x_t) = \texttt{IDK}).
$
\section{\oursns: Online {\color{\hl}Conformal Abstention} with Feedback Unlocking}
\label{sec:method}

Following the conventional conformal and selective prediction literature \citep{vovk2005algorithmic,geifman2017selective}, we instantiate the abstainer class $\Hs$ of Section~\ref{sec:problem} through the {\color{\hl}model-free design of \emph{conformal} abstainers}: a scalar parameterization of $\sh_t$ via a \emph{conformity score} function $f_t: \Xs \times \Ys \to [0, 1)$ measuring the confidence of $G(\x_t)$ being an answer of $\x_t$, \ie
$
    \sh_t(\x_t, G(\x_t)) \coloneqq \mathbbm{1}\( f_t(\x_t, G(\x_t)) \ge \tau_t \),
$
so the {\color{\hl}conformal abstainer} is parameterized by a threshold $\tau_t$, of often denoted $\Sh_t(\cdot; \tau_t)$.
We take $\tau_t$ from the finely-quantized grid $\Hs = \left\{ k / (H-1) : k = 0,\dots, H-1 \right\}$, where $H$ is the number of hypotheses, and identify each $\Sh_t$ with its threshold $\tau_t$.

We leverage a regret perspective from bandit problems to design a learning algorithm for {\color{\hl}conformal abstention} that controls the SR under adversarial bandit feedback. After recalling regret minimization in adversarial bandits (Section \ref{sec:advfeedback:regretmin}), we reduce online {\color{\hl}conformal abstention} to an adversarial bandit (Section \ref{sec:advfeedback:reduction}), which enables us to turn any regret bound into an SR bound through a novel conversion lemma (Section \ref{sec:conversion}). However, simply leveraging existing bandit algorithms is sample-inefficient under partial feedback. We therefore extend \eeeix \citep{neu2015explore} by exploiting the structure of {\color{\hl}conformal abstainers}---\emph{feedback unlocking}---for a tighter regret, and hence a tighter SR, bound (Section \ref{sec:advfeedback:algorithm}). Adversarial \emph{full} feedback is handled analogously via exponential weighting (\ewns, Appendix \ref{sec:osg-fullfeedback}).

\subsection{Regret Minimization and Adversarial Bandits}
\label{sec:advfeedback:regretmin}
The adversarial bandit problem \citep{auer2002nonstochastic} is formulated as an interactive game between a learner and an adversary. At each round $t \in {1,\dots,T}$, the \emph{learner} selects an arm $\tau_t \in \Hs$ among a set of arms $\Hs$, while the \emph{adversary} simultaneously chooses a loss function $\ell_t: \Hs \to \realnum_{\ge 0}$. The learner then observes only the incurred loss $\ell_t(\tau_t)$ as partial feedback.
The learner's performance is evaluated by \emph{regret} \citep{bubeck2012regret}
\begin{equation}
\label{eq:fullfeedback:regret}
\Reg_T \coloneqq \sum_{t=1}^T \ell_t(\tau_t) - \min_{\tau \in \Hs} \sum_{t=1}^T \ell_t(\tau),
\end{equation}
and the goal of the learner is to minimize $\Reg_T$ so that it grows sublinearly in $T$.
Here, the \emph{adaptive adversary} maximizes $\Reg_T$ by choosing $\ell_t$ based on the previously chosen arms by the learner, \ie $\tau_1,\dots,\tau_{t-1}$. This makes $\Reg_T$ a random variable, so we analyze high-probability or expected regret bounds.
For further details on regret minimization and an adversarial bandit method (\eg \eeeix \citep{neu2015explore}), see Appendix~\ref{sec:prelim:regretmin} and \ref{sec:prelim:adversarialbandit}.

\subsection{Reduction: From {\color{\hl}Conformal Abstention} to Adversarial Bandits}
\label{sec:advfeedback:reduction}

\begin{wraptable}{r}{0.68\textwidth}
\vspace{-4ex}
\caption{From online {\color{\hl}conformal abstention} to adversarial bandits}
\label{tab:advfeedback:reduction}
\centering
\vspace{1ex}
\begin{tabular}{c | c c}
\toprule
 & online {\color{\hl}conformal abstention} & adversarial bandits \\
\midrule
models &
\makecell{{\color{\hl}conformal abstainers} $\Hs$} &
\makecell{finite arms $\Hs$} \\
\midrule
feedback &
$e_t$ &
$\ell_t(\tau_t,\alpha)$ \\
\midrule
metric &
$\Vs_T$ and $\Ine_T$ &
$\Reg_T$ \\
\bottomrule
\end{tabular}
\vspace{-1.2ex}
\end{wraptable}

Here, we map the components of online {\color{\hl}conformal abstention} with partial feedback to those of adversarial bandits to leverage existing algorithms and regret bounds. 
In particular, a parameter $\tau \in \Hs$ of our learner corresponds to a bandit arm in adversarial bandits.

We first define a special loss $\ell_t(\tau, \alpha)$ and connect this to feedback $e_t$. In particular, we introduce the following two key components of loss:
$a_t(\tau) \coloneqq \mathbbm{1}( \Sh(\x_t; \tau) = \texttt{IDK} ) \in \{0, 1\}$, which measures the selection inefficiency at time $t$, called \emph{inefficiency loss},
and
$d_t(\tau, \alpha) \coloneqq \mathbbm{1}( \Sh(\x_t; \tau) \neq \texttt{IDK} ) {\color{\hl}\cdot e_t} - \alpha\mathbbm{1}( \Sh(\x_t; \tau) \neq \texttt{IDK}) + \alpha \in
{\color{\hl} \[0, 1\]}$,
which measures
the SR violation
at time $t$ with a 
margin by $\alpha$ to penalize the \texttt{IDK} response, called \emph{SR loss} with a margin.
Based on these, we define the loss function for bandits as follows:
\begin{equation}
    \ell_t(\tau, \alpha) \coloneqq \frac{a_t(\tau) + \lambda d_t(\tau, \alpha)}{1 + \lambda}
    \in [0, 1],
\label{eq:loss-sg}
\end{equation}
where $\lambda \in \realnum_{> 0}$
is a hyperparameter that controls the trade-off between the SR loss and the efficiency loss.
Note that from the bandits' perspective, the learner receives feedback in two steps: get feedback $e_t$ from the adversary 
and
uses this to compute the loss $\ell_t(\tau_{t}, \alpha)$ for the chosen arm $\tau_t$.

Finally, the link between two metrics for online {\color{\hl}conformal abstention} and adversarial bandits remains to be established. 
In the following section, we propose a novel conversion lemma to connect $\SR_T$ and $\Ine_T$ to $\Reg_T$.
See Figure \ref{fig:method_pipeline} for an illustration of our full learning pipeline from online {\color{\hl}conformal abstention} to adversarial bandits and back again.

\subsection{Regret-to-SR Conversion}
\label{sec:conversion}

We mainly leverage algorithms that minimize $\Reg_T$ for learning conformal abstainers. 
Yet, the key requirement for conformal abstention is the SR guarantee of a learner at a desired level.
To this end, we introduce a novel perspective on the connection of a regret bound to an SR bound.
In particular, 
\emph{any learner}, including learners for bandits that minimizes the regret (\ref{eq:fullfeedback:regret}) with the loss (\ref{eq:loss-sg}), controls the SR (\ref{eq:fdrrisk}).
This is achievable as the designed loss (\ref{eq:loss-sg}) penalizes  when \emph{SR loss} at each $t$ is higher than $\alpha$, weighted by a parameter $\lambda$.
See Appendix \ref{proof:lem:conversion} for a proof and details on choosing $\lambda$. 

\begin{lemma}
    \label{lem:conversion}
    Let $T \in \naturalnum$ and $\alpha \in (0, 1)$.
    For any $(\x_t, \y_t)$ sequences, leading to any loss sequences $\ell_t$ of (\ref{eq:loss-sg}) with any $a_t:\Hs \to [0, 1]$ and $e_t \in [0, 1]$,  
    we have
    \begin{equation}
        \Vs_T 
        \le
        \frac{T( 1 - \Ine_T) + (1+\lambda) \Reg_T}{\lambda} 
        =
        \Os(T/\lambda + \Reg_T)
        = \Os(\sqrt{T}),
        \label{eq:lem:conversion}
    \end{equation}
    where the second equality (\ref{eq:lem:conversion}) holds if we take $\lambda = \sqrt{T}$ and $\Reg_T = \Os(\sqrt{T})$.
\end{lemma}
Intuitively, as $\lambda\to\infty$ the loss (\ref{eq:loss-sg}) reduces to the risk term $d_t$ and (\ref{eq:lem:conversion}) becomes to $\Vs_T\le\Reg_T$. Thus, controlling regret directly controls the
SR violation.
Importantly, this lemma is applicable to learning for both full and partial feedback, as it is agnostic to feedback mechanisms. 
Also, the lemma implies that if $\Reg_T$ has a sublinear bound,
satisfied by most bandit learners,
then 
$\frac{1}{T} \Vs_T \le \Os(1/\sqrt{T})$,
under our adaptive adversarial environment, \ie an adversary maximizes $\Vs_T$.

\para{The Connection between $\Vs_T$ and $\SR_T$.}
Dividing by $T(1-\Ine_T)$ if $\Ine_T \neq 1$, (\ref{eq:lem:conversion}) becomes
\begin{equation}
\label{eq:fdrboundfromlemma}
    \SR_T -\alpha
    \le
    \frac{1}{\sqrt{T}}
    + \frac{(1+\sqrt{T})\Reg_T}{T\sqrt{T}(1 - \Ine_T)},
\end{equation}
which converges to $\Os(1/\sqrt{T})$ if $\Ine_T$ does not converge to $1$, where the condition holds in general (see Appendix \ref{sec:discussion} for discussion).

\begin{algorithm}[tb]
    \caption{\eeeix for Online Conformal Abstention with Feedback Unlocking (\oursns)}
    \label{alg:exp3-sg-unlocking}
    \begin{algorithmic}[1]
        \Procedure {ExAUL}{$T, \Hs, \alpha, \lambda, \eta, f, G, \gamma$} 
        \State $w_1(\tau) \gets 1/|\Hs|$ for all $\tau \in \Hs$
        \For{$t = 1, \ldots, T$}
            
            
            \State Choose $\tau_t \sim p_t(\tau) = \frac{w_t(\tau)}{\sum_{\bar\tau \in \Hs} w_t(\bar\tau)}$
            \State \text{Observe} $\x_t$
            \State Observe $e_t$

            \State $\tilde\ell_t(\cdot, \alpha) \gets \textsc{Unlocking}(\Hs, \alpha, \lambda, \gamma, \tau_t, \x_t, e_t)$
            \hfill
            $(\triangleright)$
            {(\ref{eq:betteradvbandit:ourunbiasedestimator})}
            
            \State
            $w_{t+1}(\tau)
            \gets w_t(\tau) \exp (-\eta \tilde \ell_t(\tau,\alpha) )
            $ for all $\tau \in \Hs$
        \EndFor
        \EndProcedure

        \Procedure {Unlocking}{$\Hs, \alpha, \lambda, \gamma, \tau_t, \x_t, e_t$} 
        \State $f_t \gets f(\x_t, G(\x_t))$
        \State $\Hs_t(\tau_t) \gets 
          \begin{cases}
            \{ \tau \in \Hs \mid \tau \le f_t \} & \!\!\!\! \text{if~} \Sh(\x_t; \tau_t) \neq \texttt{IDK}, \\
            \{ \tau \in \Hs \mid \tau > f_t \} & \!\!\!\! \text{otherwise.}
          \end{cases}$
        
        \State $\ell_t\left(\tau, \alpha \right) \gets \textsc{ComputeLoss}(\Sh(\x_t; \tau), e_t, \alpha, \lambda)$ for all $\tau \in \Hs_t(\tau_t)$
        \hfill $(\triangleright)$ Algorithm \ref{alg:computeloss}


        \State $\ell_t(\cdot,\alpha\mid\Hs_t(\tau_t)) \gets \mathbf{0}$
        \State $\Ps_t
        \gets
        \gamma + \sum_{\bar\tau \in \Hs_t(\tau_t)}
        p_t(\bar\tau)$
        \hfill $(\triangleright)$ (\ref{eq:normalizer})
        \State $\ell_t(\tau,\alpha\mid\Hs_t(\tau_t))
        \gets \ell_t(\tau,\alpha) / \Ps_t$
        for all $\tau \in \Hs_t(\tau_t)$
        \hfill $(\triangleright)$ due to $\mathbbm{1}\{\tau \in \Hs_t(\tau_t)\}=1$
        

        \State \textbf{return} $\ell_t\left({\color{red}\cdot}, \alpha \mid \Hs_t(\tau_t) \right)$
        \EndProcedure
    \end{algorithmic}
\end{algorithm}

\subsection{\oursns: \eeeix for Online {\color{\hl}Conformal Abstention} with Feedback Unlocking}
\label{sec:advfeedback:algorithm}

As for the adversarial bandit learner, we leverage \eeeix \citep{neu2015explore},
{\color{\hl}which builds on \ew \citep{littlestone1994weighted} for adversarial bandit feedback and provides guarantees against \emph{adaptive} adversaries}.
The key challenge in moving from full to partial feedback is that limited feedback makes sample complexity of learning high.
To address this, we exploit the unique structure of the feedback function $\Es_t$ for {\color{\hl}conformal abstainers} (\ref{eq:loss-sg}) to unlock feedback of other arms by a chosen arm, dubbed by \emph{feedback unlocking}.
This is closely related to bandits with side observations~\citep{mannor2011bandits,neu2015explore,luo2023improved}, but the observation structure here is induced by conformal abstainers (Appendix~\ref{sec:additionalrel}).
In particular,
by the monotonicity of a selection function in $\tau$, \ie $\sh(\x) \coloneqq \mathbbm{1}(f_t \ge \tau)$ where $f_t \coloneqq f(\x_t, G(\x_t))$,
{\color{\hl}the following equality relation between the observed feedback $e_t$ of a chosen arm $\tau_t$ and feedback of other arms $\tau$:}
{\color{\hl}
\begin{equation}
  e_t
  \coloneqq
  \Es_t(\tau_t)
  =
  \begin{cases}
    \Es_t(\tau) \text{~for~} \tau \le f_t
    &\text{if~} f_t \ge \tau_t 
    \\
    \Es_t(\tau) \text{~for~} \tau > f_t
    &\text{if~} f_t < \tau_t 
  \end{cases}.
\end{equation}
This means that we can reuse the observed feedback $e_t$ of a chosen arm $\tau_t$ for all unchosen arm $\tau$ for $\tau \le f_t$ if the abstainer does not reject an answer, \ie $f_t \ge \tau_t$. 
}

To exploit this feedback unlocking within \eeeixns,
we propose a novel loss estimator, {\color{\hl}which estimates loss over all arms $\tau$ given unlocked feedback $\Hs_t(\tau_t)$ via importance weighting:}
\begin{equation}
    \ell_t(\tau, \alpha \mid \Hs_t({\tau_t}))
    \coloneqq \frac{\ell_t(\tau, \alpha)}{\gamma_t + \sum_{\bar\tau \in \Hs_t({\tau_t})} \mathbbm{1}(\tau \in \Hs_t({\bar\tau})) \cdot p_t(\bar\tau) }
    \cdot \mathbbm{1}(\tau \in \Hs_t({\tau_t})) \in [0, \infty),
    \label{eq:betteradvbandit:ourunbiasedestimator}
\end{equation}
where 
$\gamma_t > 0$ and
{\color{\hl}$\Hs_t({\tau_t})$ is a set of feedback unlocked arms by the $\tau_t$'s feedback,} \ie
$\Hs_t({\tau_t}) \coloneqq \{ \tau \in \Hs \mid \tau \le f_t \}$ 
if $f_t \ge \tau_t$ and $\Hs_t(\tau_t) \coloneqq \{ \tau \in \Hs \mid \tau > f_t \}$ otherwise; see Figure \ref{fig:unlockset} for the visualization of $\Hs_t(\tau)$.
With learning rate $\eta_t > 0$, Algorithm \ref{alg:exp3-sg-unlocking} presents our 
extension of \eeeix with partial feedback unlocking.
Note that
our algorithm can be reduced to \eeeix
by letting $\Hs_t({\tau}) = \{\tau_t\}$ for any $\tau$,
recovering the \eeeix's loss estimator in (\ref{eq:exp3ix:unbiasedestimator}).
Also, when computing the estimator~(\ref{eq:betteradvbandit:ourunbiasedestimator}) in Algorithm~\ref{alg:exp3-sg-unlocking}, the indicator $\mathbbm{1}(\tau \in \Hs_t(\bar\tau))$
need not be evaluated since it equals $1$ for all $\tau, \bar\tau \in \Hs_t(\tau_t)$
by (\ref{proof:thm1:domain:equivalent-new}) and~(\ref{proof:thm1:swapping-new}). The normalizer is then the $\tau$-independent, \ie
\begin{equation}
    \gamma + \sum_{\bar\tau \in \Hs_t(\tau_t)} p_t(\bar\tau),
    \label{eq:normalizer}
\end{equation}
which can be precomputed each round from the drawn arm $\tau_t$.
\para{Regret Bound.}
Improving the proof of \eeeix along with our novel loss estimator (\ref{eq:betteradvbandit:ourunbiasedestimator}), our Algorithm \ref{alg:exp3-sg-unlocking} achieves a sublinear regret bound, \ie $\Reg_T \le \Os(\sqrt{T \ln |\Hs| ) })$. See Theorem \ref{thm:ours:regretbound} for the regret bound and Appendix \ref{proof:thm:ours:regretbound} for a proof on our regret bound.

Interestingly, 
despite partial feedback, our algorithm achieves the same upper bound as \ew with full feedback, \ie $\Os ( \sqrt{{T\ln |\Hs|}} )$,  
due to 
our loss estimator with feedback unlocking (\ref{eq:betteradvbandit:ourunbiasedestimator})
which exploits rich information, whereas \eeeix suffers from an additional factor of $\sqrt{|\Hs|}$, as in (\ref{eq:regretbound:exp3ix}), due to its loss estimator with limited information.

\para{SR Violation Bound.}
The tighter regret bound of Algorithm \ref{alg:exp3-sg-unlocking}
further enables tighter control of the SR guarantee over time by Lemma \ref{lem:conversion}.
In particular, 
by plugging our regret bound of Algorithm \ref{alg:exp3-sg-unlocking} by Theorem \ref{thm:ours:regretbound} into our conversion lemma in Lemma \ref{lem:conversion}, 
we have the following SR violation bound.
\begin{theorem}
Let $\ell_t(\cdot) \in [0, 1]$ of the form (\ref{eq:loss-sg}).
For any $T \in \naturalnum$, finite hypotheses $\Hs$, and $\alpha, {\delta} \in (0, 1)$,
Algorithm \ref{alg:exp3-sg-unlocking} with $\eta = 2\gamma = \sqrt{{\ln|\mathcal{H}|}/{T}}$ provides the following SR violation bound
with probability at least $1-\delta$:
\begin{equation}
        \frac{1}{T}{\Vs_T} 
        \le
        \frac{ 1 - \Ine_T}{\sqrt{T}} + \(1 + \frac{1}{\sqrt{T}} \) \( 4\sqrt{\frac{\ln|\mathcal H|}{T}} + \( \frac{1}{T} +  \sqrt{\frac{1}{T \ln|\Hs|}}\)\ln\frac{2}{\delta} \).
    {\color{red}\label{eq:srbound}}
\end{equation}
\end{theorem}
Here, the time-normalized SR violation bound is $\Os ( \sqrt{{\ln |\Hs|}/{T}} )$,
which implies that
it converges to zero at the rate of $\Os(1/\sqrt{T})$ so ${\SR_T}$ also converges to $\alpha$ to the same rate with a different constant factor, \ie ${\SR_T} \le \alpha + \Os ( \frac{\sqrt{\ln |\Hs| / T}}{(1 - \Ine_T)} )$ if $\Ine_T \neq 1$, as desired.

\begin{remark*}{($\delta$-free parameters).}
As~\eqref{eq:srbound} holds for any confidence level $\delta\in(0,1)$ under a single choice of $\eta$ and $\gamma$ independent of $\delta$,
Algorithm \ref{alg:exp3-sg-unlocking} does not require $\delta$-dependent tuning.
Although the $\ln(1/\delta)$ dependence can be improved to $\sqrt{\ln(1/\delta)}$ by tuning $\eta$ and $\gamma$ to a fixed confidence level $\delta$, the algorithm is then tied to that $\delta$.
We therefore adopt the $\delta$-free choice \citep{neu2015explore}.
\end{remark*}

\section{Experiments}\label{sec:exp}

\begin{figure*}[tb]
    \centering
    \subfigure[SR over time]{
    \includegraphics[width=0.31\textwidth]{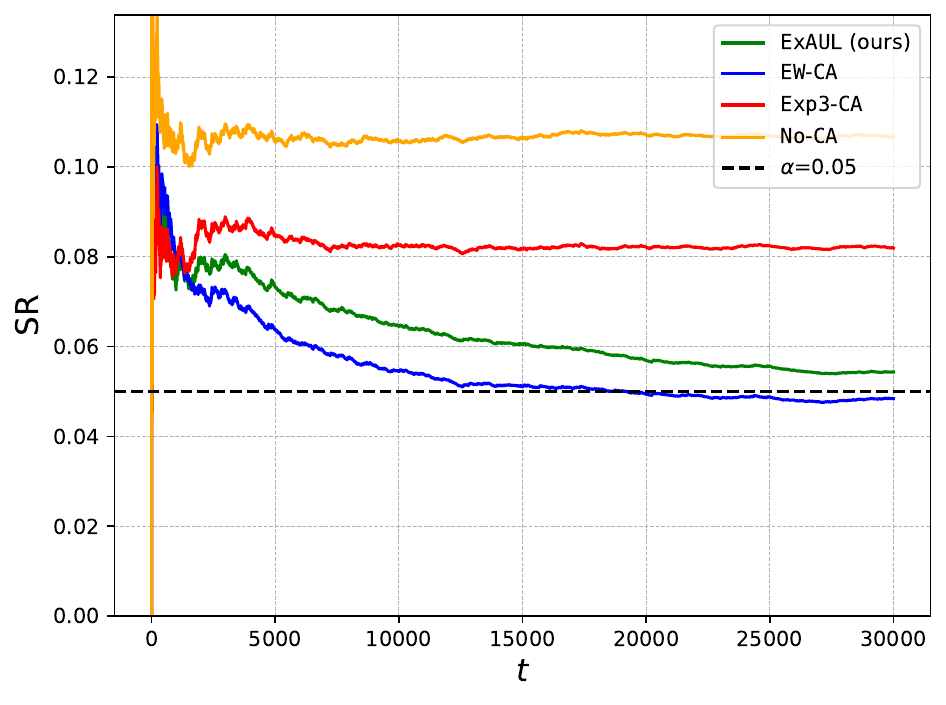}
    }
    \subfigure[Inefficiency over time]{
    \includegraphics[width=0.31\textwidth]{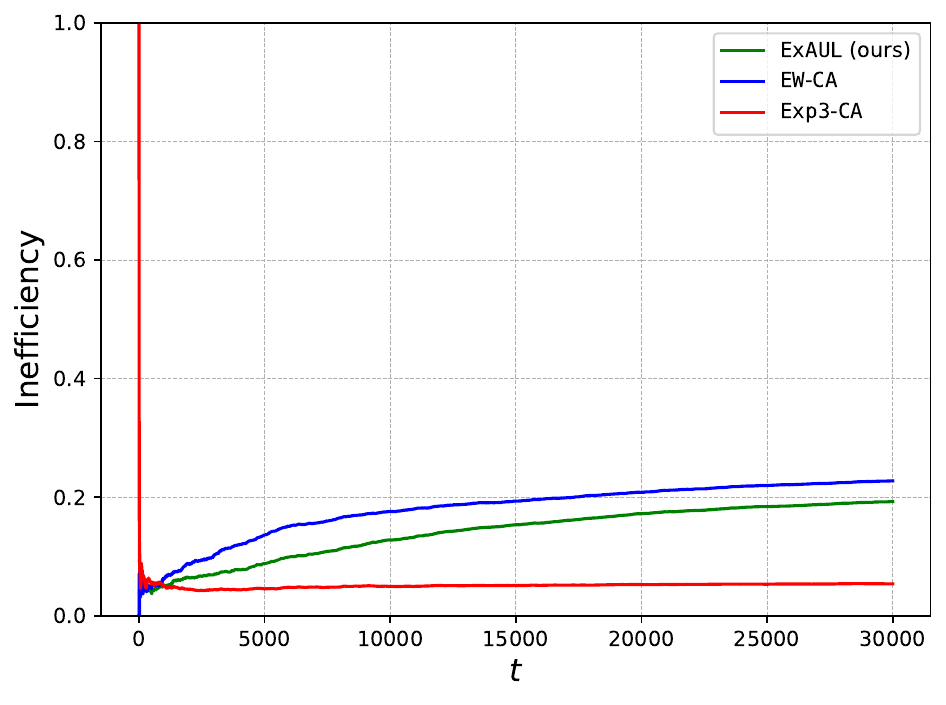}
    }
    \subfigure[SR distribution]{
    \includegraphics[width=0.31\textwidth]{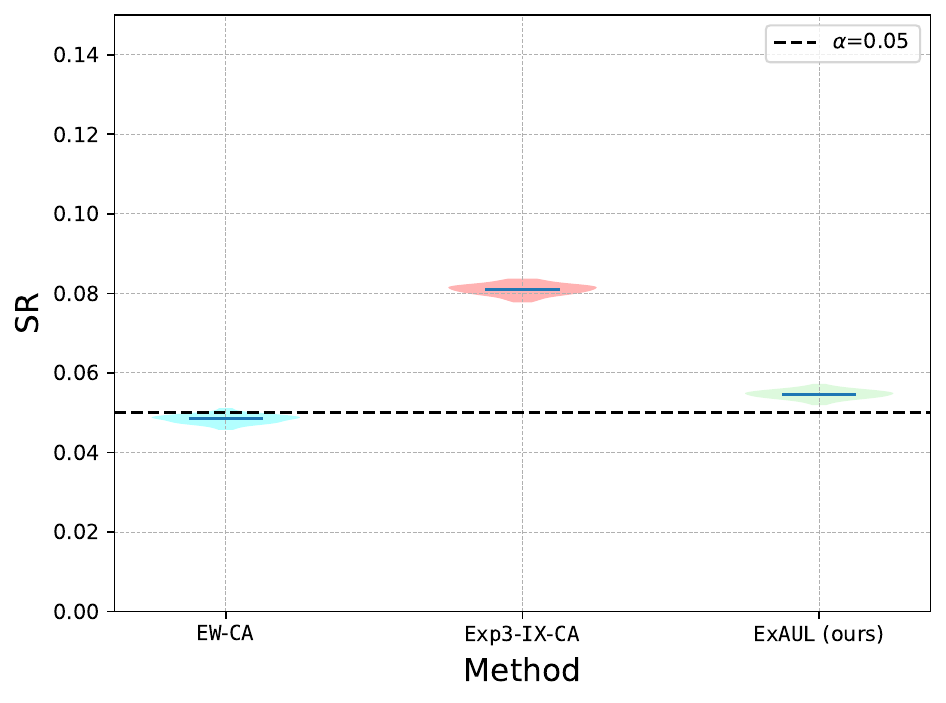}
    }
    \vspace{-1ex}
    \caption{Comparison of {\color{\hl} conformal abstention} methods under a stochastic environment with GPT-5.4 as a generator on TriviaQA
    ($T=30\mathrm{K}, {\alpha=0.05}$). The violin plots are drawn with randomly chosen $30\mathrm{K}$ samples over $100$ random trials.
    }
    \vspace{-1em}
    \label{fig:stochastic:main:gpt-triviaqa}
\end{figure*}

\begin{figure*}[tb]
    \centering
    \subfigure[SR over time]{
    \includegraphics[width=0.31\textwidth]{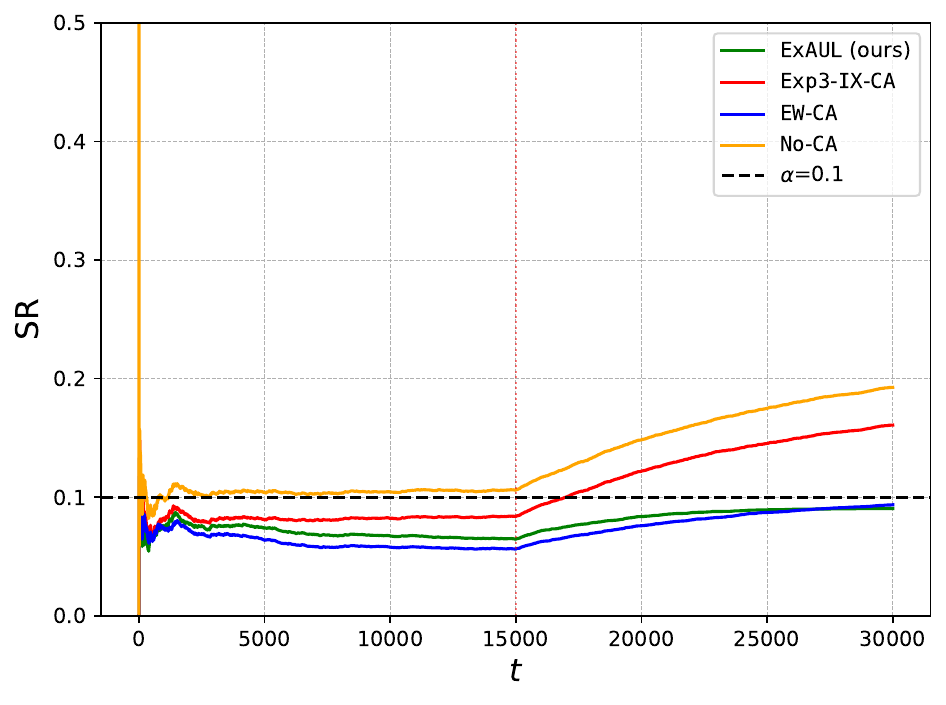}
    }
    \subfigure[Inefficiency over time]{\label{fig:shift:gpt-tri-simple-c}
    \includegraphics[width=0.31\textwidth]{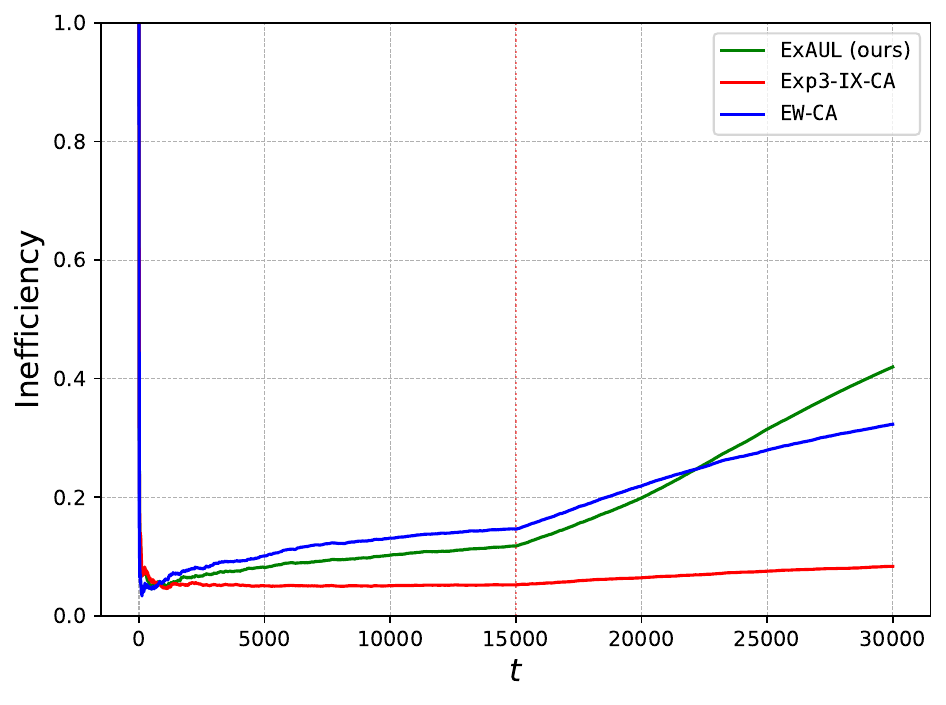}
    }
    \subfigure[SR distribution]{
    \includegraphics[width=0.31\textwidth]{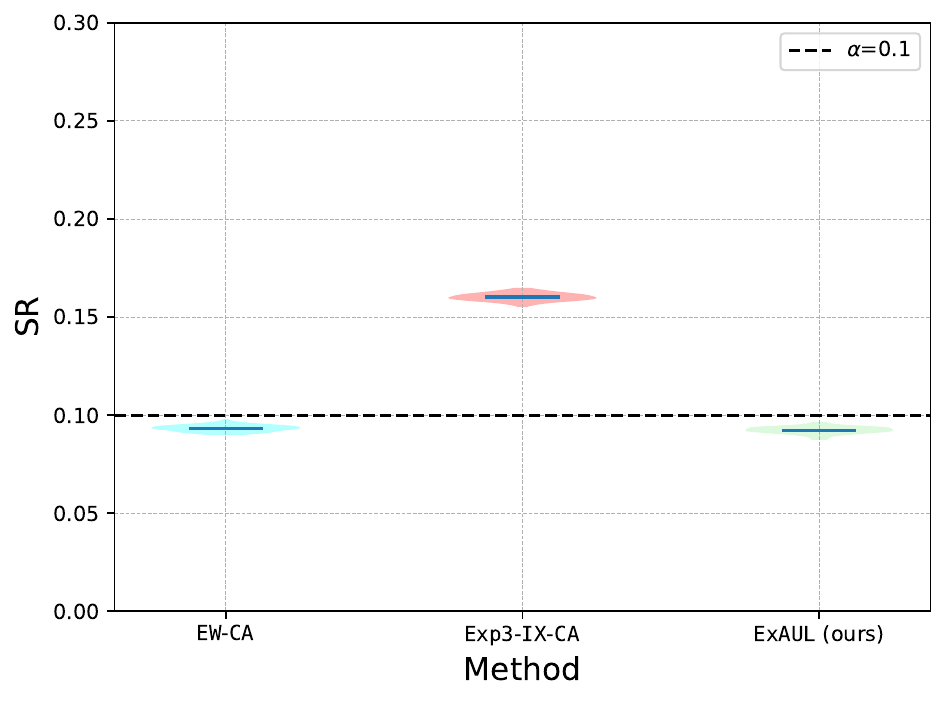}
    }
    \vspace{-1.5ex}
    \caption{Comparison of {\color{\hl} conformal abstention} methods under a single distribution-shift environment  
    with GPT-5.4 as a generator ($T = 30\mathrm{K}, \alpha = 0.10$), from TriviaQA to NQ. The violin plots are drawn with randomly chosen $30\mathrm{K}$ samples with $100$ random trials.
    }
    \vspace{-1.5ex}
    \label{fig:shift:gpt-tri-simple}
\end{figure*}


We empirically justify that \ours controls the SR while maximizing selection efficiency under four diverse environments:(1) stochastic, (2) distribution-shift, (3) adversarial, and (4) interactive environments.

\para{Datasets and Models.}
We use two datasets for stochastic and distribution-shift environments, $79\mathrm{K}$ Natural Question (NQ) \citep{kwiatkowski2019natural} and $93\mathrm{K}$ TriviaQA \citep{joshi2017triviaqa} with two base models, {\color{\hl}GPT-5.4} \citep{openai2026gpt54} and LLaMA3.1-8B-Instruct \citep{meta2024llama31-8binstruct}.
To obtain feedback, we employ {\color{\hl}GPT-5.4-nano} to generate feedback conditioned on the ($\x, \y, G(\x)$), which agrees with human annotations on $91.69\%$ of about $4.3\mathrm{K}$ NQ responses from \citet{lee2024selective}.
We simulate distribution-shift environments by mixing two datasets and an 
adversarial and interactive environment by generating interaction between {\color{\hl}GPT-5.4} and GPT-{\color{\hl}5.4-mini} models
(see Section \ref{sec:distshiftenvsetup} and \ref{sec:intenvsetup}, respectively for details). 

\para{Scoring functions.}
We fix a scoring function as $f_t = f$,
considering two $f_{\texttt{std}}$ and $f_{\texttt{con}}$, introduced in Section \ref{prelim:lg}.
In short, 
$f_{\texttt{std}}$ is a standard log-likelihood score defined as the conditional probability of an answer given a question.
$f_{\texttt{con}}$ is a self-consistency score of an answer computed via entailment scores (from an entailment model) across samples.
In particular, we compute self-consistency scores over 5 sampled responses \citep{manakul2023selfcheckgpt}.
We use $f_{\texttt{con}}$ unless specified.

\para{Methods.}
We consider two baselines, \texttt{Exp3-IX-{\color{\hl}CA}} and \texttt{No-{\color{\hl}CA}}, and one performance upper bound algorithm, \texttt{EW-{\color{\hl}CA}}.
In particular, 
(1) \texttt{Exp3-IX-{\color{\hl}CA}} (Algorithm \ref{alg:exp3ix-sg}) is an online conformal abstainer with partial feedback by adapting \eeeix \citep{neu2015explore} via our regret-to-SR conversion lemma.
(2) \texttt{No-{\color{\hl}CA}} is a non-conformal abstainer (\ie $\tau_t=0$ for all $t$), to serve as a standard use of a generator without abstention.
(3) \texttt{EW-{\color{\hl}CA}} (Algorithm \ref{alg:fullfeedback:ewsg}) is an online conformal abstainer with full feedback.
As it exploits full feedback, we use it as our empirical performance upper bound. 
%


\para{Metrics.}
We measure performance of ours along with comparing methods via 
the empirical SR, \ie $\SR_t$ and selection inefficiency, \ie $\Ine_t$.
Note that \texttt{{\color{\hl}No-CA}} achieves zero selection inefficiency by definition, so we do not explicitly add in figures.

\para{Parameters.}
We set $|\Hs|=1\mathrm{K}$ and $\lambda= \sqrt{T}$ by default.
Also, a desired SR level $\alpha$ and 
time horizon $T$ depend on tasks,
where $\alpha$ is provided by users.



\subsection{Results on Stochastic Environments}
We consider a stationary setting where question–answer pairs arrive from a fixed distribution.
By default, we follow the common experimental described above.
In Figure \ref{fig:stochastic:main:gpt-triviaqa} and \ref{fig:stochastic:main:llama3.1-8b-inst-triviaqa}, we observe that \ours has clearly better SR controllability to the desired SR compared to \texttt{Exp3-IX-{\color{\hl}CA}}, since \texttt{Exp3-IX-{\color{\hl}CA}} only observes $\ell_t(\tau_t)$, requiring a much longer horizon $T$ (Figure \ref{fig:exp3:convergence}).
These results demonstrate that our partial feedback unlocking is beneficial in SR control to the desired SR, supporting the regret bound analysis discussed in Section \ref{sec:advfeedback:algorithm}.

In addition, Figure~\ref{fig:stochastic:main:plot_over_alpha} shows that our algorithm effectively controls the SR across different $\alpha$. Moreover, Figure~\ref{fig:fdroverlambda} and \ref{fig:ineffoverlambda} indicate that the inefficiency loss $a_t$ and $\lambda$ enables the algorithm to control efficient SR guarantees.
See additional results in Section \ref{exp:addstochastic}.


\subsection{Results on Distribution-shift Environments} 
\label{sec:exp:shift}


To evaluate the robustness of the methods, we consider three distribution-shift scenarios: (1) a single instantaneous change by concatenating two datasets, (2) frequent shifts by interleaving fixed-size chunks randomly drawn from each dataset, and (3) progressive shifts by sampling each example according to a linearly increasing mixing probability. See Appendix~\ref{sec:distshiftenvsetup} for details on shift setups.

In the TriviaQA-to-NQ shift, Figure \ref{fig:shift:gpt-tri-simple}, \ref{fig:shift:gpt-tri-align}, and \ref{fig:shift:gpt-tri-linear} show that 
\texttt{Exp3-IX-{\color{\hl}CA}} exhibits a sharp increase  in SR immediately after the shift, whereas \ours maintains SR control.
In addition, Figure \ref{fig:exsul:shift:poa} further shows that \ours consistently controls the SR across different target levels $\alpha$. 
Moreover, Figure \ref{fig-regret-analysis} indicates that the regret increases in response to distribution shifts, mirroring the trend of the SR. 
Additional results across different generators and shift types are provided in Section~\ref{exp:adddistribuitionshift}.

\begin{table}[tb]
\centering
\def\arraystretch{0.8}
\caption{SR of \ours under imbalanced NQ--TriviaQA streams with NQ fraction $p_{\mathrm{NQ}}$ ($T=60\mathrm{K}$, $\alpha=0.1$). See Appendix~\ref{sec:imbalance} for details.}
\label{tab:main:imbalance}
\begin{tabular}{llcccc}
\toprule
 & & \multicolumn{4}{c}{$p_{\mathrm{NQ}}$} \\
\cmidrule(lr){3-6}
Stream & Model & 0.2 & 0.4 & 0.6 & 0.8 \\
\midrule
\multirow{2}{*}{Steady} & GPT-5.4 & 0.070 & 0.087 & 0.105 & 0.138 \\
 & Llama-3.1-8B & 0.082 & 0.093 & 0.108 & 0.130 \\
\multirow{2}{*}{Discrete} & GPT-5.4 & 0.072 & 0.083 & 0.093 & 0.111 \\
 & Llama-3.1-8B & 0.084 & 0.094 & 0.107 & 0.124 \\
\bottomrule
\end{tabular}
\vspace{-1ex}
\end{table}

\para{Imbalanced streams.}
We draw each round from NQ with probability $p_{\mathrm{NQ}}$ (steady) or place a TriviaQA block before an NQ block that makes up a fraction $p_{\mathrm{NQ}}$ of the stream (discrete).
In Table~\ref{tab:main:imbalance}, the SR stays below $\alpha$ when TriviaQA dominates and mostly exceeds $\alpha$ once NQ, the harder dataset, forms the majority.
The per-dataset SR remains roughly constant across $p_{\mathrm{NQ}}$ (about $0.16$--$0.21$ on NQ and $0.05$--$0.07$ on TriviaQA), so the minority dataset shows no additional degradation, although our guarantee concerns the SR over the whole stream rather than per dataset.

\para{Domain and score shifts.}
Shifting between open-domain QA and the medical benchmark MedQA-USMLE~\citep{jin2020disease} keeps the SR of \ours close to $\alpha$, while moving to MedQA increases inefficiency (Appendix~\ref{sec:medqa}).
Our analysis also allows $f_t$ to vary over time: when we switch the score from $f_{\texttt{con}}$ to $f_{\texttt{std}}$ in the middle of the stream, or replace it with $\mathcal{U}(0,1)$ noise with probability $0.5$, \ours still brings the SR toward $\alpha$, at the cost of more abstention and a longer horizon (Appendix~\ref{sec:analysisScore}).




\subsection{Results on Agent-Simulated Environments}
\label{sec:exp:interactive}
\label{sec:exp:adaptive:interactive}

\begin{figure*}[tb]
    \centering
    \subfigure[{SR (adaptive adversary)}]{
    \includegraphics[width=0.33\textwidth,clip,trim=0 0.3cm 0 0]{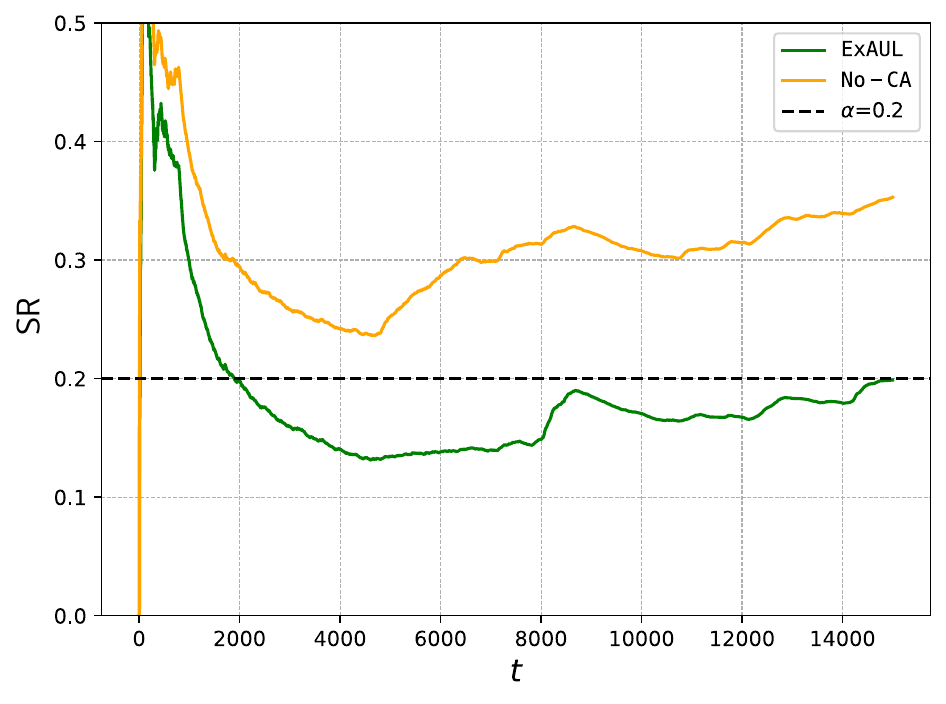}
    \label{fig:dialog:sr-adv}
    }\hspace{2em}
    \subfigure[{Inefficiency (adaptive adversary)}]{
    \includegraphics[width=0.33\textwidth,clip,trim=0 0.3cm 0 0]{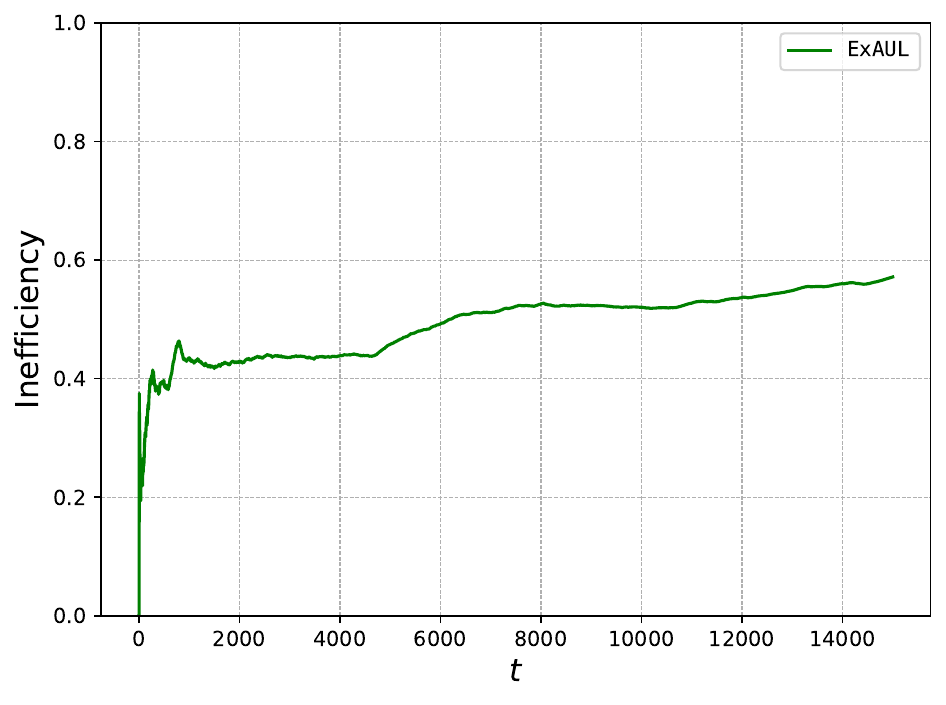}
    \label{fig:dialog:ineff-adv}
    }\\[-1ex]
    \subfigure[{SR (interactive)}]{
    \includegraphics[width=0.33\textwidth]{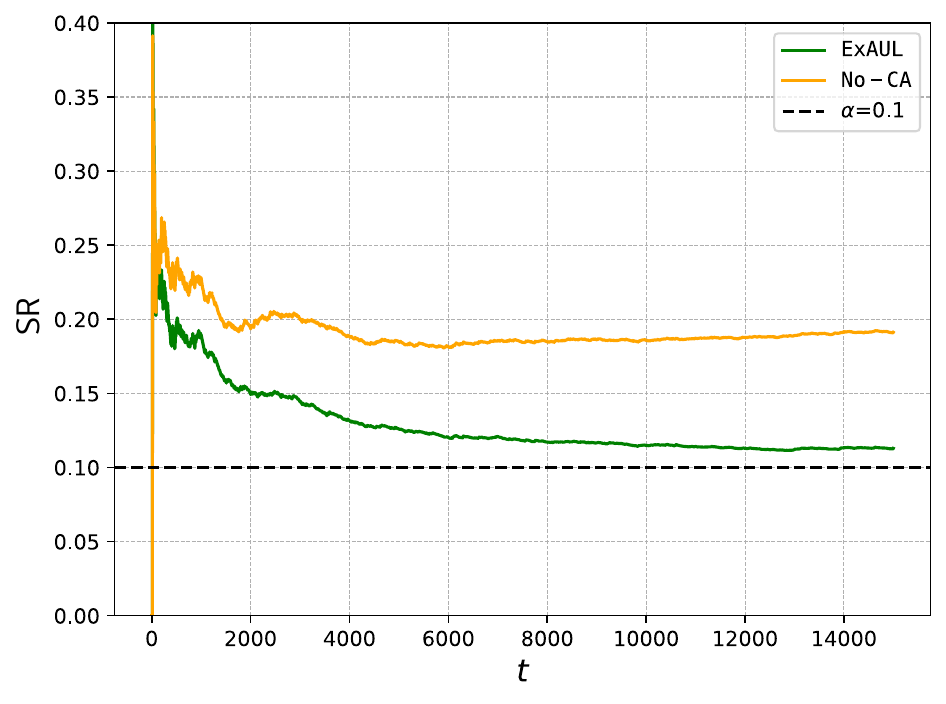}
    \label{fig:dialog:sr-int}
    }\hspace{2em}
    \subfigure[{Inefficiency (interactive)}]{
    \includegraphics[width=0.33\textwidth]{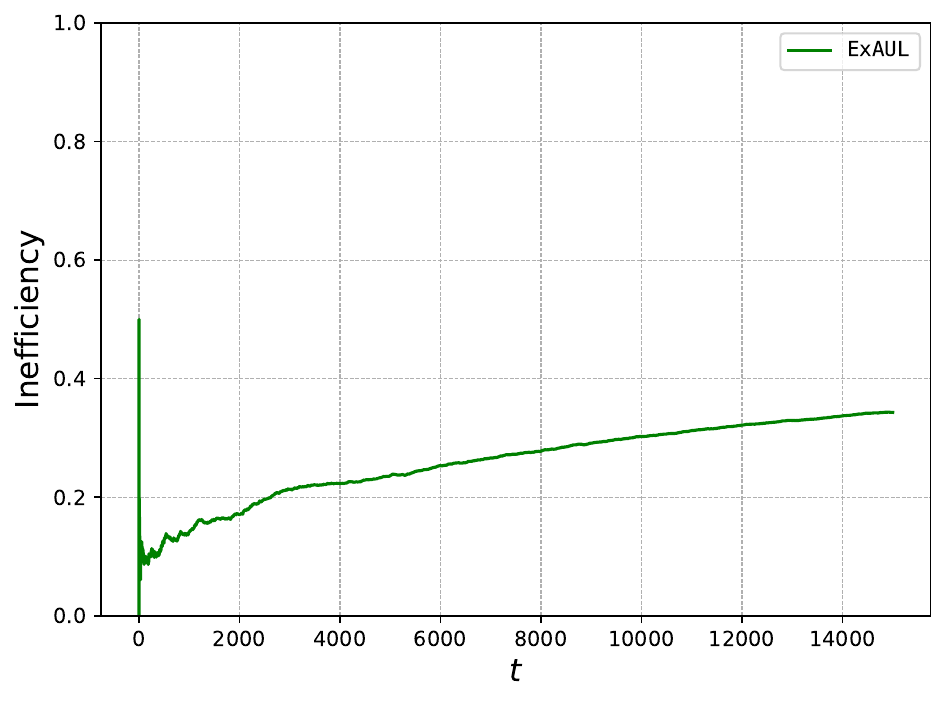}
    \label{fig:dialog:ineff-int}
    }
    \vspace{-1ex}
    \caption{$\SR_t$ and $\Ine_t$ for \ours with GPT-5.4 as an answering agent under (a)-(b) an adaptive-adversarial environment on TriviaQA and NQ ($T=15\mathrm{K}$, $\alpha=0.2$) and (c)-(d) an interactive environment ($T=15\mathrm{K}$, $\alpha=0.1$).}
    \vspace{-1.5ex}
    \label{fig:dialog}
\end{figure*}


\para{Adaptive-adversarial environment.}
In the adaptive setup, a user-acting agent observes the learner's decision history and adaptively selects $(\x_t, \y_t)$ to challenge the learner, consistent with the adaptive-adversary model in Section~\ref{sec:conversion}.
Please refer to Appendix~\ref{sec:adaptive_adversary} for the full experimental details and prompts.
As illustrated in Figure~\ref{fig:dialog:sr-adv} and \ref{fig:dialog:ineff-adv}, \ours demonstrates robust SR control even against a strategy-aware adversary, effectively adapting to shifting adversarial tactics. {\color{\hl}At the very first stage}, the adversary attempts to maximize the risk by exclusively presenting difficult questions.
However, this aggressive strategy drives the learner toward a conservative, frequently-abstaining policy, which reduces the SR since abstention incurs zero violation.
Consequently, in the later stages, the adversary adapts by mixing incorrect (hard but overconfident) and correct (easy but under-confident) questions to exploit the learner.
However, despite this strategic shift creating a highly non-stationary environment, \ours continuously adapts its policy, successfully controlling the SR under the target $\alpha$. This observation empirically validates our theoretical robustness (Lemma \ref{lem:conversion}), confirming that \ours remains effective even when the adversary actively manipulates the feedback distribution to induce failures.

\para{Interactive environment.}
We simulate a real-world interactive application, whose question stream is non-stationary, with three agents (Appendix~\ref{sec:intenvsetup}).
A user-acting agent (GPT-5.4-mini) asks questions about a context passage from SQuAD~\citep{rajpurkar2016squad} and switches to a new passage when it is done or after at most 10 turns, GPT-5.4 answers each question, and an evaluating agent (GPT-5.4-mini) provides the feedback; \ours decides whether to abstain from each answer (Figure~\ref{fig:method_overview}).
As shown in Figure~\ref{fig:dialog:sr-int} and \ref{fig:dialog:ineff-int}, the SR of \ours decreases steadily toward $\alpha$ and reaches about $0.11$ at $T=15\mathrm{K}$, while \texttt{No-CA} remains around $0.19$.


\section{Conclusion} \label{sec:conc}

We propose a learning algorithm for online {\color{\hl}conformal abstention} with  adversarial bandit feedback, which provides an SR controllability guarantee
by leveraging the adversarial bandit problem.
In particular, 
we reduce {\color{\hl}conformal abstention} to adversarial bandits for leveraging any bandit methods for learning with adversarial bandit feedback and their regret bounds. 
To interpret the regret bound back to the SR bound, we introduce a novel \emph{Regret-to-SR conversion lemma}, so the any adversarial bandit methods for conformal abstention can be used to control an SR at a desired level.
However, the existing bandit methods do not fully exploit the problem structure of conformal abstention so they suffer from sample inefficiency. To address this issue, we 
propose a novel \ours algorithm for adversarial bandits with \emph{partial feedback unlocking}, which addresses the lack of feedback information due to the nature of partial feedback.
We theoretically and empirically justify our method by providing an efficient SR violation bound and evaluating diverse learning environments, 
including stochastic, distribution-shift, interactive, and adaptive adversarial environments. 

\para{Limitation.} While the SR guarantee holds for any generator and scoring function, its practical efficiency depends on their quality: with a poorly calibrated score or a weak generator, \ours may abstain more often and need a longer horizon for the SR to approach $\alpha$ (Appendix~\ref{sec:discussion:ineff}).
In particular, our method uses the context only through the scoring function $f$, so extending it to context-aware abstention, \eg via contextual bandits, is a promising direction to reduce this dependence.

\clearpage
\section*{Acknowledgment}
We appreciate constructive feedback by anonymous reviewers. 
This work was supported by Institute of Information \& communications Technology Planning \& Evaluation (IITP) and the National Research Foundation of Korea (NRF) grant, funded by the Korea government (MSIT) (RS-2019-II191906, Artificial Intelligence Graduate School Program (POSTECH) (5\%); RS-2024-00457882, National AI Research Lab Project (20\%); No. RS-2024-00509258 and No. RS-2024-00469482, Global AI Frontier Lab (30\%); RS-2025-00560062 (20\%); RS-2026-25531812, Safety Shielding(20\%); RS-2026-25616313, Star Fellowship (5\%)).
\bibliographystyle{abbrvnat}
\bibliography{tml-lab,llm,sml}

@inproceedings{
kuhn2023semantic,
title={Semantic Uncertainty: Linguistic Invariances for Uncertainty Estimation in Natural Language Generation},
author={Lorenz Kuhn and Yarin Gal and Sebastian Farquhar},
booktitle={The Eleventh International Conference on Learning Representations },
year={2023},
url={https://openreview.net/forum?id=VD-AYtP0dve}
}

@inproceedings{joshi2017triviaqa,
    title = "{T}rivia{QA}: A Large Scale Distantly Supervised Challenge Dataset for Reading Comprehension",
    author = "Joshi, Mandar  and
      Choi, Eunsol  and
      Weld, Daniel  and
      Zettlemoyer, Luke",
    booktitle = "Proceedings of the 55th Annual Meeting of the Association for Computational Linguistics (Volume 1: Long Papers)",
    month = jul,
    year = "2017",
    address = "Vancouver, Canada",
    publisher = "Association for Computational Linguistics",
    url = "https://aclanthology.org/P17-1147",
    doi = "10.18653/v1/P17-1147",
    pages = "1601--1611",
}

@inproceedings{manakul2023selfcheckgpt,
  title={SelfCheckGPT: Zero-Resource Black-Box Hallucination Detection for Generative Large Language Models},
  author={Manakul, Potsawee and Liusie, Adian and Gales, Mark},
  booktitle={The 2023 Conference on Empirical Methods in Natural Language Processing},
  year={2023}
}

@inproceedings{bowman2015large,
  title={A large annotated corpus for learning natural language inference},
  author={Bowman, Samuel and Angeli, Gabor and Potts, Christopher and Manning, Christopher D},
  booktitle={Proceedings of the 2015 Conference on Empirical Methods in Natural Language Processing},
  pages={632--642},
  year={2015}
}

@article{mohri2023learning,
  title={Learning to reject with a fixed predictor: Application to decontextualization},
  author={Mohri, Christopher and Andor, Daniel and Choi, Eunsol and Collins, Michael},
  journal={arXiv preprint arXiv:2301.09044},
  year={2023}
}

@inproceedings{stelmakh2022asqa,
  title={ASQA: Factoid questions meet long-form answers},
  author={Stelmakh, Ivan and Luan, Yi and Dhingra, Bhuwan and Chang, Ming-Wei},
  booktitle={Proceedings of the 2022 Conference on Empirical Methods in Natural Language Processing},
  pages={8273--8288},
  year={2022}
}

@inproceedings{min2023factscore,
  title={FActScore: Fine-grained atomic evaluation of factual precision in long form text generation},
  author={Min, Sewon and Krishna, Kalpesh and Lyu, Xinxi and Lewis, Mike and Yih, Wen-tau and Koh, Pang and Iyyer, Mohit and Zettlemoyer, Luke and Hajishirzi, Hannaneh},
  booktitle={Proceedings of the 2023 conference on empirical methods in natural language processing},
  pages={12076--12100},
  year={2023}
}

@article{wei2024long,
  title={Long-form factuality in large language models},
  author={Wei, Jerry and Yang, Chengrun and Song, Xinying and Lu, Yifeng and Hu, Nathan and Huang, Jie and Tran, Dustin and Peng, Daiyi and Liu, Ruibo and Huang, Da and others},
  journal={Advances in Neural Information Processing Systems},
  volume={37},
  pages={80756--80827},
  year={2024}
}

@inproceedings{yuan2025fact,
  title={Fact-Level Calibration and Correction for Long-Form Generations},
  author={Yuan, Yige and Xu, Bingbing and Tan, Hexiang and Sun, Fei and Xiao, Teng and Li, Wei and Shen, Huawei and Cheng, Xueqi},
  booktitle={Proceedings of the 48th International ACM SIGIR Conference on Research and Development in Information Retrieval},
  pages={2807--2811},
  year={2025}
}

@inproceedings{zhang2025atomic,
  title={Atomic calibration of llms in long-form generations},
  author={Zhang, Caiqi and Yang, Ruihan and Zhang, Zhisong and Huang, Xinting and Yang, Sen and Yu, Dong and Collier, Nigel},
  booktitle={Proceedings of the 14th International Joint Conference on Natural Language Processing and the 4th Conference of the Asia-Pacific Chapter of the Association for Computational Linguistics},
  pages={148--169},
  year={2025}
}

@article{goren2026should,
  title={When Should LLMs Be Less Specific? Selective Abstraction for Reliable Long-Form Text Generation},
  author={Goren, Shani and Galil, Ido and El-Yaniv, Ran},
  journal={arXiv preprint arXiv:2602.11908},
  year={2026}
}

@misc{gibbs2021adaptive,
	archiveprefix = {arXiv},
	author = {Isaac Gibbs and Emmanuel Cand{\`e}s},
	eprint = {2106.00170},
	primaryclass = {stat.ME},
	title = {Adaptive Conformal Inference Under Distribution Shift},
	year = {2021}}

@book{cesa2006prediction,
	author = {Cesa-Bianchi, Nicolo and Lugosi, G{\'a}bor},
	publisher = {Cambridge university press},
	title = {Prediction, learning, and games},
	year = {2006}}

@book{mohri2012foundations,
	author = {Mohri, Mehryar and Rostamizadeh, Afshin and Talwalkar, Ameet},
	publisher = {MIT press},
	title = {Foundations of machine learning},
	year = {2012}}

@book{vovk2005algorithmic,
	author = {Vovk, Vladimir and Gammerman, Alex and Shafer, Glenn},
	publisher = {Springer Science \& Business Media},
	title = {Algorithmic learning in a random world},
	year = {2005}}

@article{geifman2017selective,
  title={Selective classification for deep neural networks},
  author={Geifman, Yonatan and El-Yaniv, Ran},
  journal={Advances in neural information processing systems},
  volume={30},
  year={2017}
}

@article{kwiatkowski2019natural,
  title={Natural questions: a benchmark for question answering research},
  author={Kwiatkowski, Tom and Palomaki, Jennimaria and Redfield, Olivia and Collins, Michael and Parikh, Ankur and Alberti, Chris and Epstein, Danielle and Polosukhin, Illia and Devlin, Jacob and Lee, Kenton and others},
  journal={Transactions of the Association for Computational Linguistics},
  volume={7},
  pages={453--466},
  year={2019},
  publisher={MIT Press One Rogers Street, Cambridge, MA 02142-1209, USA journals-info~…}
}

@article{goren2024hierarchical,
  title={Hierarchical Selective Classification},
  author={Goren, Shani and Galil, Ido and El-Yaniv, Ran},
  journal={arXiv preprint arXiv:2405.11533},
  year={2024}
}

@article{angelopoulos2021learn,
  title={Learn then test: Calibrating predictive algorithms to achieve risk control},
  author={Angelopoulos, Anastasios N and Bates, Stephen and Cand{\`e}s, Emmanuel J and Jordan, Michael I and Lei, Lihua},
  journal={arXiv preprint arXiv:2110.01052},
  year={2021}
}

@article{mahdavi2012trading,
  title={Trading regret for efficiency: online convex optimization with long term constraints},
  author={Mahdavi, Mehrdad and Jin, Rong and Yang, Tianbao},
  journal={The Journal of Machine Learning Research},
  volume={13},
  number={1},
  pages={2503--2528},
  year={2012},
  publisher={JMLR. org}
}

@article{chen2017online,
  title={An online convex optimization approach to proactive network resource allocation},
  author={Chen, Tianyi and Ling, Qing and Giannakis, Georgios B},
  journal={IEEE Transactions on Signal Processing},
  volume={65},
  number={24},
  pages={6350--6364},
  year={2017},
  publisher={IEEE}
}

@article{sid2024exponentially,
  title={Exponentially Weighted Algorithm for Online Network Resource Allocation with Long-Term Constraints},
  author={Sid-Ali, Ahmed and Lambadaris, Ioannis and Zhao, Yiqiang Q and Shaikhet, Gennady and Asgharnia, Amirhossein},
  journal={arXiv preprint arXiv:2405.02373},
  year={2024}
}

@article{ge2024stochastic,
  title={Stochastic Online Conformal Prediction with Semi-Bandit Feedback},
  author={Ge, Haosen and Bastani, Hamsa and Bastani, Osbert},
  journal={arXiv preprint arXiv:2405.13268},
  year={2024}
}

@article{sinha2024optimal,
  title={Optimal algorithms for online convex optimization with adversarial constraints},
  author={Sinha, Abhishek and Vaze, Rahul},
  journal={Advances in Neural Information Processing Systems},
  volume={37},
  pages={41274--41302},
  year={2024}
}

@article{littlestone1994weighted,
  title={The weighted majority algorithm},
  author={Littlestone, Nick and Warmuth, Manfred K},
  journal={Information and computation},
  volume={108},
  number={2},
  pages={212--261},
  year={1994},
  publisher={Elsevier}
}

@misc{foster2023foundations,
      title={Foundations of Reinforcement Learning and Interactive Decision Making}, 
      author={Dylan J. Foster and Alexander Rakhlin},
      year={2023},
      eprint={2312.16730},
      archivePrefix={arXiv},
      primaryClass={cs.LG},
      url={https://arxiv.org/abs/2312.16730}, 
}

@article{auer2002nonstochastic,
  title={The nonstochastic multiarmed bandit problem},
  author={Auer, Peter and Cesa-Bianchi, Nicolo and Freund, Yoav and Schapire, Robert E},
  journal={SIAM journal on computing},
  volume={32},
  number={1},
  pages={48--77},
  year={2002},
  publisher={SIAM}
}

@article{sankararaman2016semi,
  title={Semi-bandit feedback: A survey of results},
  author={Sankararaman, Karthik Abinav},
  journal={ARXIV CoRR},
  year={2016}
}

@misc{openai2026gpt54,
  title        = {{GPT-5.4} Model Card},
  author       = {{OpenAI}},
  howpublished = {\url{https://developers.openai.com/api/docs/models/gpt-5.4}},
  year         = {2026},
}

@misc{meta2024llama31-8binstruct,
  title        = {{LLaMA 3.1 8B-Instruct} Model Card},
  author       = {{Meta AI}},
  howpublished = {\url{https://huggingface.co/meta-llama/Llama-3.1-8B-Instruct}},
  year         = {2024},
}

@article{agrawal1995sample,
  title={Sample mean based index policies by o (log n) regret for the multi-armed bandit problem},
  author={Agrawal, Rajeev},
  journal={Advances in applied probability},
  volume={27},
  number={4},
  pages={1054--1078},
  year={1995},
  publisher={Cambridge University Press}
}

@article{russo2013eluder,
  title={Eluder dimension and the sample complexity of optimistic exploration},
  author={Russo, Daniel and Van Roy, Benjamin},
  journal={Advances in Neural Information Processing Systems},
  volume={26},
  year={2013}
}

@article{rajpurkar2016squad,
  title={Squad: 100,000+ questions for machine comprehension of text},
  author={Rajpurkar, Pranav and Zhang, Jian and Lopyrev, Konstantin and Liang, Percy},
  journal={arXiv preprint arXiv:1606.05250},
  year={2016}
}

@misc{Chase_LangChain_2022,
author = {Chase, Harrison},
month = oct,
title = {{LangChain}},
url = {https://github.com/langchain-ai/langchain},
year = {2022}
}

@article{neu2015explore,
  title={Explore no more: Improved high-probability regret bounds for non-stochastic bandits},
  author={Neu, Gergely},
  journal={Advances in Neural Information Processing Systems},
  volume={28},
  year={2015}
}

@article{bubeck2012regret,
  title={Regret analysis of stochastic and nonstochastic multi-armed bandit problems},
  author={Bubeck, S{\'e}bastien and Cesa-Bianchi, Nicolo and others},
  journal={Foundations and Trends{\textregistered} in Machine Learning},
  volume={5},
  number={1},
  pages={1--122},
  year={2012},
  publisher={Now Publishers, Inc.}
}

@article{jin2023selection,
  title={Selection by prediction with conformal p-values},
  author={Jin, Ying and Cand{\`e}s, Emmanuel J},
  journal={Journal of Machine Learning Research},
  volume={24},
  number={244},
  pages={1--41},
  year={2023}
}

@inproceedings{cortes2016learning,
  title={Learning with rejection},
  author={Cortes, Corinna and DeSalvo, Giulia and Mohri, Mehryar},
  booktitle={International conference on algorithmic learning theory},
  pages={67--82},
  year={2016},
  organization={Springer}
}

@article{wang2022self,
  title={Self-consistency improves chain of thought reasoning in language models},
  author={Wang, Xuezhi and Wei, Jason and Schuurmans, Dale and Le, Quoc and Chi, Ed and Narang, Sharan and Chowdhery, Aakanksha and Zhou, Denny},
  journal={arXiv preprint arXiv:2203.11171},
  year={2022}
}

@article{kadavath2022language,
  title={Language models (mostly) know what they know},
  author={Kadavath, Saurav and Conerly, Tom and Askell, Amanda and Henighan, Tom and Drain, Dawn and Perez, Ethan and Schiefer, Nicholas and Hatfield-Dodds, Zac and DasSarma, Nova and Tran-Johnson, Eli and others},
  journal={arXiv preprint arXiv:2207.05221},
  year={2022}
}

@article{kamath2020selective,
  title={Selective question answering under domain shift},
  author={Kamath, Amita and Jia, Robin and Liang, Percy},
  journal={arXiv preprint arXiv:2006.09462},
  year={2020}
}

@misc{yadkori2024mitigatingllmhallucinationsconformal,
      title={Mitigating LLM Hallucinations via Conformal Abstention}, 
      author={Yasin Abbasi Yadkori and Ilja Kuzborskij and David Stutz and András György and Adam Fisch and Arnaud Doucet and Iuliya Beloshapka and Wei-Hung Weng and Yao-Yuan Yang and Csaba Szepesvári and Ali Taylan Cemgil and Nenad Tomasev},
      year={2024},
      eprint={2405.01563},
      archivePrefix={arXiv},
      primaryClass={cs.LG},
      url={https://arxiv.org/abs/2405.01563}, 
}

@inproceedings{yu2011unimodal,
  title={Unimodal Bandits.},
  author={Yu, Jia Yuan and Mannor, Shie},
  booktitle={ICML},
  pages={41--48},
  year={2011}
}

@article{chen2021understanding,
  title={Understanding bandits with graph feedback},
  author={Chen, Houshuang and Huang, Zengfeng and Li, Shuai and Zhang, Chihao},
  journal={Advances in Neural Information Processing Systems},
  volume={34},
  pages={24659--24669},
  year={2021}
}

@inproceedings{kveton2015cascading,
  title={Cascading bandits: Learning to rank in the cascade model},
  author={Kveton, Branislav and Szepesvari, Csaba and Wen, Zheng and Ashkan, Azin},
  booktitle={International conference on machine learning},
  pages={767--776},
  year={2015},
  organization={PMLR}
}

@inproceedings{weed2016online,
  title={Online learning in repeated auctions},
  author={Weed, Jonathan and Perchet, Vianney and Rigollet, Philippe},
  booktitle={Conference on Learning Theory},
  pages={1562--1583},
  year={2016},
  organization={PMLR}
}

@article{yang2026online,
  title={Online Conformal Prediction with Adversarial Semi-bandit Feedback via Regret Minimization},
  author={Yang, Junyoung and Kim, Kyungmin and Park, Sangdon},
  journal={arXiv preprint arXiv:2604.17984},
  year={2026}
}

@article{mannor2011bandits,
  title={From bandits to experts: On the value of side-observations},
  author={Mannor, Shie and Shamir, Ohad},
  journal={Advances in neural information processing systems},
  volume={24},
  year={2011}
}

@article{alon2013bandits,
  title={From bandits to experts: A tale of domination and independence},
  author={Alon, Noga and Cesa-Bianchi, Nicolo and Gentile, Claudio and Mansour, Yishay},
  journal={Advances in Neural Information Processing Systems},
  volume={26},
  year={2013}
}

@article{jin2020disease,
  title={What Disease does this Patient Have? A Large-scale Open Domain Question Answering Dataset from Medical Exams},
  author={Jin, Di and Pan, Eileen and Oufattole, Nassim and Weng, Wei-Hung and Fang, Hanyi and Szolovits, Peter},
  journal={arXiv preprint arXiv:2009.13081},
  year={2020}
}

@inproceedings{luo2023improved,
  title={Improved high-probability regret for adversarial bandits with time-varying feedback graphs},
  author={Luo, Haipeng and Tong, Hanghang and Zhang, Mengxiao and Zhang, Yuheng},
  booktitle={International Conference on Algorithmic Learning Theory},
  pages={1074--1100},
  year={2023},
  organization={PMLR}
}

@inproceedings{lee2024selective,
    title={Selective Generation for Controllable Language Models},
    author={Minjae Lee and Kyungmin Kim and Taesoo Kim and Sangdon Park},
    booktitle={The Thirty-eighth Annual Conference on Neural Information Processing Systems},
    year={2024},
    url={https://openreview.net/forum?id=glfYOAzh2f}
}

\clearpage
\appendix

\onecolumn


\vspace{-1.5ex}
\section{Preliminaries}\label{sec:bg}
\vspace{-1.5ex}
We introduce preliminaries on language generation, selective prediction (or conformal abstention), and adversarial bandits. 
To this end, 
let 
$\Xs$ be a set of inputs (\eg examples or questions) 
and
$\Ys$ be a set of outputs (\eg labels or answers).

\vspace{-1.5ex}
\subsection{Language Generation}
\label{prelim:lg}
\vspace{-1.5ex}
We mainly consider language generators as our model to control hallucination. 
In particular, 
let $G: \Xs \to \Ys$ be a language generator, where $\Ws$ is a set of tokens and $\Xs = \Ys \coloneqq \cup_{i=0}^\infty \Ws^{i}$. Here, each $i$-th token $\hat\y_i$ of a generated answer $\hat\y \in \Ys$ is decoded from 
an underlying probability distribution $p(\y \mid \x)$, where $p$ is usually learned via language data.
For the decoding strategy, we consider greedy decoding, \ie
$\hat\y_i = \arg\max_{w \in \Ws} p( w \mid \x, \hat\y_{1:i-1})$, where $\y_{a:b}\coloneqq (\y_a, \dots, \y_b)$.
Given a generated answer $\hat\y$, there are multiple ways to measure its likelihood of correctness $f: \Xs \times \Ys \to \realnum$. 
A standard way $f_{\texttt{std}}$ considers a length-normalized token probability, \ie
$f_{\texttt{std}}(\x, \hat\y) \coloneqq (p(\hat\y_1 \mid \mathbf{x})\prod_{i=2}^{\vert\hat{\mathbf{y}}\vert}p(\hat\y_i \mid \mathbf{x}, \mathbf{\hat{y}}_{1:i-1}))^{\nicefrac{1}{|\hat{\y}|}}$. A better alternative is to consider consistency of $\hat\y$ to multiple generated answers with sampling \citep{manakul2023selfcheckgpt}, \ie  $f_{\texttt{con}}(\x, \hat\y) = \hat\Exp_{\y' \sim G(x)} f_E(\y', \hat\y) $, denoted by a consistency score, where we use sampling for decoding $\y'$, $f_E(\y', \hat\y)$ measures an entailment score of $\hat\y$ given $\y'$ (\ie whether $\y'$ entails $\hat\y$) via an entailment model (\eg {\color{\hl}GPT-5.4-nano}). 


\vspace{-1.5ex}
\subsection{Selective Prediction {\color{\hl}and Conformal Abstention}} \label{sec:selective_prediction}
\vspace{-1.5ex}

Selective prediction \citep{geifman2017selective} and generation \citep{lee2024selective} {\color{\hl}(also known as conformal abstention)} provide certified control over the risk of incorrect predictions by abstaining  answers (saying \texttt{IDK}) when the prediction is uncertain.
Given a predictor $\hat\y: \Xs \to \Ys$, a selective predictor $\Sh: \Xs \to \Ys \cup \{\texttt{IDK}\}$ abstains from returning $\hat\y(\x)$ if a selection function $\sh: \Xs \times \Ys \to \{0, 1\}$ deems the prediction uncertain, \ie
$
    \Sh(\x) \coloneqq \begin{cases}
        \hat\y(\x) &\text{if~} \sh(\x, \hat\y(\x)) = 1 \\
        \texttt{IDK} & \text{otherwise}
    \end{cases}.
$
In learning, the selection function $\sh$ is chosen to possibly satisfy a desired level of a false discovery rate \ie $\Prob(\Sh(\x) \neq_E \y \mid \Sh(\x) \neq \texttt{IDK})$ over the i.i.d. samples of $(\x, \y)$.
In selective classification \citep{geifman2017selective}, the equality relation $\neq_E$ is usually the standard exact matching, which is extended to capture semantic correctness via entailment relation in selective generation \citep{lee2024selective}. 
In contrast to the standard stochastic setup as in the previous literature, we consider online learning under partial feedback.


\vspace{-1.5ex}
\subsection{Regret Minimization}\label{sec:prelim:regretmin}
\vspace{-1.5ex}

Sequential prediction in multi-armed bandit problems is formulated as an interactive game between a learner and an adversary (also called an environment). 
In particular, for each time $t \in \{ 1, \dots, T \}$, a learner selects an arm $\tau_t \in \Hs$ among a set of arms $\Hs$ and an adversary selects a loss function $\ell_t: \Hs \to \realnum_{\ge 0}$, where a loss associated to an arm $\tau_t$ is denoted by $\ell_t(\tau_t)$ and provided to the learner as partial feedback (\ie only the loss for the selected arm is provided). Note that this interactive game often describes in rewards, but we consider losses, instead of rewards. 

The main goal of this interactive game is to find a learner whose performance is as good as the best learner in hindsight.
Here, the performance is measured by \emph{regret}, following standard bandits literatures \citep{bubeck2012regret}, \ie
\begin{equation}
    \label{apdx:eq:fullfeedback:regret}
    \Reg_T \coloneqq  \sum_{t=1}^T \ell_t(\tau_t) - \min_{\tau \in \Hs} \sum_{t=1}^T \ell_t(\tau).
\end{equation}
Thus, the goal becomes to find a learner that minimizes the regret such that an achieved regret bound is sublinear in $T$. 
Here, we say that the adversary is \emph{oblivious} if the sequence of loss functions are chosen regardless of learner's previous selections
(or chosen in advance before the game),
and the adversary is \emph{adaptive} if the loss function at time $t$, \ie $\ell_t$, is drawn a distribution that depends on the previously chosen arms, \ie $\tau_1, \dots, \tau_{t-1}$. 
In this paper, we consider the adaptive adversary which is stronger than the oblivious one. Moreover, under the adaptive adversary setup, a deterministic learner cannot win the game (\ie it does not achieve a sub-linear regret bound), so we consider a randomized learner. 
This implies that $\tau_t$ and $\ell_t$ are two sources of randomness, so $\Reg_T$ is a random variable. Thus, we consider the high-probable regret bound or a bound of an expected regret $\Exp \Reg_T$.  
Note that we only consider the learner's randomization in the analysis, since our bounds are derived to hold for any adversary  \citep{bubeck2012regret,neu2015explore}.

In the following section, we review 
one regret minimization algorithm for bandits with an adaptive adversary, called \eeeixns, and see Appendix \ref{sec:algorithm} for other algorithms.

\subsection{Adversarial Bandits and \eeeix Algorithm}\label{sec:prelim:adversarialbandit}

Traditional online learning typically assumes full feedback, \ie loss for any hypothesis is computable. 
However, practical applications may only have partial feedback, \ie loss for a chosen hypothesis is only given. 
We model this via the adversarial multi-armed bandit problem. The regret is as in online learning, but the learner only receives for a chosen arm, while the sequence $\ell_t(\tau_t)$ is chosen by an oblivious or adaptive adversary  that may condition on the history of chosen arms but not on the learner’s fresh randomness at time $t$.


To obtain the high-probability control of the $\Reg_T$ over an adaptive adversary,  Exponential-weights for Exploration and Exploitation with Implicit Exploration (\eeeixns)\citep{neu2015explore} is a natural choice. It employs implicit-exploration (IX) loss estimates within an exponential-weights scheme and does not require explicit mixing with the uniform distribution like \texttt{Exp3.P} \citep{auer2002nonstochastic}. In particular, \eeeix uses the variance-reduced loss estimator for each arm $\tau\in\Hs$ constructed from the observed loss of the chosen arm $\tau_t$, \ie
\begin{align}
    \tilde\ell_t(\tau \mid \{\tau_t\}) \coloneqq \frac{\ell_t(\tau)}{\gamma_t + p_t(\tau)} \cdot \mathbbm{1}(\tau_t = \tau) \in [0, \infty).
    \label{eq:exp3ix:unbiasedestimator}
\end{align}
where $\gamma_t>0$ is properly selected by an analysis; thus by using \eqref{eq:exp3ix:unbiasedestimator} in $\Reg_T$, with non-increasing learning rate $\eta_t$, we have the following regret with high probability for \eeeix. 
See Appendix\ref{proof:thm:exp3ix:regretbound} for a complete proof.

\begin{theorem}{\citep{neu2015explore}} \label{thm:exp3ix:regretbound}
    Let $\ell_t(\cdot) \in [0, \ell_\text{max}]$.
    For any $T \in \naturalnum$,  finite arms $\Hs$,
    and
    $\delta \in (0, 1)$,
    Algorithm \ref{alg:exp3ix} provides the following regret bound with probability at least$1-\delta$ if $\eta_t = \frac{2\gamma_t}{\ell_\text{max}} = \sqrt{{2 \ln|\mathcal{H}|}/{(\ell_{\max}^2T|\mathcal{H}|)}}$:
    \begin{equation}
        \Reg_T
        \le 
        \ell_\text{max}\(2\sqrt{2T|\Hs|\ln|\Hs|} + \(1 + \sqrt{\frac{T|\Hs|}{2\ln|\Hs|}}\)\ln\delta^{-1}\).
        \label{eq:regretbound:exp3ix}
    \end{equation}
    
\end{theorem}
We will exploit \eeeix for learning selective generators with partial feedback under an adaptive adversary. See Appendix \ref{sec:additionalmethod} for other online learning and bandit algorithms.

\subsection{Online Learning and \ew Algorithm}
\label{sec:additionalprelim}

Online learning considers to design a learner that adapts to any sequence of full feedback on hypotheses without any or with little assumption on data generation process. In particular, 
the goal of an online learner is to
find a distribution $p_t$ over hypotheses with full feedback that minimizes the regret. 

Along with many online learning algorithms, we introduce
Exponential Weighting (\ewns) \citep{littlestone1994weighted}, which minimizes the regret with finite hypotheses (Algorithm \ref{alg:ew}).
In particular, 
\ew maintains the goodness of weights for each hypothesis in terms of cumulative loss and updates them from feedback. The feedback of a hypothesis $\tau_t$ at time $t$ is represented by loss $\ell_t(\tau_t)$ but we can compute this loss for all hypotheses other than a chosen hypothesis $\tau_t$ as we directly observe a true label $\mathbbm{1} (G(\x_t) =_E \y_t)$ at the same time $t$. This \ew algorithm satisfies the following regret bound. See Appendix \ref{proof:thm:ew:regretbound} for a proof. 

\begin{theorem}{\citep{littlestone1994weighted,foster2023foundations}} \label{thm:ew:regretbound}
    Let $\ell_t(\cdot) \in [0, \ell_\text{max}]$.
    For any $T \in \naturalnum$ and finite hypotheses $\Hs$,
    Algorithm \ref{alg:ew} provides the following expected regret bound if $\eta = \sqrt{{8 \ln |\Hs|}/{(\ell_\text{max}^2 T)}}$:
    \begin{equation}
        \sum_{t=1}^T \Exp_{\tau_t \sim p_t} \ell_t(\tau_t)
        - \min_{\tau \in \Hs} \sum_{t=1}^T \ell_t(\tau)
        \le \ell_\text{max} \sqrt{{T\ln |\Hs|} / {2}} = \Os(\ell_\text{max} \sqrt{T \ln |\Hs|}),
    \end{equation}
    which holds for arbitrary loss sequences,
    thus, implying that $\Exp \Reg_T$ is also bounded by $\Os(\ell_\text{max}\sqrt{T\ln |\Hs|})$ as the linearity of expectation and Tower property.
    Moreover, for any $0 < \delta < 1$, the following regret bound also holds:
    \begin{equation*}
        \sum_{t=1}^T \ell_t(\tau_t) - \min_{\tau \in \Hs} \sum_{t=1}^T \ell_t(\tau) \le
        \ell_\text{max}\sqrt{\frac{T\ln|\Hs|}{2}} + \ell_\text{max}\sqrt{\frac{T\ln\delta^{-1}}{2}},
    \end{equation*}
    with probability at least $1-\delta$.
\end{theorem}

We will leverage and extend this \ew for learning selective generators with full feedback.


\section{Additional Related Work}
\label{sec:additionalrel}

\para{Constrained Online Learning}
Several works have studied online learning with constraints. \citet{mahdavi2012trading} first proposes long-term constraints in online convex optimization with time-invariant constraints, which allows to have the constraint function $g_t$ to be violated at each time step, \ie $g(\x_t) > 0$, but eventually satisfied, \ie $\sum_{t=1}^T g(\x_t) \le 0$.
Some other works solve a similar problem but with time-varying constraints, which is more challenging.
\citet{chen2017online, sid2024exponentially} solve a network resource allocation task by viewing this as an constrained optimization problem. Especially, \citet{sid2024exponentially} solves the problem without a convex assumption on loss.
\citet{sinha2024optimal} gives analysis on optimal algorithms for online convex optimization with adversarial constraints.
We similarly consider a SR guarantee to be satisfied over time but with bandit feedback. 

\para{Conformal Prediction with Bandit Feedback}
There is little work about online conformal prediction \citep{gibbs2021adaptive} in bandit settings. 
\citet{ge2024stochastic} proposes an online conformal prediction method with semi-bandit feedback, where we can observe the true label only if it is predicted. 
{\color{\hl}In a concurrent work, \citet{yang2026online} also explored online conformal prediction under a similar semi-bandit feedback setting. 
Our method shares the practical motivation of utilizing bandit feedback, but we specifically focus on {\color{\hl} conformal abstention}.}

{\color{\hl}
\para{Feedback Unlocking and Structured Bandits.}
Feedback unlocking is related to structured bandits, which improve sample efficiency by exploiting structure among arms, either in the expected rewards, \eg unimodality over a graph of arms~\citep{yu2011unimodal}, or in the feedback, \eg cascading feedback, where a click reveals the attraction of all preceding items~\citep{kveton2015cascading}.
Closest to ours are bandits with feedback graphs~\citep{mannor2011bandits,alon2013bandits,chen2021understanding}, where pulling an arm also reveals the losses of its neighbors and the regret scales with graph quantities such as the independence number instead of the number of arms; \citet{neu2015explore} extends \eeeix to this setting, and \citet{luo2023improved} obtain high-probability bounds for time-varying graphs.
Inspired by this line of work, we tailor graph feedback to conformal abstention: the graph is not given externally but induced by the monotonic threshold structure, and consists of two cliques, $\{\tau \le f_t\}$ and $\{\tau > f_t\}$, that change with $f_t$.
Its independence number is thus at most two, which yields regret of the full-information order (Theorem~\ref{thm:ours:regretbound}), and our estimator~(\ref{eq:betteradvbandit:ourunbiasedestimator}) takes the form of the IX estimator of \citet{neu2015explore}.
A similar one-sided feedback appears in repeated auctions, where the bidder observes its value only when it wins~\citep{weed2016online}.

}

\section{Additional Method}
\label{sec:additionalmethod}

\subsection{Online {\color{\hl} Conformal Abstention} with Full Feedback}
\label{sec:osg-fullfeedback}

Here, we specifically leverage the conventional exponential weighting (\ewns) algorithm \citep{littlestone1994weighted} for online learning under adaptive adversaries to re-purpose it for addressing online {\color{\hl} conformal abstention} with full feedback. This is an oracle method with the performance upper bound of our method with partial feedback. 
To this end, we provide the reduction of online {\color{\hl} conformal abstention} with full feedback to the online learning problem. 
Then, 
we introduce a modified \ew algorithm for {\color{\hl} conformal abstention}, called \texttt{EW-{\color{\hl}CA}}, followed by its regret bound. 
Finally, 
we leverage the proposed conversion lemma, which converts the regret bound of any online learning algorithms into the SR bound, to show the SR controllability of \texttt{EW-{\color{\hl}CA}} at a desired level.


\para{Reduction: From Online {\color{\hl} Conformal Abstention} to Online Learning.}
We convert online {\color{\hl} conformal abstention} with full feedback to conventional online learning  
(See Table \ref{tab:fullfeedback:reduction} for a summary).
In {\color{\hl} conformal abstention}, 
models are defined as a set of {\color{\hl} conformal abstainer}s $\Sh$ or equivalently a set of selection parameters $\Hs$, considered as hypotheses $\Hs$ in online learning. 
This model is sequentially updated by leveraging online feedback. In particular, we consider feedback whether a generated answer $G(\x_t)$ is semantically incorrect with respect to a true answer $\y_t$, \ie $\mathbbm{1}( G(\x_t) \neq_E \y_t )$. 
This essentially provides full feedback for any $\tau \in \Hs$ as it is the same as $\mathbbm{1}(\Sh(\x_t; \tau) \neq_E \y_t )$ by the definition of $\Sh$ if $\Sh(\x_t; \tau)\neq \texttt{IDK}$.
This feedback is converted into loss in online learning, as defined in (\ref{eq:loss-sg}).
Note that $d_t$ in the loss leverages the feedback $\mathbbm{1}( G(\x_t) \neq_E \y_t )$ by the definition.
Finally, our goal of finding a learning algorithm, which controls the SR by $\alpha$ and minimizes selection inefficiency in online {\color{\hl} conformal abstention}, is reduced to finding an online learning algorithm that minimizes the regret $\Reg_T$ with any loss sequences $\ell_t(\tau, \alpha)$ in (\ref{eq:loss-sg}).

\begin{table}[th]
    \caption{From online {\color{\hl} conformal abstention} to online learning}
    \label{tab:fullfeedback:reduction}
    \centering
    
    \begin{tabular}{c | c c}
        \toprule
        & 
        online {\color{\hl} conformal abstention} 
        & 
        online learning 
        \\
        \midrule
        models & 
        \makecell{{\color{\hl} conformal abstainers} $\Hs$} 
        & 
        \makecell{hypotheses $\Hs$}
        \\
        \midrule
        feedback &
        $\mathbbm{1}( G(\x_t) \neq_E \y_t )$
        &
        $\ell_t(\tau, \alpha)$ 
        \\
        \midrule
        metric
        &
        $\Vs_T$ and $\Ine_T$
        &
        $\Reg_T$
        \\
        \bottomrule
    \end{tabular}
\end{table}

\para{Algorithm and Its Regret Bound.}
We re-purpose \ew for {\color{\hl} conformal abstention} called \texttt{EW-{\color{\hl}CA}}. 
This mainly uses 
the special loss
$\ell_t(\tau, \alpha)$ in (\ref{eq:loss-sg}) for {\color{\hl} conformal abstention}. 
See Algorithm \ref{alg:fullfeedback:ewsg} for details. 
The corresponding regret bound is directly obtained from Theorem \ref{thm:ew:regretbound} where $\ell_\text{max} \coloneqq \max(\lambda, 1 + \lambda\alpha) \le 1+\lambda$.

\para{Regret-to-SR Conversion.}
Once we devise a regret minimization algorithm, \ie \texttt{EW-CA}, the regret guarantee is simply interpreted as the SR guarantee by Lemma \ref{lem:conversion}. 
This eventually provides the SR bound for \texttt{EW-CA}.

\section{Algorithms}
\label{sec:algorithm}

\begin{algorithm}[h]
    \caption{Procedure for Computing the Loss (\ref{eq:loss-sg}) with Partial Feedback}
    \label{alg:computeloss}
    \begin{algorithmic}[1]
        \Procedure{ComputeLoss}{$\hat S, e, \alpha, \lambda$} 

            \State $a \gets \mathbbm{1}( \hat S = \texttt{IDK} )$
            
            \State $d \gets \mathbbm{1}( \hat S \neq\texttt{IDK})\cdot e -\alpha \mathbbm{1}(\hat S\neq\texttt{IDK})
            +\alpha$
            \State $\ell \gets \frac{a + \lambda d}{1 + \lambda}$

            \State \textbf{return} $\ell$
            
        \EndProcedure
    \end{algorithmic}
\end{algorithm}

\begin{algorithm}[h]
    \caption{Exponential Weighting (\ewns)  \citep{littlestone1994weighted,foster2023foundations}}
    \label{alg:ew}
    \begin{algorithmic}[1]
        \Procedure{EW}{$T, \Hs, \eta$}
        \State $w_1 \leftarrow (1/|\Hs|, \dots, 1/|\Hs|)$
        \For{$t = 1, \ldots, T$}
            \State Observe $\x_t$
            \State Predict $\hat\y(\x_t; \tau_t)$ for $\tau_t \sim p_t(\tau) = w_t(\tau) / \sum_{\bar\tau \in \Hs} w_t(\bar\tau)$
            \State \text{Observe} $\y_t$
            \State \text{Update} $w_{t+1}(\tau) \propto w_t(\tau) \exp(-\eta \ell(\y_t, \hat\y(\x_t; \tau)) )$ \text{for all} $\tau \in \Hs$
        \EndFor
        \EndProcedure
    \end{algorithmic}
\end{algorithm}

            

            
            
            

\begin{algorithm}[h]
    \caption{Exp3 with Implicit eXploration (\eeeixns)  \citep{neu2015explore}}
    \label{alg:exp3ix}
    \begin{algorithmic}[1]
        \Procedure{Exp3-IX}{$T, \Hs, \eta_t, \gamma_t$}
        \State $w_1 \leftarrow (1/|\Hs|, \dots, 1/|\Hs|)$
        \For{$t = 1, \ldots, T$}
            

            \State Choose an arm $\tau_t \sim p_t(\tau) = w_t(\tau) / \sum_{\bar\tau \in \Hs} w_t(\bar\tau)$
            
            \State Observe $\ell_t(\tau_t)$
            
            \State $\hat{\ell}_t(\tau) \gets \frac{\ell_t(\tau_t)}{\gamma_t + p_t(\tau)}\mathbbm{1}(\tau_t = \tau)$ for all $\tau \in \Hs$
            
            \State Update $w_{t+1}(\tau) \propto 
            \exp(-\eta_t \sum_{s=1}^t \hat{\ell}_s(\tau))$
            for all $\tau \in \Hs$
        \EndFor
        \EndProcedure
    \end{algorithmic}
\end{algorithm}

\begin{algorithm}[h]
    \caption{Exponential Weighting for Online {\color{\hl} Conformal Abstention} (\texttt{EW-{\color{\hl}CA}})
    }
    \label{alg:fullfeedback:ewsg}
    \begin{algorithmic}[1]
        \Procedure{EW-{\color{\hl}CA}}{$T, \Hs, \alpha, \lambda, \eta$} 

        \State $w_1(\tau) \gets 1/|\Hs|$ for all $\tau \in \Hs$
        \For{$t = 1, \ldots, T$}
            \State Observe $\x_t$
            \State Predict $\Sh_t(\x_t; \tau_t)$ \text{where} $\tau_t \sim p_t(\tau) = 
            w_t(\tau) / \sum_{\bar\tau \in \Hs} w_t(\bar\tau)$
            \State Observe $\y_t$ thus observe $\mathbbm{1}\(G(\x_t) \neq_E \y_t\)$

            \State $a_t(\tau) \gets \mathbbm{1}( \Sh(\x_t; \tau) = \texttt{IDK} )$

            \State $d_t(\tau, \alpha) \gets \mathbbm{1}( \Sh(\x_t; \tau) \neq \texttt{IDK} \wedge G(\x_t) \neq_E \y_t ) - \alpha\mathbbm{1}( \Sh(\x_t; \tau) \neq \texttt{IDK}) + \alpha$
            
            \State $\ell_t\left(\tau, \alpha \right) \gets \frac{a_t(\tau) + \lambda d_t(\tau, \alpha)}{1+\lambda}$ for all $\tau \in \Hs$

            
            \State Update $w_{t+1}(\tau) \propto w_{t}(\tau) \exp \left\{- \eta \cdot \ell_t\left(\tau, \alpha \right) \right\}$ for all $\tau \in \Hs$
        \EndFor
        \EndProcedure
    \end{algorithmic}
\end{algorithm}

\begin{algorithm}[h]
    \caption{\eeeix for Online {\color{\hl} Conformal Abstention} with Partial Feedback (\texttt{Exp3-IX-{\color{\hl}CA}})}
    \label{alg:exp3ix-sg}
    \begin{algorithmic}[1]
        \Procedure{Exp3-IX-{\color{\hl}CA}}{$T, \Hs, \alpha, \lambda, \eta, \gamma$} 
        \State $w_1(\tau) \gets 1/|\Hs|$ for all $\tau \in \Hs$
        \For{$t = 1, \ldots, T$}
            \State \text{Observe} $\x_t$
            \State Predict $\Sh_t(\x_t; \tau_t)$ \text{where} $\tau_t \sim p_t(\tau) = \frac{w_t(\tau)}{\sum_{\bar\tau \in \Hs} w_t(\bar\tau)}$

            \State Observe $e_t$
            \State $\ell_t\left(\tau, \alpha \right) \gets \textsc{ComputeLoss}(\Sh(\x_t; \tau), e_t, \alpha, \lambda)$ for $\tau_t$
            

            \State $\ell_t\left(\tau, \alpha \mid \{\tau_t\} \right) \gets \frac{\ell_t(\tau, \alpha) }{\gamma + p_t(\tau)}\cdot\mathbbm{1}(\tau_t = \tau)$ for all $\tau \in \Hs$
            

            \State Update $w_{t+1}(\tau) \propto {\exp(-\eta \sum_{s=1}^t \ell_s\left(\tau, \alpha \mid \{\tau_t\} \right))}$
            for all $\tau \in \Hs$
            
        \EndFor
        \EndProcedure
    \end{algorithmic}
\end{algorithm}

\clearpage
\section{Experiment Setup}
\label{sec:expsetup}
\subsection{Computing Environment}
We use 4 NVIDIA A100 80GB with 128 CPUs for experiments.

\subsection{Distribution-shift Environment Setup}
\label{sec:distshiftenvsetup}

We describe the experimental setup used to evaluate the robustness of {\color{\hl} conformal abstention} by three distinct distribution-shift scenarios. We detail the chunking strategy and the procedure for constructing each shift environment in the following.

\para{Single Shift.} 
The dataset is composed of two large homogeneous segments, each containing 15$\mathrm{K}$ examples randomly sampled from NQ and TriviaQA. A single distribution shift occurs at the boundary between these two segments.

\para{Alternating Shift.} The dataset constructed by a sequence of ten 3$\mathrm{K}$-example chunks that alternate between NQ and TriviaQA (e.g., NQ, TriviaQA, NQ, TriviaQA and so forth). Each chunk induces a distribution transition at its endpoint, resulting in frequent and periodic shifts.

\para{Gradual Shift.} This setup generates a smooth distribution change over time by drawing individual samples probabilistically between NQ and TriviaQA, with the sampling probability gradually shifting from pure NQ (or TriviaQA) at the start to pure TriviaQA (or NQ) at the end. 
In particular, given a multinomial distribution $[\frac{t}{T}, 1 - \frac{t}{T}]$ for each dataset that changes over time, we first randomly choose a dataset which follows this multinomial distribution, and then we sample one question-answering pair in the sampled dataset.
Over the full sequence length, this creates linear interpolation between the two datasets, enabling analysis of model performance under a progressive distribution shift.

\subsection{Interactive Environment Setup}
\label{sec:intenvsetup}

For the interactive environment, we consider dialog-based conversations. 
In particular,
{\color{\hl}GPT-5.4} serves as a question-answering agent, while we utilize two {\color{\hl}GPT-5.4-mini} APIs, where one is a user-acting agent to generate questions, and another is an evaluating agent, having access to the responses. 

To instruct the user-acting agent to generate questions,
we use the SQuAD \citep{rajpurkar2016squad} dataset,
which consists of many question-answer pairs, each associated with a gold context.
Here, we discard the QA pairs and only use $20,962$ context passages, which vary over a range of topics (\eg science, history, celebrities, entertainment, etc.), forming a diverse set of natural distributions.
During the simulation, we randomly select a context from the dataset and provide it to the user-acting agent, which is prompted to continuously generate questions based on the given context.
(\eg Figure \ref{fig:method_overview}).
To mimic a dynamically shifting environment, \eg the user's interest shifts over time, we further instruct the user-acting agent to change the context during the interaction when it determines it is done with the current one (\eg \texttt{"When you are done with the current context, say 'SHIFT' to request a new context."}) or after at most 10 question-answering turns.
To design this custom workflow, we implement the core functions of the simulation, including memory handling and multi-turn conversation, using LangGraph, a framework built on LangChain \citep{Chase_LangChain_2022}.

\subsection{Adaptive-Adversary Environment Setup}
\label{sec:adaptive_adversary}

We simulated a strategy-aware adaptive adversary ({\color{\hl}GPT-5.4-mini}) targeting a learner employing {\color{\hl}GPT-5.4} as the generator, over the combined NQ and TriviaQA datasets. In this setup, a user-acting agent observes the learner's decision history and strategically selects questions to maximize the learner's failure—specifically targeting accepted incorrect answers and rejected correct answers—in alignment with the adaptive adversary defined in Section \ref{sec:conversion}.

Note that this simulation employs an LLM assigned an adversarial role to actively induce failures, serving as a realistic proxy rather than a theoretically optimal adversary (which implies omniscience). In other words, this black-box adversary does not possess full parametric knowledge of the learner—such as the specific algorithm (\oursns) or the scoring function $f$—but relies solely on observed decisions, and the prompt that the learner adapts over time. The specific prompts used to enforce this adversarial objective are provided below.

\clearpage
\lstdefinestyle{promptstyle}{
    basicstyle=\small\ttfamily,
    frame=single,
    breaklines=true,
    backgroundcolor=\color{gray!5},
    columns=fullflexible
}
\begin{lstlisting}[style=promptstyle]
# System Prompt
You are an adversary engaged in a game with a learner model. Your goal is to expose the learner's weaknesses.

Below is the list of candidate questions you can ask, with the decision history:
{history}

Current Status:
- the current step: {step}
- The total mean of incorrect answers not rejected (IDK==False and label==0) over steps: {total_incorrect_not_rejected}
- The total mean of correct answers rejected (IDK==True and label==1) over steps: {total_correct_rejected}

{instructions}

When you want to find more proper questions for maximizing your objective, say 'SHIFT' to get a new question.
You must decide this **yourself** - do **not** ask whether to 'SHIFT', simply respond with 'SHIFT' by yourself when needed.
\end{lstlisting}
\begin{lstlisting}[style=promptstyle]
# Instructions
**Your Objective:**
You are an adversary trying to **maximize the total mean of incorrect answer not rejected (IDK==False and label==0)** by the learner.
However, you must know that the learner trains a threshold to reject incorrect answers based on your questions and their confidence score of answers,
thus you need to be strategic in asking questions, i.e., you should also **maximize the total mean of correct answers rejected (IDK==True and label==1)** by the learner.
E.g., if the learner starts to reject incorrect answers more often, you should also try to ask other questions that may fool the learner.

Keep in mind that:
- The learner updates its rejection threshold over time.
- Look for IDK patterns in the history list.
- Select questions that are likely to maximize your objective.

**Output Format:**
- If you select a question, respond with **ONLY the index number** (e.g., 5).
- Do not write any other explanation or text.
\end{lstlisting}

\clearpage
\section{Additional Experiments}
\subsection{Stochastic Environment}
\label{exp:addstochastic}

\begin{figure*}[h]
    \centering
    \subfigure[SR over time]{
    \includegraphics[width=0.31\textwidth]{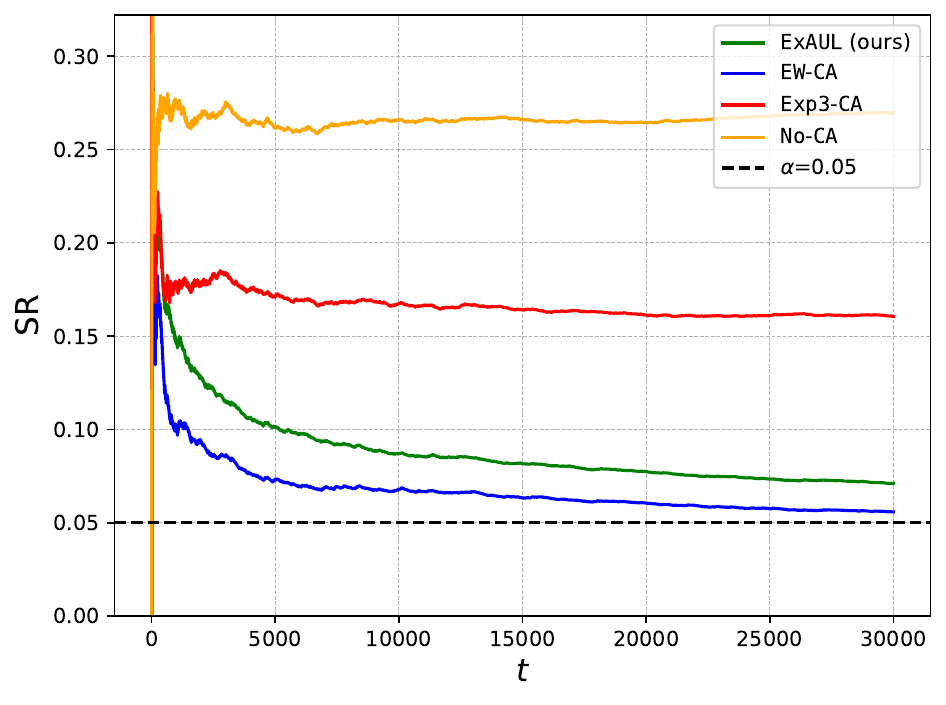}
    }
    \subfigure[Inefficiency over time]{
    \includegraphics[width=0.31\textwidth]{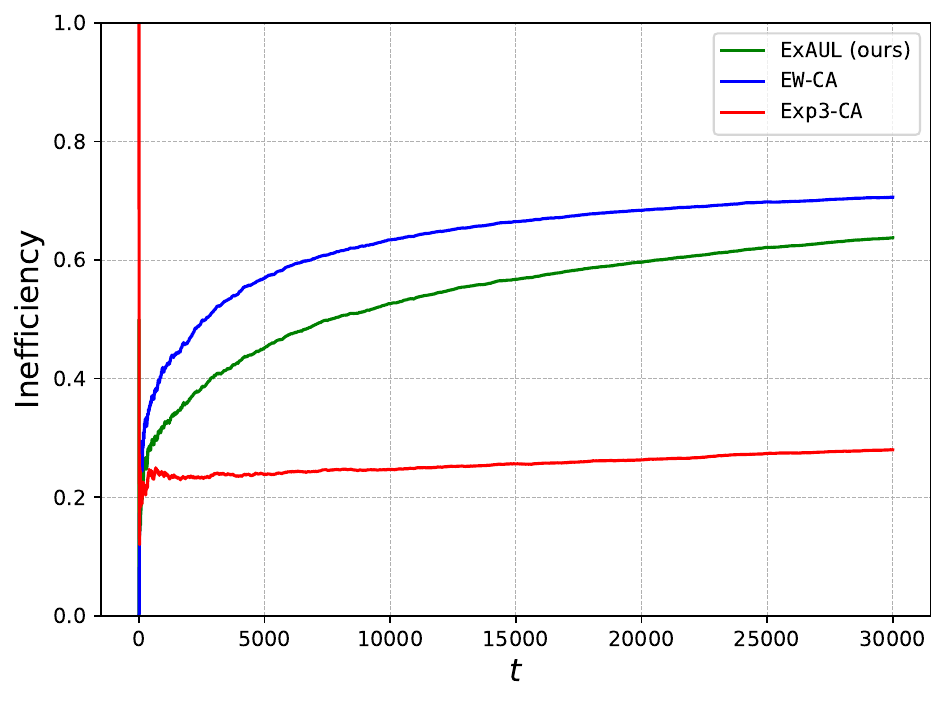}
    }
    \subfigure[SR distribution]{
    \includegraphics[width=0.31\textwidth]{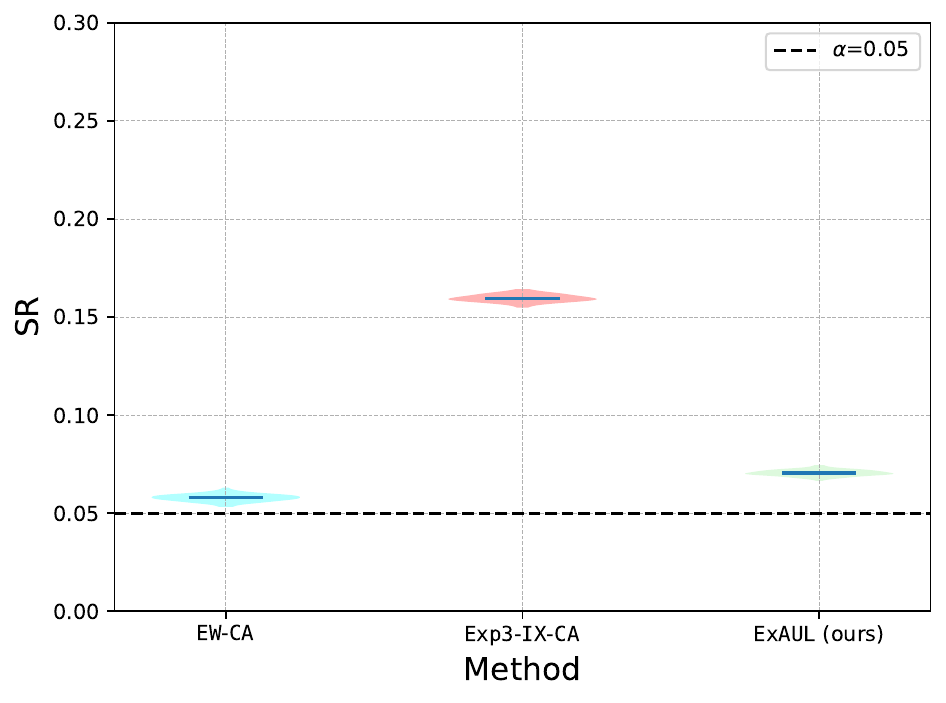}
    }
    \vspace{-1.5ex}
    \caption{Comparison of {\color{\hl}conformal abstention} methods under a stochastic environment with LLaMA3.1-8B-Instruct as a generator on TriviaQA ($T=30\mathrm{K}, \alpha=0.05$). 
    The violin plots are drawn with randomly chosen $30\mathrm{K}$ samples over $100$ random trials.
    }
    \vspace{-1.5ex}
    \label{fig:stochastic:main:llama3.1-8b-inst-triviaqa}
\end{figure*}

\begin{figure*}[h]
    \centering
    \subfigure[SR over time]{
    \includegraphics[width=0.31\textwidth]{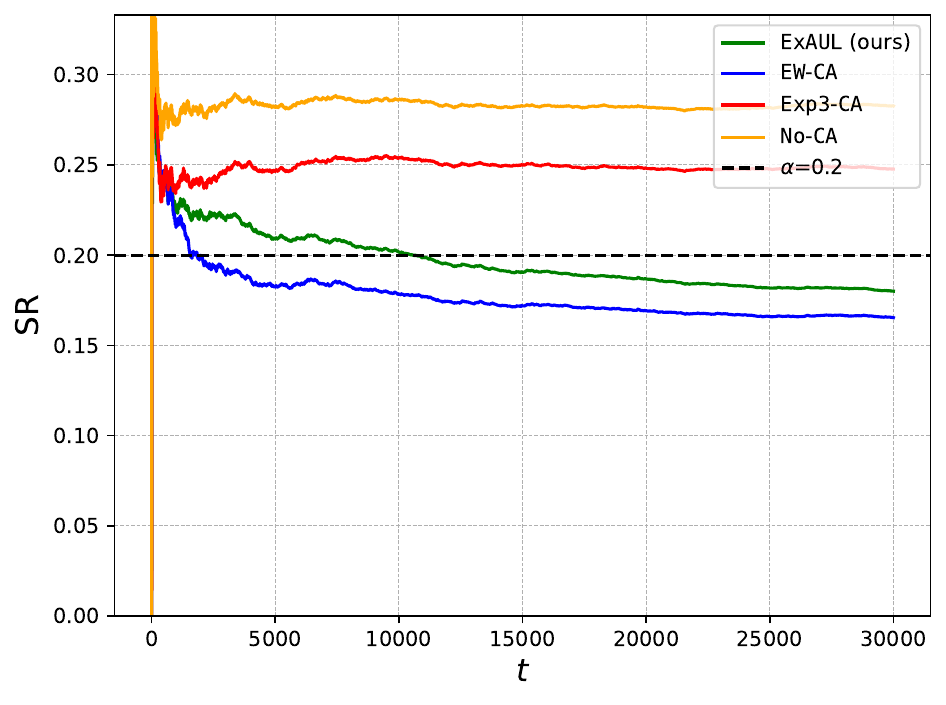}
    }
    \subfigure[Inefficiency over time]{
    \includegraphics[width=0.31\textwidth]{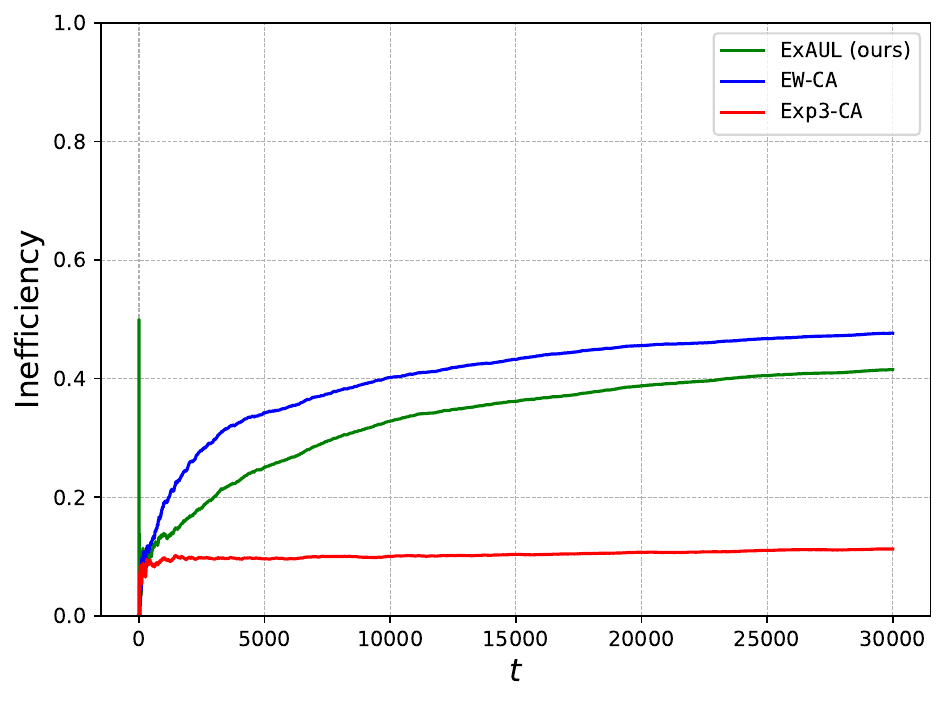}
    }
    \subfigure[SR distribution]{
    \includegraphics[width=0.31\textwidth]{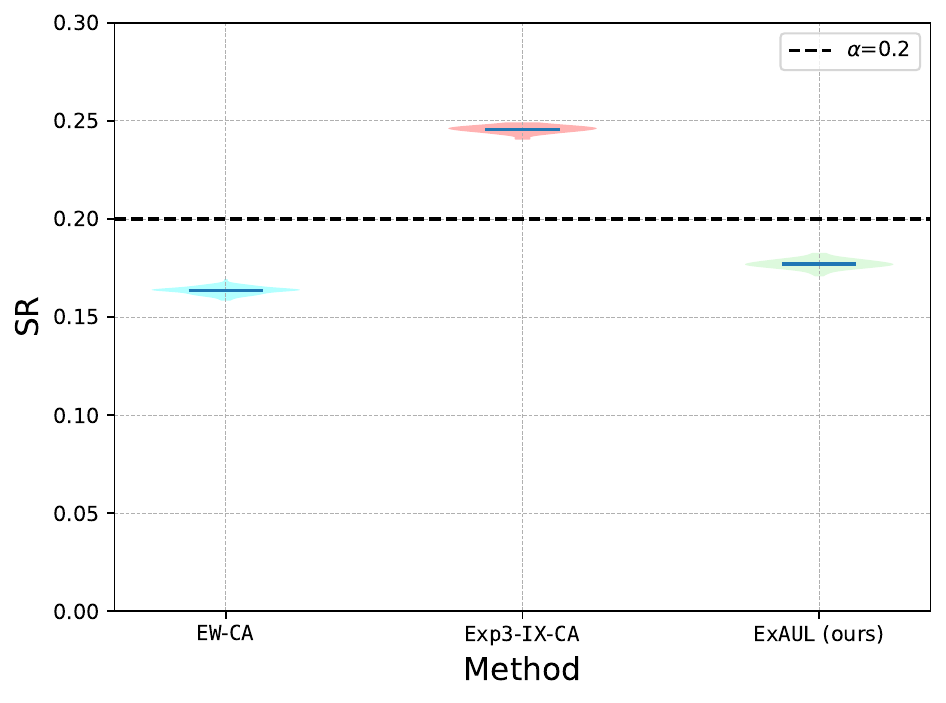}
    }
    \caption{Comparison of conformal abstention methods under a stochastic environment with GPT-5.4 as a generator on NQ ($T=30\mathrm{K}, {\alpha=0.20}$). The violin plots are drawn with randomly chosen $30\mathrm{K}$ samples with $100$ random trials.
    }
    \label{fig:stochastic:main:gpt-nq}
\end{figure*}

\begin{figure*}[h]
    \centering
    \subfigure[SR over time]{
    \includegraphics[width=0.31\textwidth]{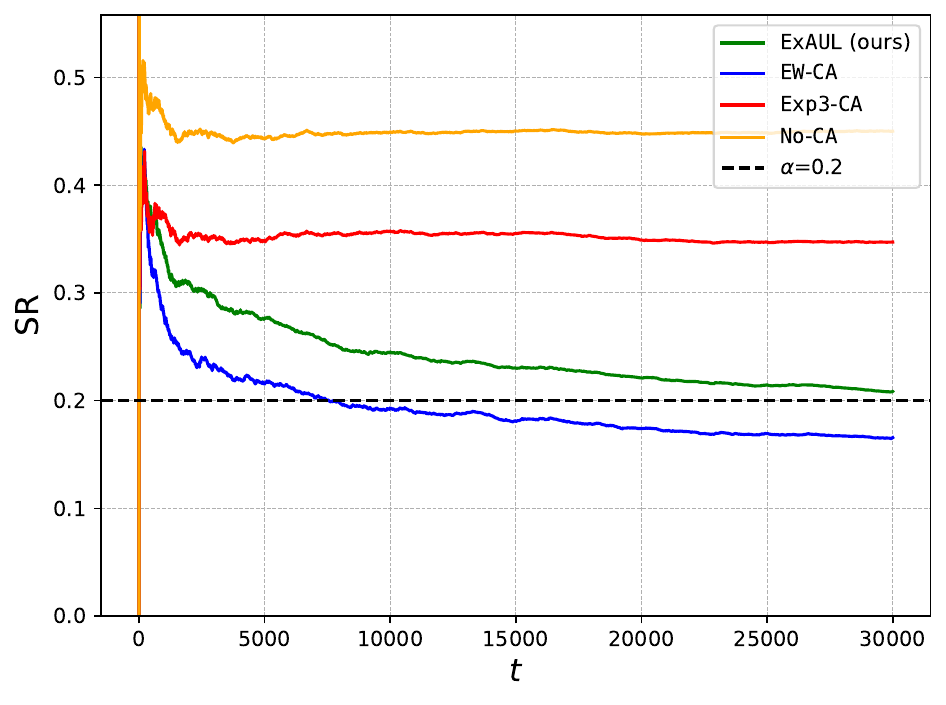}
    }
    \subfigure[Inefficiency over time]{
    \includegraphics[width=0.31\textwidth]{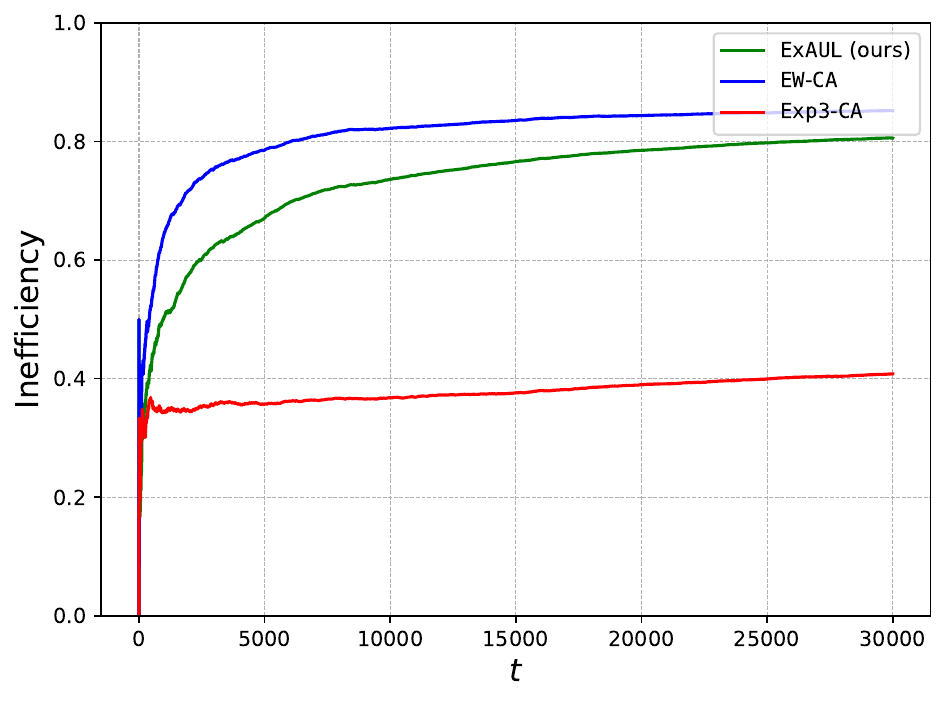}
    }
    \subfigure[SR distribution]{
    \includegraphics[width=0.31\textwidth]{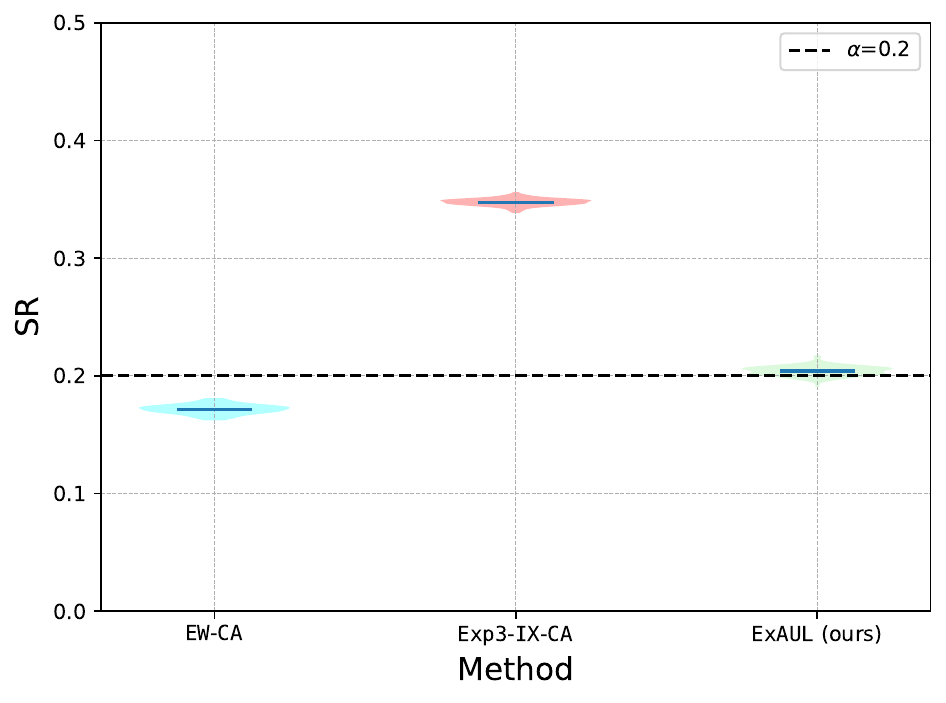}
    }
    \caption{Comparison of conformal abstention methods under a stochastic environment with LLaMA3.1-8B-Instruct as a generator on NQ ($T=30\mathrm{K}, {\alpha=0.20}$). The violin plots are drawn with randomly chosen $30\mathrm{K}$ samples with $100$ random trials. 
    }
    \label{fig:stochastic:main:llama3.1-8b-inst-nq}
\end{figure*}

\begin{figure*}[h]
    \centering
    \subfigure[SR over time]{
    \includegraphics[width=0.31\textwidth]{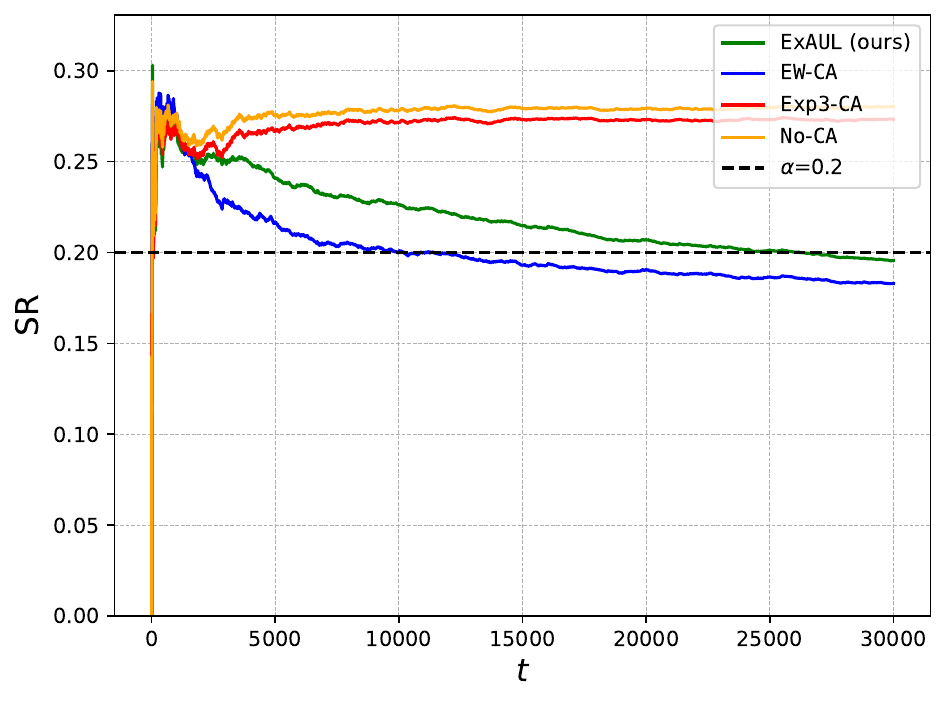}
    }
    \subfigure[Inefficiency over time]{
    \includegraphics[width=0.31\textwidth]{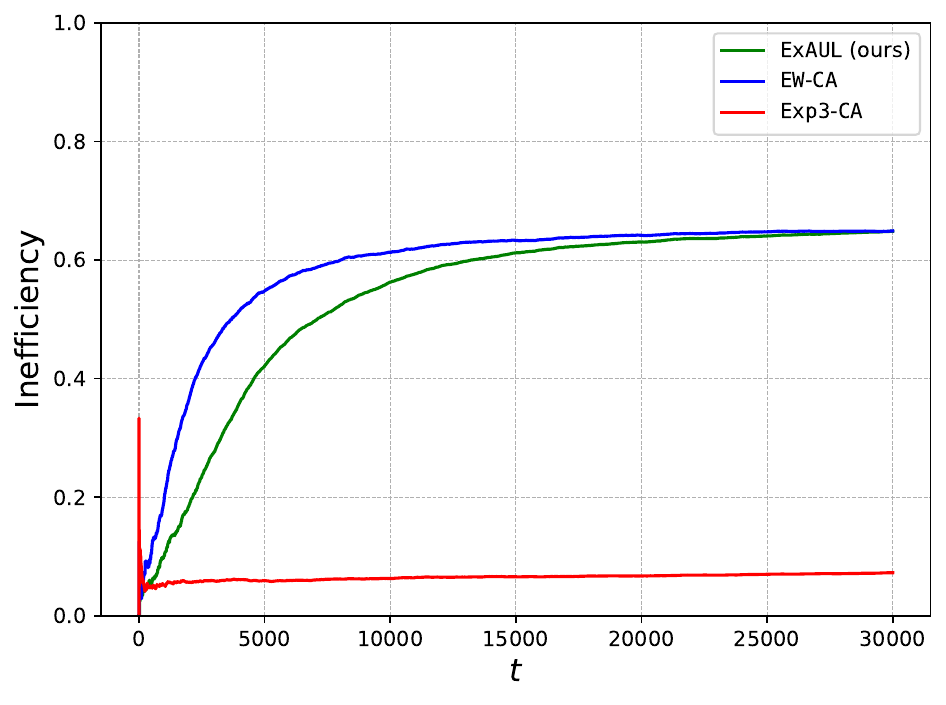}
    }
    \subfigure[SR distribution]{
    \includegraphics[width=0.31\textwidth]{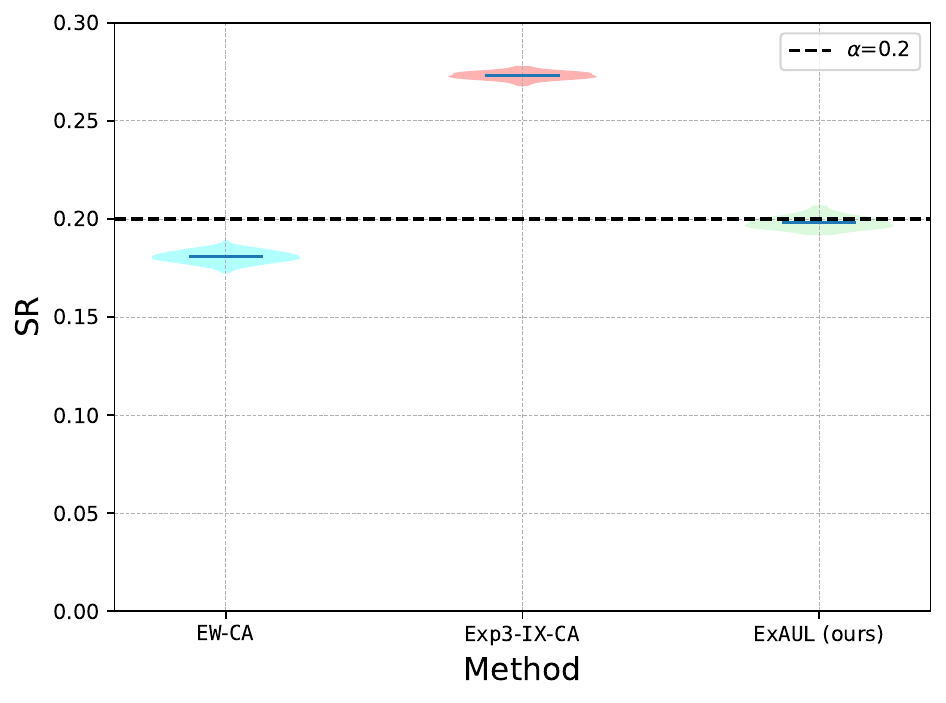}
    }
    \caption{Comparison of conformal abstention methods under a stochastic environment with GPT-5.4 as a generator on NQ ($T=30\mathrm{K}, {\alpha=0.20}$) and $f_{\texttt{std}}$ as our scoring function. The violin plots are drawn with randomly chosen $30\mathrm{K}$ samples with $100$ random trials.
    }
    \label{fig:stochastic:main:gpt-3.5-turbo-nq:f_std}
\end{figure*}

\begin{figure*}[h]
    \centering
    \subfigure[SR over time]{
    \includegraphics[width=0.31\textwidth]{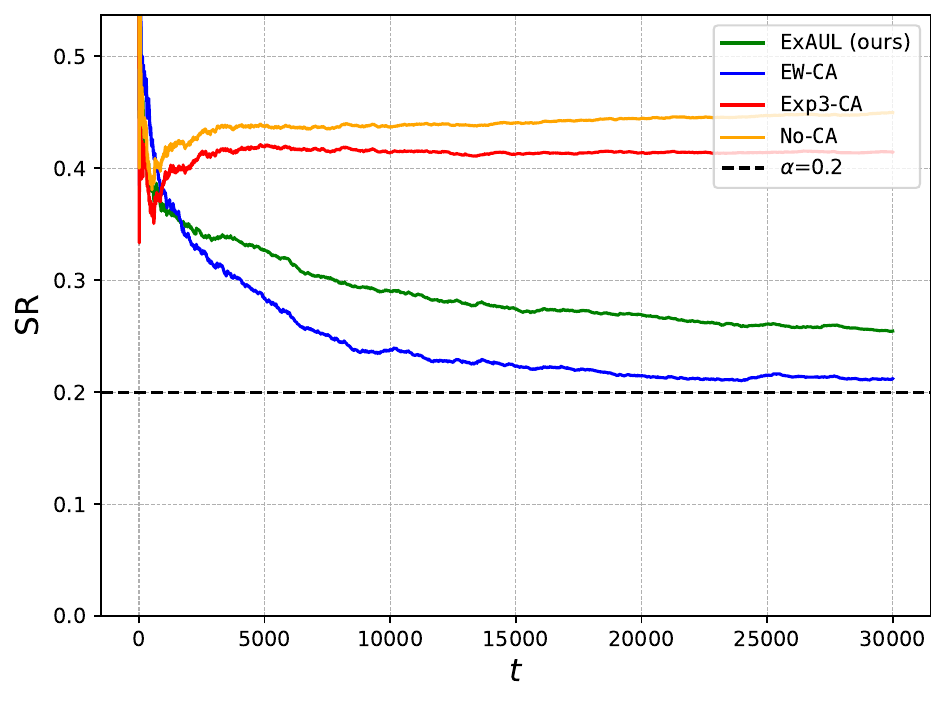}
    }
    \subfigure[Inefficiency over time]{
    \includegraphics[width=0.31\textwidth]{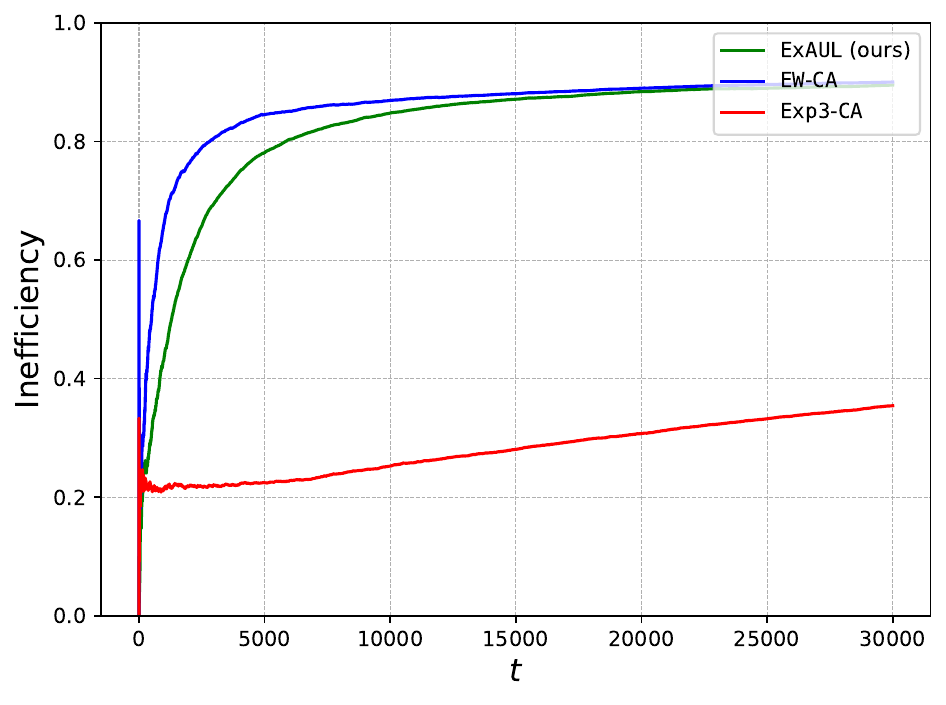}
    }
    \subfigure[SR distribution]{
    \includegraphics[width=0.31\textwidth]{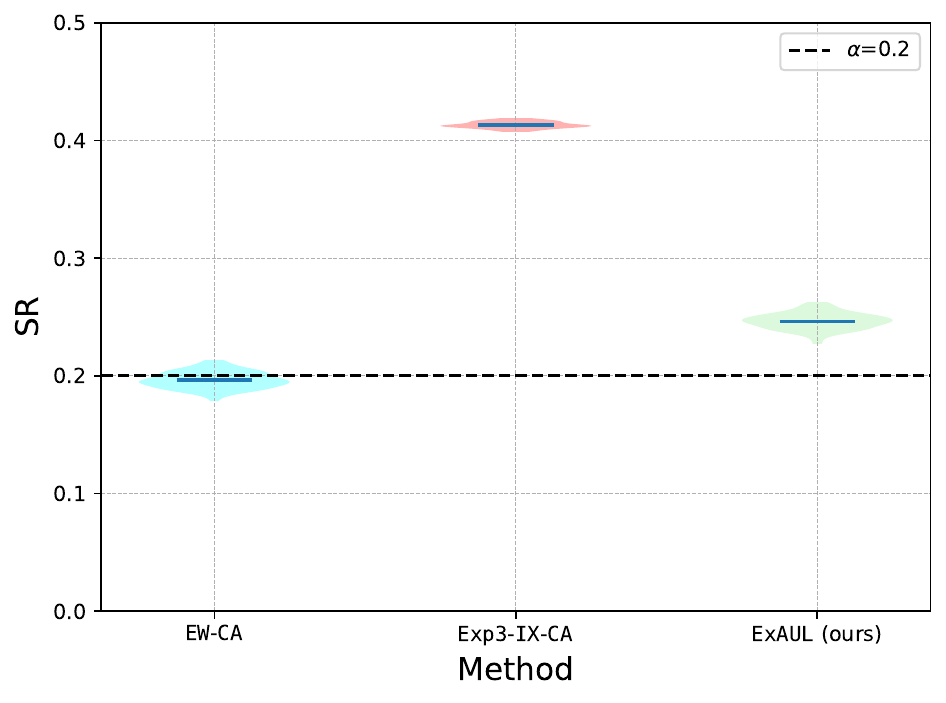}
    }
    \caption{Comparison of conformal abstention methods under a stochastic environment with LLaMA3.1-8B-Instruct as a generator on NQ ($T=30\mathrm{K}, {\alpha=0.20}$) and $f_{\texttt{std}}$ as our scoring function. The violin plots are drawn with randomly chosen $30\mathrm{K}$ samples with $100$ random trials.
    }
    \label{fig:stochastic:main:llama3.1-8b-inst-nq:f_std}
\end{figure*}

\clearpage

\subsection{Distribution-shift Environment}
\label{exp:adddistribuitionshift}

\subsubsection{Single Shift}
\begin{figure*}[h]
    \centering
    \subfigure[SR over time]{
    \includegraphics[width=0.31\textwidth]{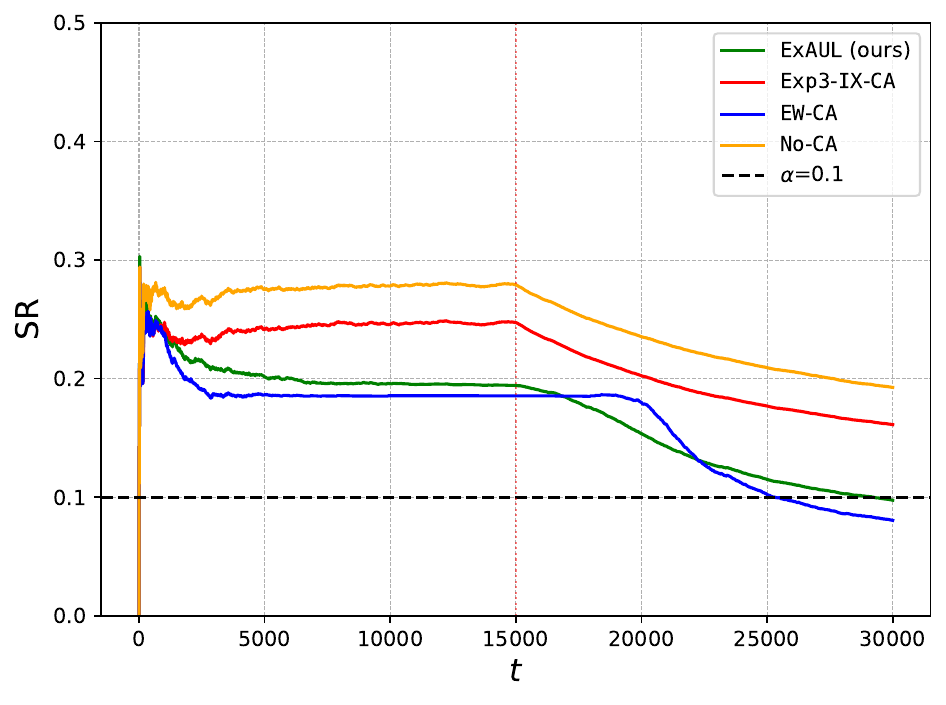}
    }
    \subfigure[Inefficiency over time]{
    \includegraphics[width=0.31\textwidth]{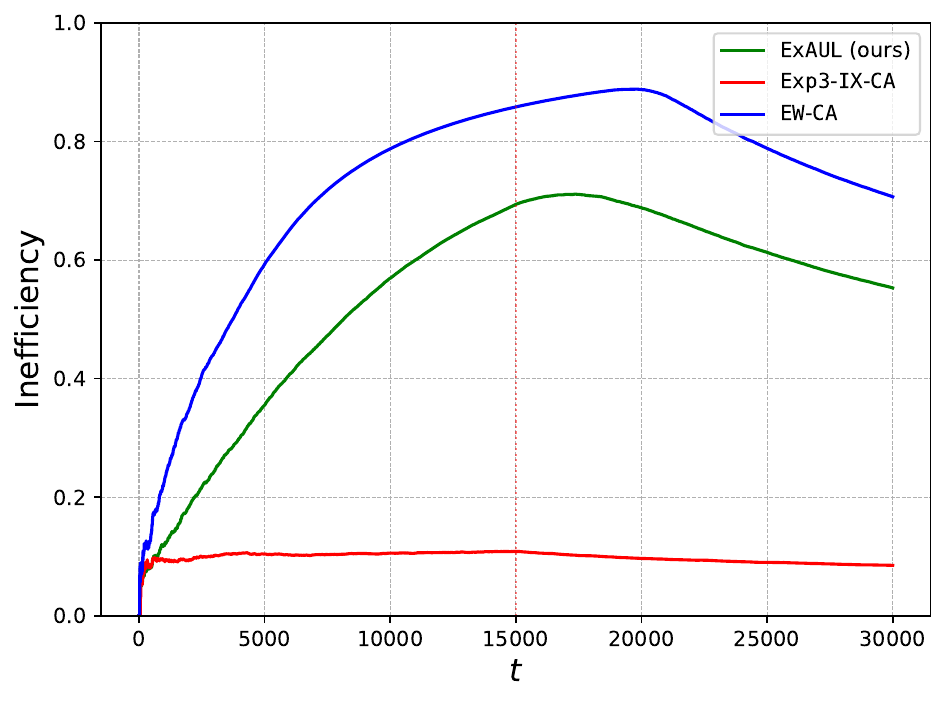}
    }
    \subfigure[SR distribution]{
    \includegraphics[width=0.31\textwidth]{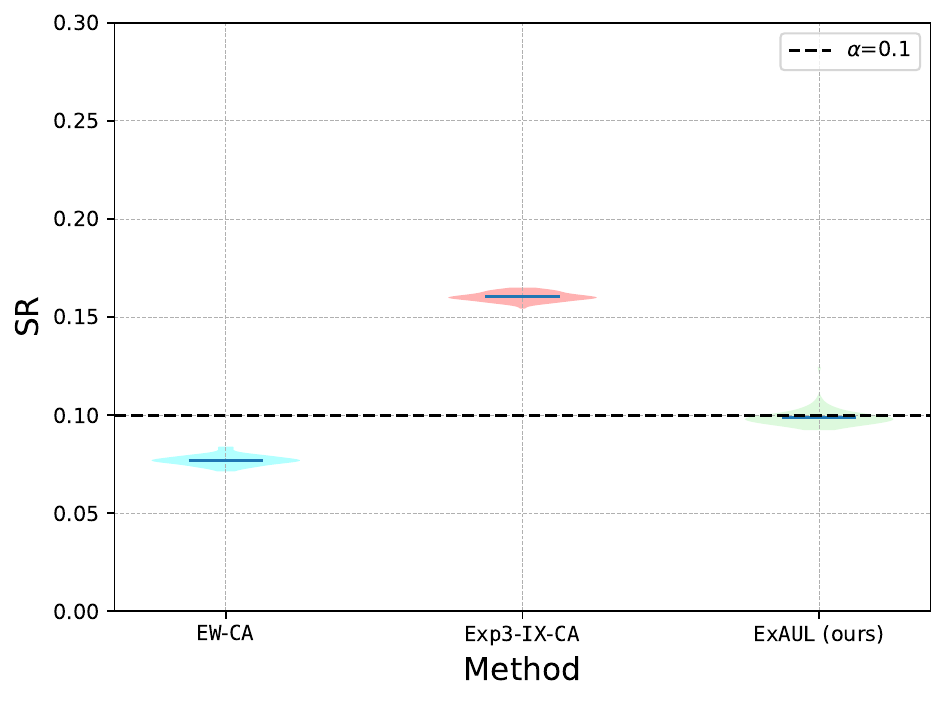}
    }
    \caption{Comparison of conformal abstention methods under a distribution-shift environment with GPT-5.4 as a generator ($T = 30\mathrm{K}, \alpha = 0.10$). 
    We consider a single distribution shift from NQ to TriviaQA. 
    The violin plots are drawn with randomly chosen $30\mathrm{K}$ samples with $100$ random trials.
    }
    \label{fig:shift:gpt-nq-simple}
\end{figure*}

\begin{figure*}[h]
    \centering
    \subfigure[SR over time]{
    \includegraphics[width=0.31\textwidth]{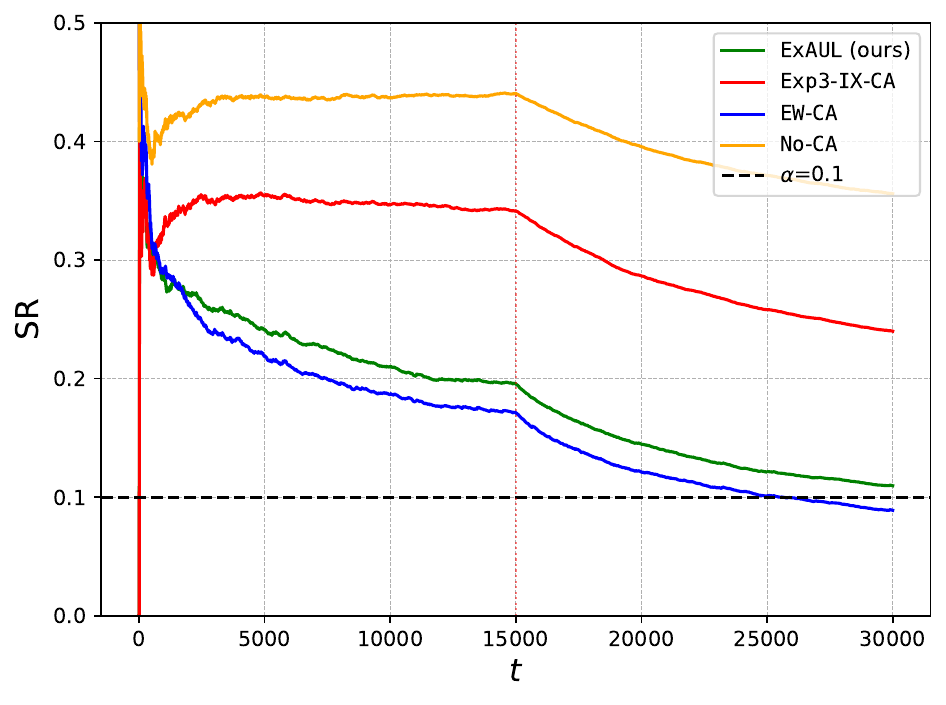}
    }
    \subfigure[Inefficiency over time]{
    \includegraphics[width=0.31\textwidth]{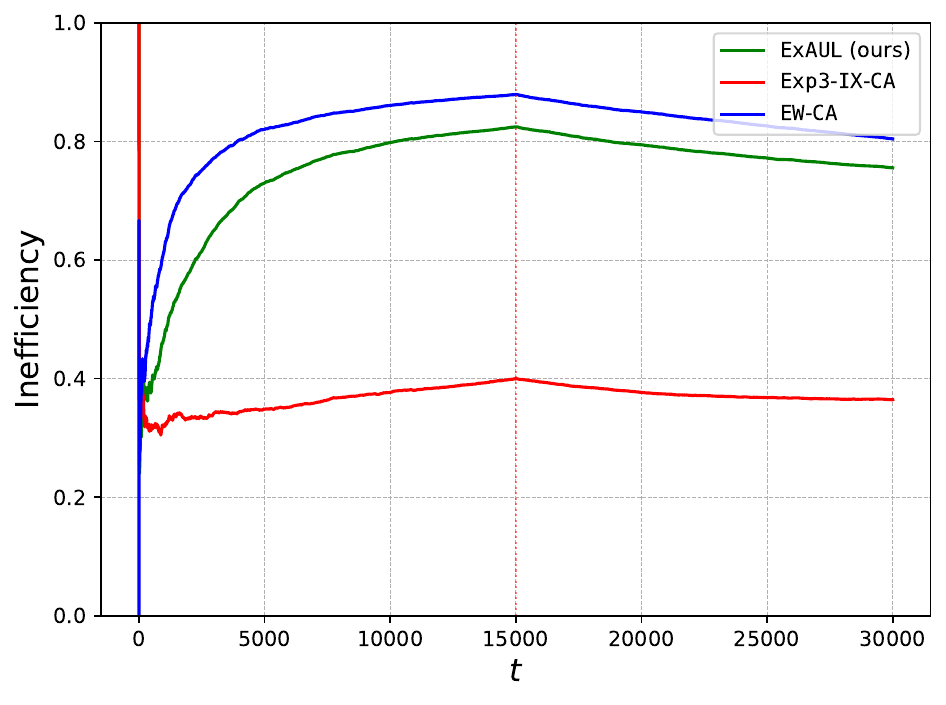}
    }
    \subfigure[SR distribution]{
    \includegraphics[width=0.31\textwidth]{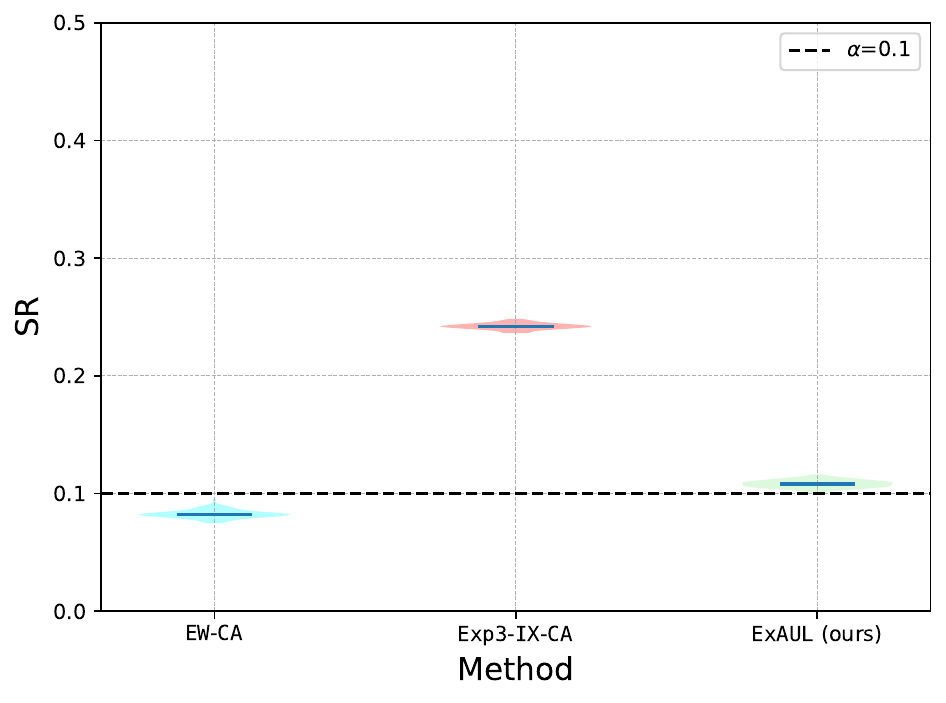}
    }
    \caption{Comparison of conformal abstention methods under a  distribution-shift environment with LLaMA3.1-8B-Instruct as a generator ($T = 30\mathrm{K}, \alpha = 0.10$), in a single distribution shift from NQ to TriviaQA.
    The violin plots are drawn with randomly chosen $30\mathrm{K}$ samples with $100$ random trials.
    }
    \label{fig:shift:llama-nq-simple}
\end{figure*}

\begin{figure*}[h]
    \centering
    \subfigure[SR over time]{
    \includegraphics[width=0.31\textwidth]{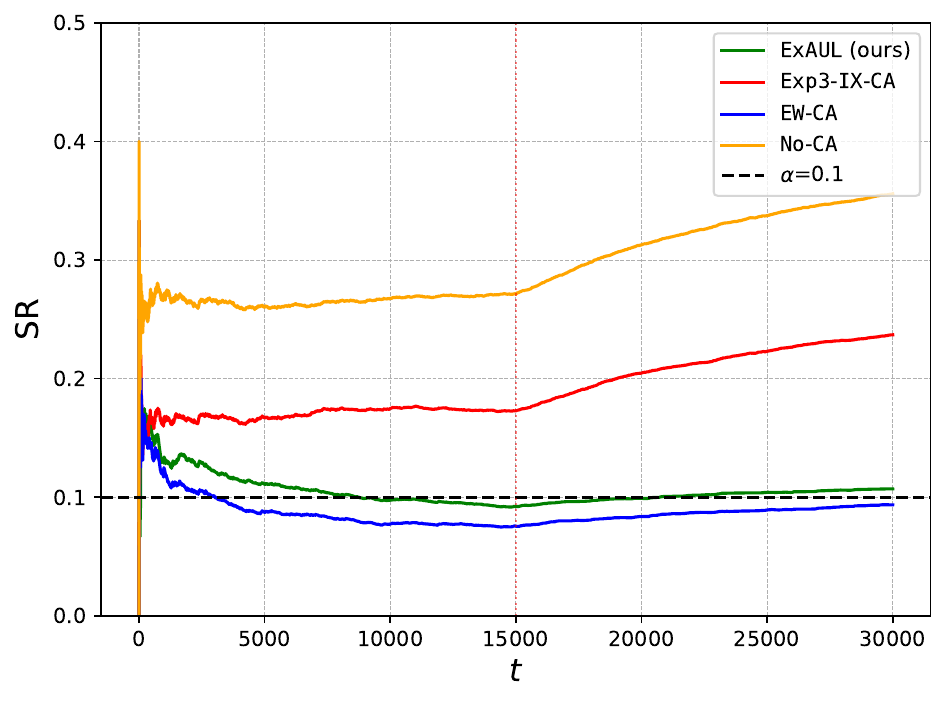}
    }
    \subfigure[Inefficiency over time]{
    \includegraphics[width=0.31\textwidth]{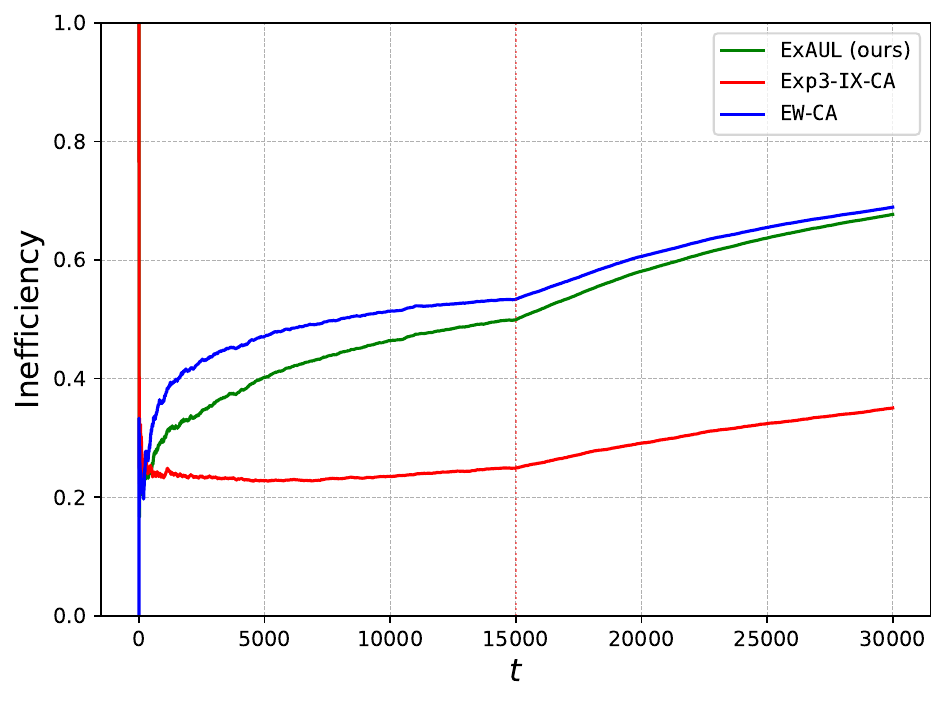}
    }
    \subfigure[SR distribution]{
    \includegraphics[width=0.31\textwidth]{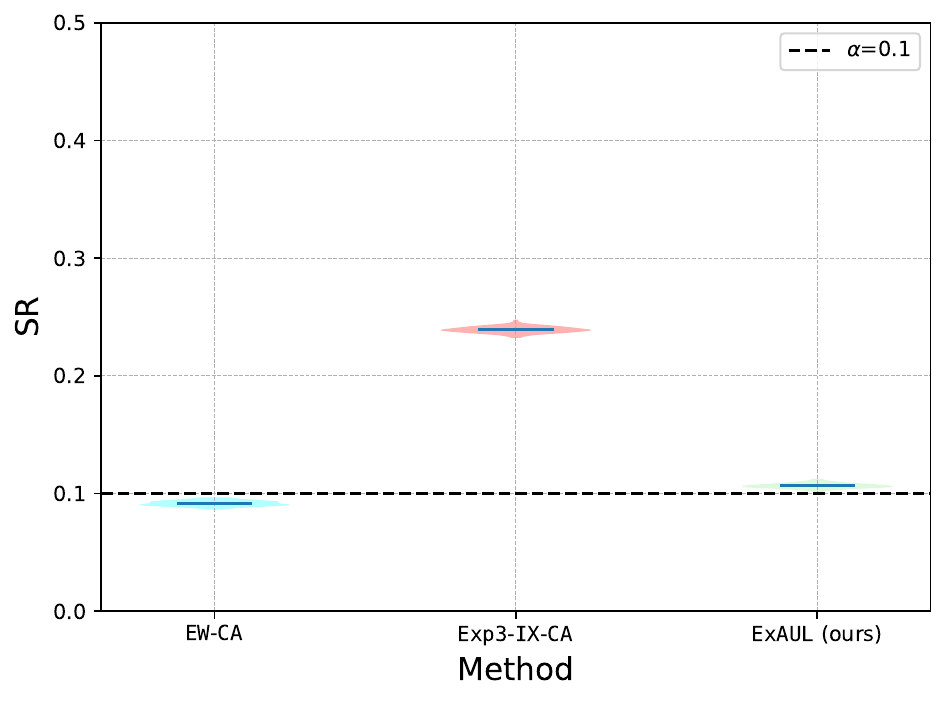}
    }
    \caption{Comparison of conformal abstention methods under a distribution-shift environment with LLaMA3.1-8B-Instruct as a generator ($T = 30\mathrm{K}, \alpha = 0.10$), in a single distribution shift from TriviaQA to NQ.
    The violin plots are drawn with randomly chosen $30\mathrm{K}$ samples with $100$ random trials.
    }
    \label{fig:shift:llama-tri-simple}
\end{figure*}

\clearpage

\subsubsection{Alternating Shift}
\begin{figure*}[h]
    \centering
    \subfigure[SR over time]{
    \includegraphics[width=0.31\textwidth]{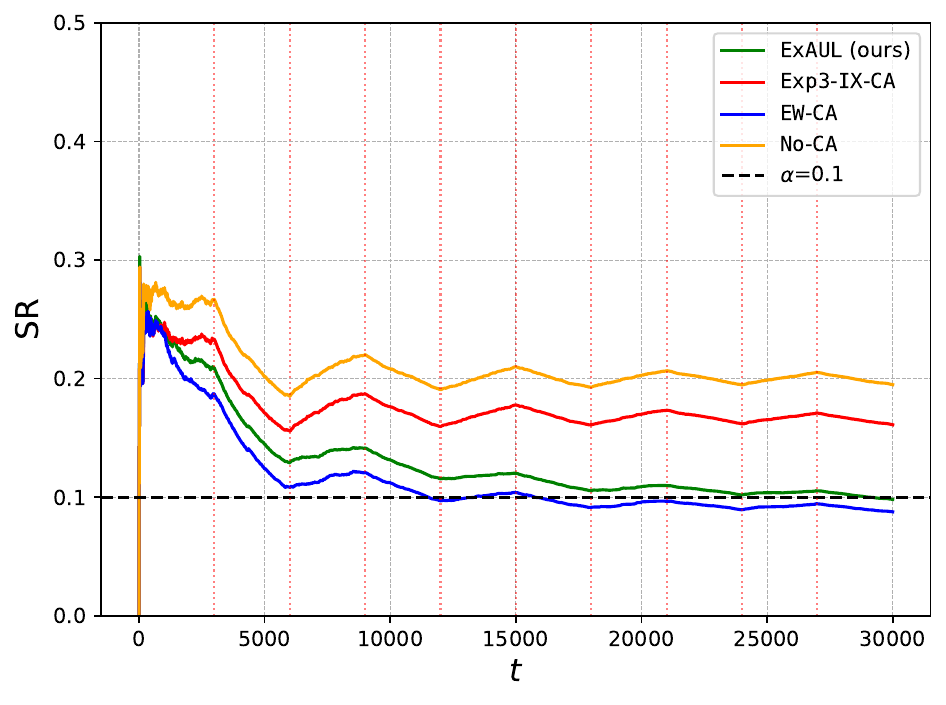}
    }
    \subfigure[Inefficiency over time]{
    \includegraphics[width=0.31\textwidth]{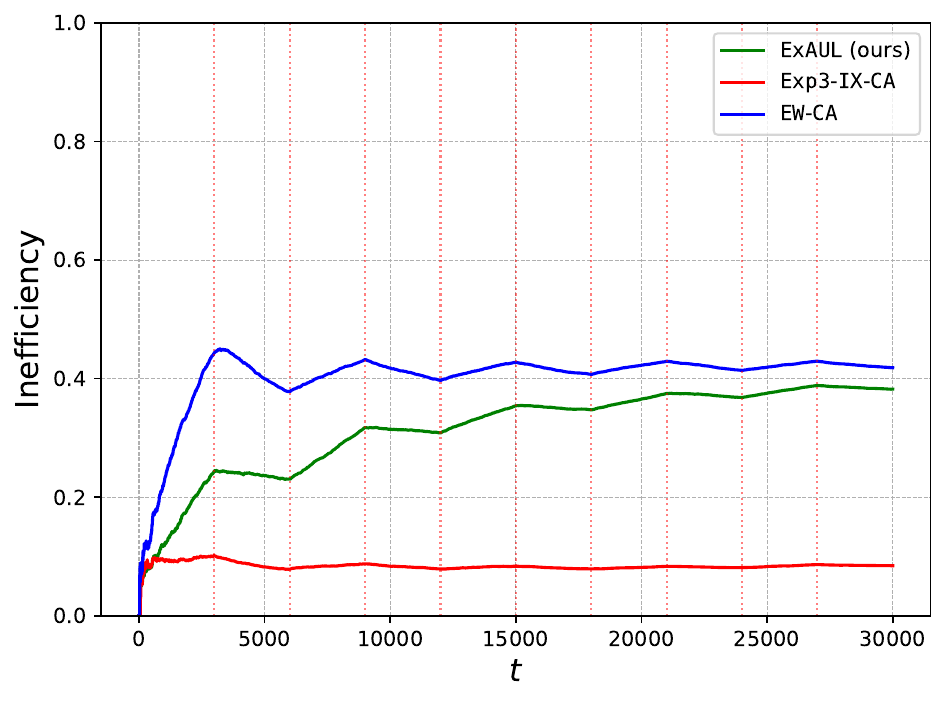}
    }
    \subfigure[SR distribution]{
    \includegraphics[width=0.31\textwidth]{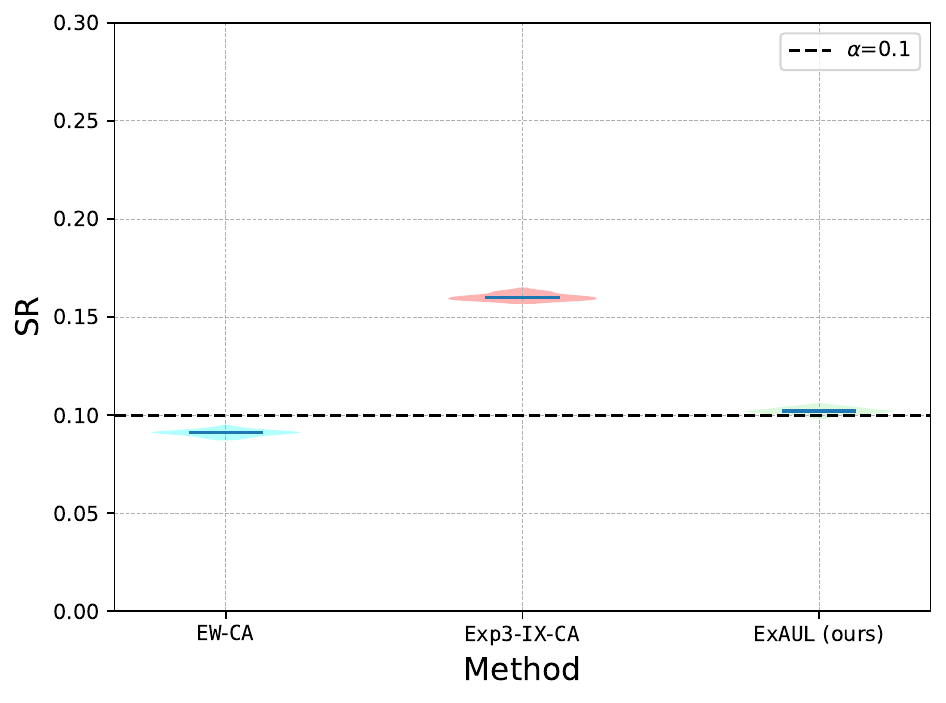}
    }
    \caption{Comparison of conformal abstention methods under a distribution-shift environment with
    GPT-5.4 as a generator ($T=30\mathrm{K}, \alpha=0.10$), in a distribution shift by alternating between NQ and TriviaQA over time, beginning with NQ.
    The violin plots are drawn with randomly chosen $30\mathrm{K}$ samples with $100$ random trials.
    }
    \label{fig:shift:gpt-nq-align}
\end{figure*}

\begin{figure*}[h]
    \centering
    \subfigure[SR over time]{
    \includegraphics[width=0.31\textwidth]{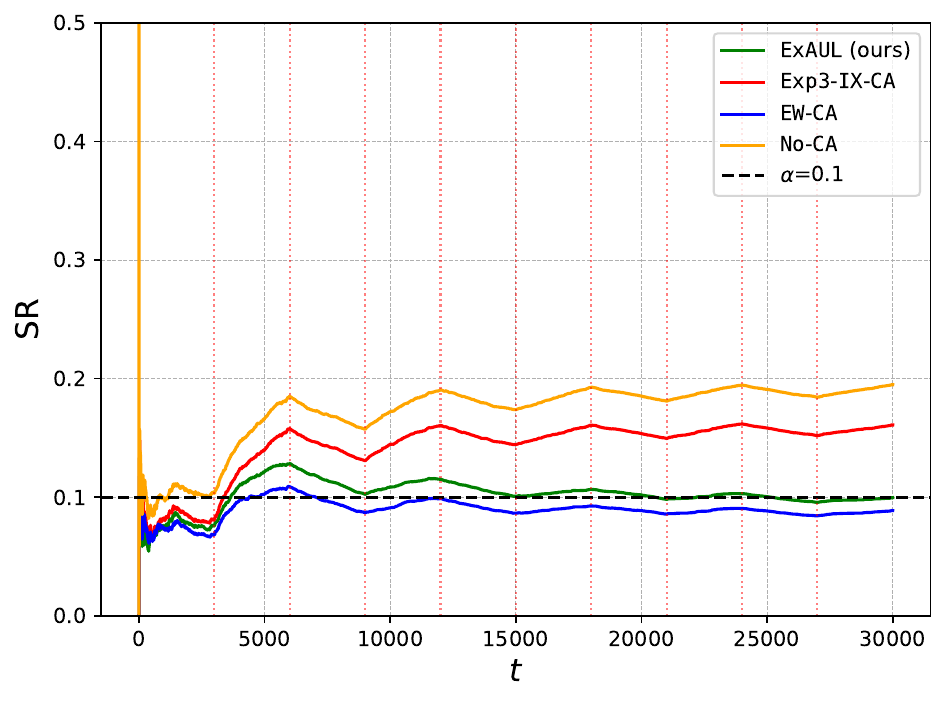}
    }
    \subfigure[Inefficiency over time]{
    \includegraphics[width=0.31\textwidth]{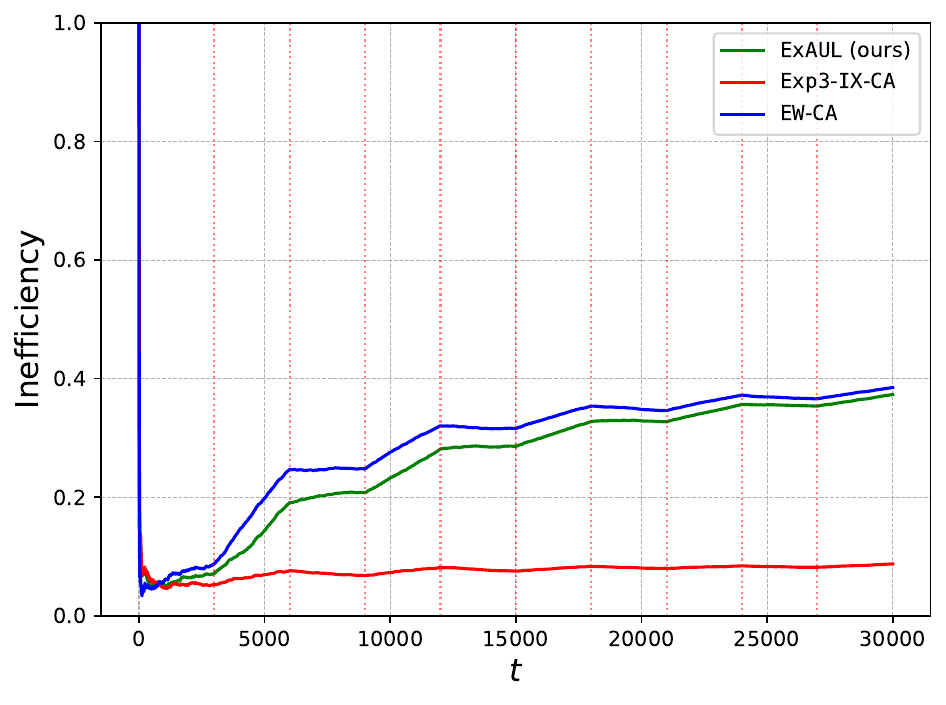}
    }
    \subfigure[SR distribution]{
    \includegraphics[width=0.31\textwidth]{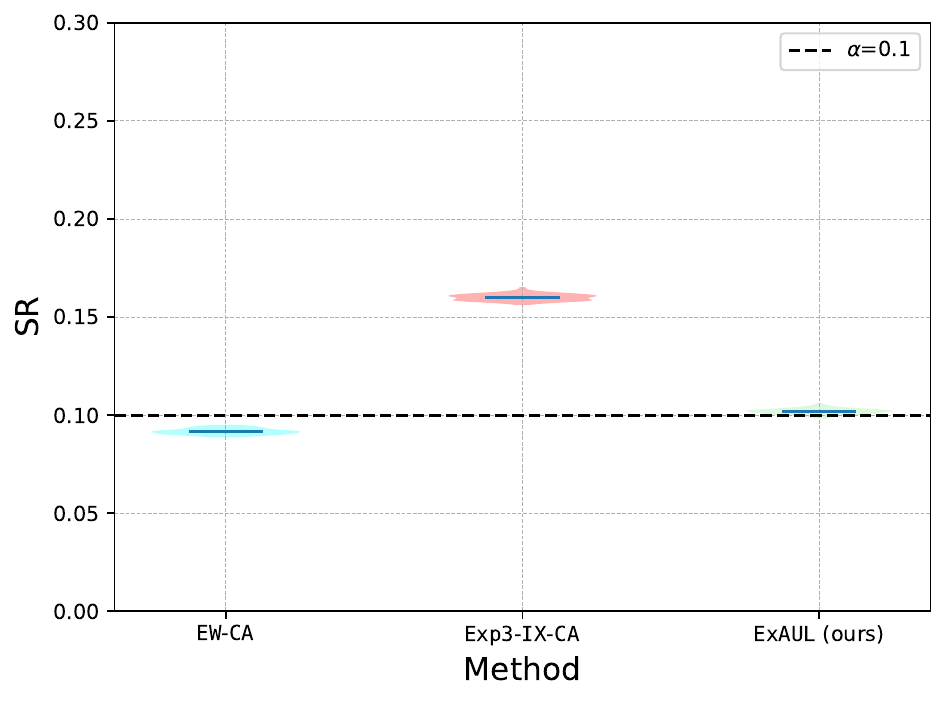}
    }
    \caption{Comparison of conformal abstention methods under an alternating distribution-shift environment with 
    GPT-5.4 as a generator ($T=30\mathrm{K}, \alpha=0.10$), alternating between TriviaQA and NQ, starting with TriviaQA.
    The violin plots are drawn with randomly chosen $30\mathrm{K}$ samples with $100$ random trials.
    }
    \label{fig:shift:gpt-tri-align}
\end{figure*}

\begin{figure*}[h]
    \centering
    \subfigure[SR over time]{
    \includegraphics[width=0.31\textwidth]{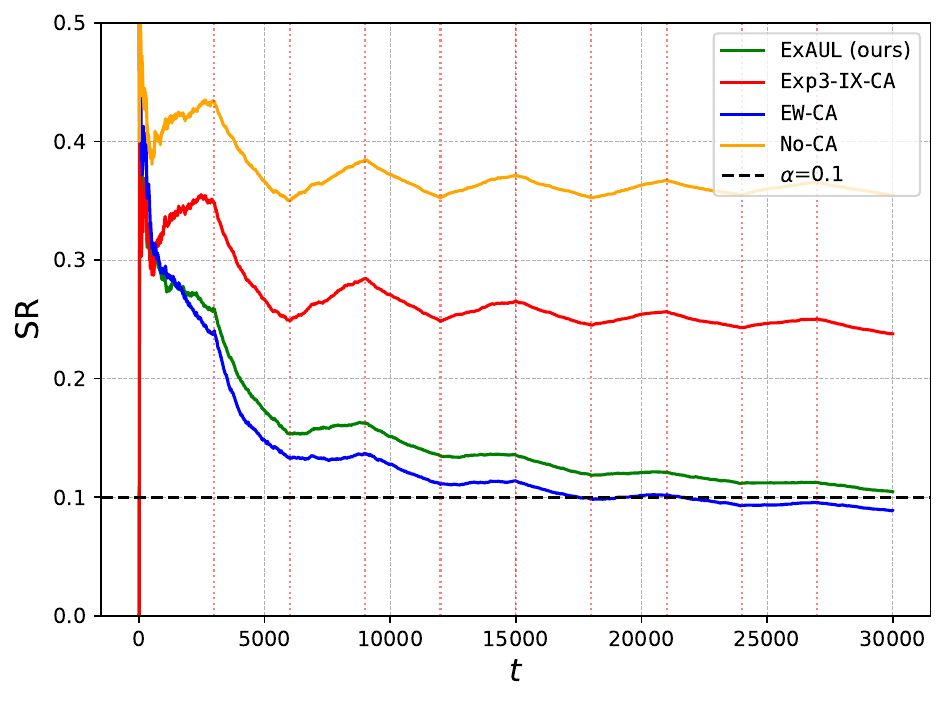}
    }
    \subfigure[Inefficiency over time]{
    \includegraphics[width=0.31\textwidth]{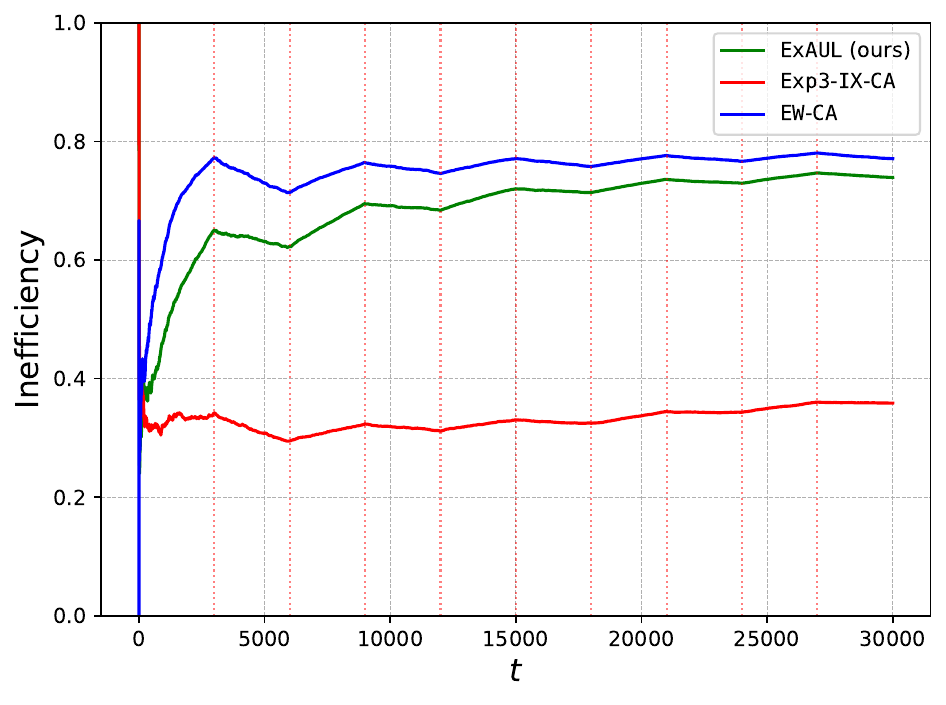}
    }
    \subfigure[SR distribution]{
    \includegraphics[width=0.31\textwidth]{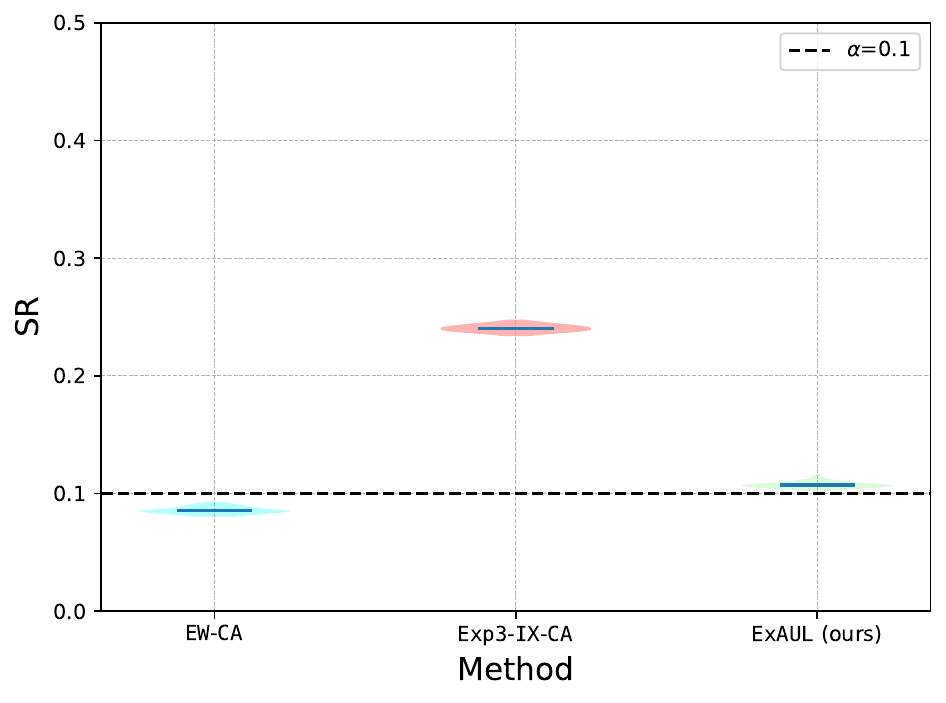}
    }
    \caption{Comparison of conformal abstention methods under a distribution-shift environment with LLaMA3.1-8B-Instruct as a generator ($T=30\mathrm{K}, \alpha=0.10$), in a distribution shift by alternating between NQ and TriviaQA over time, beginning with NQ.
    The violin plots are drawn with randomly chosen $30\mathrm{K}$ samples with $100$ random trials.
    }
    \label{fig:shift:llama-nq-align}
\end{figure*}

\begin{figure*}[h]
    \centering
    \subfigure[SR over time]{
    \includegraphics[width=0.31\textwidth]{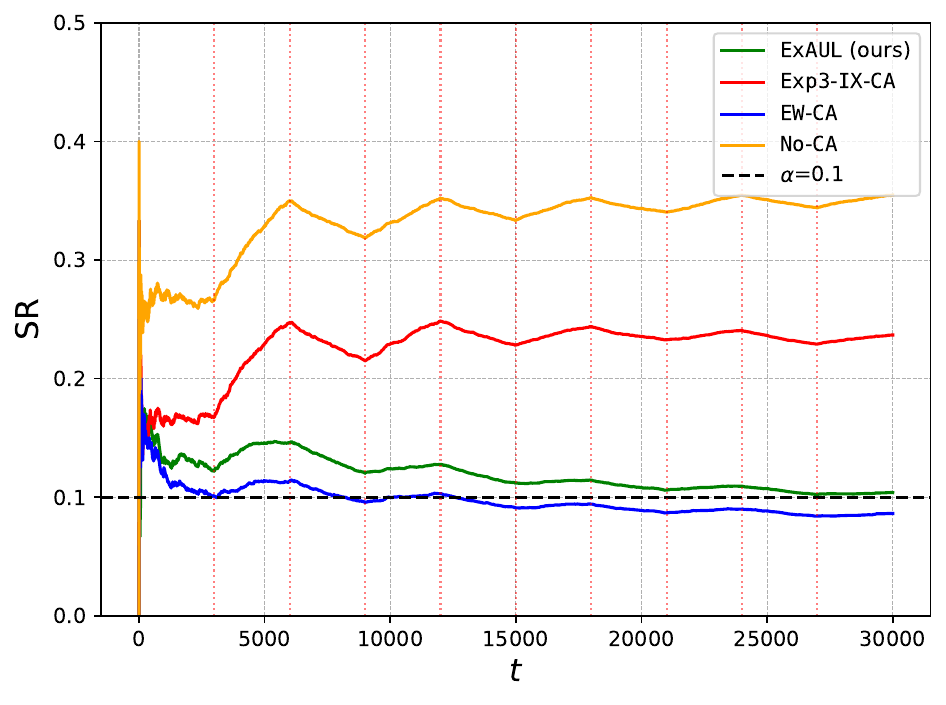}
    }
    \subfigure[Inefficiency over time]{
    \includegraphics[width=0.31\textwidth]{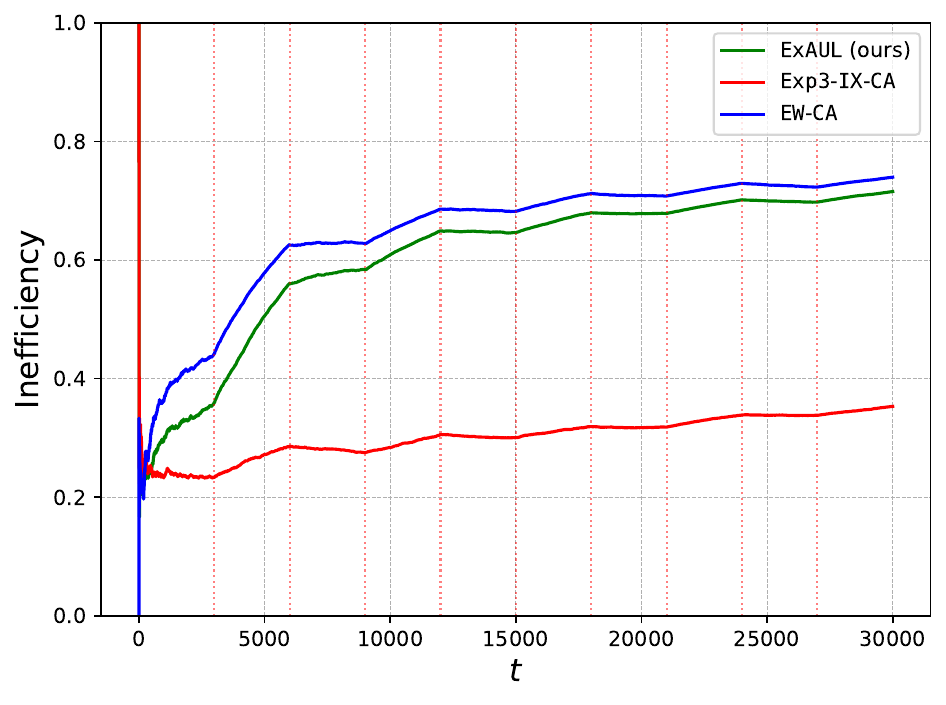}
    }
    \subfigure[SR distribution]{
    \includegraphics[width=0.31\textwidth]{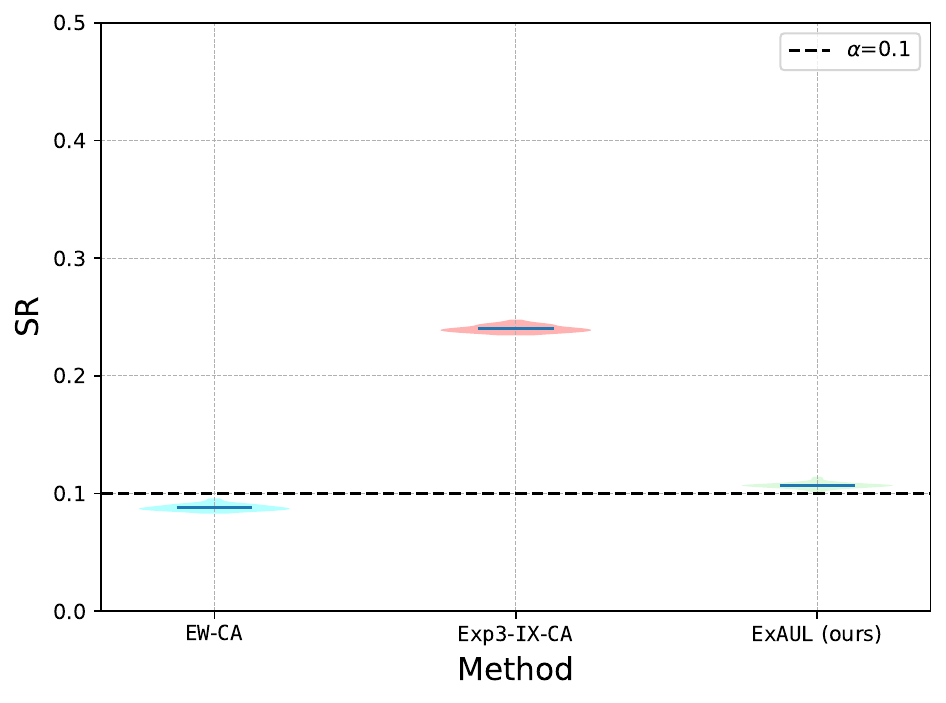}
    }
    \caption{Comparison of conformal abstention methods under a distribution-shift environment with
    LLaMA3.1-8B-Instruct as a generator ($T=30\mathrm{K}, \alpha=0.10$), in a distribution shift by alternating between TriviaQA and NQ over time, beginning with TriviaQA.
    The violin plots are drawn with randomly chosen $30\mathrm{K}$ samples with $100$ random trials.
    }
    \label{fig:shift:llama-tri-align}
\end{figure*}


\subsubsection{Gradual Shift}
\begin{figure*}[h]
    \centering
    \subfigure[SR over time]{
    \includegraphics[width=0.31\textwidth]{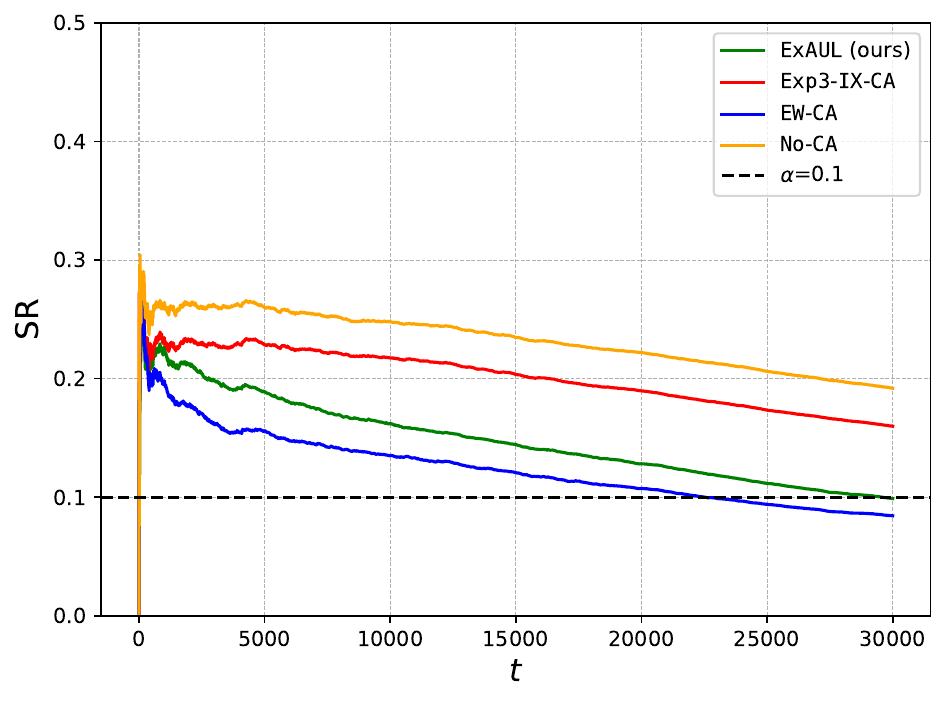}
    }
    \subfigure[Inefficiency over time]{
    \includegraphics[width=0.31\textwidth]{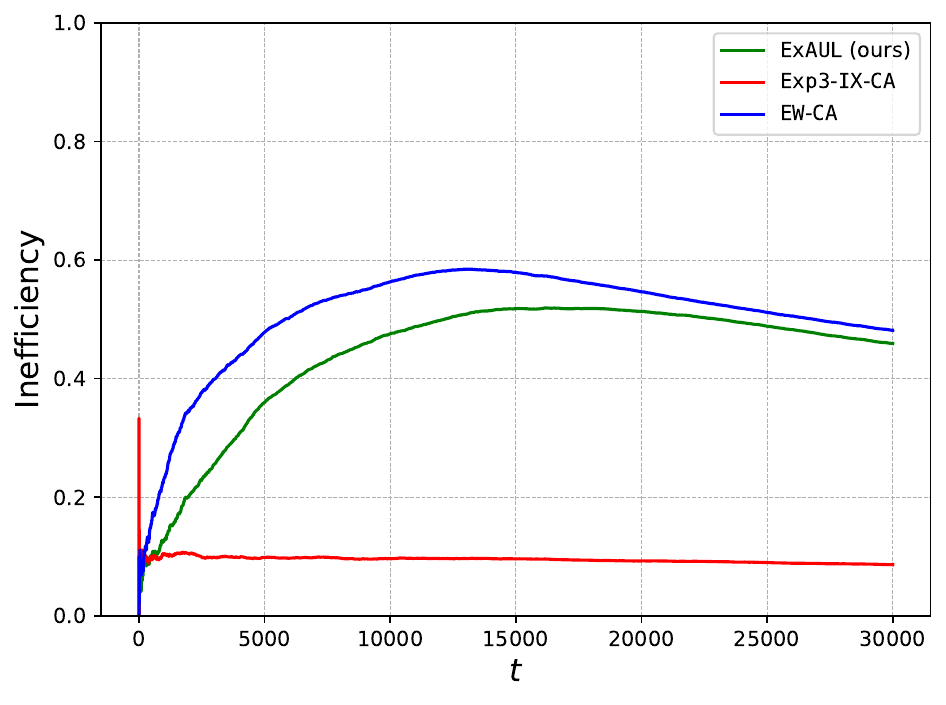}
    }
    \subfigure[SR distribution]{
    \includegraphics[width=0.31\textwidth]{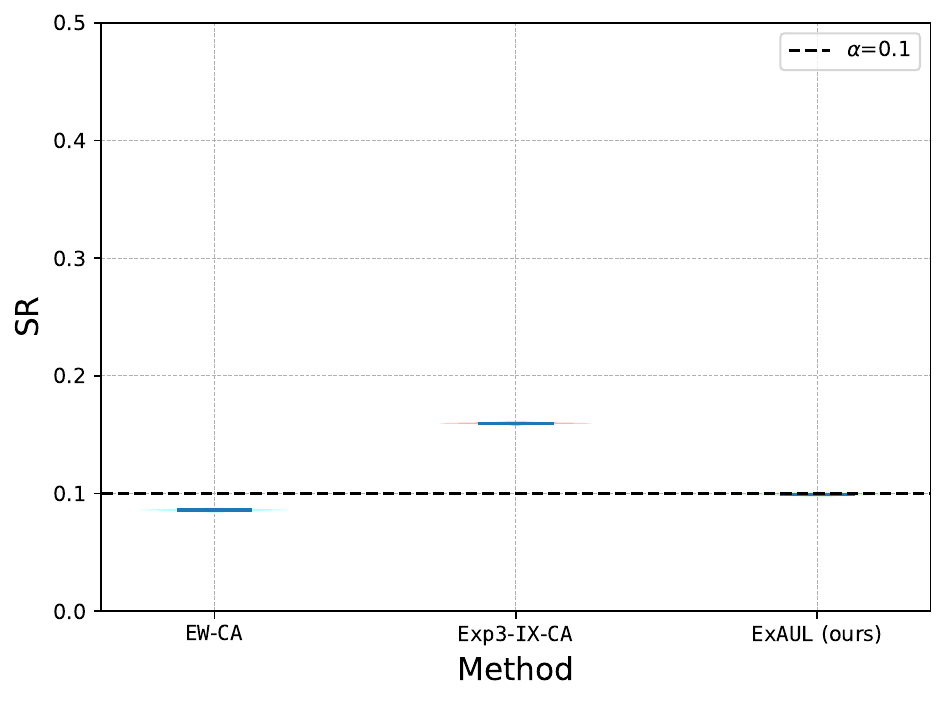}
    }
    \caption{Comparison of conformal abstention methods under a distribution-shift environment with GPT-5.4 as a generator ($T = 30\mathrm{K}, \alpha = 0.10$), in a gradual distribution shift from NQ to TriviaQA over time.
    The violin plots are drawn with randomly chosen $30\mathrm{K}$ samples with $100$ random trials.
    }
    \label{fig:shift:gpt-nq-linear}
\end{figure*}

\begin{figure*}[h]
    \centering
    \subfigure[SR over time]{
    \includegraphics[width=0.31\textwidth]{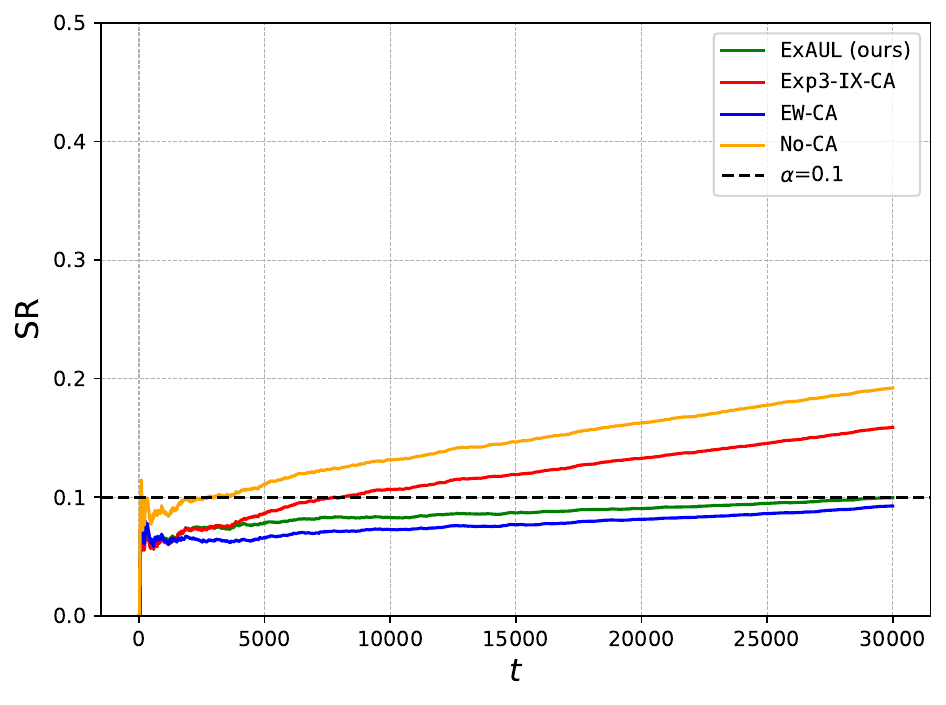}
    }
    \subfigure[Inefficiency over time]{
    \includegraphics[width=0.31\textwidth]{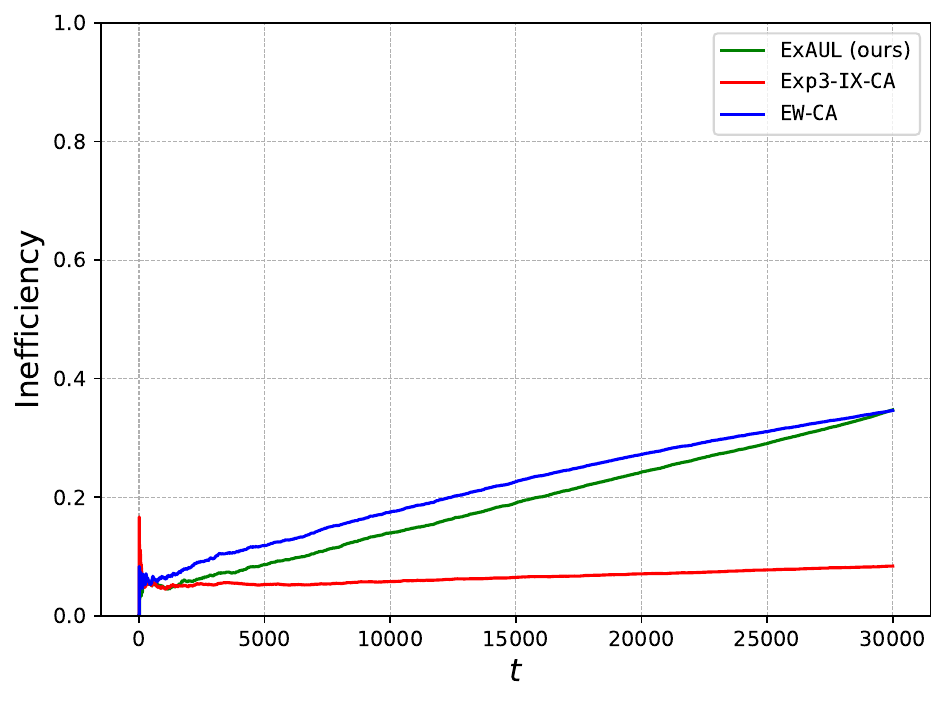}
    }
    \subfigure[SR distribution]{
    \includegraphics[width=0.31\textwidth]{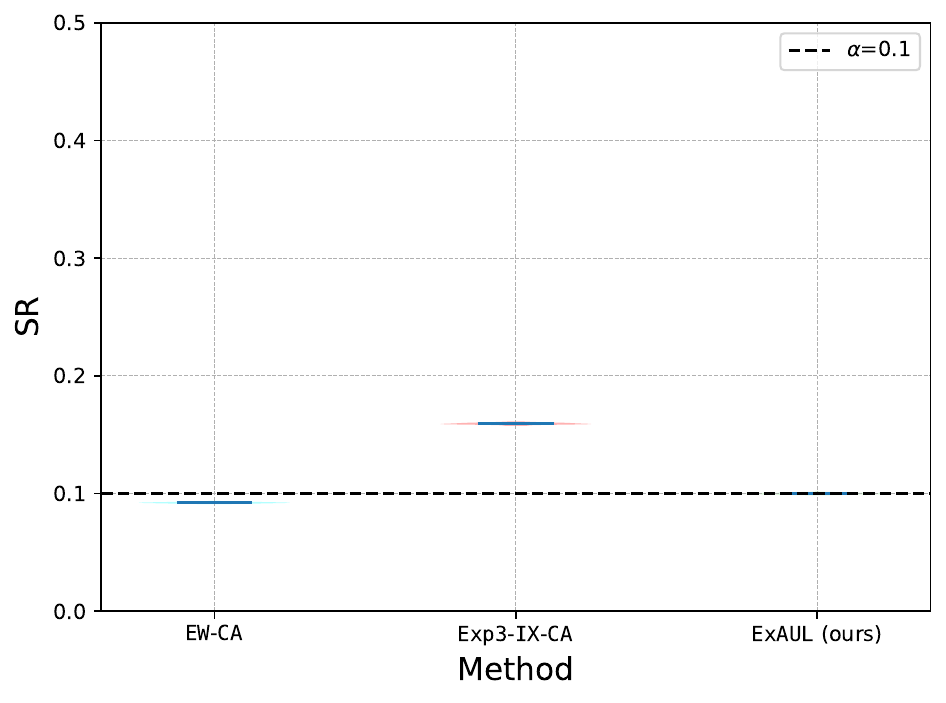}
    }
    \vspace{-1.5ex}
    \caption{Comparison of conformal abstention methods under a gradual distribution-shift environment with GPT-5.4 as a generator ($T = 30\mathrm{K}, \alpha = 0.10$), from TriviaQA to NQ over time.
    The violin plots are drawn with randomly chosen $30\mathrm{K}$ samples with $100$ random trials.
    }
    \label{fig:shift:gpt-tri-linear}
\end{figure*}

\begin{figure*}[h]
    \centering
    \subfigure[SR over time]{
    \includegraphics[width=0.31\textwidth]{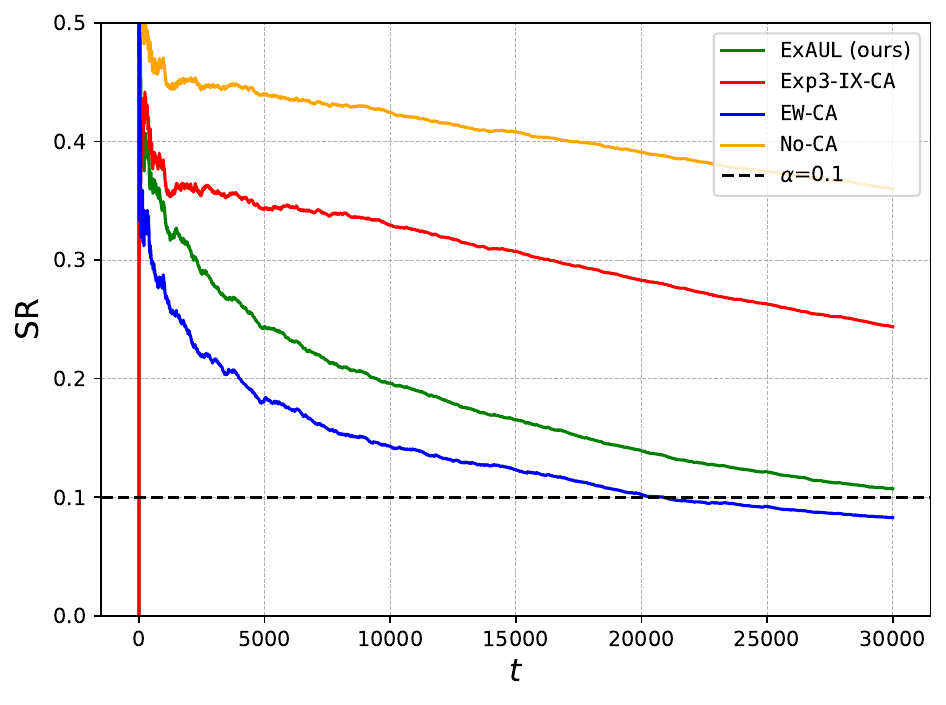}
    }
    \subfigure[Inefficiency over time]{
    \includegraphics[width=0.31\textwidth]{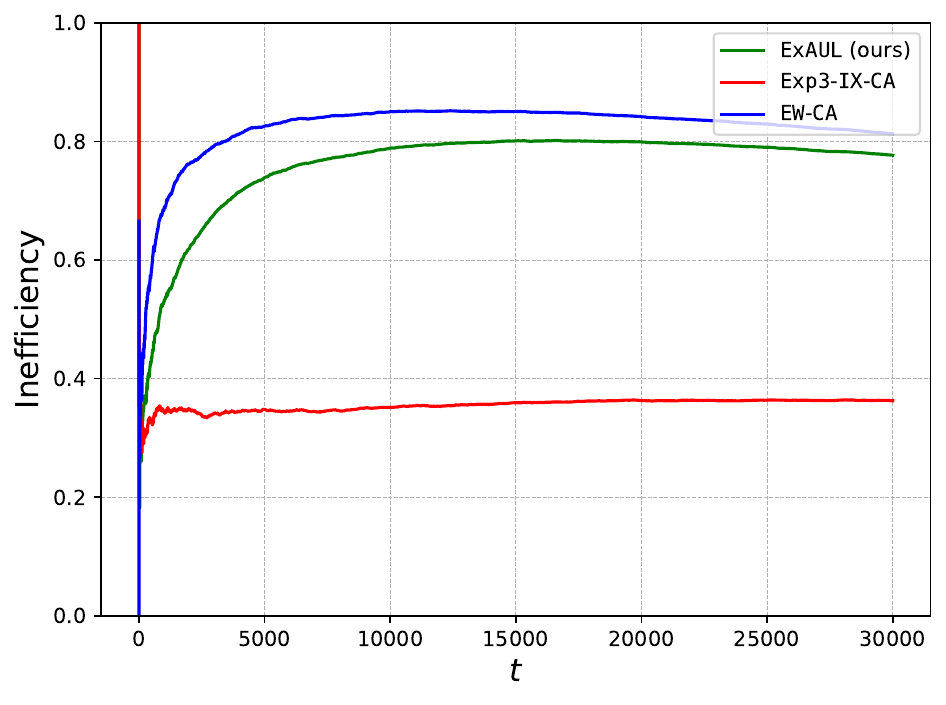}
    }
    \subfigure[SR distribution]{
    \includegraphics[width=0.31\textwidth]{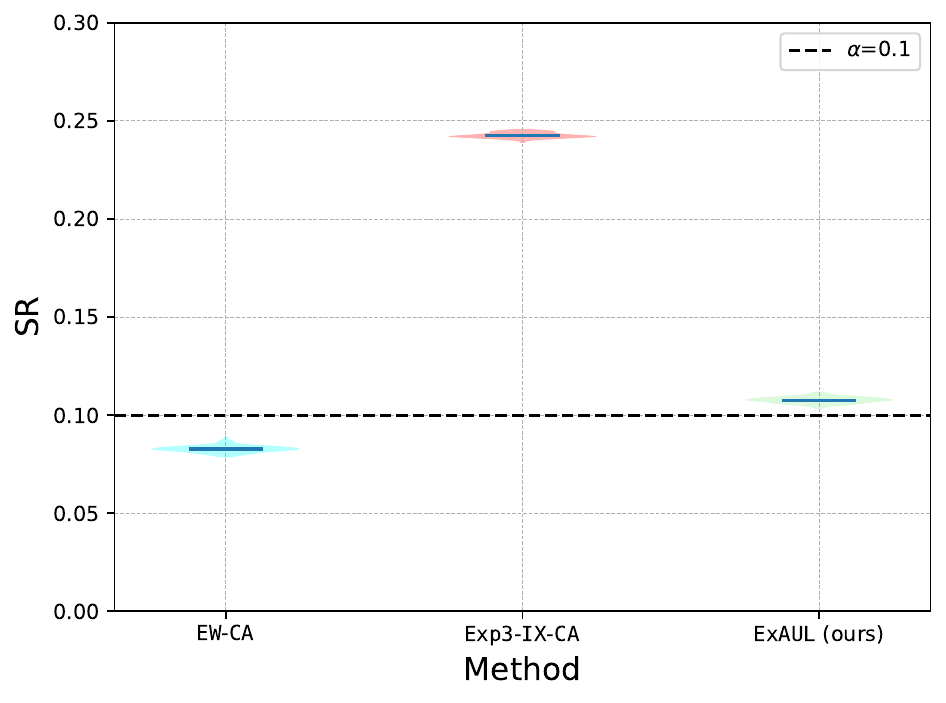}
    }
    \caption{Comparison of conformal abstention methods under a distribution-shift environment with LLaMA3.1-8B-Instruct as a generator ($T = 30\mathrm{K}, \alpha = 0.10$), in a gradual distribution shift from NQ to TriviaQA over time.
    The violin plots are drawn with randomly chosen $30\mathrm{K}$ samples with $100$ random trials.
    }
    \label{fig:shift:llama-nq-linear}
\end{figure*}

\begin{figure*}[h]
    \centering
    \subfigure[SR over time]{
    \includegraphics[width=0.31\textwidth]{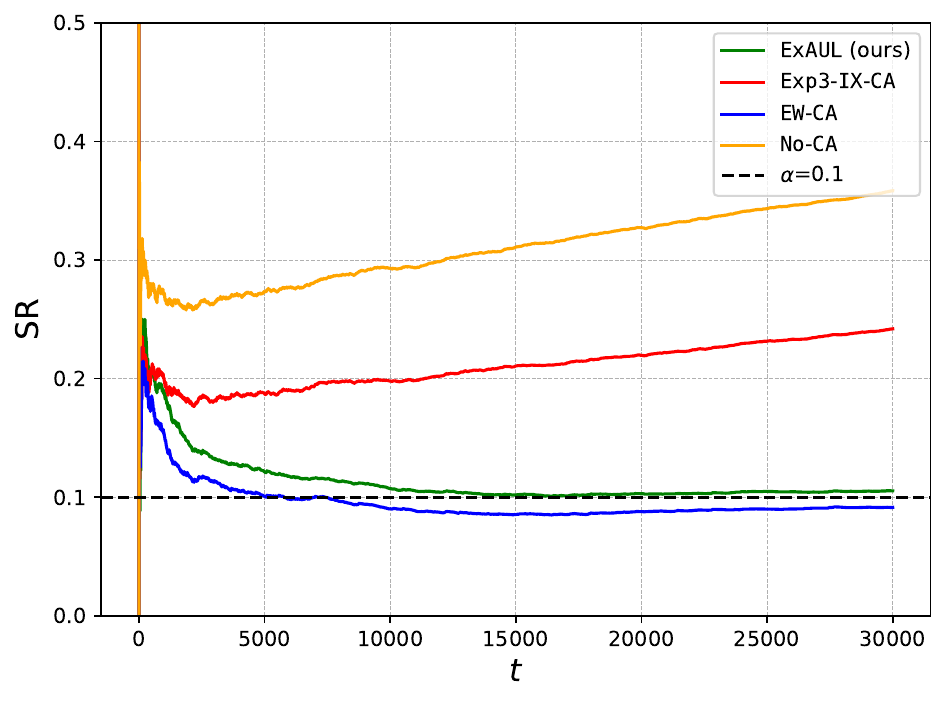}
    }
    \subfigure[Inefficiency over time]{
    \includegraphics[width=0.31\textwidth]{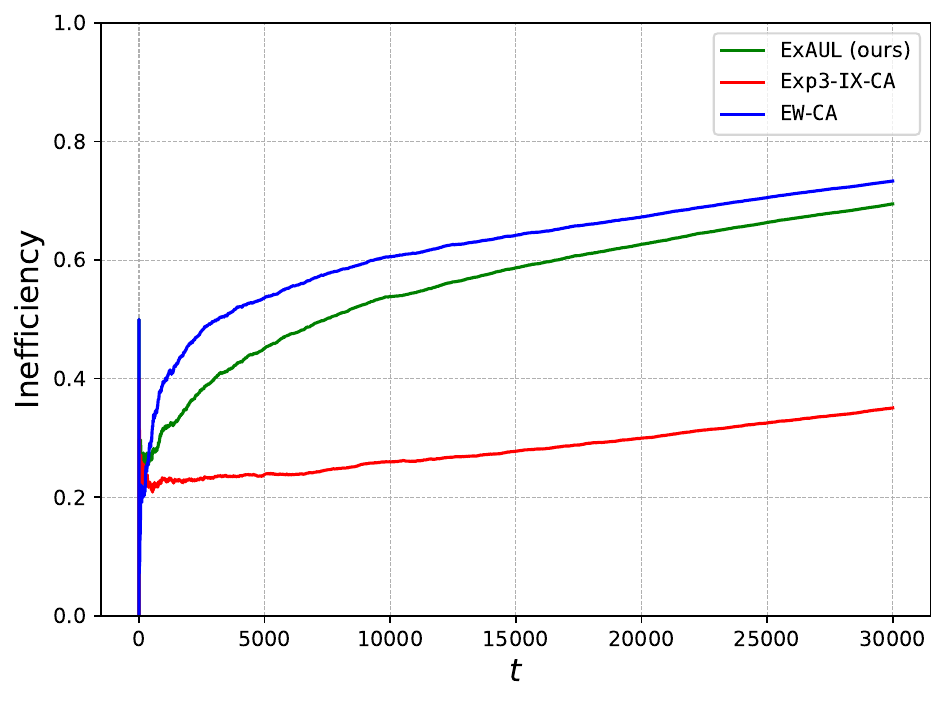}
    }
    \subfigure[SR distribution]{
    \includegraphics[width=0.31\textwidth]{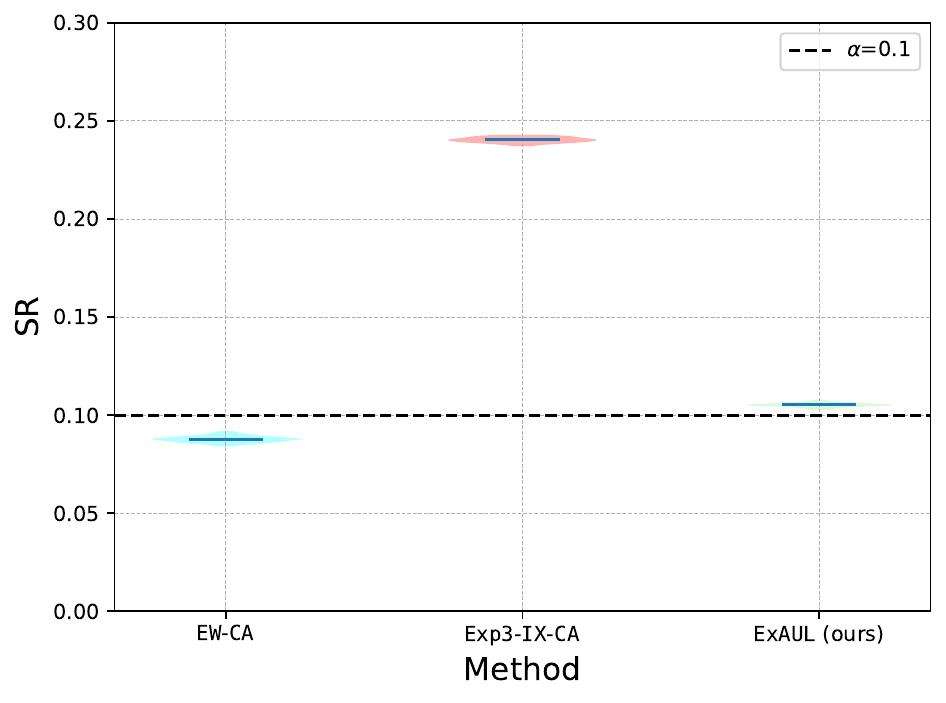}
    }
    \caption{Comparison of conformal abstention methods under a distribution-shift environment with LLaMA3.1-8B-Instruct as a generator ($T = 30\mathrm{K}, \alpha = 0.10$), in a gradual distribution shift from TriviaQA to NQ over time.
    The violin plots are drawn with randomly chosen $30\mathrm{K}$ samples with $100$ random trials.
    }
    \label{fig:shift:llama-tri-linear}
\end{figure*}

\subsubsection{Domain Shift: Open-domain QA to Medical QA}
\label{sec:medqa}
To evaluate a larger domain shift than NQ$\leftrightarrow$TriviaQA, we consider MedQA-USMLE~\citep{jin2020disease}, a medical QA benchmark, and construct streams that shift between an open-domain dataset and MedQA in both directions. 
\begin{figure*}[h]
    \centering
    \subfigure[SR over time]{
    \includegraphics[width=0.31\textwidth]{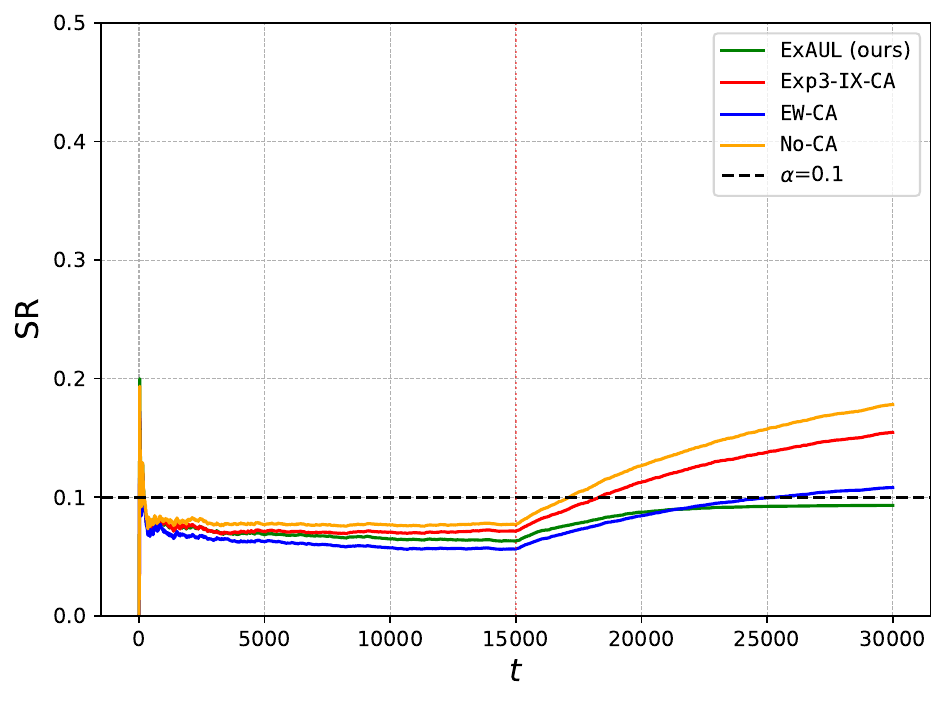}
    }
    \subfigure[Inefficiency over time]{
    \includegraphics[width=0.31\textwidth]{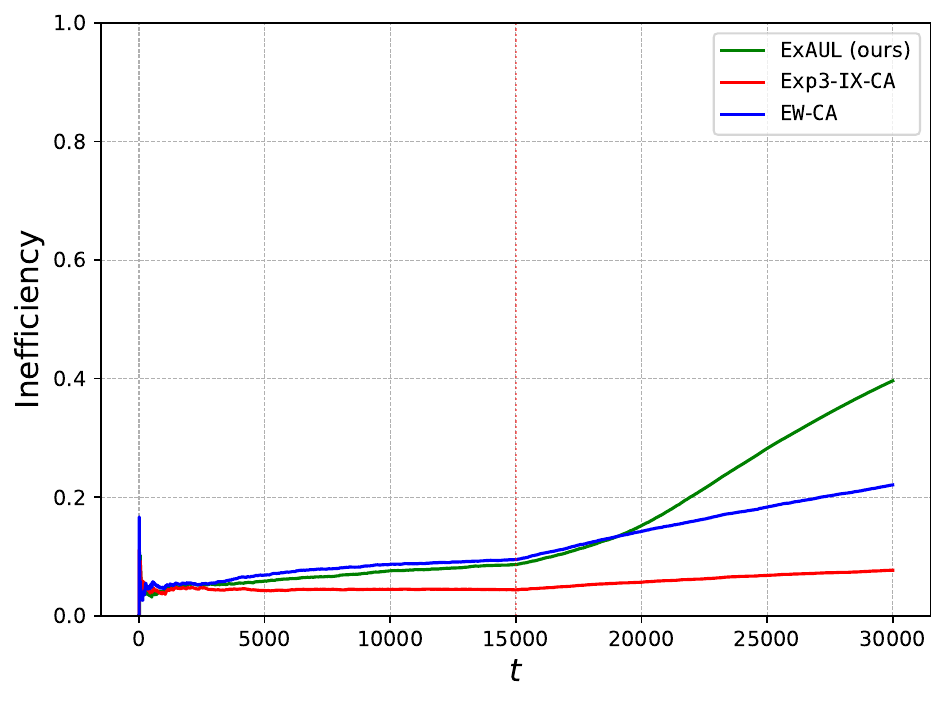}
    }
    \subfigure[SR distribution]{
    \includegraphics[width=0.31\textwidth]{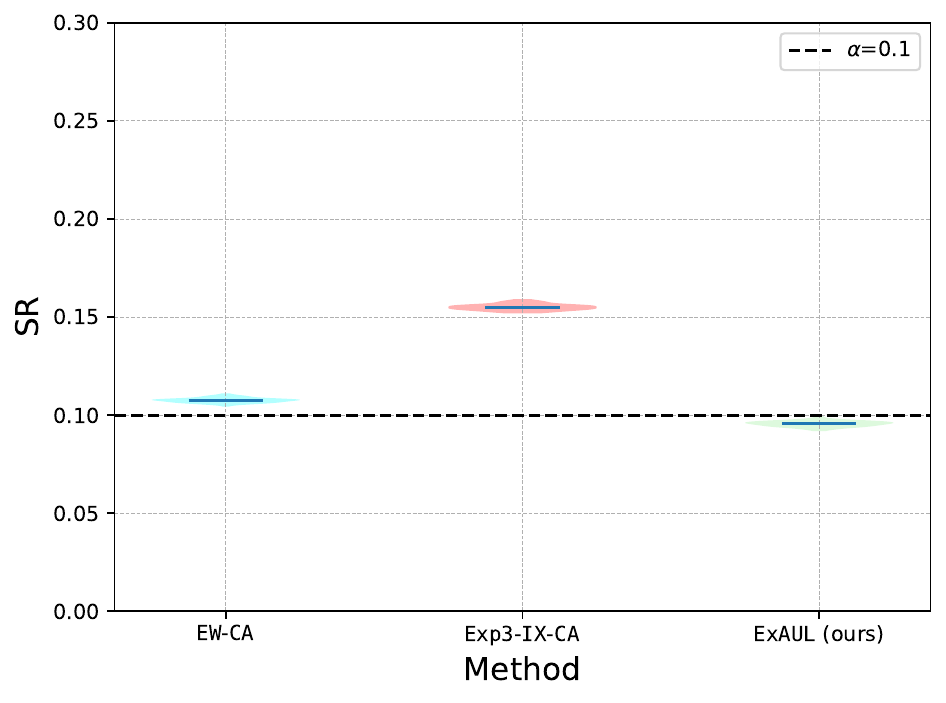}
    }
    \caption{Comparison of conformal abstention methods under a distribution-shift environment with GPT-5.4 as a generator ($T = 30\mathrm{K}, \alpha = 0.1$), in a single distribution shift from MedQA to NQ over time.
    The violin plots are drawn with randomly chosen $30\mathrm{K}$ samples with $100$ random trials.
    }
    \label{fig:shift:medqa-nq}
\end{figure*}
\begin{figure*}[h]
    \centering
    \subfigure[SR over time]{
    \includegraphics[width=0.31\textwidth]{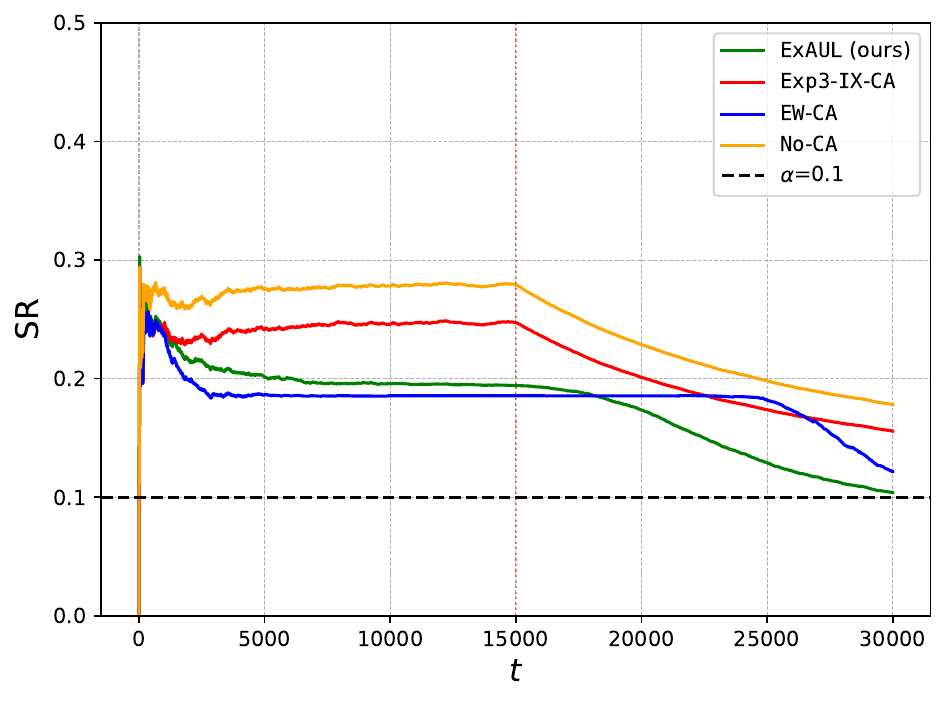}
    }
    \subfigure[Inefficiency over time]{
    \includegraphics[width=0.31\textwidth]{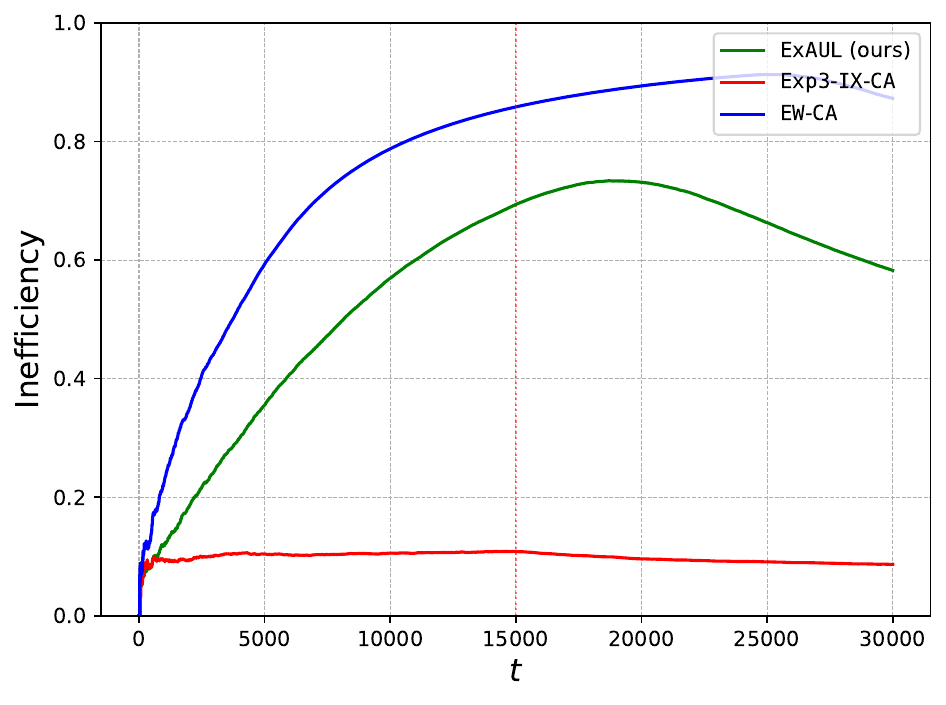}
    }
    \subfigure[SR distribution]{
    \includegraphics[width=0.31\textwidth]{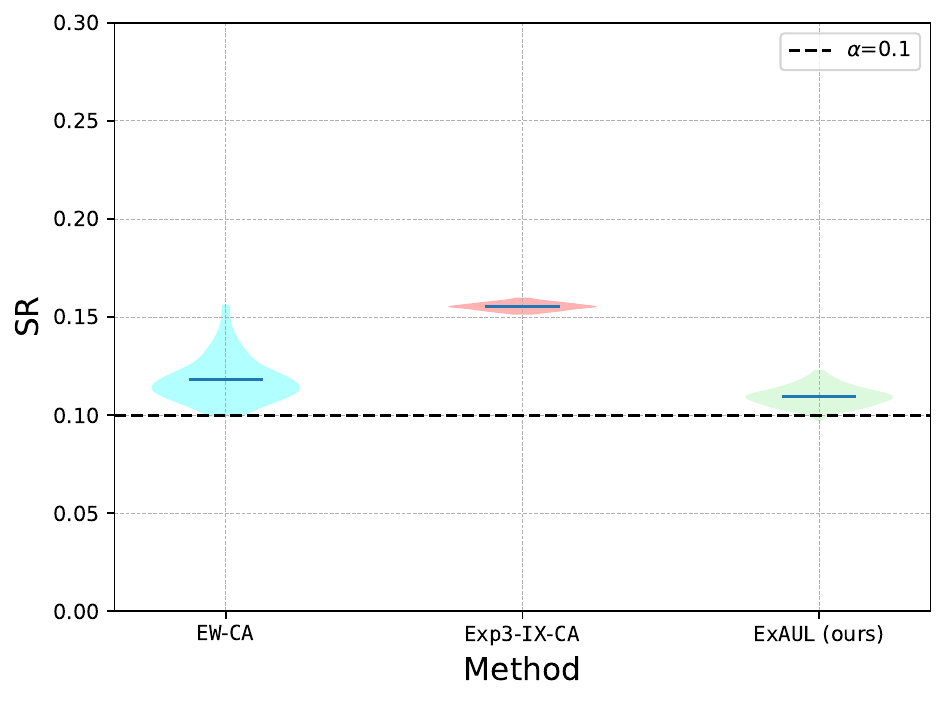}
    }
    \caption{Comparison of conformal abstention methods under a distribution-shift environment with GPT-5.4 as a generator ($T = 30\mathrm{K}, \alpha = 0.1$), in a single distribution shift from NQ to MedQA over time.
    The violin plots are drawn with randomly chosen $30\mathrm{K}$ samples with $100$ random trials.
    }
    \label{fig:shift:nq-medqa}
\end{figure*}
\begin{figure*}[h]
    \centering
    \subfigure[SR over time]{
    \includegraphics[width=0.31\textwidth]{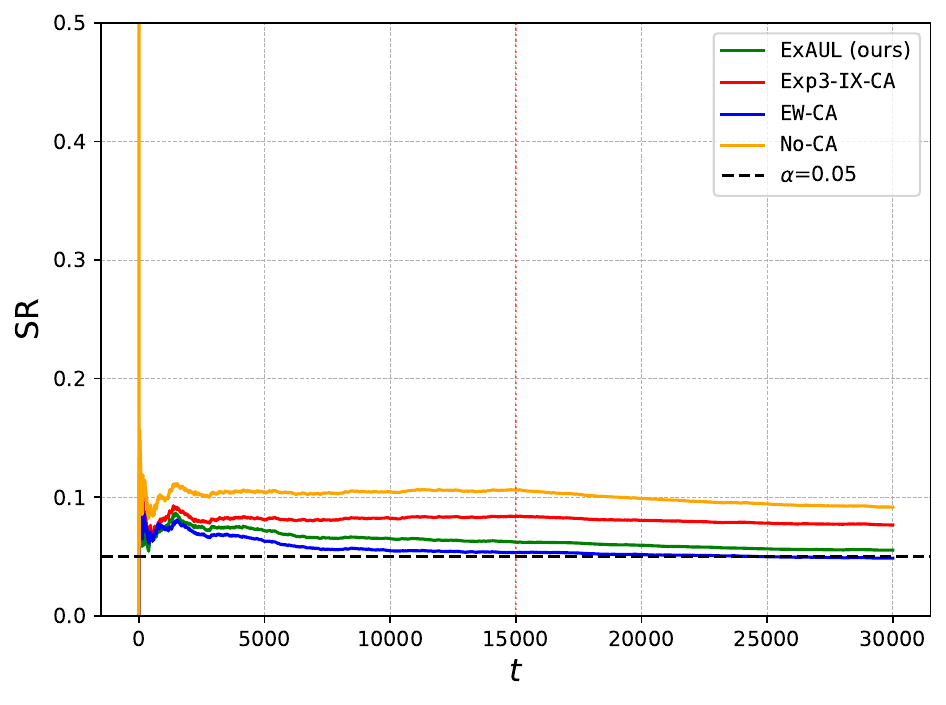}
    }
    \subfigure[Inefficiency over time]{
    \includegraphics[width=0.31\textwidth]{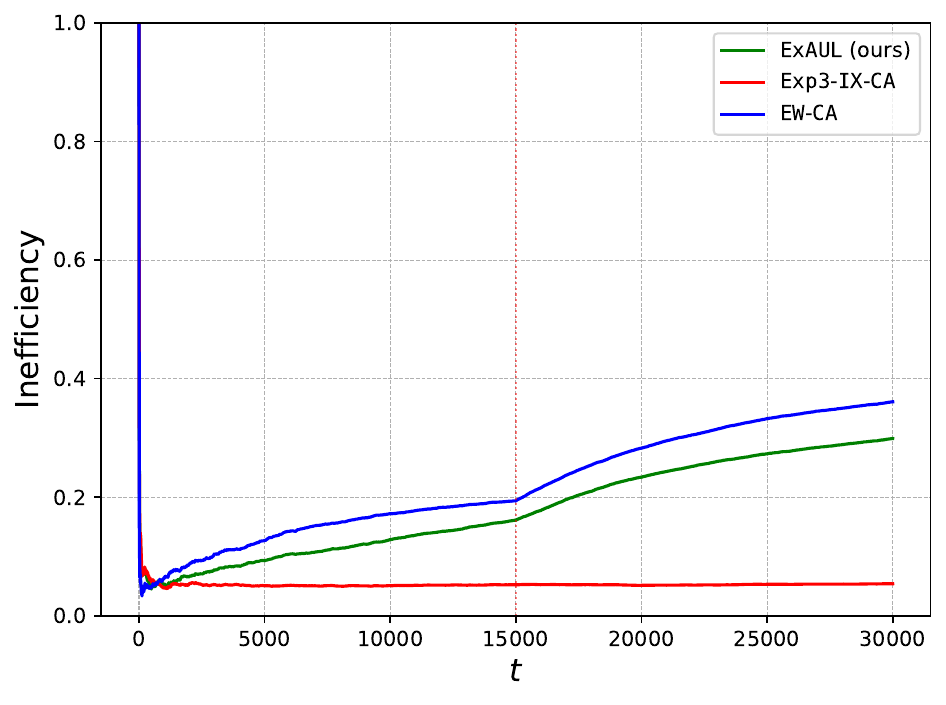}
    }
    \subfigure[SR distribution]{
    \includegraphics[width=0.31\textwidth]{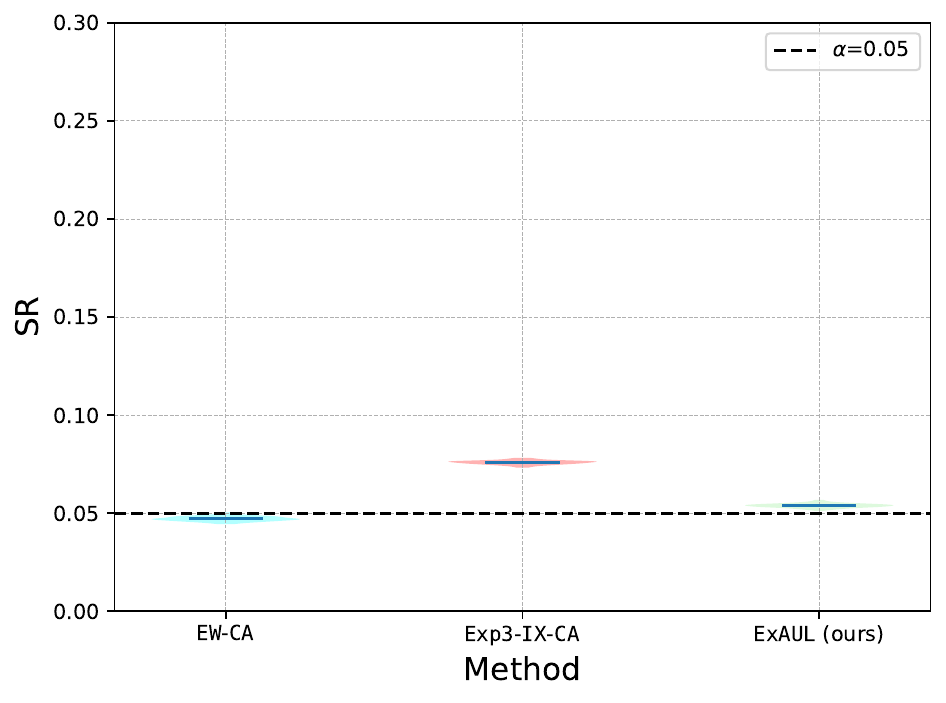}
    }
    \caption{Comparison of conformal abstention methods under a distribution-shift environment with GPT-5.4 as a generator ($T = 30\mathrm{K}, \alpha = 0.05$), in a single distribution shift from TriviaQA to MedQA over time.
    The violin plots are drawn with randomly chosen $30\mathrm{K}$ samples with $100$ random trials.
    }
    \label{fig:shift:tri-medqa}
\end{figure*}
\begin{figure*}[h]
    \centering
    \subfigure[SR over time]{
    \includegraphics[width=0.31\textwidth]{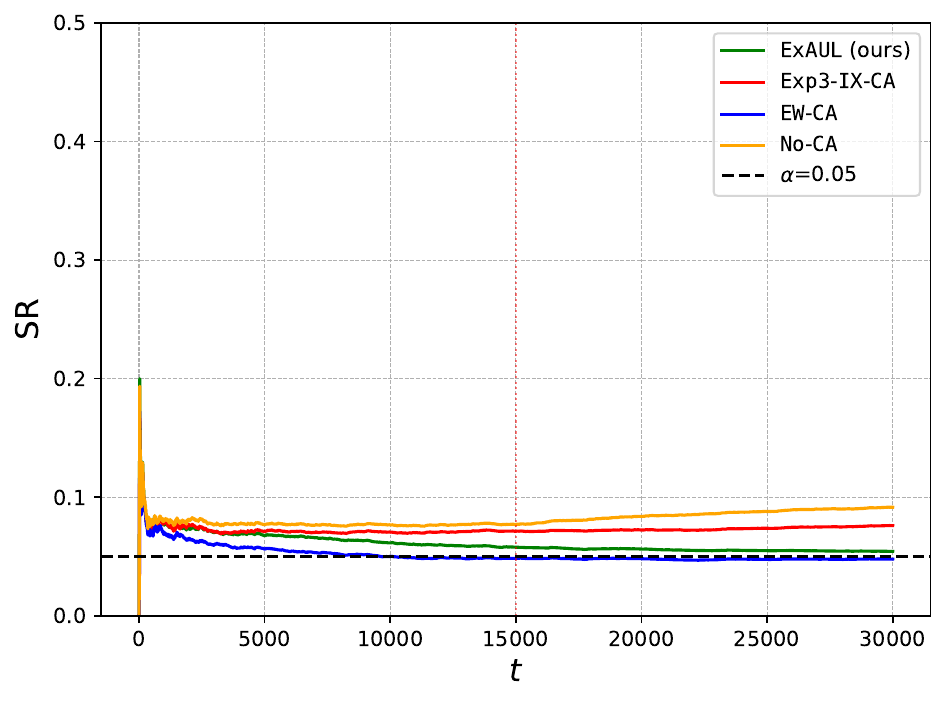}
    }
    \subfigure[Inefficiency over time]{
    \includegraphics[width=0.31\textwidth]{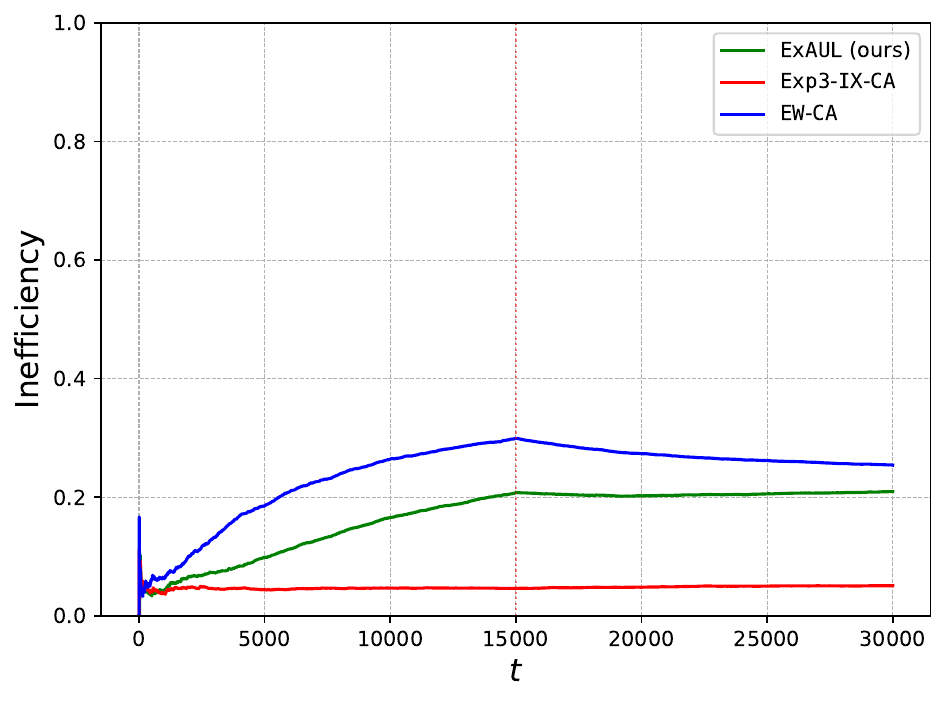}
    }
    \subfigure[SR distribution]{
    \includegraphics[width=0.31\textwidth]{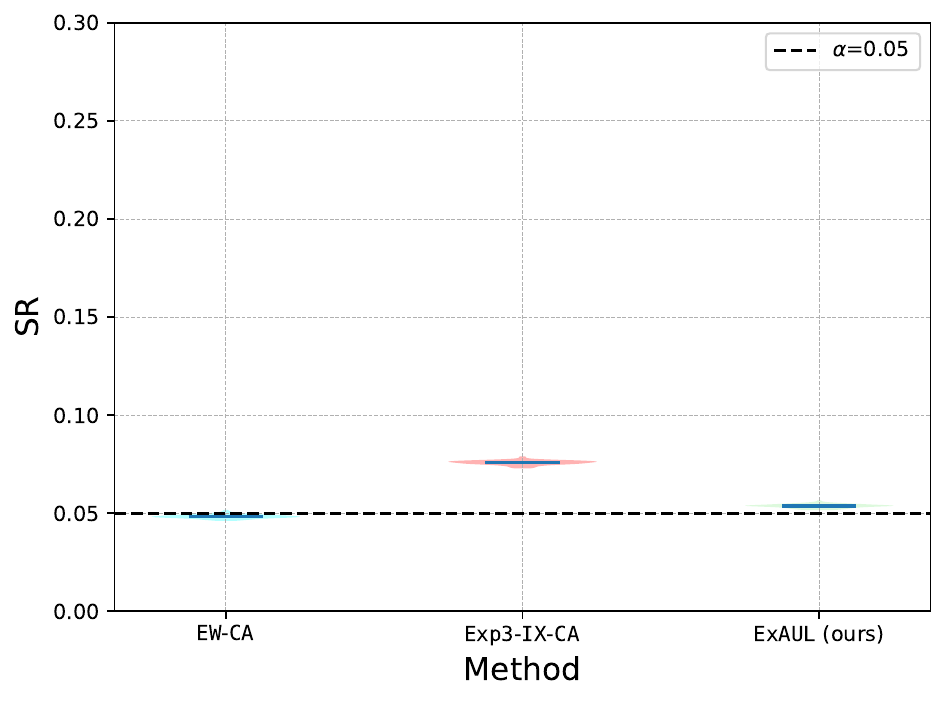}
    }
    \caption{Comparison of conformal abstention methods under a distribution-shift environment with GPT-5.4 as a generator ($T = 30\mathrm{K}, \alpha = 0.05$), in a single distribution shift from MedQA to TriviaQA over time.
    The violin plots are drawn with randomly chosen $30\mathrm{K}$ samples with $100$ random trials.
    }
    \label{fig:shift:medqa-tri}
\end{figure*}

Despite the substantial domain shift, $\mathtt{ExAUL}$ keeps the SR close to $\alpha$ in all four settings (Figures~\ref{fig:shift:medqa-nq}--\ref{fig:shift:tri-medqa}).
Shifting from open-domain QA to MedQA increases inefficiency, while SR control is preserved.





\subsection{Imbalanced Streams}
\label{sec:imbalance}
The shift experiments above are balanced between the two datasets.
To test persistent imbalance, we consider (i) a \emph{steady-state imbalance} setting, where each round is drawn from NQ with probability $p_{\mathrm{NQ}}$ and from TriviaQA otherwise, and (ii) an \emph{imbalanced discrete shift}, where a TriviaQA block is followed by an NQ block whose lengths are proportional to $(p_{\mathrm{TriviaQA}}, p_{\mathrm{NQ}})$.
All experiments use $T=60\mathrm{K}$ and $\alpha=0.1$ with GPT-5.4 and Llama-3.1-8B.

\begin{table}[h]
\centering
\caption{SR of \texttt{ExAUL} under imbalanced streams ($T=60\mathrm{K}$, $\alpha=0.1$). Each column is the mixing ratio $(p_{\mathrm{NQ}}, p_{\mathrm{TriviaQA}})$.}
\label{tab:imbalance}
\small
\setlength{\tabcolsep}{4pt}
\begin{tabular}{llcccc}
\toprule
Setting & Model & (0.2,0.8) & (0.4,0.6) & (0.6,0.4) & (0.8,0.2) \\
\midrule
\multirow{2}{*}{Steady-state}
& GPT-5.4      & 0.070 & 0.087 & 0.105 & 0.138 \\
& Llama-3.1-8B & 0.082 & 0.093 & 0.108 & 0.130 \\
\midrule
\multirow{2}{*}{Discrete shift (TriviaQA$\to$NQ)}
& GPT-5.4      & 0.072 & 0.083 & 0.093 & 0.111 \\
& Llama-3.1-8B & 0.084 & 0.094 & 0.107 & 0.124 \\
\bottomrule
\end{tabular}
\end{table}

In both settings, the SR stays near $\alpha$ for balanced and TriviaQA-dominated streams and rises gradually as NQ dominates (Table~\ref{tab:imbalance}).
This reflects the difference in dataset difficulty rather than a failure of the learner: the per-dataset SR is roughly constant across ratios (about $0.16$--$0.21$ on NQ and $0.05$--$0.07$ on TriviaQA), so the overall SR tracks the share of NQ in the stream, and the minority dataset shows no additional degradation.
Thus \texttt{ExAUL} degrades gracefully rather than failing abruptly under persistent imbalance.

\clearpage

\subsection{Analysis on Time Horizon and Desired SR Parameters}
\label{sec:analysisonTandalpha}
\begin{figure*}[h]
  \subfigure[NQ dataset along with GPT-5.4]{
    \includegraphics[width=0.4\linewidth]{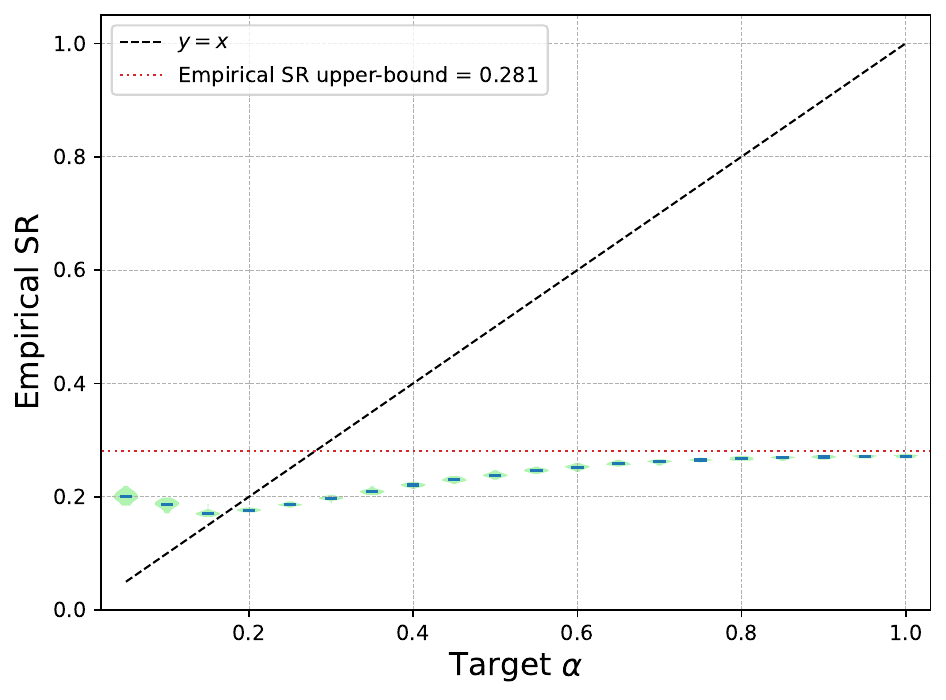}
    \label{fig:plot_over_alpha:nq:gpt-3.5:fdr}
  }
  \hfill
  \subfigure[NQ dataset along with LLaMA3.1-8B]{
    \includegraphics[width=0.4\linewidth]{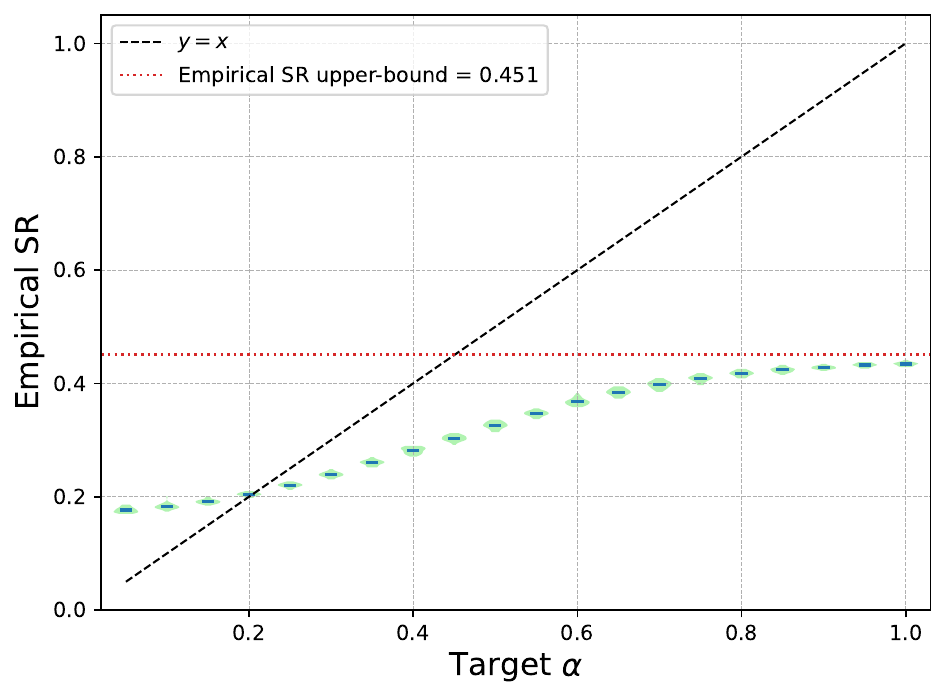}
    \label{fig:plot_over_alpha:nq:llama3.1:fM2:1500000:fdr}
  }
  \subfigure[TriviaQA dataset along with GPT-5.4]{
    \includegraphics[width=0.4\linewidth]{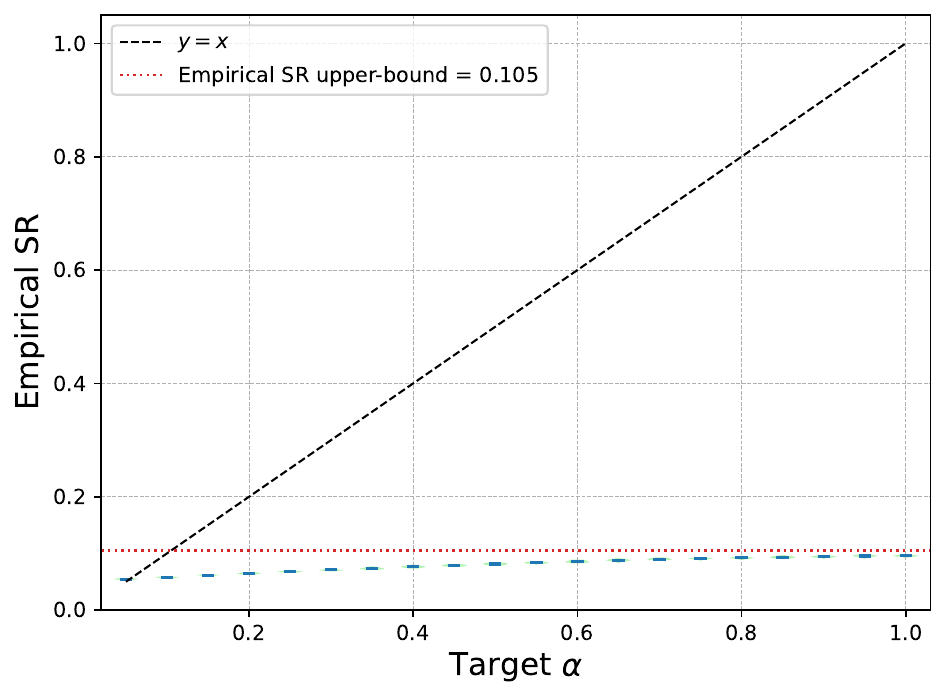}
    \label{fig:plot_over_alpha:triviaqa:gpt-3.5:fM2:1500000:fdr}
  }
  \hfill
  \subfigure[TriviaQA dataset along with LLaMA3.1-8B]{
    \includegraphics[width=0.4\linewidth]{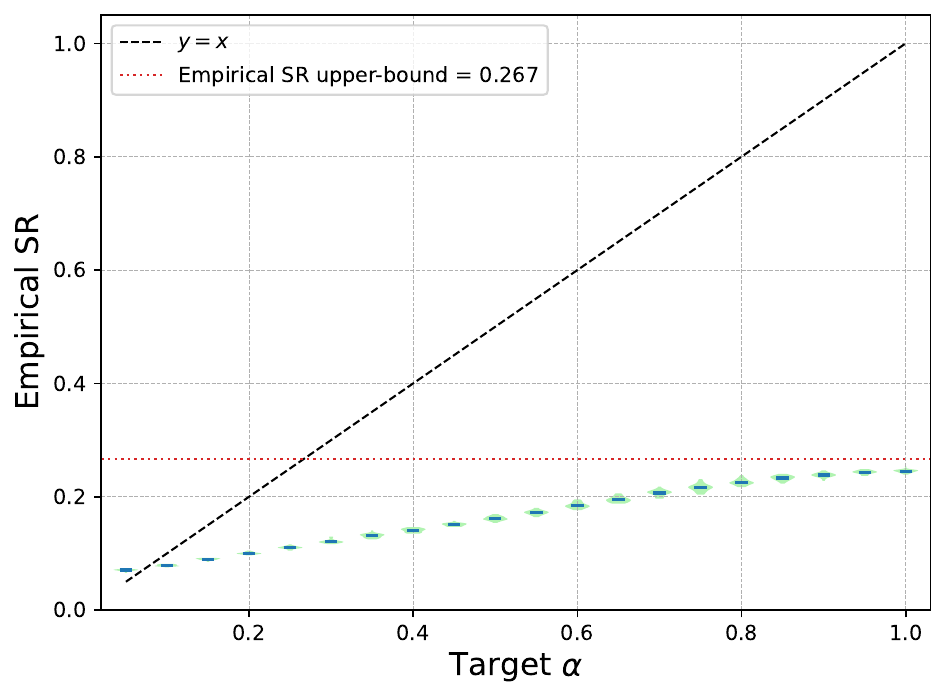}
    \label{fig:plot_over_alpha:triviaqa:llama3.1:fM2:1500000:fdr}
  }
  \vspace{-1ex}
  \caption{The violin plots for the ablation study over varying $\alpha$ on datasets under stochastic environments. The violin plots are drawn with randomly chosen $30\mathrm{K}$ samples over 30 random trials.
  Here, converged SRs mostly go along with target risk $\alpha$ (\ie  violin plots are below of the diagonal line), which empirically demonstrates that our algorithm finds efficient conformal abstainers with SR guarantees.
  Note that $\textbf{Err}_T$ refers to the error rate on a given dataset, \ie $\textbf{Err}_T \coloneqq \sum_{t=1}^T \mathbbm{1}\( G(\x_t) \neq_E \y_t \)/T$
  and the SRs of conformal abstainers should be lower than $\textbf{Err}_T$ to be meaningful in controlling the SRs.
  In (a), the SR exceeds $\alpha$ for small $\alpha$ due to the poor calibration of $f_{\texttt{con}}$, not a failure of the guarantee.
  For NQ with GPT-5.4, $44.5\%$ of the samples have the maximum score $f_{\texttt{con}} \approx 1$ (the largest floating-point value below $1$, as $f_t \in [0,1)$), and $14.8\%$ of them are incorrect.
  As no threshold can separate these tied samples, every $\tau < 1$ answers all of them, so for $\alpha$ below $14.8\%$, \ours is driven toward the always-abstaining threshold $\tau = 1$.
  The SR violation $\Vs_T$ in~(\ref{eq:fdrrisk}), which our theory bounds, then remains controlled since abstention incurs no violation, whereas the empirical $\SR_T$ is a ratio over only a few answered samples and thus becomes large.
  This is consistent with our SR bound
  in Section~\ref{sec:advfeedback:algorithm}, which depends on $1-\Ine_T$, and thus becomes vacuous when \ours rarely answers (see Appendix~\ref{sec:discussion:ineff} for further discussion).
  }
  \label{fig:stochastic:main:plot_over_alpha}
\end{figure*}

\begin{figure*}
  \subfigure[TriviaQA shifts to NQ with GPT-5.4]{
    \includegraphics[width=0.43\linewidth]{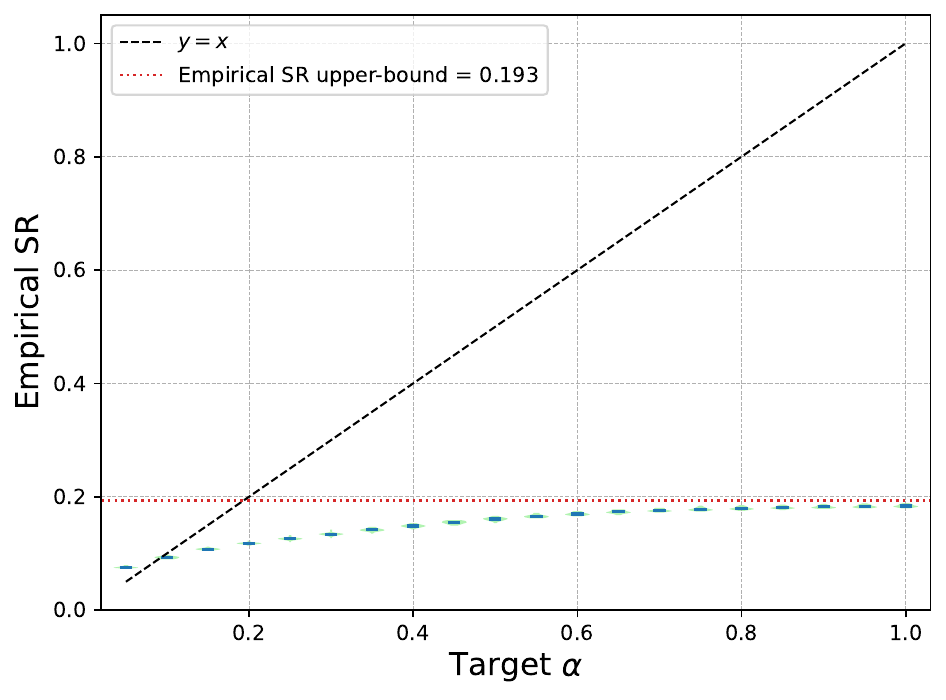}
    \label{fig:tri:shift:gpt-3.5:fM2:fdr}
  }
  \hfill
  \subfigure[TriviaQA shifts to NQ  with LLaMA3.1-8B]{
    \includegraphics[width=0.43\linewidth]{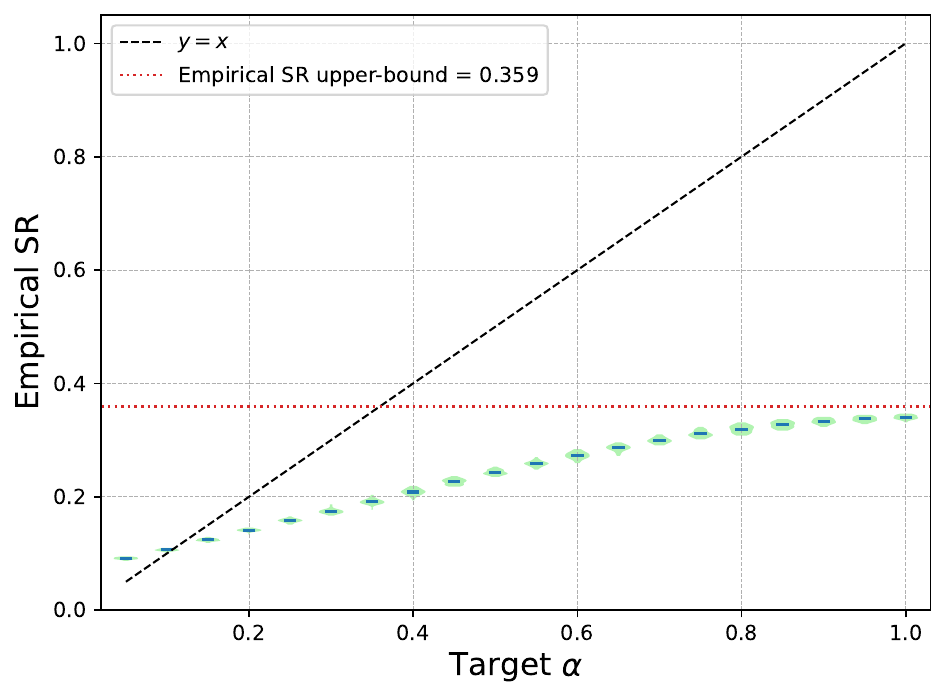}
    \label{fig:tri:shift:llama3.1:fM2:fdr}
  }
  \subfigure[NQ shifts to TriviaQA with GPT-5.4]{
    \includegraphics[width=0.43\linewidth]{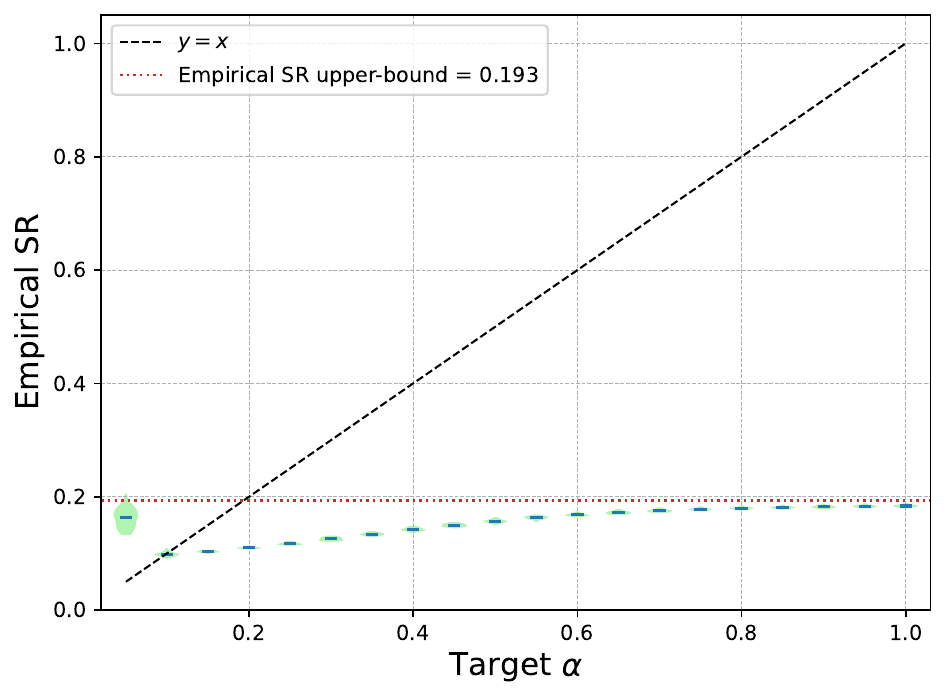}
    \label{fig:nq:shift:gpt-3.5:fM2:fdr}
  }
  \hfill
  \subfigure[NQ shifts to TriviaQA with LLaMA3.1-8B]{
    \includegraphics[width=0.43\linewidth]{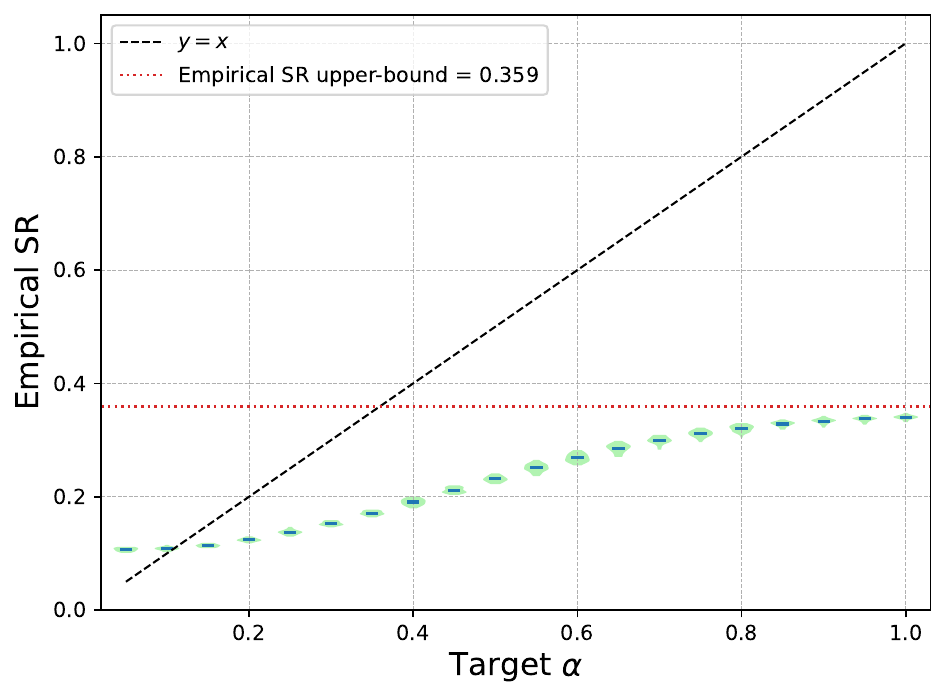}
    \label{fig:nq:shift:llama3.1:fM2:fdr}
  }
  \vspace{-1ex}
  \caption{The violin plots for the ablation study over varying $\alpha$ on datasets under distribution-shift environments. The violin plots are drawn with randomly chosen dataset with single distribution shift between $15 \mathrm{K}$-sized  TriviaQA and $15\mathrm{K}$-sized NQ, over 30 random trials. Here, converged SRs mostly go along with target risk $\alpha$ (\ie  violin plots are below the diagonal line), which empirically demonstrates that our algorithm finds efficient conformal abstainers with SR guarantees.
  Note that $\textbf{Err}_T$ refers to the error rate on a given dataset, \ie $\textbf{Err}_T \coloneqq \sum_{t=1}^T \mathbbm{1}\( G(\x_t) \neq_E \y_t \)/T$
  and the SRs of conformal abstainers should be lower than $\textbf{Err}_T$ to be meaningful in controlling the SRs.
  The SR above $\alpha$ for small $\alpha$ in (c) has the same cause as in Figure~\ref{fig:plot_over_alpha:nq:gpt-3.5:fdr}: the stream starts with NQ with GPT-5.4, where many samples tie at the maximum score of $f_{\texttt{con}}$, so \ours is driven toward always abstaining and $\SR_T$ is computed over only a few answered samples, while $\Vs_T$ remains controlled.
  }
  
  \label{fig:exsul:shift:poa}
\end{figure*}

\begin{figure*}
  \subfigure[NQ dataset along with GPT-5.4]{
    \includegraphics[width=0.43\linewidth]{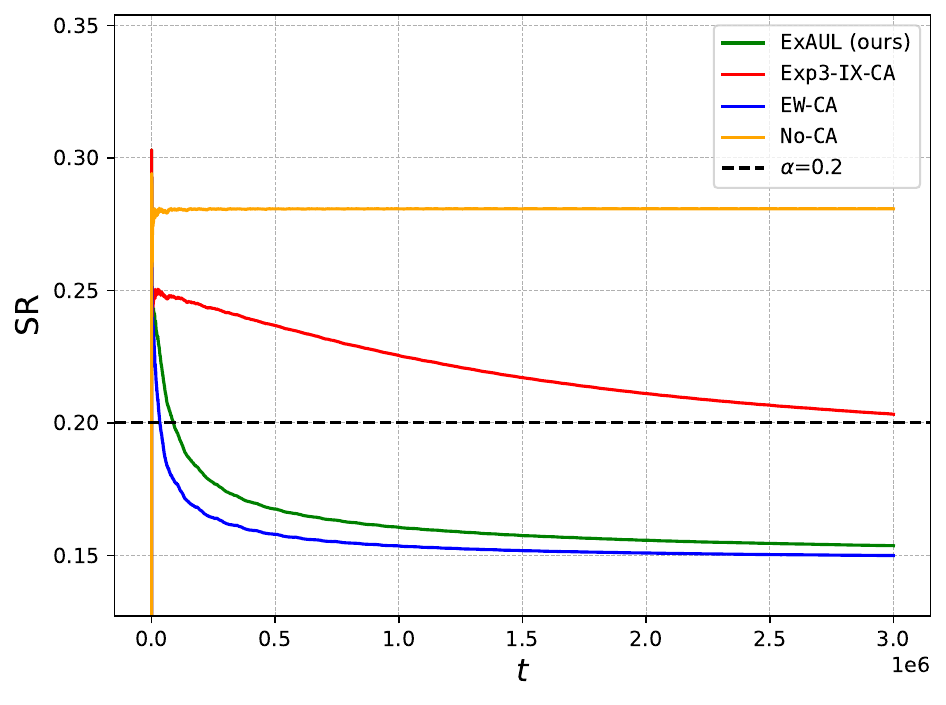}
    \label{fig:nq:gpt-3.5:fM2:1500000:fdr}
  }
  \hfill
  \subfigure[NQ dataset along with LLaMA3.1-8B]{
    \includegraphics[width=0.43\linewidth]{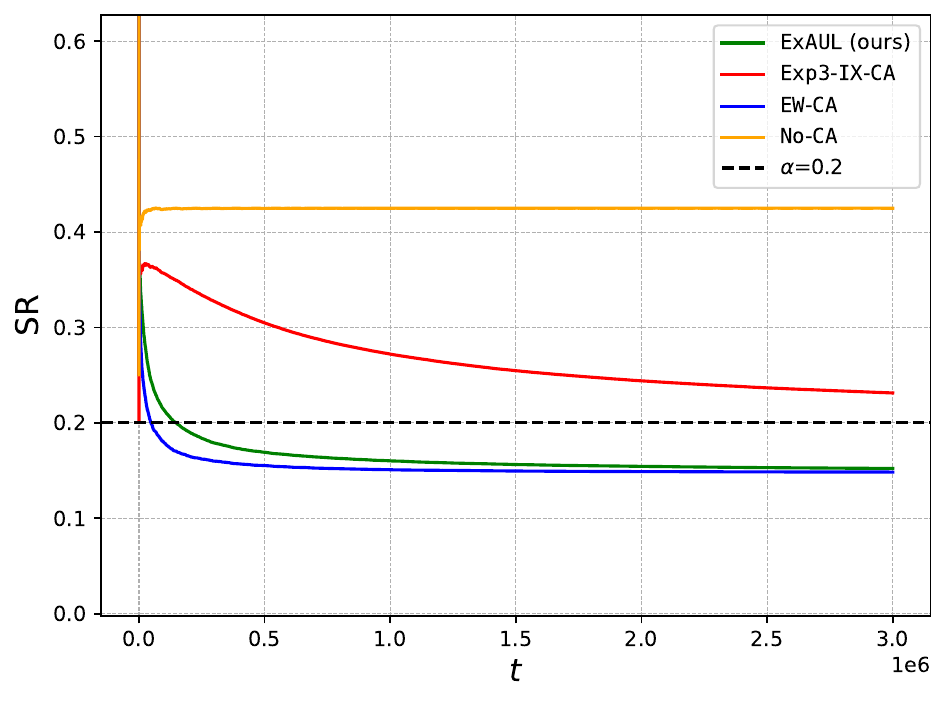}
  }
  \subfigure[TriviaQA dataset along with GPT-5.4]{
    \includegraphics[width=0.43\linewidth]{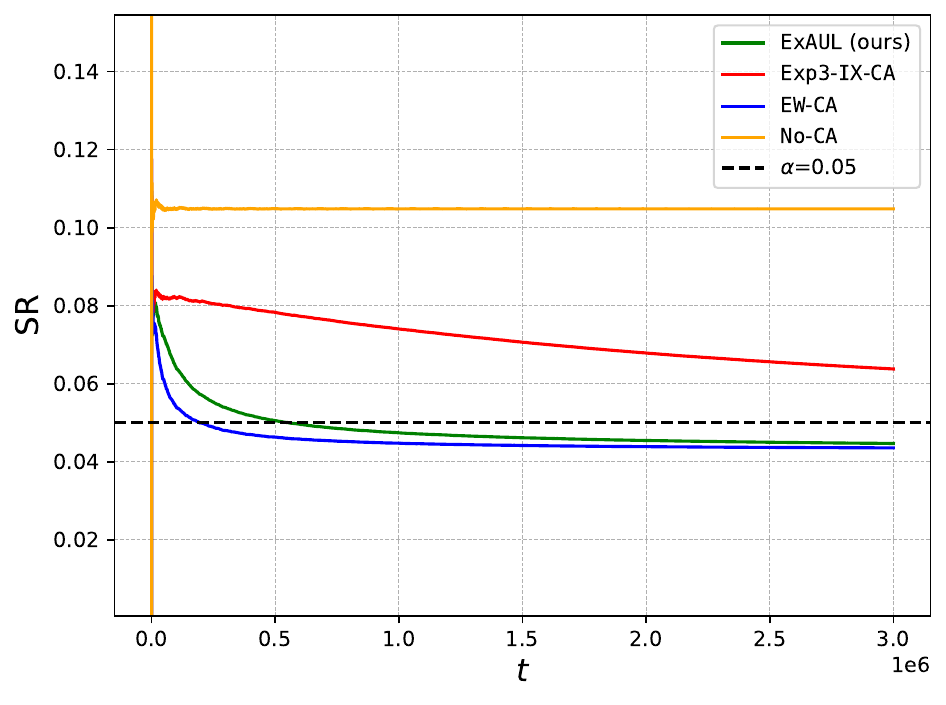}
    \label{fig:triviaqa:gpt-3.5:fM2:1500000:fdr}
  }
  \hfill
  \subfigure[TriviaQA dataset along with LLaMA3.1-8B]{
    \includegraphics[width=0.43\linewidth]{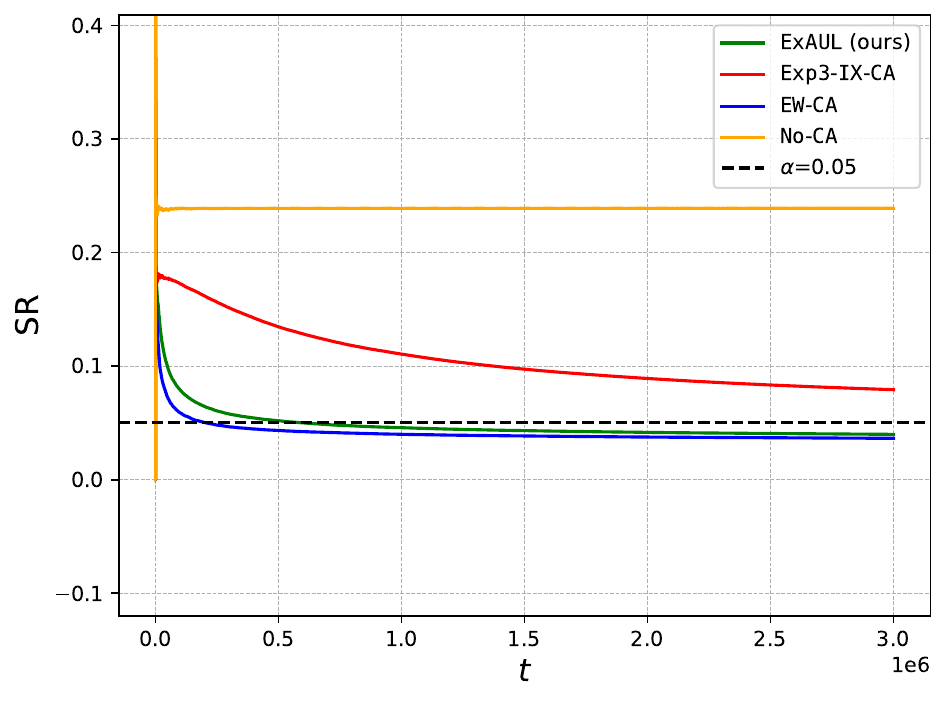}
    \label{fig:triviaqa:llama3.1:fM2:1500000:fdr}
  }
  \vspace{-1ex}
  \caption{Additional $\SR_t$ comparisons of conformal abstention methods under stochastic environments with a longer time horizon. Here, we repeat the datasets to evaluate whether \texttt{Exp3-IX-CA} converges over a long horizon {\color{\hl}($T = 3,000\mathrm{K}$).} 
  {\color{\hl}As shown in the plots, \texttt{Exp3-IX-CA} requires a long horizon $T$ to achieve reliable $\SR_t$ control.}
  }
  \label{fig:exp3:convergence}
\end{figure*}

\begin{figure*}
  \centering
  \subfigure[NQ dataset shift to TriviaQA dataset along with GPT-5.4]{
    \includegraphics[width=0.43\linewidth]{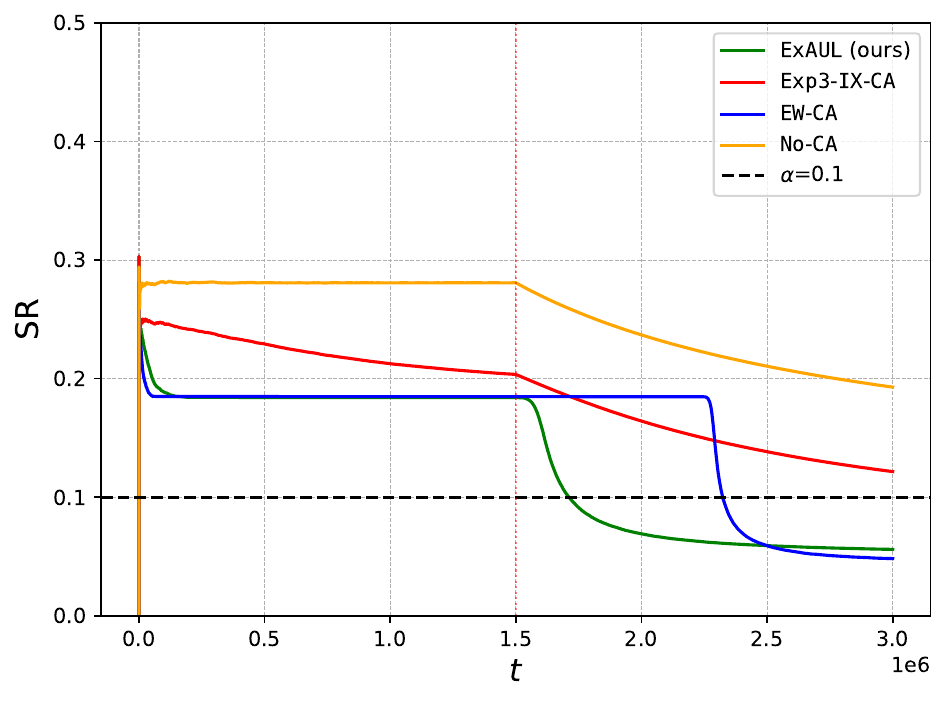}
  }
  \hfill
  \subfigure[NQ dataset shift to TriviaQA dataset along with LLaMA3.1-8B]{
    \includegraphics[width=0.43\linewidth]{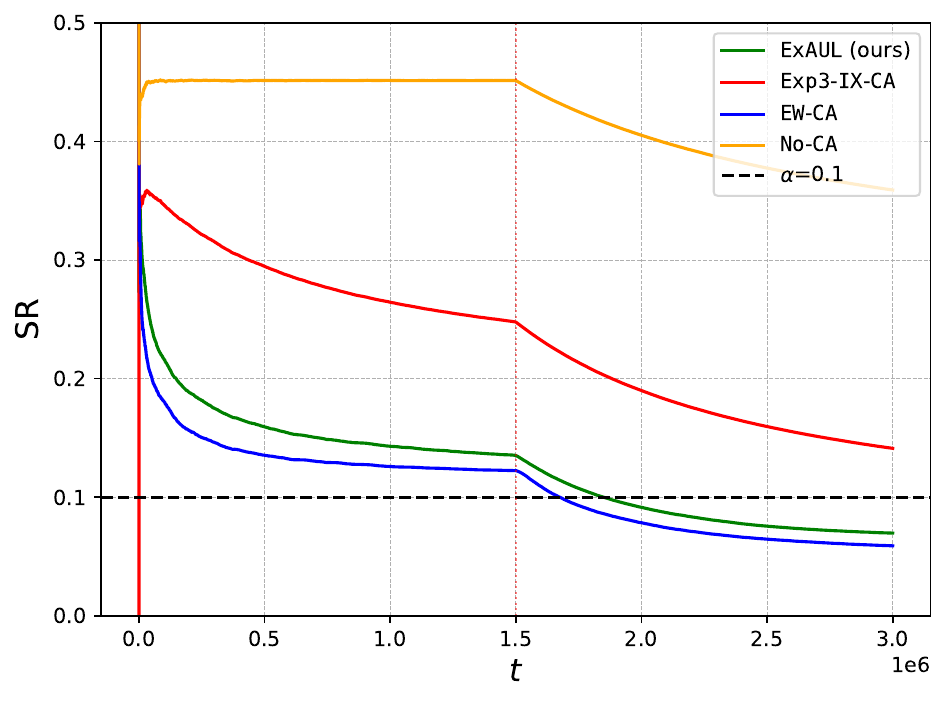}
  }
  \subfigure[TriviaQA dataset shift to NQ dataset along with GPT-5.4]{
    \includegraphics[width=0.43\linewidth]{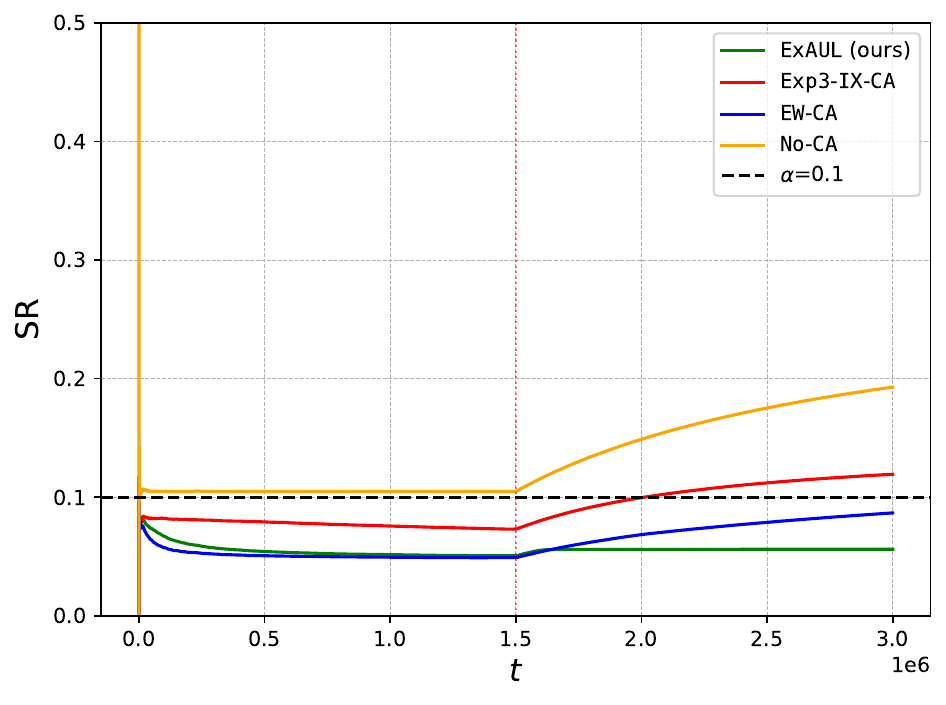}
  }
  \hfill
  \subfigure[TriviaQA dataset shift to NQ dataset along with LLaMA3.1-8B]{
    \includegraphics[width=0.43\linewidth]{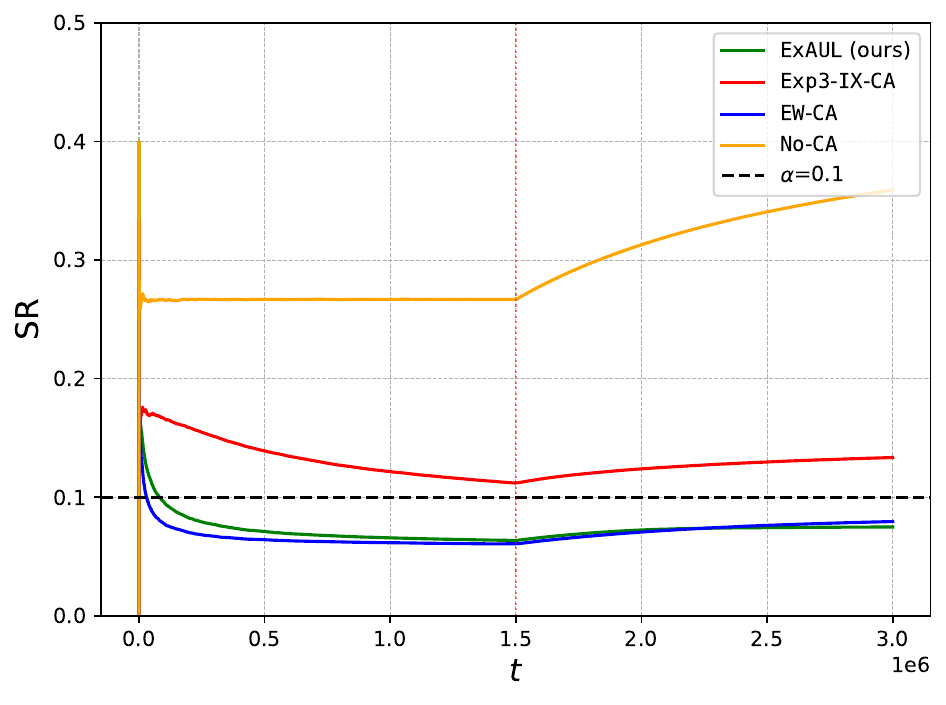}
  }
  \vspace{-1ex}
  
  \caption{Additional $\SR_t$ comparison of conformal abstention methods under distribution-shift environments with a longer time horizon. Here, we repeat the datasets to evaluate whether \texttt{Exp3-IX-CA} 
  maintains stable control of $\SR_t$ over a long horizon.
  We consider a single distribution shift between two {\color{\hl}$1,500 \mathrm{K}$-sized datasets} (NQ, triviaQA) denoted in a  dotted vertical line. 
  {\color{\hl}As shown in the plots, \texttt{Exp3-IX-CA} requires a long horizon $T$ to achieve reliable $\SR_t$ control.
  In (a), the SR converges to a value above $\alpha$ during the first phase. This is due to the poor calibration of $f_{\texttt{con}}$ for GPT-5.4 on NQ, which leads \texttt{EW-CA} and \ours to always abstain (see Appendix \ref{sec:discussion:ineff} and Figure \ref{fig:stochastic:main:plot_over_alpha}).}
  }
  \label{fig:exp3:shift_convergence}
\end{figure*}

\clearpage
\subsection{Analysis on Varying $\lambda$}
\label{sec:varyinglambda}
\begin{figure*}[h]
  \subfigure[NQ dataset along with GPT-5.4]{
    \includegraphics[width=0.43\linewidth]{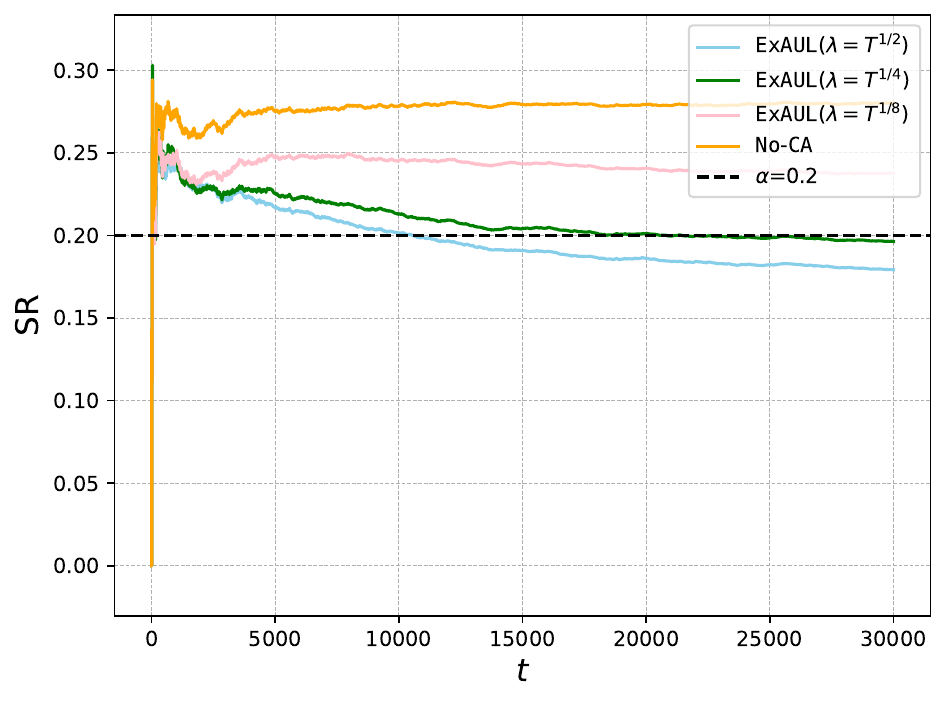}
  }
  \hfill
  \subfigure[NQ dataset along with LLaMA3.1-8B]{
    \includegraphics[width=0.43\linewidth]{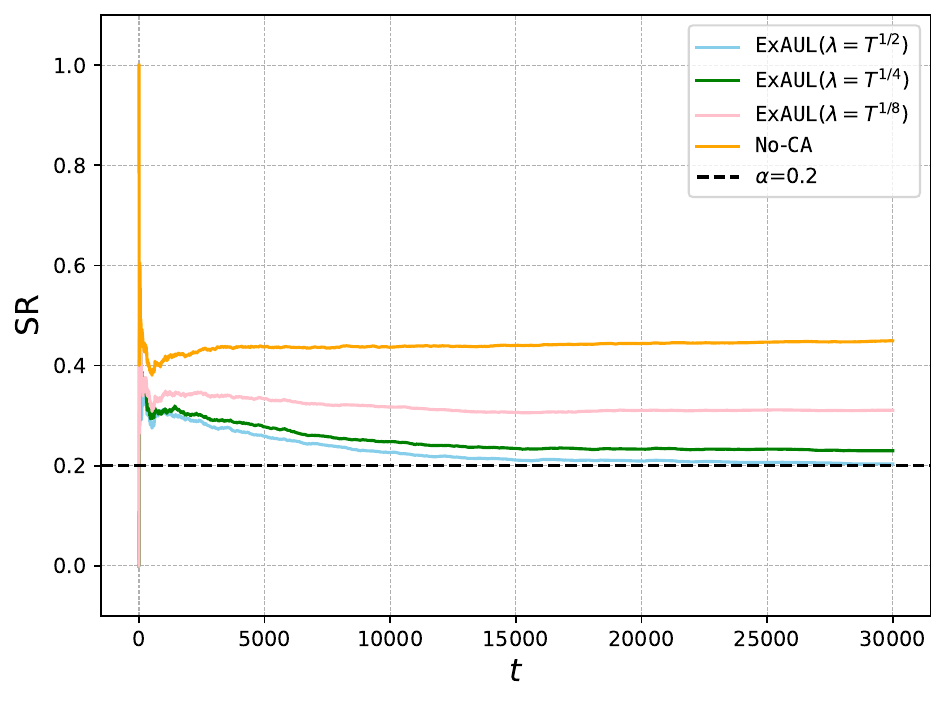}
  }
  \subfigure[TriviaQA dataset along with GPT-5.4]{
    \includegraphics[width=0.43\linewidth]{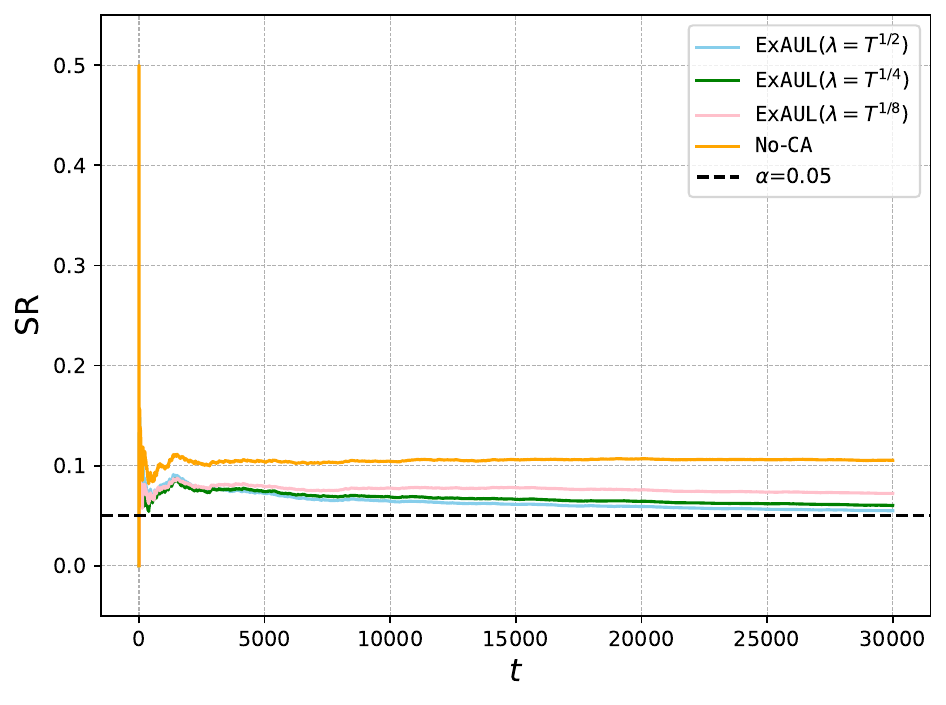}
  }
  \hfill
  \subfigure[TriviaQA dataset along with LLaMA3.1-8B]{
    \includegraphics[width=0.43\linewidth]{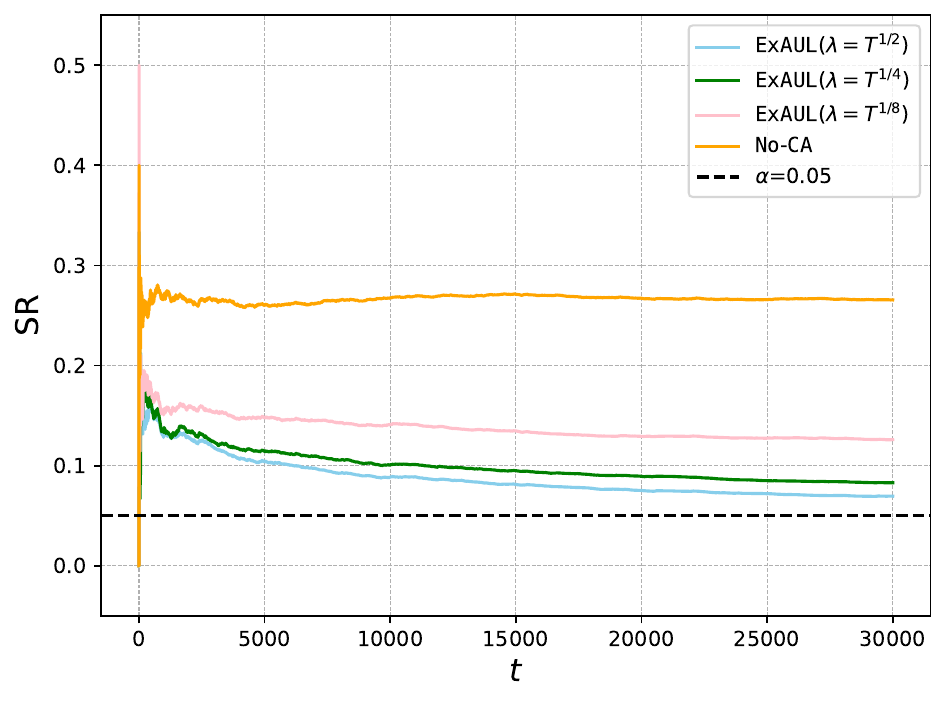}
  }
  \vspace{-1ex}
  \caption{Additional $\SR_t$ comparisons of conformal abstention methods under stochastic environments with varying $\lambda$ for $T = 30\mathrm{K}$ datasets.
  }
  \label{fig:fdroverlambda}
\end{figure*}

\begin{figure*}[h]
  \subfigure[NQ dataset along with GPT-5.4]{
    \includegraphics[width=0.43\linewidth]{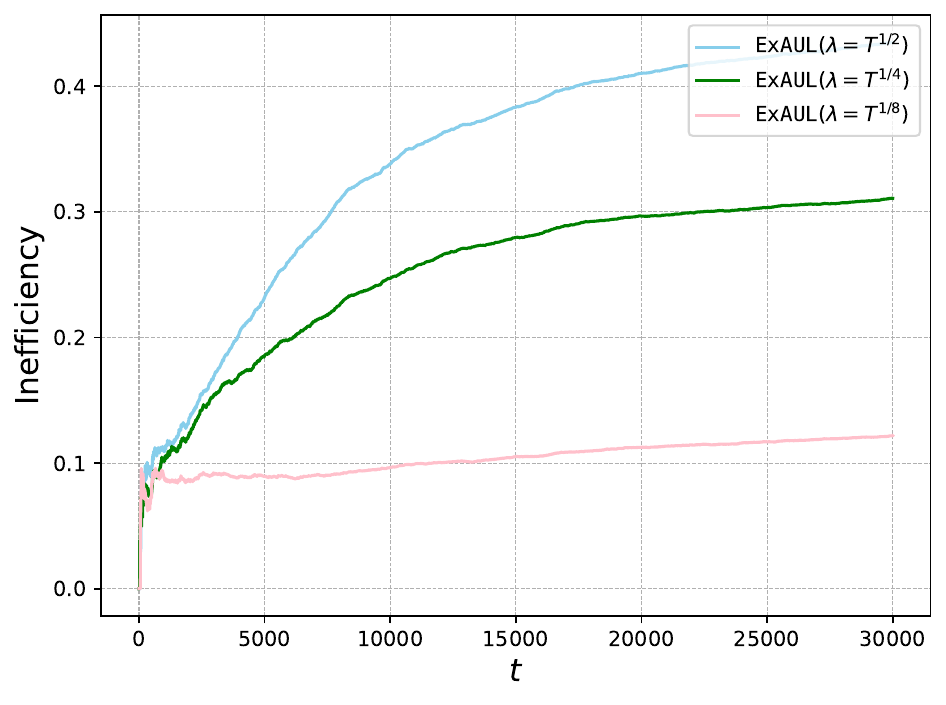}
  }
  \hfill
  \subfigure[NQ dataset along with LLaMA3.1-8B]{
    \includegraphics[width=0.43\linewidth]{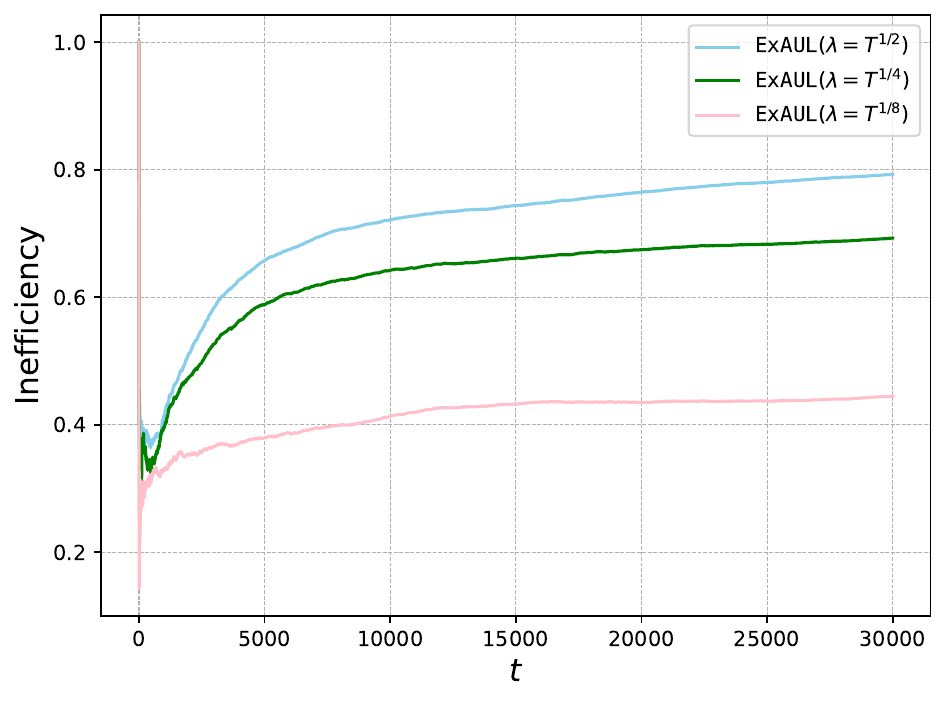}
  }
  \subfigure[TriviaQA dataset along with GPT-5.4]{
    \includegraphics[width=0.43\linewidth]{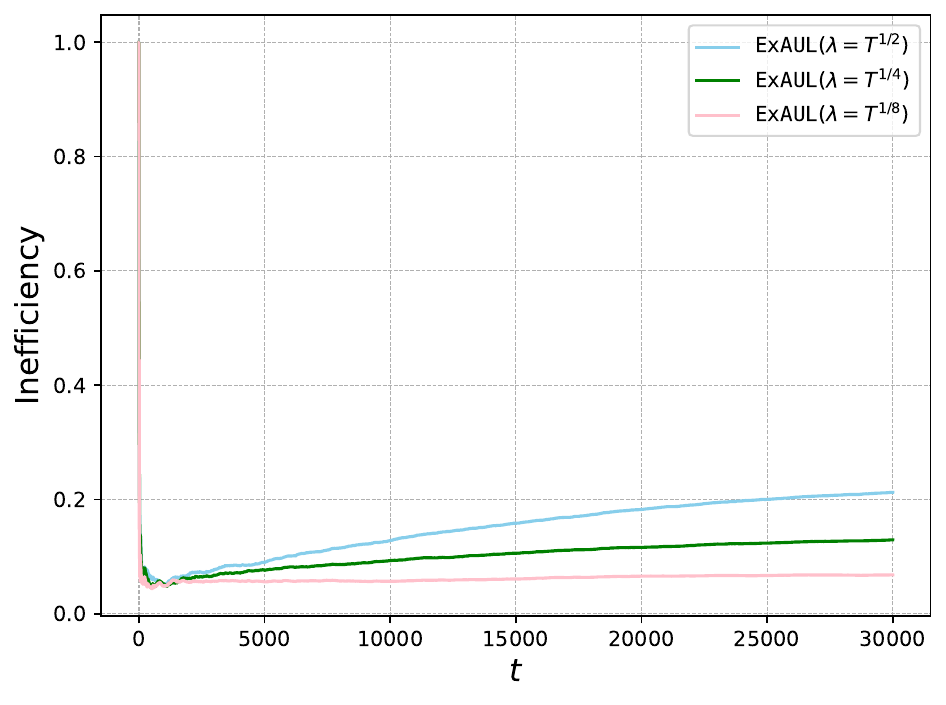}
  }
  \hfill
  \subfigure[TriviaQA dataset along with LLaMA3.1-8B]{
    \includegraphics[width=0.43\linewidth]{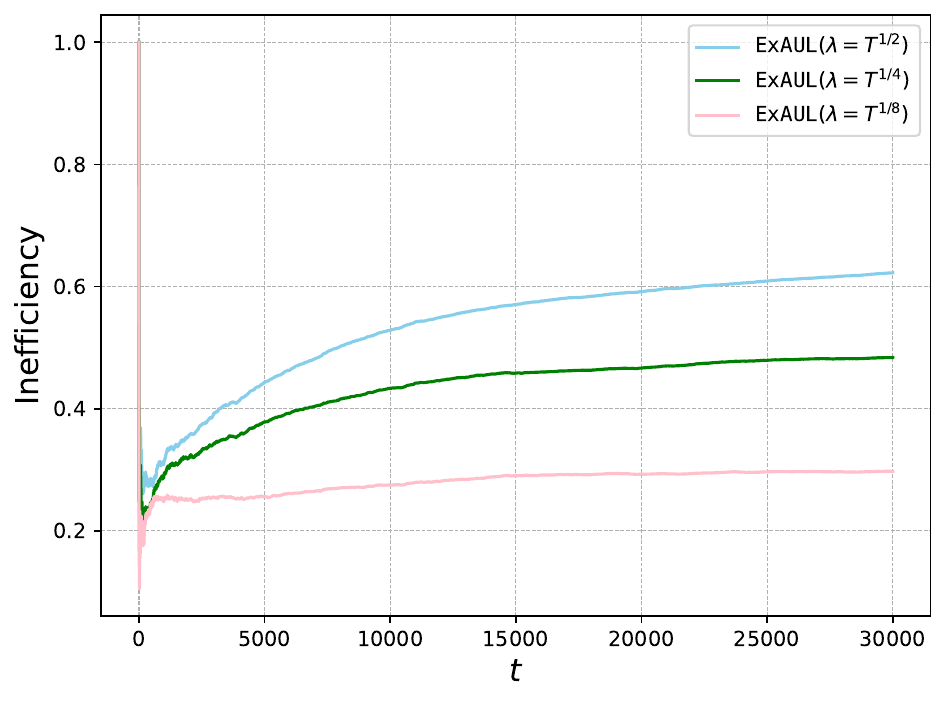}
  }
  \vspace{-1ex}
  \caption{Additional $\Ine_t$ comparisons of conformal abstention methods under stochastic environments with varying $\lambda$ for $T = 30\mathrm{K}$ datasets.
  }
  \label{fig:ineffoverlambda}
\end{figure*}

\clearpage
\subsection{Analysis on the Average Regret in Distribution-shift Environments}
\label{sec:analysisRegret}
\begin{figure*}[h]
    \centering
    \subfigure[Regret in single shift]{
    \includegraphics[width=0.31\textwidth]{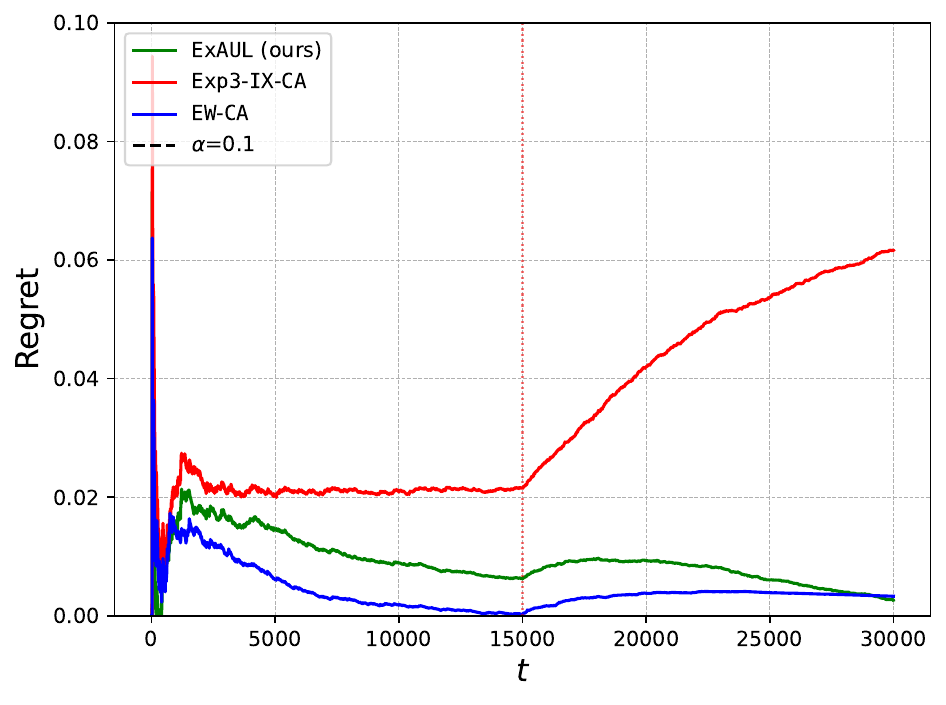}
    }
    \subfigure[Regret in alternating shift]{
    \includegraphics[width=0.31\textwidth]{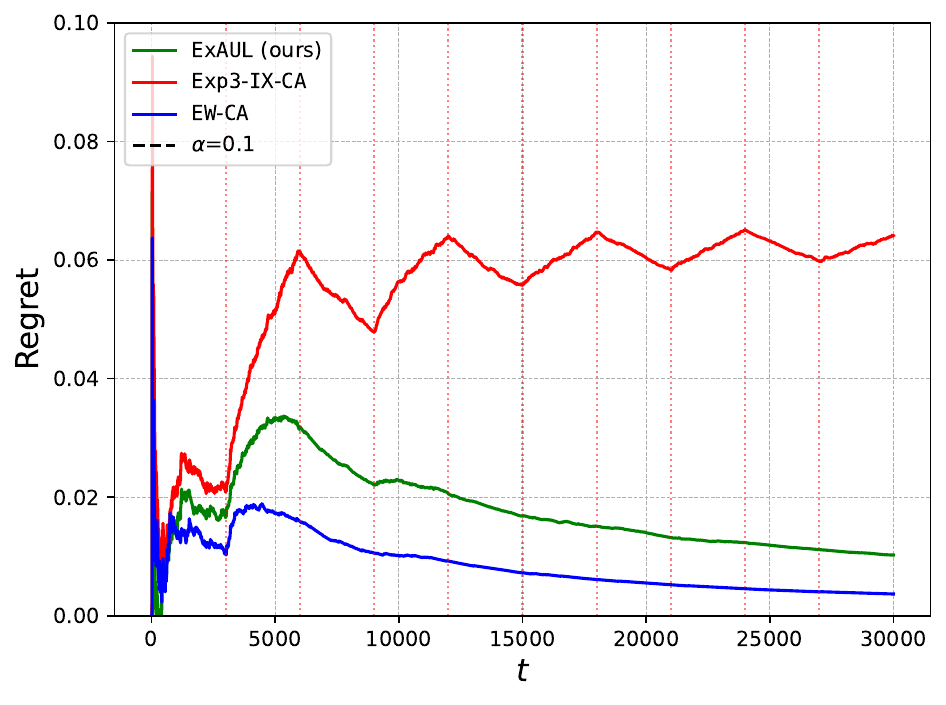}
    }
    \subfigure[Regret in gradual shift]{
    \includegraphics[width=0.31\textwidth]{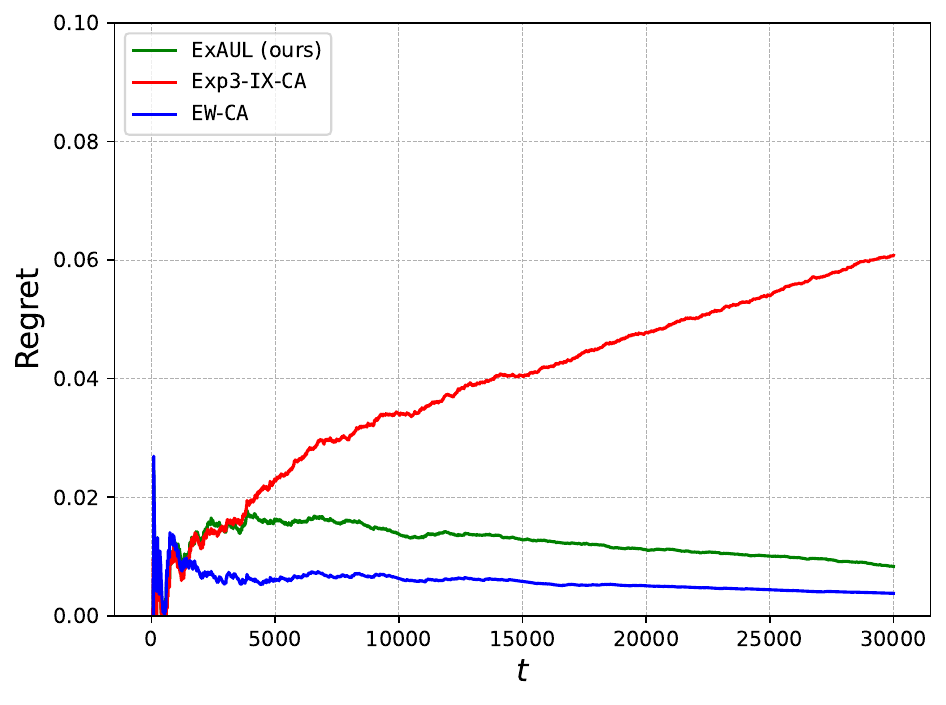}
    }
    \caption{Comparison of normalized regret trends during training. Similar to SR, regret also shows an increase when a distribution shift occurs. The experimental setups for (a), (b), and (c) correspond to those in Figure \ref{fig:shift:gpt-tri-simple}, \ref{fig:shift:gpt-tri-align}, and \ref{fig:shift:gpt-tri-linear}, respectively.}
    \label{fig-regret-analysis}
\end{figure*}

{\subsection{Analysis on Imperfect Feedback}
\label{sec:analysisImperfect}
\label{sec:noisyfeedback}}

While our framework provides a guarantee for the given feedback,
we examine for completeness how \ours behaves when the feedback is perturbed.
In all experiments below, \ours learns from the perturbed feedback, while we report the SR against the original labels from GPT-5.4-nano.

\begin{figure*}[h]
  \subfigure[TriviaQA dataset along with GPT-5.4]{
    \includegraphics[width=0.43\linewidth]{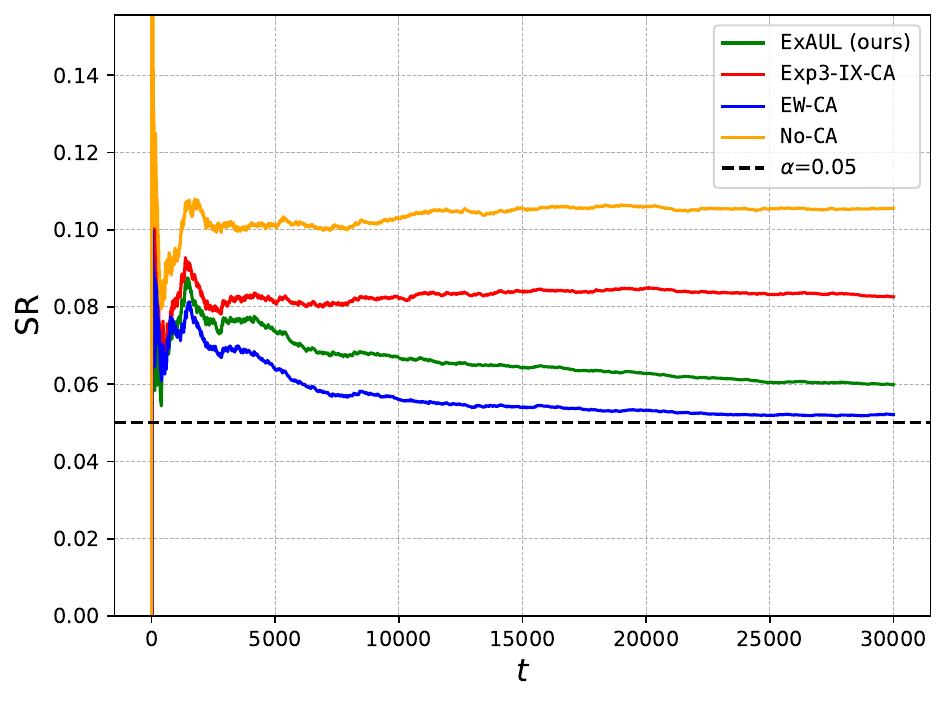}
  }
  \hfill
  \subfigure[TriviaQA dataset along with LLaMA3.1-8B]{
    \includegraphics[width=0.43\linewidth]{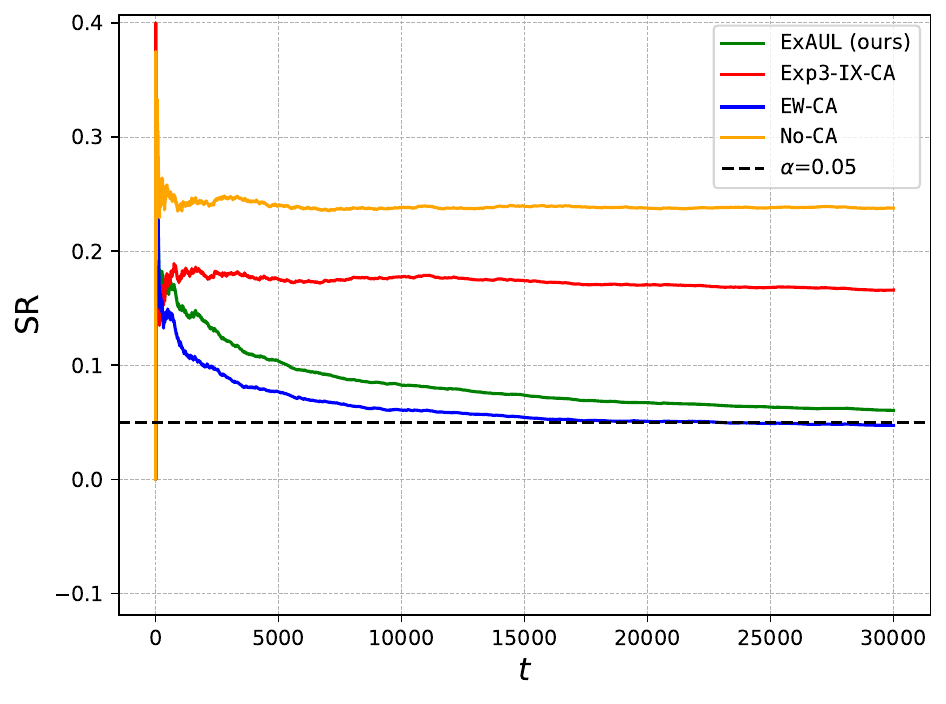}
  }
  \subfigure[NQ dataset along with GPT-5.4]{
    \includegraphics[width=0.43\linewidth]{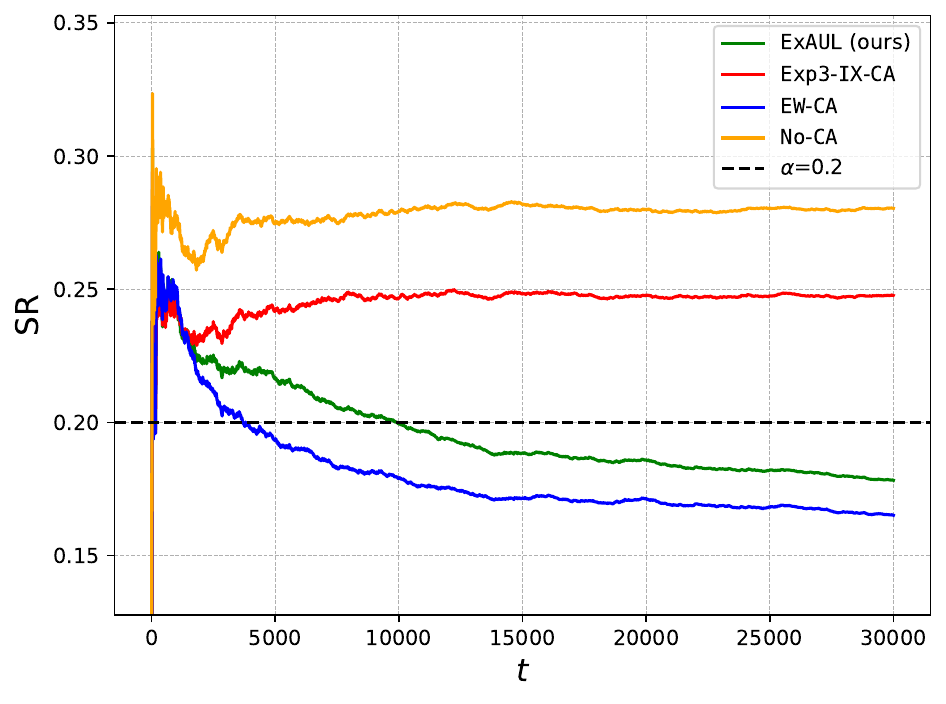}
  }
  \hfill
  \subfigure[NQ dataset along with LLaMA3.1-8B]{
    \includegraphics[width=0.43\linewidth]{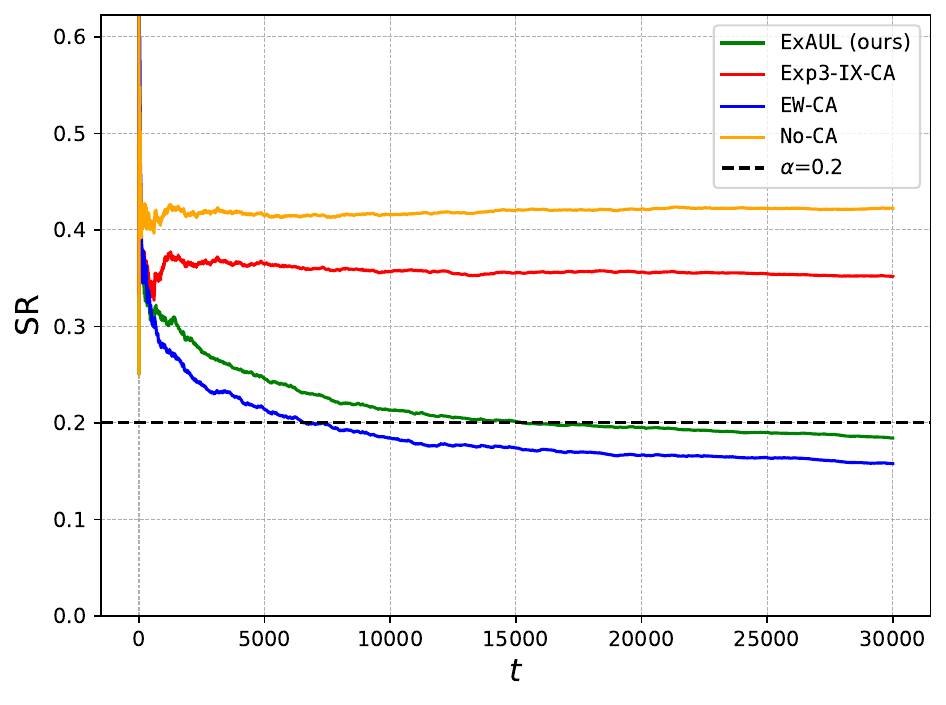}
  }
    \caption{Comparison of conformal abstention methods under an imperfect supervision environment on the NQ ($T=30K$, $\alpha=0.2$) and TriviaQA ($T=30K, \alpha=0.05$) along with GPT-5.4 and LLaMA3.1-8B, same as Figure 
    \ref{fig:stochastic:main:gpt-triviaqa}, \ref{fig:stochastic:main:llama3.1-8b-inst-triviaqa}, \ref{fig:stochastic:main:gpt-nq}, and \ref{fig:stochastic:main:llama3.1-8b-inst-nq}. Here, "noisy" feedback is generated by randomly flipping 5\% of the feedback labels.
    For a fair comparison, we use probability $0.05 / (2 * \texttt{base error rate})$ for incorrect labels and $0.05 / [2 (1 - \texttt{base error rate})]$ for correct ones. Similar to the ROUGE-L experiment (Figure \ref{fig:imperfect:rouge}), \ours demonstrates robustness, successfully learning to control the SR even under conditions of label noise.}
  \label{fig:imperfect:random5}
\end{figure*}

\begin{figure*}[h]
  \subfigure[TriviaQA dataset along with GPT-5.4]{
    \includegraphics[width=0.43\linewidth]{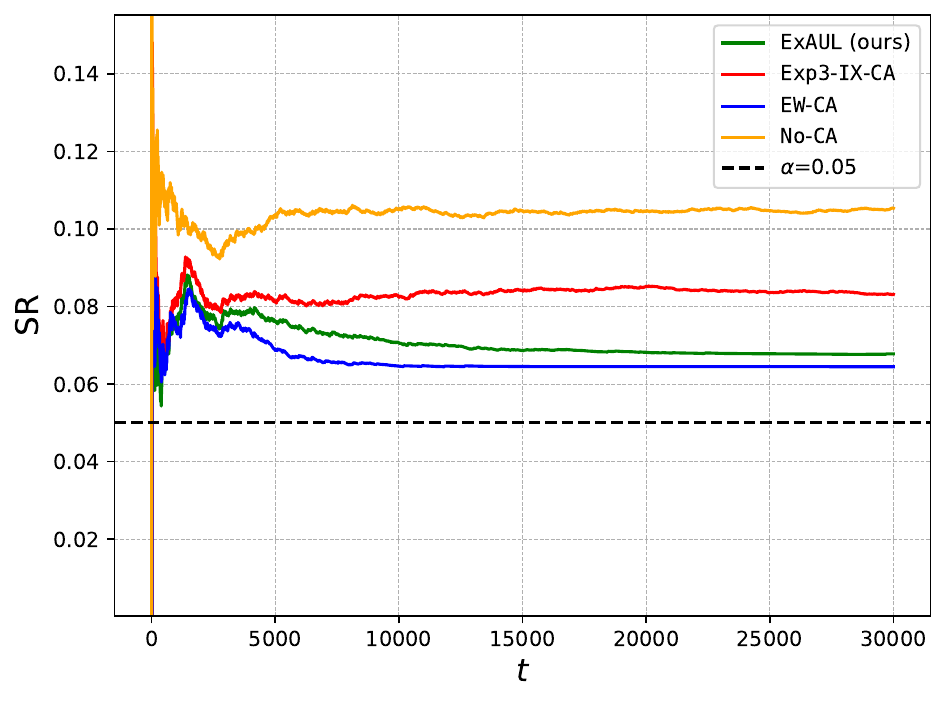}
  }
  \hfill
  \subfigure[TriviaQA dataset along with LLaMA3.1-8B]{
    \includegraphics[width=0.43\linewidth]{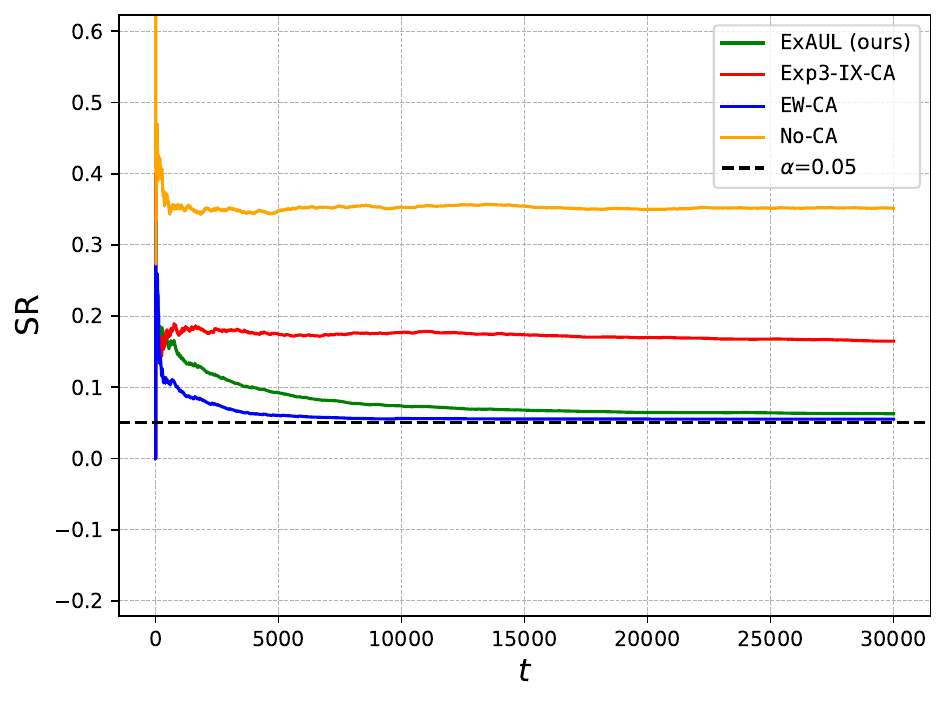}
  }
  \subfigure[NQ dataset along with GPT-5.4]{
    \includegraphics[width=0.43\linewidth]{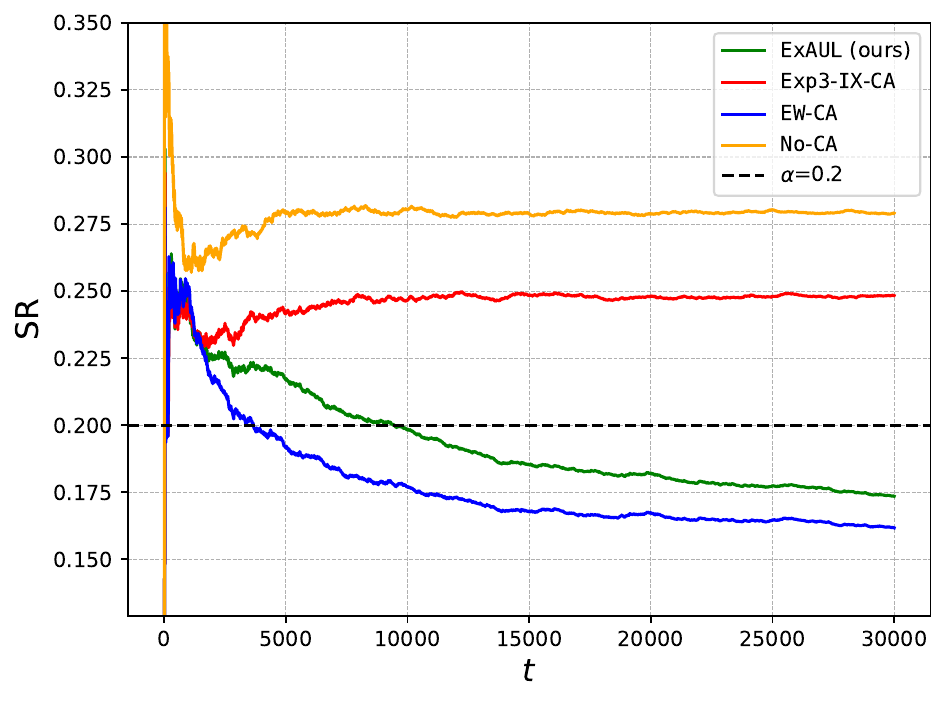}
  }
  \hfill
  \subfigure[NQ dataset along with LLaMA3.1-8B]{
    \includegraphics[width=0.43\linewidth]{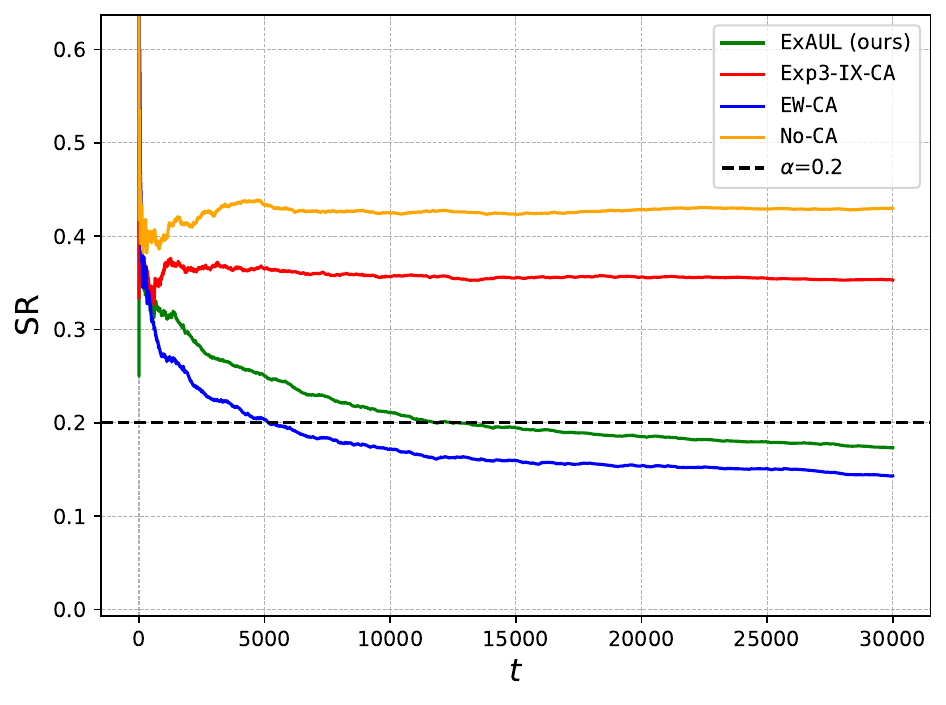}
  }
  \caption{Comparison of conformal abstention methods under an imperfect supervision environment on the NQ ($T=30K$, $\alpha=0.2$) and TriviaQA ($T=30K, \alpha=0.05$) along with GPT-5.4 and LLaMA3.1-8B, same as Figure  \ref{fig:stochastic:main:gpt-triviaqa}, \ref{fig:stochastic:main:llama3.1-8b-inst-triviaqa}, \ref{fig:stochastic:main:gpt-nq}, and \ref{fig:stochastic:main:llama3.1-8b-inst-nq}. Here, "noisy" feedback is simulated using thresholded ROUGE-L scores as the supervision signal.
  For a fair comparison, we use different thresholds for each (dataset, model) pair to maintain the original error rate.
  The plots demonstrate that \ours maintains its performance advantage, successfully controlling the SR near the target $\alpha$ despite the weaker, heuristic-based feedback.}
  \label{fig:imperfect:rouge}
\end{figure*}

\para{Weak supervision.}
We first replace the feedback with weaker proxies, namely thresholded ROUGE-L scores and randomly flipped labels (Figures~\ref{fig:imperfect:random5} and~\ref{fig:imperfect:rouge}).


\para{Label noise.}
We further flip the feedback with symmetric noise, false-negative noise (correct-to-incorrect flips), and false-positive noise (incorrect-to-correct flips) at rates of $5\%$, $10\%$, and $20\%$, on NQ with GPT-5.4 and Llama-3.1-8B ($T=30\mathrm{K}$, $\alpha=0.2$, $f_{\texttt{con}}$).

\begin{table}[h]
\centering
\caption{Empirical SR under imperfect feedback.}
\label{tab:imperfect_feedback}
\small
\setlength{\tabcolsep}{3.5pt}
\begin{tabular}{ll ccc ccc}
\toprule
& & \multicolumn{3}{c}{GPT-5.4} & \multicolumn{3}{c}{Llama-3.1-8B} \\
\cmidrule(lr){3-5} \cmidrule(lr){6-8}
Noise & Rate & \texttt{Exp3-IX-CA} & \textbf{\texttt{ExAUL}} & \texttt{EW-CA} & \texttt{Exp3-IX-CA} & \textbf{\texttt{ExAUL}} & \texttt{EW-CA} \\
\midrule
\multicolumn{2}{l}{\texttt{No-CA}} & \multicolumn{3}{c}{0.282} & \multicolumn{3}{c}{0.448} \\
\midrule
Clean & --   & 0.246 & \textbf{0.181} & 0.166 & 0.344 & \textbf{0.199} & 0.172 \\
\cmidrule(lr){1-8}
\multirow{3}{*}{Sym.}
      & 5\%  & 0.246 & \textbf{0.179} & 0.161 & 0.345 & \textbf{0.197} & 0.159 \\
      & 10\% & 0.246 & \textbf{0.191} & 0.184 & 0.345 & \textbf{0.193} & 0.151 \\
      & 20\% & 0.246 & \textbf{0.229} & 0.214 & 0.345 & \textbf{0.197} & 0.165 \\
\cmidrule(lr){1-8}
\multirow{3}{*}{FN}
      & 5\%  & 0.245 & \textbf{0.177} & 0.160 & 0.343 & \textbf{0.193} & 0.155 \\
      & 10\% & 0.245 & \textbf{0.199} & 0.195 & 0.342 & \textbf{0.187} & 0.146 \\
      & 20\% & 0.245 & \textbf{0.224} & 0.208 & 0.340 & \textbf{0.183} & 0.155 \\
\cmidrule(lr){1-8}
\multirow{3}{*}{FP}
      & 5\%  & 0.246 & \textbf{0.182} & 0.168 & 0.346 & \textbf{0.204} & 0.175 \\
      & 10\% & 0.246 & \textbf{0.185} & 0.172 & 0.348 & \textbf{0.208} & 0.177 \\
      & 20\% & 0.247 & \textbf{0.195} & 0.181 & 0.351 & \textbf{0.223} & 0.193 \\
\bottomrule
\end{tabular}
\end{table}

\para{Analysis.}
As shown in Table~\ref{tab:imperfect_feedback}, the SR of \ours exceeds $\alpha$ by at most $0.008$ for flip rates up to $10\%$ and by at most $0.029$ at $20\%$, and the most harmful noise type depends on the generator.
\texttt{EW-CA} attains a lower SR than \ours in every setting, while \texttt{Exp3-IX-CA} does not reach $\alpha$ in any setting.
Obtaining reliable feedback from noisy or biased signals, \eg by denoising or aggregating user feedback, is a separate research direction that is complementary to ours, since any improved feedback can be directly used by \ours.

\clearpage
\subsection{Analysis on scoring function}
\label{sec:analysisScore}
Our theoretical guarantee is formulated over a class of thresholds applied to a given confidence score and does not require the score to be calibrated or informative. To empirically evaluate the robustness of our method to changes in score quality, we consider two settings: switching between different scoring functions and corrupting confidence scores with random noise.

\paragraph{Robustness to Different Scoring Functions}
We first investigate whether our method remains effective when the scoring function changes during deployment. Specifically, we replace the original scoring function $f_{\texttt{con}}$ with $f_{\texttt{std}}$ at the midpoint of the stream ($t=T/2$), while keeping the generator, feedback, and evaluation labels unchanged.

\begin{figure*}[h]
  \subfigure[NQ dataset along with GPT-5.4]{
    \includegraphics[width=0.45\linewidth]{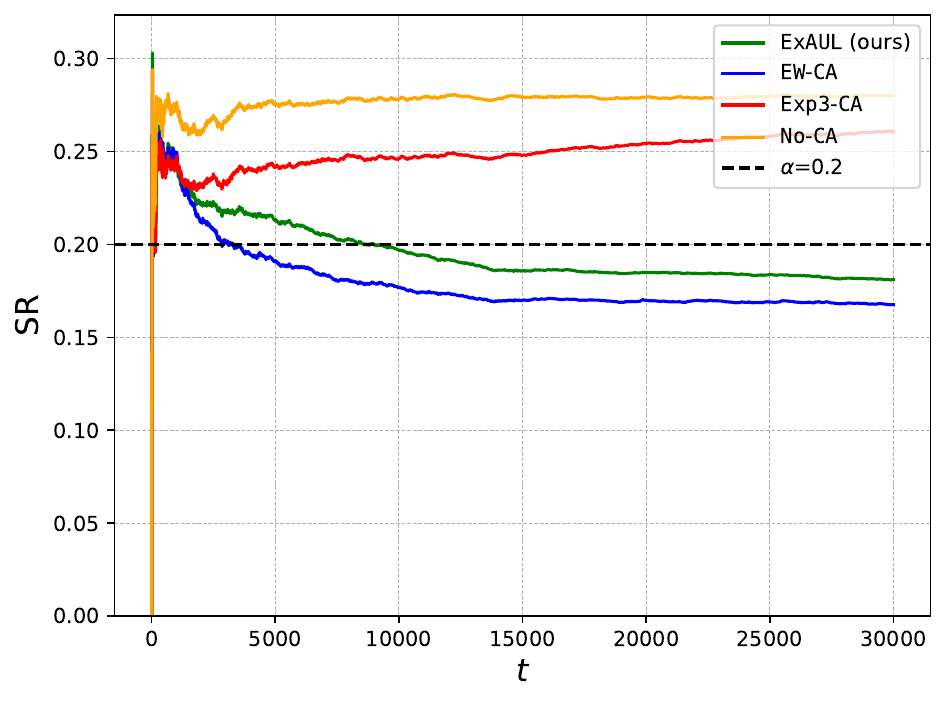}
  }
  \hfill
  \subfigure[NQ dataset along with LLaMA3.1-8B]{
    \includegraphics[width=0.45\linewidth]{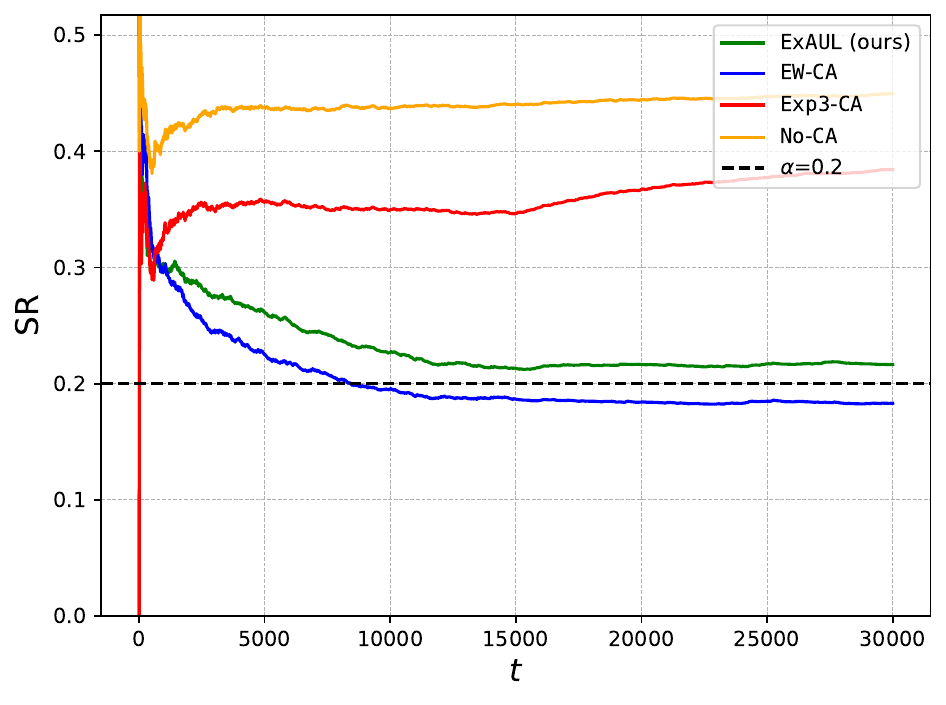}
  }
  \subfigure[TriviaQA dataset along with GPT-5.4]{
    \includegraphics[width=0.45\linewidth]{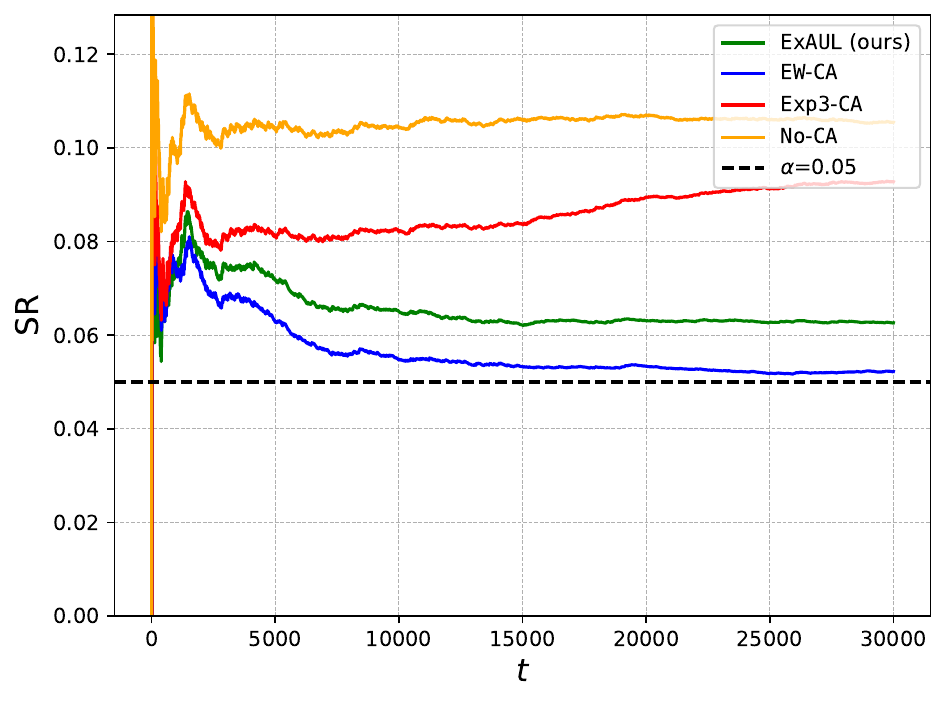}
  }
  \hfill
  \subfigure[TriviaQA dataset along with LLaMA3.1-8B]{
    \includegraphics[width=0.45\linewidth]{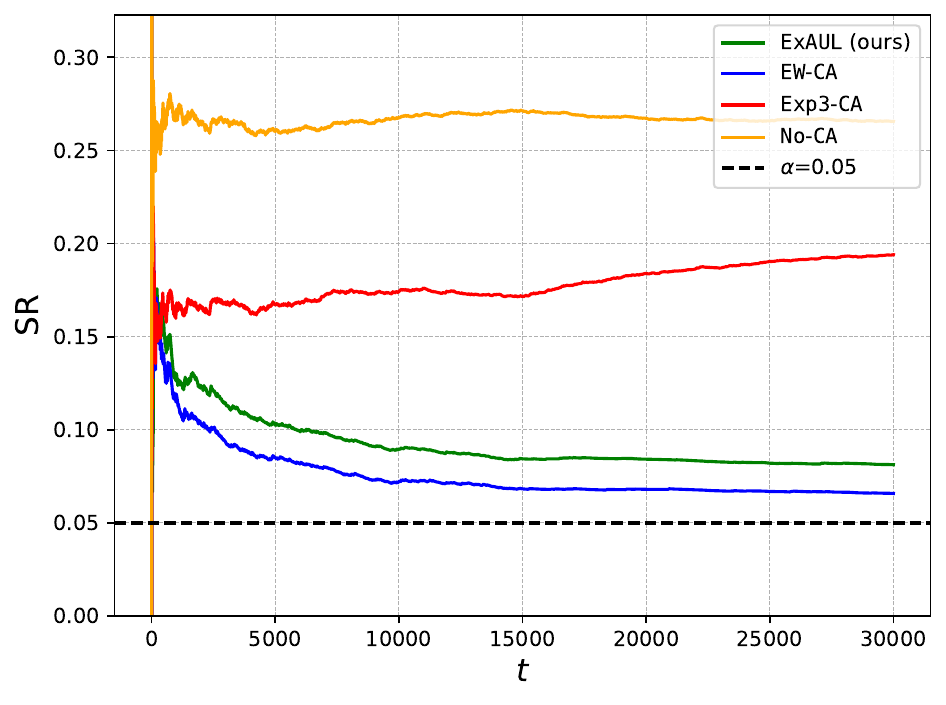}
  }
  \caption{Comparison of $\SR_t$ across conformal abstention methods when the scoring function changes from $f_{\texttt{con}}$ to $f_{\texttt{std}}$ over $T = 30\mathrm{K}$ datasets.
  }
  \label{fig:score-switch}
\end{figure*}

\begin{figure*}[h]
  \subfigure[NQ dataset along with GPT-5.4]{
    \includegraphics[width=0.45\linewidth]{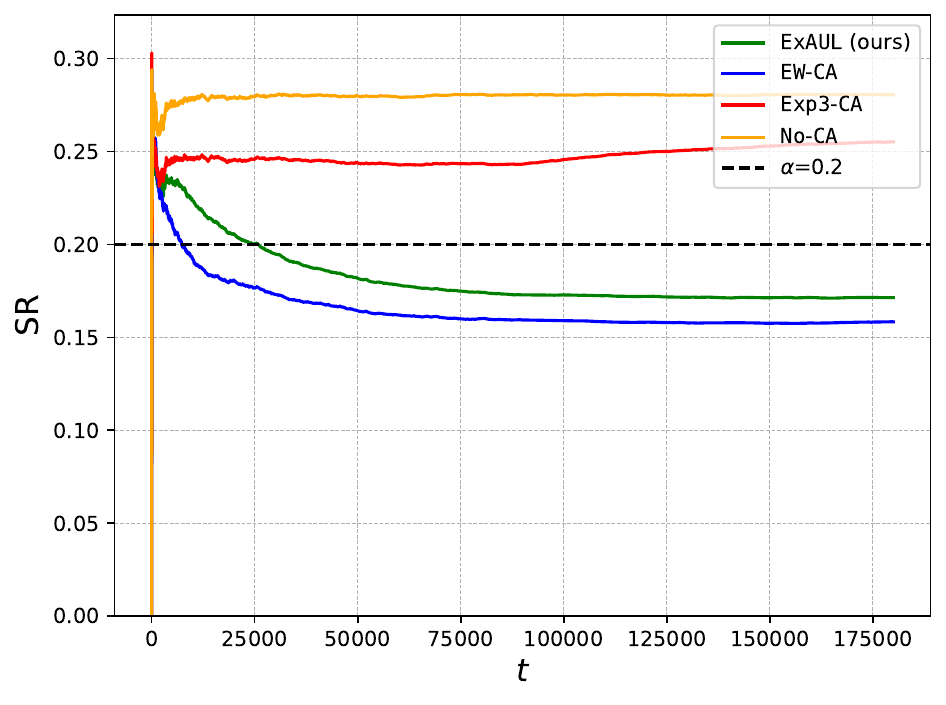}
  }
  \hfill
  \subfigure[NQ dataset along with LLaMA3.1-8B]{
    \includegraphics[width=0.45\linewidth]{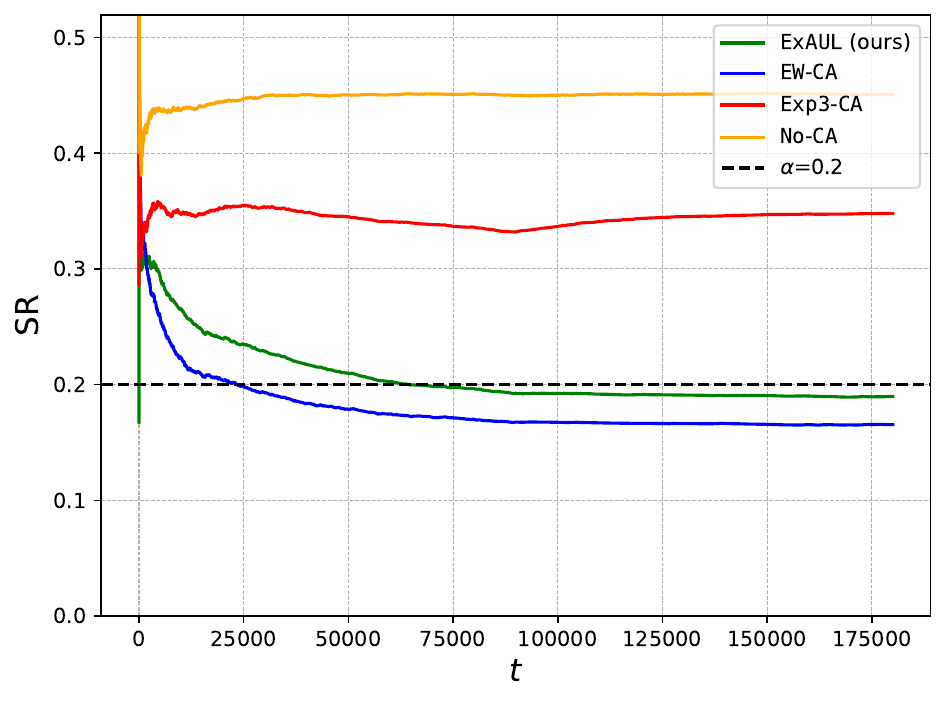}
  }
  \subfigure[TriviaQA dataset along with GPT-5.4]{
    \includegraphics[width=0.45\linewidth]{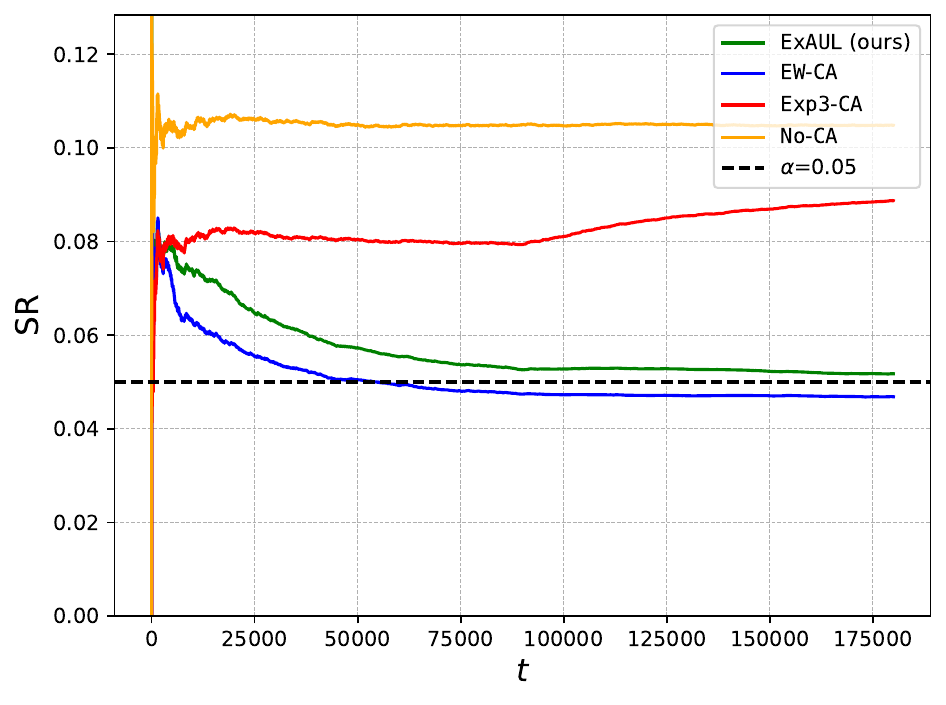}
  }
  \hfill
  \subfigure[TriviaQA dataset along with LLaMA3.1-8B]{
    \includegraphics[width=0.45\linewidth]{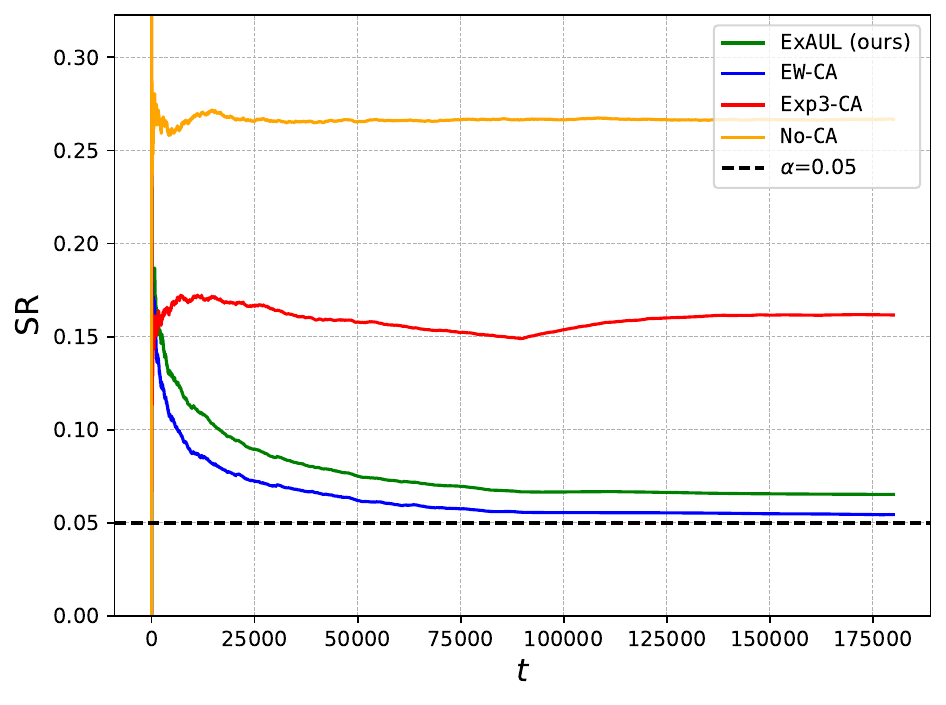}
  }
  \vspace{-1ex}
  \caption{Comparison of $\SR_t$ across conformal abstention methods when the scoring function changes from $f_{\texttt{con}}$ to $f_{\texttt{std}}$ over $T = 180\mathrm{K}$ datasets.
  }
  \label{fig:score-switch-long}
\end{figure*}

As shown in Fig.~\ref{fig:score-switch}, $\texttt{ExAUL}$ maintains its SR relatively close to the target level $\alpha$ across different datasets and generators, despite the change in scoring function. Although the empirical SR does not always remain below $\alpha$ for $T=30\mathrm{K}$, Fig.~\ref{fig:score-switch-long} shows that extending the horizon to $T=180\mathrm{K}$ further reduces the SR, bringing it closer to the target level. These results suggest that $\texttt{ExAUL}$ can adapt to changes in the scoring function and progressively approach the target risk level over longer horizons.

\paragraph{Robustness to Uncalibrated Confidence Scores}
We further investigate robustness to uninformative confidence scores by randomly corrupting the original scores. At each round, independently with probability $\varepsilon$, the original confidence score is replaced by a random value $u_t\sim\mathcal{U}(0,1)$, independent of the correctness of the generated answer:

\begin{equation*}
  \tilde{s}_t =
  \begin{cases}
    u_t, & \text{with probability } \varepsilon,\\
    f_{\texttt{con}}(x_t), & \text{otherwise.}
  \end{cases}
\end{equation*}
Here, $\varepsilon=0$ recovers the original score stream, while $\varepsilon=1$ makes the scores entirely non-informative. We set $\varepsilon=0.5$ while keeping the generator, feedback, and evaluation labels unchanged.

\begin{figure*}[h]
    \centering
    \subfigure[SR over time]{
    \includegraphics[width=0.45\textwidth]{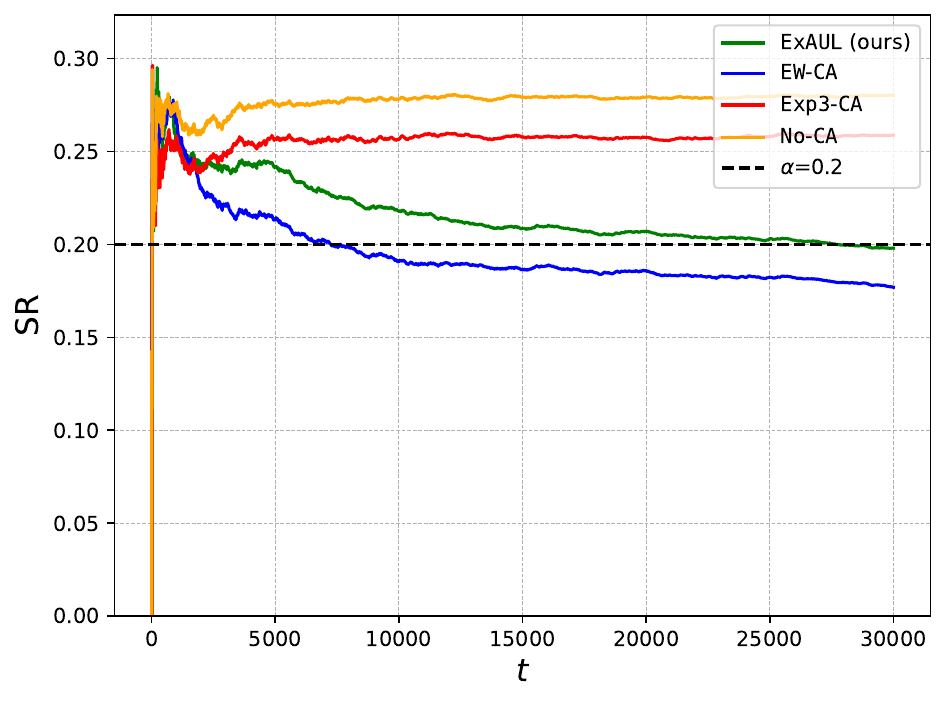}
    }
    \subfigure[Inefficiency over time]{
    \includegraphics[width=0.45\textwidth]{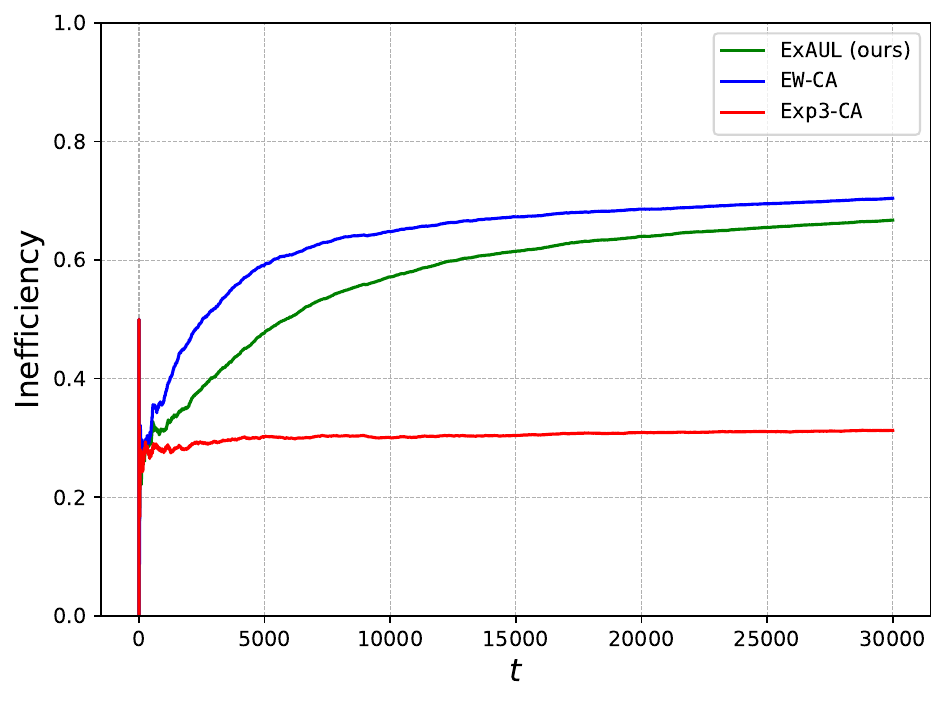}
    }
    \vspace{-1ex}
    \caption{Robustness to confidence-score corruption on NQ ($T=30\mathrm{K}, {\alpha=0.20}$) with GPT-5.4.
    }
    \label{fig:score-robust:gpt-nq}
\end{figure*}

As shown in Fig.~\ref{fig:score-robust:gpt-nq}, $\texttt{ExAUL}$ maintains its SR near the target level $\alpha$ despite the corruption of confidence scores. Moreover, compared with the original setting using only $f_{\texttt{con}}$ in Fig.~\ref{fig:stochastic:main:gpt-nq}, we observe a reduction in efficiency, indicating a shift in the risk--efficiency trade-off. These results suggest that $\texttt{ExAUL}$ can remain effective even when the confidence scores become less informative.

\subsection{Long-Form Generation}
\label{sec:longform}
Recent long-form methods decompose a response into atomic claims and estimate confidence or verify correctness per claim~\citep{min2023factscore,wei2024long,yuan2025fact,zhang2025atomic,goren2026should}.
Given such claim-level scores and feedback, \ours applies without modification by treating each claim as one online round; independence between claims is not required, since our protocol permits an adaptive adversary (Section~\ref{sec:problem}).

\para{Setup.}
On ASQA~\citep{stelmakh2022asqa}, GPT-5.4-nano generates 3--5 sentence responses, extracts atomic claims, and judges their factuality, and the score is $P(\mathrm{True})$ from the first-token probabilities of ``True'' and ``False''~\citep{yuan2025fact}.
We exclude claims judged unverifiable, since absence from the reference does not imply incorrectness, which yields $3{,}528$ claims from $987$ responses.
We set $e_t = \mathbbm{1}(\text{claim is contradicted})$, $\alpha=0.3$, and $|\Hs|=1\mathrm{K}$, with five epochs over the claim stream ($T\approx18\mathrm{K}$).

\begin{table}[t]
\centering
\caption{Claim-level SR and inefficiency on ASQA with GPT-5.4-nano ($\alpha=0.3$), as mean $\pm$ standard deviation over five learner seeds.}
\label{tab:longform}
\begin{tabular}{lcc}
\toprule
Method & SR & Inefficiency \\
\midrule
\texttt{No-CA} & $0.4776$ & $0.0000$ \\
\texttt{Exp3-IX-CA} & $0.3564 \pm 0.0020$ & $0.4148 \pm 0.0026$ \\
\textbf{\oursns} & $\mathbf{0.2535 \pm 0.0043}$ & $0.6816 \pm 0.0125$ \\
\texttt{EW-CA} & $0.2110 \pm 0.0018$ & $0.7375 \pm 0.0020$ \\
\bottomrule
\end{tabular}
\end{table}

\para{Results.}
As shown in Table~\ref{tab:longform}, \ours keeps the claim-level SR below $\alpha$ while retaining $31.8\%$ of the claims, whereas \texttt{Exp3-IX-CA} does not reach the target, and \texttt{EW-CA} with full feedback abstains even more often.
This check only shows that \ours operates at the claim level; a complete long-form system would also require reliable claim extraction, verification, and reconstruction, which we leave to future work.

\section{Theorem on Our Main Regret Bound}

\begin{theorem} \label{thm:ours:regretbound}
    Let $\ell_t(\cdot) \in [0, 1]$ of the form (\ref{eq:loss-sg}).
    For any $T \in \naturalnum$, $\alpha \in (0, 1)$, and finite hypotheses $\Hs$,
    Algorithm \ref{alg:exp3-sg-unlocking} with $\eta = 2\gamma = \sqrt{\frac{\ln|\mathcal{H}|}{T}}$ provides the following regret bound
    with probability at least $1-\delta$:
    \begin{equation*}
        \Reg_T
        \le
        4\sqrt{T\ln|\mathcal H|} + \( 1 +  \sqrt{\frac{T}{\ln|\Hs|}}\)\ln\frac{2}{\delta}.
    \end{equation*}
    
\end{theorem}

\clearpage
\section{Proofs}

\subsection{A Proof of Theorem \ref{thm:ew:regretbound}}
\label{proof:thm:ew:regretbound} 

We follow the standard proof \citep{foster2023foundations}, followed by extending it for a loss range with $[0, \ell_\text{max}]$.

Let the cumulative loss of the $i$-th hypothesis $\tau_i \in \Hs$ be $L_t^i \coloneqq \sum_{s=1}^t \ell_s (\tau_i)$,
$W_t \coloneqq \sum_{i=1}^{|\Hs|} \exp\left\{ -\eta L_t^i \right\}$.


\para{The upper bound of $\ln\frac{W_t}{W_{t-1}}$.}
We have the following upper bound of $\ln\frac{W_t}{W_{t-1}}$:
\begin{align}
    \ln\frac{W_t}{W_{t-1}}
    &= \ln\frac
    {\sum_{i=1}^{|\Hs|} \exp\left\{-\eta L_t^i \right\}
    }
    {\sum_{i=1}^{|\Hs|} \exp\{-\eta L_{t-1}^i\}} \nonumber\\
    &= \ln
    \sum_{i=1}^{|\Hs|} \exp\left\{-\eta \ell_t(\tau_i) \right\}
    \frac{\exp\{-\eta L_{t-1}^i\}}{\sum_{i=1}^{|\Hs|} \exp\{-\eta L_{t-1}^i\}} \nonumber\\
    &= \ln \Exp_{\tau_t \sim p_t} \exp\left\{-\eta \ell_t(\tau_t) \right\} \nonumber\\
    &\le -\eta \Exp_{\tau_t \sim p_t} \ell_t(\tau_t) 
    + \frac{\eta^2 \ell_\text{max}^2}{8},
    \label{proof:ew:hoeffding}
\end{align}
where (\ref{proof:ew:hoeffding}) holds by the Hoeffding's lemma and the fact that $\ell_t(\tau_i) \in [0, \ell_\text{max}]$.
Thus, we have
\begin{equation*}
    \ln\frac{W_T}{W_{0}}
    = \sum_{t=1}^T \ln \frac{W_t}{W_{t-1}}
    \le -\eta \sum_{t=1}^T \Exp_{\tau_t \sim p_t} \ell_t(\tau_t) + \frac{\eta^2 \ell_\text{max}^2 T}{8}.
\end{equation*}

\para{The lower bound of $\ln\frac{W_t}{W_{t-1}}$.}
We have the lower bound of $\ln\frac{W_t}{W_{t-1}}$ as follows:
\begin{align*}
    \ln{\frac {W_T}{W_0}}
    = \ln \sum_{i=1}^{|\Hs|} \exp\{-\eta L_T^i\} - \ln |\Hs|
    &\ge \ln \left( \max_{i \in \{1, \dots, |\Hs|\}} \exp\{-\eta L^i_T\} \right) - \ln|\Hs| \\
    &= -\eta \min_{i \in \{1, \dots, |\Hs|\}} L^i_T - \ln|\Hs| \\
    &= -\eta \min_{\tau \in \Hs} \sum_{t=1}^T \ell_t(\tau)
    - \ln|\Hs|.
\end{align*}

\para{Combine the lower and upper bounds.}
If we combine the lower bound and the upper bound, we obtain the following inequality:
\begin{equation*} \label{eq:combine}
    \sum_{t=1}^T \Exp_{\tau_t \sim p_t} \ell_t(\tau_t)
    - \min_{\tau \in \Hs} \sum_{t=1}^T \ell_t(\tau)
    \le \frac{\eta \ell_\text{max}^2T} {8} + \frac{\ln |\Hs|} \eta.
\end{equation*}
{
If $\eta = \sqrt{8\ln|\Hs|/\ell_\text{max}^2T}$, 
\begin{align}
    \sum_{t=1}^T \Exp_{\tau_t \sim p_t} \ell_t(\tau_t)
    - \min_{\tau \in \Hs} \sum_{t=1}^T \ell_t(\tau)
    &\le
    \ell_\text{max}\sqrt{\frac{T\ln|\Hs|}{2}}.
    \label{eq:EW:org-bound}
\end{align}
}


{Since the guarantee in (\ref{eq:EW:org-bound}) holds for any sequence of loss functions in \ewns, the bound is also valid against an adaptive adversary. Furthermore, the expected regret is given by $\Exp\[\Reg_T\] = \Exp\left[\sum_{t=1}^T \ell_t(\tau_t) - \min_{\tau \in \Hs} \sum_{t=1}^T \ell_t(\tau)\right],$ which follows from the tower property of expectation.}

\para{High probability bound.}
Moreover, we have
\begin{align}
    \sum_{t=1}^T \ell_t(\tau_t) - \min_{\tau \in \Hs} \sum_{t=1}^T \ell_t(\tau)
    &=
    \( \sum_{t=1}^T \ell_t(\tau_t) - \sum_{t=1}^T \Exp_{\tau_t \sim p_t} \ell_t(\tau_t)\)
    + \(\sum_{t=1}^T \Exp_{\tau_t \sim p_t} \ell_t(\tau_t) - \min_{\tau \in \Hs} \sum_{t=1}^T \ell_t(\tau)\).
    \nonumber
\end{align}

Since $\Exp_t \[\ell_t(\tau_t) - \Exp_{\tau_t \sim p_t} \ell_t(\tau_t)\] = 0$, which implies that $\sum_{t=1}^T \[\ell_t(\tau_t) - \Exp_{\tau_t \sim p_t} \ell_t(\tau_t)\]$ is a martingale,
we can use Hoeffding-Azuma inequality
\begin{align*}
    \Prob \(\sum_{t=1}^T \[\ell_t(\tau_t) - \Exp_{\tau_t \sim p_t} \ell_t(\tau_t)\] \ge \ep \) \le \exp\(-\frac{2\ep^2}{\sum_{t=1}^T c_t^2}\)
\end{align*}
for all $\ep\ge0$ where $c_t = \ell_\text{max}$ since $\ell_t(\tau_t) - \Exp_{\tau_t \sim p_t} \ell_t(\tau_t) \in [-\Exp_{\tau_t \sim p_t} \ell_t(\tau_t), \ell_\text{max} - \Exp_{\tau_t \sim p_t} \ell_t(\tau_t)]$.

Thus, if we put $\delta = \exp \(-\frac{2\ep^2}{T\ell_\text{max}^2}\)$ we get
\begin{equation*}
    \sum_{t=1}^T \ell_t(\tau_t) - \sum_{t=1}^T \Exp_{\tau_t \sim p_t} \ell_t(\tau_t)
    \le \ell_\text{max}\sqrt{\frac{T \ln \delta^{-1}}{2}},
\end{equation*}
with probability at least $1-\delta$,
which suggests that
\begin{align}
    \sum_{t=1}^T \ell_t(\tau_t) - \min_{\tau \in \Hs} \sum_{t=1}^T \ell_t(\tau)
    &\le \frac{\eta \ell_\text{max}^2T} {8} + \frac{\ln |\Hs|} {\eta}
    + \ell_\text{max}\sqrt{\frac{T \ln \delta^{-1}}{2}}
    \nonumber
\end{align}
with probability at least $1-\delta$.

{
If $\eta = \sqrt{8\ln|\Hs|/\ell_\text{max}^2T}$, 
\begin{align}
    \sum_{t=1}^T \ell_t(\tau_t) - \min_{\tau \in \Hs} \sum_{t=1}^T \ell_t(\tau)
    &\le
    \ell_\text{max}\sqrt{\frac{T\ln|\Hs|}{2}} + \ell_\text{max}\sqrt{\frac{T\ln\delta^{-1}}{2}}.
    \label{eq:EW:hprob}
\end{align}
}

\subsection{A Proof of Theorem \ref{thm:exp3ix:regretbound}}
\label{proof:thm:exp3ix:regretbound} 

This proof follows standard proof \citep{neu2015explore} but with a non-trivial loss range $[0,\ell_\text{max}]$.
Recall that $\ell_t(\tau)\in[0, \ell_\text{max}]$ and its biased estimator $\ell_t(\tau\mid\{\tau_t\}) \in [0,\infty)$, as defined in (\ref{eq:exp3ix:unbiasedestimator}).

We first introduce new Lemma \ref{lem:exp3} and Corollary \ref{cor:exp3} that establish a high probability bound between the loss estimator and the empirical loss. Corollary \ref{cor:exp3} provides a bound for a fixed value of $\tau$, and its proof follows directly from the Lemma \ref{lem:exp3} by applying a union bound. Together, Lemma \ref{lem:exp3} and Corollary \ref{cor:exp3} allow us to translate the upper bound on the estimator equation into a high-probability regret bound. 
\begin{lemma}
Let $\gamma_t \ge 0$ for $t \in \{1, 2, \dots, T\}$ be a fixed non-increasing sequence and $\alpha_{t, \tau}$ be a real value that satisfies $\alpha_{t, \tau} \le  2\gamma_t$ for all $t \in \naturalnum$ and $\tau \in \Hs$. Then, we have
\begin{equation}
    \sum_{t=1}^{T}  \sum_{\tau\in\Hs} \alpha_{t, \tau} \(\ell_t(\tau \mid \{\tau_t\}) - \ell_t(\tau)\)
    \le \ell_\text{max} \ln\frac{1}{\delta}, \nonumber
\end{equation}
\label{lem:exp3}
with probability $1-\delta$.
\end{lemma}

\begin{proof}
First, letting $\beta_t = \frac{2\gamma_t}{\ell_\text{max}}$, we have 
\begin{align}
    \ell _t(\tau \mid \{\tau_t\})
    &=\frac{\ell_t(\tau)}{p_t(\tau) +  \gamma_t} \cdot \mathbbm{1}(\tau_t = \tau)
    \nonumber \\
    &= \ell_\text{max} \cdot \frac{\ell_t(\tau)/\ell_\text{max}}{p_t(\tau) +  \gamma_t} \cdot \mathbbm{1}(\tau_t = \tau)
    \nonumber \\
    &\le \ell_\text{max} \cdot \frac{\ell_t(\tau)/\ell_\text{max}}{p_t(\tau) +  \gamma_t \cdot \ell_t(\tau)/\ell_\text{max}} \cdot \mathbbm{1}(\tau_t = \tau)
    \nonumber \\
    &= \frac{\ell_\text{max}}{2\gamma_t} \cdot \frac{2\gamma_t\cdot\frac{\ell_t(\tau)}{\ell_\text{max}p_t(\tau)}}{1 +  \gamma_t\cdot\frac{\ell_t(\tau)}{\ell_\text{max}p_t(\tau)}}\mathbbm{1}(\tau_t = \tau)
    \nonumber \\
    &\le \frac{\ell_\text{max}}{2\gamma_t} \cdot\ln\(1 + 2\gamma_t\cdot\frac{\ell_t(\tau)}{\ell_\text{max}p_t(\tau)}\mathbbm{1}(\tau_t = \tau)\)
    \label{eq:exp3:ln}
    \\
    &= \frac{1}{\beta_t}\cdot\ln\(1 + \beta_t\frac{\ell_t(\tau)}{p_t(\tau)}\mathbbm{1}(\tau_t = \tau)\)
    \label{eq:exp3:lemma1}
\end{align}
where the inequality (\ref{eq:exp3:ln}) holds by $\frac{x}{1+x/2} \le \ln(1+x)$ for $x\ge0$ and the range $\{0, 1\}$ of the indicator function.

Second, we explicitly define the following auxiliary quantities:
\begin{equation}
    \tilde{S}_t = \sum_{\tau\in\Hs} \frac{\alpha_{t, \tau}}{\ell_\text{max}} \ell _t(\tau \mid \{\tau_t\}) 
    \quad \text{and} \quad 
    S_t = \sum_{\tau\in\Hs} \frac{\alpha_{t, \tau}}{\ell_\text{max}} \ell_t(\tau).
    \nonumber
\end{equation}

Denoting by $\Exp_t$ the conditional expectation with respect to  $\tau_t$, given the past learner's choices $\tau_1, \ldots, \tau_{t-1}$, we have the following relation between $\tilde{S}_t$ and $S_t$:
\begin{align}
    \Exp_t\[\exp\tilde{S}_t\]
    &\le 
    \Exp_t\[\exp\(\sum_{\tau\in\Hs} \frac{\alpha_{t, \tau}}{\ell_\text{max}\beta_t} \cdot \ln\(1 + \beta_t\cdot\frac{\ell_t(\tau)}{p_t(\tau)}\mathbbm{1}(\tau_t = \tau)\)\)\]
    \label{eq:exp3:ln2}
    \\
    &= 
    \Exp_t\[\prod_{\tau\in\Hs}\(1 + \beta_t\frac{\ell_t(\tau)}{p_t(\tau)}\mathbbm{1}(\tau_t = \tau)\)^{\frac{\alpha_{t, \tau}}{\ell_\text{max}\beta_t}}\]
    \nonumber \\
    &\le
    \Exp_t\[1 + \sum_{\tau\in\Hs}\frac{\alpha_{t, \tau}}{\ell_\text{max}}\cdot\frac{\ell_t(\tau)}{p_t(\tau)}\cdot\mathbbm{1}(\tau_t = \tau)\]
    \label{eq:exp3:ber}\\
    &\le
    1 + \sum_{\tau\in\Hs}\frac{\alpha_{t, \tau}}{\ell_\text{max}}\cdot\ell_t(\tau)
    \nonumber 
    \\
    &\le
    \exp\(\sum_{\tau\in\Hs}\frac{\alpha_{t, \tau}}{\ell_\text{max}}\cdot\ell_t(\tau)\) 
    = \exp\(S_t\).
    \label{eq:exp3:exp}
\end{align}
Here, the inequality (\ref{eq:exp3:ln2}) follows directly from (\ref{eq:exp3:lemma1}). Subsequently, (\ref{eq:exp3:ber}) is obtained by applying the Bernoulli inequality $(1+z)^a \le (1+az)$ which is valid for $z\ge0$ and $0\le a \le1$ and the condition of $\alpha_t \le 2\gamma_t$, followed by the property of the indicator function.

Finally we define:
\begin{align}
    W_t := \exp\left(\sum_{i=1}^{t} \big(\tilde S_i - S_i\big)\right). \nonumber
\end{align}
This immediately implies that $W_t = W_{t-1} \cdot \exp\(\tilde S_t - S_t\)$. 
Taking the conditional expectation with respect to $\tau_t$, we obtain
\begin{align}
    \Exp_t[W_t]  
    & = \Exp_t \[ W_{t-1} \cdot \exp\(\tilde S_t - S_t\) \]  
    \nonumber \\
    &= W_{t-1} \Exp_t \[ \exp\(\tilde S_t - S_t\) \] 
    \nonumber \\
    &=W_{t-1}\exp\( - S_t\) \Exp_t \[\exp\(\tilde S_t\)\]
    \label{eq:exp3:martingale} \\
    &\le W_{t-1}\exp\( - S_t\) \exp\( S_t \)
    \label{eq:exp3:eses}
    \\
    &\le W_{t-1},
    \label{eq:exp3:mart}
\end{align}
where (\ref{eq:exp3:martingale}) $W_{t-1}$ and $S_t$ are dependent on the history up to time $t-1$, we pull $W_{t-1}$ and $\exp(-S_t)$ outside of the conditional expectation.
Also, (\ref{eq:exp3:eses}) holds due to (\ref{eq:exp3:exp}).

From 
(\ref{eq:exp3:mart}), 
we have $\Exp_T[W_T] \le \Exp_{T-1} [W_{T-1}] \le \dots \le \Exp_{0} [W_{0}]$.
As $W_0 = 1$, we have $\Exp_T [W_T] \le 1$. 

Then, by this and the Markov’s inequality, we have
\begin{align}
    \Prob\left( \sum_{t=1}^T (\tilde S_t - S_t)>
    \ln \frac{1}{\delta} \right) 
    =
    \Prob\left( W_T >
    \frac{1}{\delta} \right) 
    \le
    \delta \Exp_T[W_T]
    \le \delta. \nonumber
\end{align}


Recall that $\ln W_T = \sum_{t=1}^T\(\tilde S_t - S_t\) = \sum_{t=1}^T\sum_{\tau\in\Hs}\frac{\alpha_{t, \tau}}{\ell_\text{max}}\(\ell_t(\tau\mid\{\tau_t\})-\ell_t(\tau)\)$ then we have
\begin{align}
    \sum_{t=1}^T \sum_{\tau\in\Hs} \alpha_{t, \tau} \(\ell_t(\tau\mid\{\tau_t\})-\ell_t(\tau)\) \le \ell_\text{max}\ln \frac{1}{\delta} \nonumber
\end{align}
with probability at least $1-\delta$, which completes the proof.
\end{proof}
\begin{corollary}
Let $\gamma_t \ge \gamma > 0$ for all $t$.
Simultaneously for all $\tau \in \Hs$, we have
\begin{equation}
    \sum_{t=1}^T\ell_t(\tau\mid\{\tau_t\}) - \sum_{t=1}^T\ell_t(\tau) \le \frac{\ell_\text{max}}{2\gamma}\ln \frac{|\Hs|}{\delta}
\end{equation}
\label{cor:exp3}
with probability $1-\delta$.
\end{corollary}
\begin{proof}
For each $\bar\tau \in \Hs$, let $\alpha_{t, \tau} \coloneqq 2 \gamma \mathbbm{1}(\tau = \bar\tau) \le 2\gamma_t$, which satisfies the condition for Lemma \ref{lem:exp3}. Then, due to Lemma \ref{lem:exp3}, we have
\begin{align*}
    \ell_\text{max}\ln \frac{|\Hs|}{\delta}
    &\ge
    \sum_{t=1}^T \sum_{\tau\in\Hs} \alpha_{t, \tau} \(\ell_t(\tau\mid\{\tau_t\})-\ell_t(\tau)\) 
    \\
    &=
    \sum_{t=1}^T \sum_{\tau\in\Hs} 2 \gamma \mathbbm{1}(\tau = \bar\tau) \(\ell_t(\tau\mid\{\tau_t\})-\ell_t(\tau)\) 
    \\
    &=
    \sum_{t=1}^T 2 \gamma  \(\ell_t(\bar\tau\mid\{\tau_t\})-\ell_t(\bar\tau)\) 
\end{align*}
with probability at least $1 - \frac{\delta}{|\Hs|}$.
By taking the union bound for all $\bar\tau \in \Hs$, we complete the proof. 
\end{proof}

Then, we provide the main proof. 
\paragraph{First step.}
We split the expected loss estimator into two logarithmic terms Eq.~(3.7) in \citep{bubeck2012regret}: a variability term and the log of an exponential expectation. We will rewrite each term in the following steps.
\begin{multline}
    \Exp_{\tau \sim p_t}\ell(\tau\mid\{\tau_t\})
    = \frac{1}{\eta_t}\underbrace{\ln\Exp_{\tau \sim p_t}\exp\Big(-\eta_t\Big( \ell_t(\tau\mid\{\tau_t\})-\Exp_{\bar\tau \sim p_t}\ell_t(\bar{\tau}\mid \{\tau_t\})\Big)\Big)}_{\text{variability term}} \\ 
    - \frac{1}{\eta_t}\underbrace{ \ln\Exp_{\tau \sim p_t}\exp\Big(-\eta_t\ell_t(\tau\mid\{\tau_t\})\Big)}_{\text{log of an exponential expectation term}}.
    \label{eq:exp3:var-decomp}
\end{multline}


\paragraph{Second step.}
We provide an upper bound on the variability term.
\begin{align}
    \ln\Exp_\tau\exp \Big(-\eta_t & \(\ell_t(\tau\mid\{\tau_t\})  -\Exp_{\bar\tau}\ell_t(\bar{\tau}\mid\{{\tau}_t\})\)\Big)
    \nonumber \\
    &= \ln\Exp_\tau \exp\Big( -\eta_t\ell_t(\tau\mid\{\tau_t\}) + \eta_t\Exp_{\bar\tau}\ell_t(\bar\tau \mid \{\tau_t\}) \Big) 
    \nonumber \\
    &= \ln\Exp_\tau\exp\Big(-\eta_t\ell_t(\tau\mid\{\tau_t\})\Big) 
    +
    \ln \exp \Big(\eta_t \Exp_{\bar\tau}\ell_t(\bar\tau \mid \{\tau_t\})\Big) 
    \nonumber \\
    &\le \Exp_\tau \exp \(-\eta_t\ell_t(\tau\mid\{\tau_t\})\) - 1 + 
    \eta_t \Exp_{\bar\tau} \ell_t(\bar\tau \mid \{\tau_t\})
    \label{eq:exp3:log-bound} \\
    &\le \Exp_\tau\Big[\exp\(-\eta_t\ell_t(\tau\mid\{\tau_t\})\) - 1 + \eta_t\ell_t(\tau\mid\{\tau_t\})\Big]
    \nonumber \\
    &\le \Exp_\tau\eta_t^2\frac{{\ell_t(\tau\mid\{\tau_t\})}^2}{2}
    \label{eq:exp3:exp-bound} \\
    &= \frac{\eta_t^2}{2}\sum_{\tau\in\Hs}p_t(\tau){\ell_t(\tau\mid\{\tau_t\})}^2
    \nonumber \\
    &= \frac{\eta_t^2}{2}\sum_{\tau\in\Hs}p_t(\tau)\(\frac{\ell_t(\tau)}{p_t(\tau) + \gamma_t}\cdot\mathbbm{1}(\tau_t=\tau)\)\ell_t(\tau\mid\{\tau_t\})
    \nonumber \\
    &\le \frac{\ell_\text{max}\eta_t^2}{2}\sum_{\tau\in\Hs}\ell_t(\tau\mid\{\tau_t\}),
    \label{eq:exp3:variance-control}
\end{align}
where
(\ref{eq:exp3:log-bound}) holds due to $\ln(x) \le x-1$ for $x\le1$, and (\ref{eq:exp3:exp-bound}) uses $\exp(x)\le 1 + x + x^2/2$ for $x\le0$.

\paragraph{Third step.}

Let $\tilde L_t(\tau) \coloneqq \sum_{i=1}^t \ell_i(\tau \mid \{\tau_i\})$ and $\tilde L_0(\tau) = 0$. Then, we have
\begin{align}
    - \frac{1}{\eta_t} \ln \Exp_{\tau\sim p_t} \exp &\bigg( -\eta_t \ell_t(\tau \mid \{\tau_t\}) \bigg)
    \nonumber
    \\
    &= - \frac{1}{\eta_t} \ln \sum_{\tau \in \Hs} p_t(\tau) \exp\bigg( -\eta_t \ell_t(\tau \mid \{\tau_t\}) \bigg)
    \nonumber
    \\
    &= - \frac{1}{\eta_t} \ln \sum_{\tau \in \Hs} \frac{\exp (-\eta_t \tilde L_{t-1}(\tau))}{\sum_{\bar \tau \in \Hs} \exp (-\eta_t \tilde L_{t-1}(\bar \tau))}
    \exp\bigg( -\eta_t \ell_t(\tau \mid \{\tau_t\}) \bigg)
    \nonumber
    \\
    &= - \frac{1}{\eta_t} \ln \frac{\sum_{\tau \in \Hs} \exp (-\eta_t \tilde L_t(\tau))}{\sum_{\tau \in \Hs} \exp (-\eta_t \tilde L_{t-1}(\tau))}.
    \label{eq:exp3:frac}
\end{align}

\paragraph{Fourth step.}
We combine rewritten forms to continue the derivation as follows:
\begin{align}
    \sum_{t=1}^T\Exp_{\tau \sim p_t}& \ell(\tau\mid\{\tau_t\}) 
    \nonumber
    \\
    &\le
    \sum_{t=1}^T\frac{\ell_\text{max}\eta_t}{2}\sum_{\tau\in\Hs}\ell_t(\tau\mid\{\tau_t\})
    -
    \sum_{t=1}^T\frac{1}{\eta_t}\ln\frac{\sum_{\tau\in\Hs}\exp\(-\eta_t\tilde{L}_t(\tau)\)}{\sum_{\tau\in\Hs}\exp\(-\eta_t\tilde{L}_{t-1}(\tau)\)}
    \label{eq:exp3:combine} \\
    &\le
    \sum_{t=1}^T\frac{\ell_\text{max}\eta_t}{2}\sum_{\tau\in\Hs}\ell_t(\tau\mid\{\tau_t\})
    -
    \frac{1}{\eta_T}\sum_{t=1}^T\ln\frac{\sum_{\tau\in\Hs}\exp\(-\eta_t\tilde{L}_t(\tau)\)}{\sum_{\tau\in\Hs}\exp\(-\eta_t\tilde{L}_{t-1}(\tau)\)}
    \label{eq:exp3:telescoping} \\
    &=
    \sum_{t=1}^T\frac{\ell_\text{max}\eta_t}{2}\sum_{\tau\in\Hs}\ell_t(\tau\mid\{\tau_t\})
    -
    \frac{1}{\eta_T}\ln\frac{\sum_{\tau\in\Hs}\exp\(-\eta_T\tilde{L}_T(\tau)\)}{\sum_{\tau\in\Hs}\exp\(-\eta_0\tilde{L}_0(\tau)\)}
    \nonumber \\
    &=
    \sum_{t=1}^T\frac{\ell_\text{max}\eta_t}{2}\sum_{\tau\in\Hs}\ell_t(\tau\mid\{\tau_t\})
    -
    \frac{1}{\eta_T}\ln\sum_{\tau\in\Hs}\exp\(-\eta_T\tilde{L}_T(\tau)\) + \frac{\ln|\Hs|}{\eta_T}
    \nonumber \\
    &\le
    \sum_{t=1}^T\frac{\ell_\text{max}\eta_T}{2}\sum_{\tau\in\Hs}\ell_t(\tau\mid\{\tau_t\})
    -
    \frac{1}{\eta_t}\ln\(\max_\tau\exp\(-\eta_T\tilde{L}_T(\tau)\)\) + \frac{\ln|\Hs|}{\eta_T}
    \nonumber \\
    &=
    \sum_{t=1}^T\frac{\ell_\text{max}\eta_t}{2}\sum_{\tau\in\Hs}\ell_t(\tau\mid\{\tau_t\})
    +
    \min_\tau\sum_{t=1}^T\ell_t(\tau\mid\{\tau_t\}) + \frac{\ln|\Hs|}{\eta_T},
    \label{eq:exp3:final}
\end{align}
where (\ref{eq:exp3:combine}) holds from 
(\ref{eq:exp3:var-decomp}),
(\ref{eq:exp3:variance-control}), and
(\ref{eq:exp3:frac}).
Also, 
(\ref{eq:exp3:telescoping}) holds as
$\eta_t$ is non-increasing
and
$\frac{\sum_{\tau\in\Hs}\exp\(-\eta_t\tilde{L}_t(\tau)\)}{\sum_{\tau\in\Hs}\exp\(-\eta_t\tilde{L}_{t-1}(\tau)\)}\le1$.

Note that 
$\Exp_{\tau\sim p_t}\ell(\tau\mid\{\tau_t\}) = \ell_t(\tau_t) - \gamma_t\sum_{\tau\in\Hs}\ell(\tau\mid\{\tau_t\})$ for the following reason:
\begin{align*}
    \Exp_{\tau\sim p_t}\ell_t(\tau\mid\{\tau_t\}) 
    &= \sum_{\tau\in\Hs}p_t(\tau)\frac{\ell_t(\tau)}{p_t(\tau) + \gamma_t}\mathbbm{1}(\tau_t = \tau)
    \nonumber \\
    &=\sum_{\tau\in\Hs} \( \frac{p_t(\tau) + \gamma_t}{p_t(\tau) + \gamma_t}\ell_t(\tau)\mathbbm{1}(\tau_t = \tau) - \frac{\gamma_t}{p_t(\tau) + \gamma_t}\ell_t(\tau)\mathbbm{1}(\tau_t = \tau) \)
    \nonumber \\
    &= \ell_t(\tau_t) - \gamma_t\sum_{\tau\in\Hs}\ell_t(\tau\mid\{\tau_t\}). 
\end{align*}

From this fact and  (\ref{eq:exp3:final}), we have
\begin{align}
    \sum_{t=1}^T \ell_t(\tau_t)
    -
    \min_\tau\sum_{t=1}^T\ell_t(\tau\mid\{\tau_t\})
    &\le 
    \sum_{t=1}^T\frac{\ell_\text{max}\eta_t}{2}\sum_{\tau\in \Hs }\ell_t(\tau\mid\{\tau_t\})
    + \sum_{t=1}^T \gamma_t\sum_{\tau\in \Hs }\ell(\tau\mid\{\tau_t\})
    + \frac{\ln|\Hs|}{\eta_T}
    \nonumber \\
    &=
    \sum_{t=1}^T\(\frac{\ell_\text{max}\eta_t}{2} + \gamma_t\)\sum_{\tau\in \Hs }\ell_t(\tau\mid\{\tau_t\})
    +
    \frac{\ln|\Hs|}{\eta_T}.
    \label{eq:exp3:prefinal}
\end{align}

If $\eta_t \le \frac{2\gamma_t}{\ell_\text{max}}$ then it satisfies the Lemma condition \ie $\alpha_{t, \tau} = \frac{\ell_\text{max}\eta_t}{2} + \gamma_t \le 2\gamma_t$, we can apply Lemma \ref{lem:exp3}, from (\ref{eq:exp3:prefinal}) to get
\begin{equation*}
    \sum_{t=1}^T \ell_t(\tau_t)
    -
    \min_\tau\sum_{t=1}^T\ell_t(\tau\mid\{\tau_t\})
    \le 
    \sum_{t=1}^T\(\frac{\ell_\text{max}\eta_t}{2} + \gamma_t\)\sum_{\tau\in \Hs }\ell_t(\tau)
    +
    \frac{\ln|\Hs|}{\eta_T}
    +
    \ell_\text{max} \ln \frac{2}{\delta}  
\end{equation*}
with probability $1-\delta/2$.

After that, by Corollary \ref{cor:exp3}, we get 
\begin{equation*}
    \sum_{t=1}^T \ell_t(\tau_t)
    -
    \min_\tau\sum_{t=1}^T\ell_t(\tau)
    \le
    \sum_{t=1}^T \ell_t(\tau_t)
    -
    \min_\tau\sum_{t=1}^T\ell_t(\tau\mid\{\tau_t\})
    + 
    \frac{\ell_\text{max}}{2\gamma}\ln\frac{2|\Hs|}{\delta}
\end{equation*}
with probability $1-\delta/2$.

Finally, we get combining the above two via the union bound as follows:
\begin{equation*}
    \Reg_T
    = 
    \sum_{t=1}^T \ell_t(\tau_t)
    -
    \min_\tau\sum_{t=1}^T \ell_t(\tau)
    \\
    \le 
    \sum_{t=1}^T\(\frac{\ell_\text{max}\eta_t}{2} + \gamma_t\)\sum_{\tau\in \Hs }\ell_t(\tau)
    +
    \frac{\ln|\Hs|}{\eta_T}
    + \ell_\text{max} \ln \frac{2}{\delta}
    + \frac{\ell_\text{max}}{2\gamma}\ln\frac{2|\Hs|}{\delta}
\end{equation*}
with probability $1-\delta$.

The concluding steps of our proof closely follow the analysis presented by \citep{neu2015explore}. 
Recall that if the learning rate $\eta_t \le \frac{2\gamma_t}{\ell_{\max}}$ it satisfies the conditions of Lemma \ref{lem:exp3}. 
Then, letting $\eta_t = \frac{2\gamma_t}{\ell_\text{max}}$ and $\gamma = \frac{\eta_T \ell_{\text{max}}}{2}$
due to Corollary \ref{cor:exp3}, we can bound the regret bound as follows:

\begin{align}
    \sum_{t=1}^T\(\frac{\ell_\text{max}\eta_t}{2} + \gamma_t\)& \sum_{\tau\in \Hs }\ell_t(\tau)
    + \frac{\ln|\Hs|}{\eta_T}
    + \ell_\text{max} \ln \frac{2}{\delta}
    + \frac{\ell_\text{max}}{2\gamma}\ln\frac{2|\Hs|}{\delta}
    \nonumber \\
    &= 
    \sum_{t=1}^T\ell_\text{max}\eta_t\sum_{\tau\in \Hs }\ell_t(\tau)
    + \frac{2\ln|\Hs|}{\eta_T}
    + \ell_\text{max} \ln \frac{2}{\delta}
    + \frac{1}{\eta_T}\ln\frac{2}{\delta}
    \nonumber \\
    &\le
    \sum_{t=1}^T\ell_\text{max}^2|\Hs|\eta_t
    + \frac{2\ln|\Hs|}{\eta_T}
    + \ell_\text{max} \ln \frac{2}{\delta}
    + \frac{1}{\eta_T}\ln\frac{2}{\delta}.
    \label{eq:exp3:pre-unkown} 
\end{align}

First, we consider the setting with a known time horizon. In this case, we use a fixed learning rate, $\eta_t = \eta_T$ for all $t$, which simplifies the general regret bound to:
\begin{equation}
    \Reg_T 
    \le
    \ell_\text{max}^2T|\Hs|\eta_T
    + \frac{2\ln|\Hs|}{\eta_T}
    + \ell_\text{max} \ln \frac{2}{\delta}
    + \frac{1}{\eta_T}\ln\frac{2}{\delta}.
    \label{eq:exp3:pre-kown}
\end{equation}

Finally, we can get the optimal learning rate $\eta_t = \sqrt{\frac{2\ln|\Hs| + \ln(2/\delta)}{\ell_{\max}^2T|\Hs|}}$. 
Here, to make the learning rate agnostic to $\delta$, we approximate 
$\eta_t \approx \sqrt{\frac{2\ln|\Hs|}{\ell_{\max}^2T|\Hs|}}$, which follows
the convention in the original analysis \citep{neu2015explore}.

Then, from (\ref{eq:exp3:pre-kown}), we obtain the following upper bound:
\begin{align}
    \Reg_T
    &\le 
    \ell_\text{max}\(2\sqrt{2T|\Hs|\ln|\Hs|} + \(1 + \sqrt{\frac{T|\Hs|}{2\ln|\Hs|}}\)\ln\frac{2}{\delta} \)
    \nonumber \\
    &= \mathcal{O}\(2\ell_\text{max}\sqrt{2T|\Hs|\ln|\Hs|/\delta}\).
    \label{eq:exp3:kownT}
\end{align}

Second, for the setting with an unknown time horizon, we use the learning rate $\eta_t = \sqrt{\frac{\ln|\Hs|}{\ell_{\max}^2t|\Hs|}}$. Then we get general regret bound from (\ref{eq:exp3:pre-unkown}), noting that $\sum_{t=1}^T\frac{1}{\sqrt{t}}\le2\sqrt{T}$
\begin{align}
    \Reg_T 
    &\le \ell_\text{max}\(4\sqrt{T|\Hs|\ln|\Hs|} + \(1 + \sqrt{\frac{T|\Hs|}{\ln|\Hs|}}\)\ln\frac{2}{\delta}\)
    \nonumber \\
    &= \mathcal{O}\(4\ell_\text{max}\sqrt{T|\Hs|\ln|\Hs|/\delta}\).
    \nonumber
\end{align}

\paragraph{Expected regret}
The expected regret can be obtained via integrating the deviations in (\ref{eq:exp3:kownT}), \ie
\begin{equation*}
    \Exp [W] \le \int_0^1 \frac{1}\delta\Prob\(W > \ln\frac{1}{\delta} \) d\delta.
\end{equation*}
Here, if we take $W = \frac{1}{\ell_\text{max}\(1+\sqrt{T|\Hs|/(2\ln|\Hs|)}\)}\(\Reg_T-2\ell_\text{max}\sqrt{2 T|\Hs|\ln|\Hs|}\)
- \ln2 > \ln \frac{1}{\delta}$, which holds
with probability at most $\delta$
then
\begin{equation*}
    \Exp [W] \le 1,
\end{equation*}
which suggests that
\begin{align}
    \Exp[\Reg_T]
    &\le 
     \ell_\text{max} 
    \(2\sqrt{2T|\Hs|\ln|\Hs|}
    +  (1+\ln2)\(1+\sqrt{\frac{T|\Hs|}{2\ln|\Hs|}}\)
    \)
    \nonumber\\
    &= \Os(\ell_\text{max} \sqrt{T|\Hs|\ln|\Hs|}).
    \label{eq:exp3:expregretbound}
\end{align}

\subsection{A Proof of Lemma \ref{lem:conversion}}
\label{proof:lem:conversion}

We derive a lower bound of $\Reg_T$ which allows us to bound the SR violation $\Vs_T$.
From the definition of regret and the loss function $\ell_t(\tau, \alpha)$ in (\ref{eq:loss-sg}), the following inequality holds for any comparator policy $\tau \in \Hs$:
\begin{align}
    (1+\lambda)\Reg_T
    &= 
    \sum_{t=1}^T \Big[ a_t(\tau_t) + \lambda d_t(\tau_t, \alpha) \Big]
    - \min_{\tau' \in \Hs} 
    \sum_{t=1}^T \left[ a_t(\tau') + \lambda d_t(\tau', \alpha) \right]
    \nonumber \\
    &\ge
    \sum_{t=1}^T \Big[ a_t(\tau_t) + \lambda d_t(\tau_t, \alpha) \Big]
    - \sum_{t=1}^T \left[ a_t(\tau) + \lambda d_t(\tau, \alpha) \right].
    \label{eq:onlineconversion:regret_def}
\end{align}
Let $\tau^* \in \Hs$ be the policy that minimizes the SR violation, \ie $\tau^* \in \argmin_{\tau \in \Hs} \Vs_T(\tau)$,
where $\Vs_T(\tau) \coloneqq \sum_{t=1}^T d_t(\tau, \alpha) - \alpha T$.
By substituting $\tau = \tau^*$ into (\ref{eq:onlineconversion:regret_def}) and rearranging the terms regarding the constraint term $d_t$, we obtain:
\begin{align}
    \sum_{t=1}^T \lambda d_t(\tau_t, \alpha) - \sum_{t=1}^T \lambda d_t(\tau^*, \alpha)
    &\le
    (1+\lambda)\Reg_T 
    + \sum_{t=1}^T \left[ a_t(\tau^*) - a_t(\tau_t) \right].
    \label{eq:onlineconversion:rearrange_step1}
\end{align}
Using the definitions $\Vs_T(\tau) = \sum_{t=1}^T d_t(\tau, \alpha) - \alpha T$ and $\sum_{t=1}^T a_t(\tau_t) = T \cdot \Ine_T$, we can rewrite (\ref{eq:onlineconversion:rearrange_step1}) as:
\begin{align}
    \lambda \left( \Vs_T - \min_{\tau \in \Hs} \Vs_T(\tau) \right)
    &\le
    (1+\lambda)\Reg_T 
    + \sum_{t=1}^T a_t(\tau^*) - T \cdot \Ine_T
    \nonumber \\
    &\le
    (1+\lambda)\Reg_T 
    + T (1 - \Ine_T),
    \label{eq:onlineconversion:rearrange}
\end{align}
where the last inequality holds since $a_t(\cdot) \in [0, 1]$, implying $\sum_{t=1}^T a_t(\tau^*) \le T$.

Finally, we observe that the minimum SR violation is non-positive, as the fallback policy $\tau_{\emptyset}$ (always abstaining) achieves zero risk:
\begin{equation}
    \min_{\tau \in \Hs} \Vs_T(\tau) \le \Vs_T(\tau_{\emptyset}) = 0.
\end{equation}
Dividing (\ref{eq:onlineconversion:rearrange}) by $\lambda$ and applying this observation yields the final bound:
\begin{align}
    \Vs_T
    &\le 
    \frac{(1+\lambda)\Reg_T}{\lambda} + \frac{T(1 - \Ine_T)}{\lambda}.
\end{align}
Setting $\lambda = T^{\nicefrac{1}{2}}$, we conclude:
\begin{align}
    \Vs_T
    &\le 
    \frac{(1+T^{\nicefrac{1}{2}})\Reg_T}{T^{\nicefrac{1}{2}}} + T^{\nicefrac{1}{2}}(1 - \Ine_T)
    = \Os(T^{\nicefrac{1}{2}}).
\end{align}

\subsection{A Proof of Theorem \ref{thm:ours:regretbound}}
\label{proof:thm:ours:regretbound}

Here we prove Theorem~\ref{thm:ours:regretbound} for an arbitrary loss range $\ell_t(\cdot) \in [0, \ell_\text{max}]$ of the form (\ref{eq:loss-sg}) and a non-increasing learning rate; the theorem is the case $\ell_\text{max}=1$. Specifically, with $\eta_t = \frac{2\gamma_t}{\ell_\text{max}} = \sqrt{\frac{\ln|\mathcal{H}|}{\ell_\text{max}^2 T}}$, Algorithm~\ref{alg:exp3-sg-unlocking} provides, with probability at least $1-\delta$,
\begin{equation*}
    \Reg_T \le \ell_\text{max}\(4\sqrt{T\ln|\mathcal H|} + \( 1 + \sqrt{\frac{T}{\ln|\Hs|}}\)\ln\frac{2}{\delta}\).
\end{equation*}

The proof technique follows that of the \eeeix \citep{neu2015explore} regret bound except that we use a novel loss estimator for feedback unlocking. 
Note that we abbreviate $\ell_t(\tau, \alpha \mid \Hs_t(\tau_t))$ as $\ell_t(\tau \mid \Hs_t(\tau_t))$ in this proof.

First, 
we introduce some properties of 
$\Hs_t(\tau_t)$ in our algorithm.
\begin{figure*}[h]
  \centering
  \subfigure[Case for $\tau > f_t$]{
    \begin{tikzpicture}[
        >=stealth,
        very thick,
        scale=0.7
    ]
    
    \draw [->] (-3, 0) -- (4, 0);
    
    \draw (0, 0) -- (0, 1.5);
    \draw [->] (0, 1.5) -- (4, 1.5);
    
    \node [below=3pt] at (0, 0) {$f_t$};
    \draw [fill=white] (0, 0) circle (2.5pt);
    
    \node [below=3pt] at (1.5, 0) {$\tau$};
    \draw (1.5, 0.1) -- (1.5, -0.1);
    
    \node [above] at (-1.5, 1.5) {$\Hs_t^C(\tau)$};
    \node [above] at (2, 1.5) {$\Hs_t(\tau)$};
    
    \end{tikzpicture}
  }
  \hspace{3em}
  \subfigure[Case for $\tau \le f_t$]{
    \begin{tikzpicture}[
        >=stealth,
        very thick,
        scale=0.7
    ]
    
    \draw [->] (-3, 0) -- (4, 0);
    
    \draw (0, 0) -- (0, 1.5);
    \draw [->] (0, 1.5) -- (-3, 1.5);
    
    \node [below=3pt] at (0, 0) {$f_t$};
    \draw [fill=black] (0, 0) circle (2.5pt);
    
    \node [below=3pt] at (-1.5, 0) {$\tau$};
    \draw (-1.5, 0.1) -- (-1.5, -0.1);
    
    \node [above] at (-1.5, 1.5) {$\Hs_t(\tau)$};
    \node [above] at (2, 1.5) {$\Hs_t^C(\tau)$};
    
    \end{tikzpicture}
  }
  \vspace{-2ex}
  \caption{
  Visualization of $\Hs_t(\tau)$.
  }
  \label{fig:unlockset}
\end{figure*}

\begin{enumerate}
\item (self-inclusion) The first trivial property is that $\tau$ is in $\Hs_t(\tau)$ itself:
\begin{equation}
    \tau \in \Hs_t(\tau).
    \label{proof:thm1:trivial}
\end{equation}

\item (partition) The second property is that if $\tau' \in \Hs_t(\tau)$, then we have
\begin{equation}
    \Hs_t(\tau') = \Hs_t(\tau).
    \label{proof:thm1:domain:equivalent-new}
\end{equation}
\noindent This holds since
if $\tau' \in \Hs_t(\tau)$ and $\tau \in \Hs_t(\tau)$ due to (\ref{proof:thm1:trivial}),
$\tau$ and $\tau'$ are in the same side from $f_t$
(\ie either $\tau \le f_t \wedge \tau' \le f_t$ or $\tau > f_t \wedge \tau' > f_t$), thus 
$\Sh(\x_t; \tau) = \Sh(\x_t; \tau')$, implying
$\Hs_t(\tau') = \Hs_t(\tau)$ by our construction of $\Hs_t(\cdot)$ in our algorithm.
The second case when $\tau \in \Hs_t(\tau')$ similarly holds as well.

\item (complement) The third property, similar to the second one, is that
if $\tau' \in \Hs_t^C(\tau)$,
then we have
\begingroup
\setlength{\abovedisplayskip}{0pt}
\setlength{\belowdisplayskip}{0pt}
\begin{gather}
    \Hs_t^C(\tau) = \Hs_t(\tau'),
    \text{ or equivalently }
    \Hs_t(\tau) = \Hs_t^C(\tau'). \label{proof:thm1:complement}
\end{gather}
\endgroup

This property holds as follows: 
$\tau' \in \Hs_t(\tau')$ holds due to (\ref{proof:thm1:trivial}).
Suppose $\tau' \in \Hs_t^C(\tau)$, then 
we have $\tau' \notin \Hs_t(\tau)$. 
This along with $\tau' \in \Hs_t(\tau')$
implies $\Hs_t(\tau') \neq \Hs_t(\tau)$.
Since $\Hs_t(\tau_t)$ is either $\{\tau \mid \tau \le f_t\}$ or $\{\tau \mid \tau > f_t\}$ depending on the value of $\tau_t$, $\Hs_t(\cdot)$ partitions the space $[0,1]$ into two disjoint sets based on $f_t$,
where one set is the complement of the other.
Therefore, $\Hs_t(\tau) \neq \Hs_t(\tau')$ is equivalent to either $\Hs_t^C(\tau) = \Hs_t(\tau')$ or $\Hs_t(\tau) = \Hs_t^C(\tau')$.

\item (swap) The final property is that
\begin{equation}
    \tau' \in \Hs_t(\tau) \quad\text{if and only if}\quad
    \tau \in \Hs_t(\tau').
    \label{proof:thm1:swapping-new}
\end{equation}
Suppose that $\tau' \in \Hs_t(\tau)$ and $\tau' \in \Hs_t(\tau')$, then the above holds as follows:
\begin{align*}
    \tau 
    \in \Hs_t(\tau) 
    = \Hs_t(\tau'),
\end{align*}
where
the inclusion holds due to (\ref{proof:thm1:trivial})
and 
the equality holds due to (\ref{proof:thm1:domain:equivalent-new}) if $\tau', \tau \in \Hs_t(\tau)$. The reverse holds similarly as well.

\end{enumerate}


With these properties, we can derive the following two useful equalities.
Let 
\begin{equation}
\hat \ell_t (\tau \mid  \Hs_t(\tau_t)) = \frac{\ell_t(\tau)}{\sum_{\bar\tau \in \Hs_t({\tau_t})} \mathbbm{1}(\tau \in \Hs_t({\bar\tau})) \cdot p_t(\bar \tau)} \mathbbm{1}(\tau \in \Hs_t({\tau_t})),
\label{eq:ourunbiasedestimator}
\end{equation} 
which is our loss estimator without $\gamma_t$. 
Then, the following equality holds:
\begin{align}
    \Exp_{\tau_t \sim p_t} \hat \ell_t(\tau \mid \Hs_t({\tau_t}))
    &=
     \sum_{\tau' \in \Hs} p_t(\tau') \[ \frac{\ell_t(\tau)}{\sum_{\bar\tau \in \Hs_t({\tau'})} \mathbbm{1}(\tau \in \Hs_t({\bar\tau})) \cdot p_t(\bar \tau)} \mathbbm{1}(\tau \in \Hs_t({\tau'})) \]
    \nonumber \\
    &=
    \ell_t(\tau) \sum_{\tau' \in \Hs} \frac{p_t(\tau')}{\sum_{\bar\tau \in \Hs_t({\tau'})} \mathbbm{1}(\tau \in \Hs_t({\bar\tau})) \cdot p_t(\bar \tau)} \mathbbm{1}(\tau \in \Hs_t({\tau'}))
    \nonumber \\
    &=
    \ell_t(\tau)
    \sum_{\tau' \in \Hs}
    \frac{p_t(\tau')}
    {\sum_{\bar\tau \in \Hs_t(\tau')}
    \mathbbm{1}(\tau \in \Hs_t({\bar\tau}))\cdot p_t(\bar \tau)} \mathbbm{1}({\tau' \in \Hs_t({\tau})})
    \label{proof:thm3:swapping}
    \\
    &=
    \ell_t(\tau)
    \sum_{\tau' \in \Hs_t(\tau)}
    \frac{p_t(\tau')}
    {\sum_{\bar\tau \in \Hs_t(\tau')}
    \mathbbm{1}(\tau \in \Hs_t({\bar\tau}))\cdot p_t(\bar \tau)} 
    \label{proof:thm3:trivial:indicator} \\
    &=
    \ell_t(\tau)
    \sum_{\tau' \in \Hs_t(\tau)}
    \frac{p_t(\tau')}
    {\sum_{\bar\tau \in \Hs_t(\tau)}
    \mathbbm{1}(\tau \in \Hs_t({\bar\tau}))\cdot p_t(\bar \tau)} 
    \label{proof:thm3:state1} \\
    &=
    \ell_t(\tau)
    \sum_{\tau' \in \Hs_t(\tau)}
    \frac{p_t(\tau')}
    {\sum_{\bar\tau \in \Hs_t(\tau)}
    \mathbbm{1}(\bar \tau \in \Hs_t({\tau}))\cdot p_t(\bar \tau)} 
    \label{proof:thm3:swapping2} \\
    &=
    \ell_t(\tau)
    \sum_{\tau' \in \Hs_t(\tau)}
    \frac{p_t(\tau')}
    {\sum_{ \bar\tau \in \Hs_t(\tau)}
    p_t(\bar \tau)} 
    \nonumber \\
    &=
    \ell_t(\tau),
    \label{eq:unbiasedlossest}
\end{align}
where 
(\ref{proof:thm3:swapping}) holds as (\ref{proof:thm1:swapping-new}),
(\ref{proof:thm3:trivial:indicator}) follows from the properties of the summation and indicator function,
(\ref{proof:thm3:state1}) holds as
(\ref{proof:thm1:domain:equivalent-new}) from $\tau'\in \Hs_t(\tau)$ in the summation,
and (\ref{proof:thm3:swapping2})
follows by the same argument as (\ref{proof:thm3:swapping}) from the property (\ref{proof:thm1:swapping-new}),
\ie $\bar \tau \in \Hs_t(\tau) \text{~if~} \tau \in \Hs_t(\bar \tau)$.


Also, the following equality with respect to our loss estimator (\ref{eq:betteradvbandit:ourunbiasedestimator}) holds similarly:
\begin{align}
    \Exp_{\tau \sim p_t} \ell_t(\tau \mid \Hs_t(\tau_t))
    &=
    \sum_{\tau \in \Hs} p_t(\tau) \frac{\ell_t(\tau)}{\gamma_t + \sum_{\bar\tau \in \Hs_t(\tau_t)} \mathbbm{1}(\tau \in \Hs_t(\bar\tau)) p_t(\bar\tau)}\cdot \mathbbm{1}(\tau \in \Hs_t(\tau_t))
    \nonumber\\
    &=
    \sum_{\tau \in \Hs_t(\tau_t)} p_t(\tau) \frac{\ell_t(\tau)}{\gamma_t + \sum_{\bar\tau \in \Hs_t(\tau_t)} \mathbbm{1}(\tau \in \Hs_t(\bar\tau)) p_t(\bar\tau)}
    \nonumber\\
    &=
    \ell_t(\tau_t)
    \sum_{\tau \in \Hs_t(\tau_t)} \frac{p_t(\tau)}{\gamma_t + \sum_{\bar\tau \in \Hs_t(\tau_t)} \mathbbm{1}(\tau \in \Hs_t(\bar\tau)) p_t(\bar\tau)}
    \label{eq:specializedloss:1}\\
    &=
    \ell_t(\tau_t)
    \sum_{\tau \in \Hs_t(\tau_t)} \frac{p_t(\tau)}{\gamma_t + \sum_{\bar\tau \in \Hs_t(\tau_t)} p_t(\bar\tau)}
    \label{eq:specializedloss2:1}\\
    &=
    \ell_t(\tau_t)
    \frac{\sum_{\tau \in \Hs_t(\tau_t)} p_t(\tau)}{\gamma_t + \sum_{\bar\tau \in \Hs_t(\tau_t)} p_t(\bar\tau)}
    \nonumber\\
    &=
    \ell_t(\tau_t)
    \frac{\gamma_t + \sum_{\tau \in \Hs_t(\tau_t)} p_t(\tau)}{\gamma_t + \sum_{\bar\tau \in \Hs_t(\tau_t)} p_t(\bar\tau)}
    -\ell_t(\tau_t)\frac{\gamma_t}{\gamma_t + \sum_{\bar\tau \in \Hs_t(\tau_t)} p_t(\bar\tau)}
    \label{eq:thm2:exp:over:tau1}\\
    &=
    \ell_t(\tau_t)
    -\gamma_t
    \frac{\ell_t(\tau_t)}
    {\gamma_t + \sum_{\bar\tau \in \Hs_t(\tau_t)} \mathbbm{1}(\tau_t \in \Hs_t(\bar\tau)) p_t(\bar\tau)}
    \mathbbm{1}(\tau_t\in\Hs_t(\tau_t))
    \label{eq:thm2:exp:over:tau1-5} \\
    &=
    \ell_t(\tau_t)
    - \gamma_t \ell_t(\tau_t \mid \Hs_t(\tau_t)),
    \label{eq:thm2:exp:over:tau2}
\end{align}
where (\ref{eq:specializedloss:1}) holds as $\ell_t(\tau)=\ell_t(\tau')$ for all $\tau\in\Hs_t(\tau')$ by the definition of the specialized loss in (\ref{eq:loss-sg}), (\ref{eq:specializedloss2:1}) holds as the same as (\ref{proof:thm3:swapping2}), (\ref{eq:thm2:exp:over:tau1}) holds as $\tau_t \in \Hs_t(\tau_t)$ and (\ref{proof:thm1:swapping-new}), 
(\ref{eq:thm2:exp:over:tau1-5}) holds due to the property of the indicator function,
and 
(\ref{eq:thm2:exp:over:tau2}) holds by the definition.

Then, we first introduce a new Lemma \ref{lem:exsul} and Corollary \ref{cor:exsul} to establish a high probability bound.
\clearpage
\raggedbottom
\begin{lemma}
    Let $\alpha_t \le 2 \gamma_t$ where $\gamma_t$ is non-increasing in $t$. Then, the following inequality holds with probability at least $1-\delta$:
    \vspace{-2ex}
    \begin{equation}
        \sum_{t=1}^{T} \alpha_t \ell_t(\tau_{t} \mid \Hs_t(\tau_t)) - \sum_{t=1}^{T} {2 \alpha_t \ell_\text{max}}
        \le \ell_\text{max} \ln (1/\delta).
        \label{eq:lem2}
    \end{equation}
    \label{lem:exsul}
\end{lemma}
\begin{proof}
    First, the following inequality holds for any $\tau \in \Hs$: 
    \begin{align}
        \ell_t(\tau \mid \Hs_t(\tau_t))
        &=
        \frac{\ell_t(\tau)}{\gamma_t + \sum_{\bar\tau \in \Hs_t({\tau_t})} \mathbbm{1}(\tau \in \Hs_t({\bar\tau})) \cdot p_t(\bar\tau) } \cdot \mathbbm{1}(\tau \in \Hs_t({\tau_t}))
        \nonumber
        \\
        &=
        \ell_\text{max} \cdot \frac{\ell_t(\tau)/\ell_\text{max}}{\gamma_t + \sum_{\bar\tau \in \Hs_t({\tau_t})} \mathbbm{1}(\tau \in \Hs_t({\bar\tau})) \cdot p_t(\bar\tau) } \cdot \mathbbm{1}(\tau \in \Hs_t({\tau_t}))
        \nonumber
        \\
        &\le
        \ell_\text{max} \cdot \frac{\ell_t(\tau)/\ell_\text{max}}{\gamma_t \ell_t(\tau)/\ell_\text{max} + \sum_{\bar\tau \in \Hs_t({\tau_t})} \mathbbm{1}(\tau \in \Hs_t({\bar\tau})) \cdot p_t(\bar\tau) } \cdot \mathbbm{1}(\tau \in \Hs_t({\tau_t}))
        \nonumber
        \\
        &=
        \frac{\ell_\text{max}}{2\gamma_t} \cdot \frac{2\gamma_t \ell_t(\tau)/\ell_\text{max}}{\gamma_t \ell_t(\tau)/\ell_\text{max} +  \sum_{\bar\tau \in \Hs_t({\tau_t})} \mathbbm{1}(\tau \in \Hs_t({\bar\tau})) \cdot p_t(\bar\tau) } \cdot \mathbbm{1}(\tau \in \Hs_t({\tau_t}))
        \nonumber
        \\
        &=
        \frac{1}{\beta_t} \cdot \frac{\frac{\beta_t \ell_t(\tau)}{\sum_{\bar\tau \in \Hs_t({\tau_t})} \mathbbm{1}(\tau \in \Hs_t({\bar\tau})) \cdot p_t(\bar\tau)}}{\frac{\beta_t \ell_t(\tau)}{2\sum_{\bar\tau \in \Hs_t({\tau_t})} \mathbbm{1}(\tau \in \Hs_t({\bar\tau})) \cdot p_t(\bar\tau)} + 1} \cdot \mathbbm{1}(\tau \in \Hs_t({\tau_t}))
        \nonumber
        \\
        &=
        \frac{1}{\beta_t} \cdot \frac{\frac{\beta_t \ell_t(\tau)}{\sum_{\bar\tau \in \Hs_t({\tau_t})} \mathbbm{1}(\tau \in \Hs_t({\bar\tau})) \cdot p_t(\bar\tau)} \cdot \mathbbm{1}(\tau \in \Hs_t({\tau_t})) }{\frac{\beta_t \ell_t(\tau)}{2\sum_{\bar\tau \in \Hs_t({\tau_t})} \mathbbm{1}(\tau \in \Hs_t({\bar\tau})) \cdot p_t(\bar\tau)}\cdot \mathbbm{1}(\tau \in \Hs_t({\tau_t})) + 1} 
        \nonumber
        \\
        &=
        \frac{1}{\beta_t} \cdot \frac{\beta_t \hat \ell_t(\tau \mid \Hs_t(\tau_t))}{\beta_t \hat \ell_t(\tau \mid \Hs_t(\tau_t))/2 + 1} 
        \nonumber
        \\
        &\le
        \frac{1}{\beta_t} \cdot \ln\(1+\beta_t \hat \ell_t(\tau \mid \Hs_t(\tau_t)) \),
        \label{eq:lem:beta}
    \end{align}
    where $\beta_t=2\gamma_t/\ell_\text{max}$, $\hat \ell_t(\tau \mid \Hs_t(\tau_t)) \coloneqq \frac{\ell_t(\tau)}{\sum_{\bar\tau \in \Hs_t({\tau_t})} \mathbbm{1}(\tau \in \Hs_t({\bar\tau})) \cdot p_t(\bar\tau) } \cdot \mathbbm{1}(\tau \in \Hs_t({\tau_t}))$, and the last inequality holds as $\frac{x}{1+x/2} \le \ln(1+x)$ for all $x\ge 0$.
    

    Let $\Exp_t$ be the expectation conditioned on $\tau_1, \dots, \tau_{t-1}$
    and $\tau'$ be any element in $\Hs$.
    Then, we have
    \begin{align}
        \Exp_t &\[ \exp\(\frac{\alpha_t}{\ell_\text{max}} \ell_t(\tau_t \mid \Hs_t(\tau_t))\) \]
        \nonumber
        \\
        &\le
        \Exp_t\[\exp\(\frac{\alpha_t}{\ell_\text{max} \beta_t} \ln \(1 + \beta_t \hat \ell_t(\tau_t \mid \Hs_t(\tau_t))\) \)\]
        \label{eq:lem:beta:taut}
        \\
        &\le
        \Exp_t\[1+\frac{\alpha_t}{\ell_\text{max}} \hat \ell_t(\tau_t \mid \Hs_t(\tau_t))\]
        \label{eq:exsulix:alphabeta}
        \\
        &=
        1+\frac{\alpha_t}{\ell_\text{max}} \Exp_t \hat \ell_t(\tau_t \mid \Hs_t(\tau_t))
        \nonumber
        \\
        &=
        1+\frac{\alpha_t}{\ell_\text{max}}\sum_{\tau_t \in \Hs} p_t(\tau_t)\frac{\ell_t(\tau_{t})}{\sum_{\bar\tau \in \Hs_t({\tau_t})} \mathbbm{1}(\tau_{t} \in \Hs_t({\bar\tau})) \cdot p_t(\bar\tau) } \cdot \mathbbm{1}(\tau_{t} \in \Hs_t({\tau_t}))
        \nonumber
        \\
        &=
        1+\frac{\alpha_t}{\ell_\text{max}}\sum_{\tau_t \in \Hs} p_t(\tau_t)\frac{\ell_t(\tau_t)}{\sum_{\bar\tau\in\Hs_t(\tau_t)}p_t(\bar\tau)}
        \nonumber
        \\
        &=
        1+\frac{\alpha_t}{\ell_\text{max}}
        \(
        \sum_{\tau \in \Hs_t(\tau')}
        \frac{p_t(\tau)\ell_t(\tau)}{\sum_{\bar\tau \in \Hs_t(\tau)}  p_t(\bar\tau)}  
        +
        \sum_{\tau \in \Hs_t^C(\tau')}
        \frac{p_t(\tau)\ell_t(\tau)}{\sum_{\bar\tau \in \Hs_t(\tau)}  p_t(\bar\tau)}
        \)
        \nonumber
        \\
        &\le
        1+\frac{\alpha_t}{\ell_\text{max}}\ell_\text{max}
        \(
        \sum_{\tau \in \Hs_t(\tau')}
        \frac{p_t(\tau)}{\sum_{\bar\tau \in \Hs_t(\tau)}  p_t(\bar\tau)}  
        +
        \sum_{\tau \in \Hs_t^C(\tau')}
        \frac{p_t(\tau)}{\sum_{\bar\tau \in \Hs_t(\tau)}  p_t(\bar\tau)}
        \)
        \nonumber
        \\
        &=
        1+\alpha_t
        \(
        \sum_{\tau \in \Hs_t(\tau')}
        \frac{p_t(\tau)}{\sum_{\bar\tau \in \Hs_t({ \tau'})}  p_t(\bar\tau)}  
        +
        \sum_{\tau \in \Hs_t^C(\tau')}
        \frac{p_t(\tau)}{\sum_{\bar\tau \in \Hs_t( \tau)}  p_t(\bar\tau)}
        \)
        \label{proof:lem3:V:left}
        \\
        &=
        1+\alpha_t
        \(
        \sum_{\tau \in \Hs_t(\tau')}
        \frac{p_t(\tau)}{\sum_{\bar\tau \in \Hs_t({ \tau'})}  p_t(\bar\tau)}  
        +
        \sum_{\tau \in \Hs_t^C(\tau')}
        \frac{p_t(\tau)}{\sum_{\bar\tau \in {\Hs_t^C ( \tau') }}  p_t(\bar\tau)}
        \)
        \label{proof:lem3:V:right}
        \\
        &=
        1+2\alpha_t
        \label{proof:lem3:final}
        \\
        &\le
        \exp(2\alpha_t), \label{eq:exsul:lemma:rel}
    \end{align}
    where
    (\ref{eq:lem:beta:taut} holds as (\ref{eq:lem:beta}).
    (\ref{eq:exsulix:alphabeta}) uses $x\ln(1+y) \le \ln(1+xy)$ for all $y > -1$ and $x \in [0,1]$ since $\alpha_t \le 2\gamma_t = \ell_\text{max} \beta_t$.
    (\ref{proof:lem3:V:left}) holds as (\ref{proof:thm1:domain:equivalent-new}), \ie $\Hs_t(\tau)=\Hs_t(\tau')$ due to $\tau\in\Hs_t(\tau')$ in the summation.
    Also,
    (\ref{proof:lem3:V:right}) holds due to (\ref{proof:thm1:complement}),
    \ie
    $\Hs_t(\tau) = \Hs_t^C(\tau')$ as $\tau \in \Hs_t^C(\tau')$ from the summation condition.
    Finally, \eqref{proof:lem3:final} holds because neither
    $\Hs_t(\tau)$ nor $\Hs_t^C(\tau)$ for $\tau \in \Hs$ can be empty under the definitions
    of $f_t$ and $\Hs_t(\tau)$; see Figure~\ref{fig:unlockset}.

    Thus, we can find that $W_t \coloneqq \exp\( \sum_{s=1}^t \frac{\alpha_s}{\ell_\text{max}} \(\ell_s(\tau_s \mid \Hs_s(\tau_s)) - 2\ell_\text{max}\) \)$ is a supermartingale with respect to $\Exp_t$, \ie $\Exp_t \[W_t\] \le W_{t-1}$ as follows:
    \begin{align}
        \Exp_t[W_t] 
        &= \Exp_t \[ W_{t-1} \exp\( \frac{\alpha_t}{\ell_\text{max}} \(\ell_t(\tau_t \mid \Hs_t(\tau_t)) - 2\ell_\text{max}\) \) \]
        \nonumber \\
        &=
        W_{t-1} \Exp_t \[  \exp\( \frac{\alpha_t}{\ell_\text{max}} \(\ell_t(\tau_t \mid \Hs_t(\tau_t)) - 2\ell_\text{max}\) \) \]
        \nonumber \\
        &=
        W_{t-1} \Exp_t \[  \exp\( \frac{\alpha_t}{\ell_\text{max}} \ell_t(\tau_t \mid \Hs_t(\tau_t)) \) \] \exp\( - 2 \alpha_t \) 
        \nonumber \\
        &\le
        W_{t-1} \exp\( 2\alpha_t\) \exp\( - 2 \alpha_t \) 
        \label{eq:exsul:lemma:connection} \\
        &= W_{t-1}, \nonumber
    \end{align}
    where (\ref{eq:exsul:lemma:connection}) holds due to (\ref{eq:exsul:lemma:rel}).
    Due to $W_0=1$ by the definition, the above implies $\Exp\[W_T\] \le 1$. 
    Then, by applying this with the Markov’s inequality, we have
    \begin{align}
        \Prob\left( 
        \sum_{t=1}^T 
        \frac{\alpha_t}{\ell_\text{max}} \(\ell_t(\tau_t \mid \Hs_t(\tau_t)) - 2\ell_\text{max}\)
        >
        \ln \frac{1}{\delta} \right) 
        =
        \Prob\left( W_T >
        \frac{1}{\delta} \right) 
        \le
        \delta \Exp_T[W_T]
        \le \delta. \nonumber
    \end{align}
    
    Thus, 
    \begin{equation*}
        \sum_{t=1}^T \frac{\alpha_t}{\ell_\text{max}} \(\ell_t(\tau_t \mid \Hs_t(\tau_t)) - 2\ell_\text{max}\)
        \le
        \ln (1/\delta),
    \end{equation*}
    with probability at least $1-\delta$, which completes the proof.
\end{proof}

\begin{corollary}
    Let $\gamma_t = \gamma > 0$ for all $t$. Then, simultaneously for all $\tau \in \Hs$,
    \begin{equation}
        \sum_{t=1}^{T} \ell_t(\tau \mid \Hs_t(\tau_t)) - \sum_{t=1}^{T} \ell_t(\tau)
        \le \frac{\ell_\text{max} \ln (|\Hs| / \delta)}{2\gamma},
        \label{eq:cor1}
    \end{equation}
    \label{cor:exsul}
    with probability at least $1-\delta$.
\end{corollary}
\begin{proof}
    The proof of this  corollary can be derived from Lemma \ref{lem:exsul}, 
    but here we provide a direct proof, which uses similar proof techniques as in Lemma \ref{lem:exsul}.

    Let $\gamma_t = \gamma \ge 0$.
    Then, we have
    \begin{align}
        \Exp_t \[ \exp\(\frac{2\gamma}{\ell_\text{max}} \ell_t(\tau \mid \Hs_t(\tau_t))\) \]
        &\le
        \Exp_t\[\exp\(\frac{2\gamma}{\ell_\text{max} \beta} \ln \(1 + \beta \hat \ell_t(\tau \mid \Hs_t(\tau_t))\) \)\]
        \label{eq:cor:beta}
        \\
        &\le
        \Exp_t\[1+\frac{2\gamma}{\ell_\text{max}} \hat \ell_t(\tau \mid \Hs_t(\tau_t))\]
        \label{eq:exsulix:gamma}
        \\
        &=
        1+\frac{2\gamma}{\ell_\text{max}} \ell_t(\tau) \label{eq:exsulix:gamma:hat}
        \\
        &\le
        \exp\(\frac{2\gamma}{\ell_\text{max}} \ell_t(\tau) \), \label{eq:cor:supermartingale}
    \end{align}
    where (\ref{eq:cor:beta}) holds same as (\ref{eq:lem:beta}) with fixed $\gamma_t=\gamma$ and $\beta_t = \beta$ for all $t$.
    Also, (\ref{eq:exsulix:gamma}) uses $x\ln(1+y) \le \ln(1+xy)$ for all $y > -1$ and $x \in [0,1]$ along with the fact that $2\gamma = \ell_\text{max} \beta$ and thus $\frac{2 \gamma}{\beta \ell_\text{max}} = 1$.
    (\ref{eq:exsulix:gamma:hat}) holds as $\Exp_t \hat \ell_t(\tau \mid \Hs_t(\tau_t)) = \ell_t(\tau)$ from \ref{eq:unbiasedlossest}.

    Thus, we can find that $W_t \coloneqq \exp\( \sum_{s=1}^t \frac{2\gamma}{\ell_\text{max}} \(\ell_s(\tau \mid \Hs_s(\tau_s)) - \ell_s(\tau)\) \)$ is a supermartingale
    with respect to $\Exp_t$, \ie $\Exp\[W_t\] \le W_{t-1}$ as follows:
    \begin{align}
        \Exp_t \[W_t\]
        &= \Exp_t\[W_{t-1} \cdot \exp\(\frac{2\gamma}{\ell_\text{max}}\(\ell_t(\tau \mid \Hs_t(\tau_t))-\ell_t(\tau)\)\)\]  
        \nonumber \\
        &= W_{t-1} \Exp_t\[ \exp\(\frac{2\gamma}{\ell_\text{max}}\(\ell_t(\tau \mid \Hs_t(\tau_t))-\ell_t(\tau)\)\)\] 
        \nonumber \\
        &= W_{t-1} \Exp_t \[ \exp\( \frac{2\gamma}{\ell_\text{max}} \ell_t (\tau \mid \Hs_t(\tau_t) ) \) \] \exp \(-\frac{2\gamma}{\ell_\text{max}}\ell_t( \tau ) \)
        \nonumber \\
        &
        \le W_{t-1}  \exp\( \frac{2\gamma}{\ell_\text{max}} \ell_t (\tau ) \)
        \exp \(-\frac{2\gamma}{\ell_\text{max}}\ell_t( \tau ) \)
        \label{eq:cor:supermartingale2}
        \\
        &= W_{t-1}.
        \nonumber 
    \end{align}
    where (\ref{eq:cor:supermartingale2}) holds as (\ref{eq:cor:supermartingale}).
    From this and $W_0=1$ by the definition, we have $\Exp\[W_T\] \le \Exp\[W_{T-1}\] \le \dots \le 1$.
    By this and the Markov’s inequality, we have
    \begin{align}
        \Prob\left( 
        \sum_{t=1}^T 
        \frac{ 2\gamma}{\ell_\text{max}} \(\ell_t(\tau_t \mid \Hs_t(\tau_t)) - \ell_t(\tau) \)
        >
        \ln \frac{1}{\delta} \right) 
        =
        \Prob\left( W_T >
        \frac{1}{\delta} \right) 
        \le
        \delta \Exp_T[W_T]
        \le \delta. \nonumber
    \end{align}
    Then, by the union bound, for all $\tau$, we have
    \begin{equation*}
        \sum_{t=1}^T \(\ell_t(\tau \mid \Hs_t(\tau_t)) - \ell_t(\tau)\)
        \le
        \frac{\ell_\text{max}\ln (|\Hs|/\delta)}{2\gamma}
    \end{equation*}
    with probability at least $1-\delta$, which completes the proof.
\end{proof}

We then prove our main theorem, which consists of four steps.

\paragraph{First step.}
We split the expected loss estimator into two logarithmic terms Eq.~(3.7) in \citep{bubeck2012regret}:
\begin{multline}
    \Exp_{\tau \sim p_t} \ell_t(\tau \mid \Hs_t(\tau_t))
    = \frac{1}{\eta_t}\underbrace{\ln \Exp_{\tau\sim p_t} \exp\bigg( -\eta_t\Big(\ell_t(\tau \mid \Hs_t(\tau_t)) - \Exp_{\bar\tau \sim p_t} \ell_t(\bar \tau \mid \Hs_t(\tau_t))\Big) \bigg)}_{\text{variability term}} \\ 
    - \frac{1}{\eta_t}\underbrace{ \ln \Exp_{\tau\sim p_t} \exp\bigg( -\eta_t \ell_t(\tau \mid \Hs_t(\tau_t)) \bigg)}_{\text{log of an exponential expectation term}},
    \label{eq:exsulix:split}
\end{multline}
where $\eta_t$ is non-increasing in $t \in \naturalnum$.
In the following steps, each term will be rewritten.

\paragraph{Second step.}
We provide an upper bound on the first term:
\begin{align}
    \ln \Exp_{\tau \sim p_t} & \exp\(-\eta_t \Big(
    \ell_t(\tau \mid \Hs_t(\tau_t))
    - \Exp_{\bar \tau \sim p_t} \ell_t(\bar \tau \mid \Hs_t(\tau_t)) \Big)\) 
    \nonumber\\
    &= 
    \ln\Exp_{\tau \sim p_t} \exp\big(-\eta_t \ell_t(\tau \mid \Hs_t(\tau_t))\big)
    + \eta_t \Exp_{\bar \tau \sim p_t} \ell_t(\bar \tau \mid \Hs_t(\tau_t))
    \nonumber\\
    &\le
    \Exp_{\tau \sim p_t} \[\exp\big(-\eta_t \ell_t(\tau \mid \Hs_t(\tau_t))\big) 
    - 1 + \eta_t \ell_t(\tau \mid \Hs_t(\tau_t)) \]
    \label{eq:exsulix:second:ln:x1}\\
    &\le
    \Exp_{\tau \sim p_t} \eta_t^2 \frac{\ell_t(\tau \mid \Hs_t(\tau_t))^2}{2}
    \label{eq:exsulix:second:exp}\\
    &\le
    \frac{\eta_t^2}{2} \sum_{\tau\in \Hs} p_t(\tau) \ell_t(\tau \mid \Hs_t(\tau_t))^2
    \nonumber\\
    &=
    \frac{\eta_t^2}{2} \sum_{\tau\in \Hs} p_t(\tau) \(\frac{\ell_t(\tau)}{\gamma_t + \sum_{\bar\tau \in \Hs_t({\tau_t})} \mathbbm{1}(\tau \in \Hs_t({\bar\tau})) \cdot p_t(\bar\tau) } \cdot \mathbbm{1}(\tau \in \Hs_t({\tau_t}))\)^2
    \label{eq:exsulix:exaul4:ref}\\
    &=
    \frac{\eta_t^2}{2} \sum_{\tau\in \Hs_t(\tau_t)} p_t(\tau) \(\frac{\ell_t(\tau)}{\gamma_t + \sum_{\bar\tau \in \Hs_t({\tau_t})} \mathbbm{1}(\tau \in \Hs_t({\bar\tau})) \cdot p_t(\bar\tau) }\)^2
    \nonumber\\
    &=
    \frac{\eta_t^2}{2} \sum_{\tau\in \Hs_t(\tau_t)} p_t(\tau) \(\frac{\ell_t(\tau)}{\gamma_t + \sum_{\bar\tau \in \Hs_t({\tau_t})} p_t(\bar\tau) }\)^2
    \label{eq:exsulix:second:swap}\\
    &=
    \frac{\eta_t^2}{2} \frac{\sum_{\tau\in \Hs_t(\tau_t)}p_t(\tau)\ell_t(\tau)^2}{\(\gamma_t + \sum_{\bar\tau \in \Hs_t({\tau_t})} p_t(\bar\tau)\)^2}
    \nonumber\\
    &\le
    \frac{\ell_\text{max} \eta_t^2}{2} \frac{\sum_{\tau\in \Hs_t(\tau_t)}p_t(\tau)}{\(\gamma_t + \sum_{\bar\tau \in \Hs_t({\tau_t})} p_t(\bar\tau)\)^2}
    \ell_t(\tau_t)
    \label{eq:specializedloss2}\\
    &\le
    \frac{\ell_\text{max} \eta_t^2}{2} \frac{\ell_t(\tau_t)}{\gamma_t + \sum_{\bar\tau \in \Hs_t({\tau_t})} p_t(\bar\tau)}
    \nonumber\\
    &=
    \frac{\ell_\text{max} \eta_t^2}{2} \frac{\ell_t(\tau_{t})}{\gamma_t + \sum_{\bar\tau \in \Hs_t({\tau_t})} \mathbbm{1}(\tau_t \in \Hs_t({\bar\tau})) \cdot p_t(\bar\tau) } \cdot \mathbbm{1}(\tau_t \in \Hs_t({\tau_t})),
    \label{eq:exsulix:second:beforefinal} \\
    &=
    \frac{\ell_\text{max} \eta_t^2}{2} \ell_t(\tau_t \mid \Hs_t(\tau_t)),
    \label{eq:exsulix:second:final}
\end{align}
where (\ref{eq:exsulix:second:ln:x1}) holds due to $\ln(x) \le x-1$ for $x \le 1$,
(\ref{eq:exsulix:second:exp}) uses $\exp(x) \le 1+x+x^2/2$ for $x \le 0$,
(\ref{eq:exsulix:second:swap}) holds as the same as (\ref{eq:specializedloss2:1}),
(\ref{eq:specializedloss2}) holds as $\ell_t(\tau)=\ell_t(\tau')$ for all $\tau\in\Hs_t(\tau')$ by the definition of the specialized loss (\ref{eq:loss-sg}) and its domain $[0, \ell_\text{max}]$,
and (\ref{eq:exsulix:second:beforefinal}) holds as the same as (\ref{eq:thm2:exp:over:tau1-5}), \ie the property of the indicator function.

\paragraph{Third step.}
Next, we reformulate 
the second term, the log of an exponential expectation term. 
Let $\tilde L_t(\tau) \coloneqq \sum_{s=1}^t \ell_s(\tau \mid \Hs_s(\tau_s))$ and $\tilde 
L_0(\tau) = 0$. Then, we have
\begin{align}
    - \frac{1}{\eta_t} \ln \Exp_{\tau\sim p_t} \exp &\bigg( -\eta_t \ell_t(\tau \mid \Hs_t(\tau_t)) \bigg)
    \nonumber \\
    &= - \frac{1}{\eta_t} \ln \sum_{\tau \in \Hs} p_t(\tau) \exp\bigg( -\eta_t \ell_t(\tau \mid \Hs_t(\tau_t)) \bigg)
    \nonumber \\
    &= - \frac{1}{\eta_t} \ln \sum_{\tau \in \Hs} \frac{\exp (-\eta_t \tilde L_{t-1}(\tau))}{\sum_{\bar \tau \in \Hs} \exp (-\eta_t \tilde L_{t-1}(\bar \tau))}
    \exp\bigg( -\eta_t \ell_t(\tau \mid \Hs_t(\tau_t)) \bigg)
    \nonumber \\
    &= - \frac{1}{\eta_t} \ln \frac{\sum_{\tau \in \Hs} \exp (-\eta_t \tilde L_t(\tau))}{\sum_{\tau \in \Hs} \exp (-\eta_t \tilde L_{t-1}(\tau))}.
    \label{eq:exsulix:second:final2}
\end{align}

\paragraph{Fourth step.}
Then, combining rewritten terms, (\ref{eq:exsulix:second:final}) and (\ref{eq:exsulix:second:final2}), the following inequality holds:
\begin{align}
    \sum_{t=1}^T \Exp_{\tau \sim p_t} \ell_t(\tau \mid \Hs_t(\tau_t))
    &\le
    \sum_{t=1}^T \frac{\ell_\text{max} \eta_t}{2} \ell_t(\tau_t \mid \Hs_t(\tau_t))
    - \sum_{t=1}^{T} \frac{1}{\eta_t} \ln \frac{\sum_{\tau \in \Hs} \exp (-\eta_t \tilde L_t(\tau))}{\sum_{\tau \in \Hs} \exp (-\eta_t \tilde L_{t-1}(\tau))}
    \nonumber\\
    &\le
    \sum_{t=1}^T \frac{\ell_\text{max} \eta_t}{2} \ell_t(\tau_t \mid \Hs_t(\tau_t))
    - \frac{1}{\eta_T} \sum_{t=1}^{T} \ln \frac{\sum_{\tau \in \Hs} \exp (-\eta_t \tilde L_t(\tau))}{\sum_{\tau \in \Hs} \exp (-\eta_t \tilde L_{t-1}(\tau))}
    \label{eq:exsulix:fourth:eta}\\
    &=
    \sum_{t=1}^T \frac{\ell_\text{max} \eta_t}{2} \ell_t(\tau_t \mid \Hs_t(\tau_t))
    - \frac{1}{\eta_T} \ln \frac{\sum_{\tau \in \Hs} \exp (-\eta_T \tilde L_T(\tau))}{\sum_{\tau \in \Hs} \exp (-\eta_1 \tilde L_{0}(\tau))}
    \nonumber\\
    &=
    \sum_{t=1}^T \frac{\ell_\text{max} \eta_t}{2} \ell_t(\tau_t \mid \Hs_t(\tau_t))
    - \frac{1}{\eta_T} \ln \sum_{\tau \in \Hs} \exp (-\eta_T \tilde L_T(\tau))
    + \frac{\ln|\Hs|}{\eta_T}
    \nonumber\\
    &\le
    \sum_{t=1}^T \frac{\ell_\text{max} \eta_t}{2} \ell_t(\tau_t \mid \Hs_t(\tau_t))
    - \frac{1}{\eta_T} \ln \(\max_{\tau} \exp (-\eta_T \tilde L_T(\tau))\)
    + \frac{\ln|\Hs|}{\eta_T}
    \nonumber\\
    &=
    \sum_{t=1}^T \frac{\ell_\text{max} \eta_t}{2} \ell_t(\tau_t \mid \Hs_t(\tau_t))
    + \min_\tau \sum_{t=1}^{T} \ell_t(\tau \mid \Hs_t(\tau_t))
    + \frac{\ln|\Hs|}{\eta_T},
    \nonumber
    \end{align}
where (\ref{eq:exsulix:fourth:eta}) holds as $\eta_t$ is non-increasing
and $\ln \frac{\sum_{\tau \in \Hs} \exp (-\eta_t \tilde L_t(\tau))}{\sum_{\tau \in \Hs} \exp (-\eta_t \tilde L_{t-1}(\tau))} \le 0$ for all $t$.

From this and the fact (\ref{eq:thm2:exp:over:tau2}) that $\Exp_{\tau \sim p_t} \ell_t(\tau \mid \Hs_t(\tau_t)) = \ell_t(\tau_t) - \gamma_t \ell_t(\tau_t \mid \Hs_t(\tau_t))$, we have
\begin{align}
    \sum_{t=1}^{T} \ell_t(\tau_t) - \min_\tau \sum_{t=1}^{T} \ell_t(\tau \mid \Hs_t(\tau_t))
    &\le
    \sum_{t=1}^T \frac{\ell_\text{max} \eta_t}{2}\ell_t(\tau_t \mid \Hs_t(\tau_t))
    + \frac{\ln|\Hs|}{\eta_T}
    + \sum_{t=1}^T \gamma_t \ell_t(\tau_t \mid \Hs_t(\tau_t))
    \nonumber
    \\
    &=
    \sum_{t=1}^T \(\frac{\ell_\text{max} \eta_t}{2} + \gamma_t\) \ell_t(\tau_t \mid \Hs_t(\tau_t))
    + \frac{\ln|\Hs|}{\eta_T}.
    \nonumber
\end{align}

Then, from Lemma \ref{lem:exsul} and Corollary \ref{cor:exsul} along with the union bound, the following inequality holds with probability at least $1-2\delta'$, where $\delta' = \delta/2$ if $\eta_t \le \frac{2\gamma_t}{\ell_\text{max}}$:
\begin{align*}
    \sum_{t=1}^{T} \ell_t(\tau_t) - \min_\tau \sum_{t=1}^{T} \ell_t(\tau)
    &\le
    \sum_{t=1}^T \(\frac{\ell_\text{max} \eta_t}{2} + \gamma_t\) \ell_t(\tau_t \mid \Hs_t(\tau_t))
    + \frac{\ln|\Hs|}{\eta_T}
    + \frac{\ell_\text{max} \ln (2|\Hs|/\delta)}{2\gamma}
    \\
    &\le
    2\ell_\text{max} \sum_{t=1}^T \(\frac{\ell_\text{max} \eta_t}{2} + \gamma_t\) 
    + \ell_\text{max}\ln{\frac{2}{\delta}}
    + \frac{\ln|\Hs|}{\eta_T}
    + \frac{\ell_\text{max} \ln (2|\Hs|/\delta)}{2\gamma},
\end{align*}
where the first inequality holds with probability at least $1 - \delta'$ from Corollary \ref{cor:exsul}
and
the second inequality holds with probability at least $1 - \delta'$ from Lemma \ref{lem:exsul} when $\alpha_t=\frac{\ell_\text{max}\eta_t}{2} + \gamma_t \le 2 \gamma_t$.

\textbf{Final Bounds.}
We have our regret bound depending on the setup of the learning rate $\eta_t$. 

If $\eta_t = \frac{2\gamma_t}{\ell_\text{max}} = \sqrt{\frac{\ln|\mathcal{H}|}{\ell_\text{max}^2 T}}$, with probability at least $1-\delta$,
\begin{align}
    \Reg_T
    &\le  
    2 \ell_\text{max} \sqrt{T\ln|\Hs|}
    + \ell_\text{max} \( 1 +  \sqrt{\frac{T}{\ln|\Hs|}}\)\ln\frac{2}{\delta}
    + 2\ell_\text{max}\sqrt{T\ln|\Hs|}
    \nonumber
    \\
    &=  
    \ell_\text{max} 
    \(4\sqrt{T\ln|\Hs|}
    + \( 1 +  \sqrt{\frac{T}{\ln|\Hs|}}\)\ln\frac{2}{\delta}
    \)
    \nonumber
    \\
    &=
    \Os(\ell_\text{max} \sqrt{T\ln(|\Hs|/\delta)}).
    \label{eq:exsulix:known:regretbound}
\end{align}

If $\eta_t = \frac{2\gamma_t}{\ell_\text{max}} = \sqrt{\frac{ \ln|\mathcal{H}|}{2\ell_\text{max}^2 t}}$, with probability at least $1-\delta$,
\begin{align}
    \Reg_T
    &\le  
    2 \ell_\text{max}\sqrt{2 T\ln|\mathcal H|}
    + \ell_\text{max} \( 1 +  \sqrt{\frac{2T}{\ln|\Hs|}}\)\ln\frac{2}{\delta}
    + 2\ell_\text{max}\sqrt{2T\ln|\mathcal H|}
    \nonumber\\
    &= 
    \ell_\text{max}
    \(4\sqrt{2T\ln|\mathcal H|} + \( 1 +  \sqrt{\frac{2T}{\ln|\Hs|}}\)\ln\frac{2}{\delta}
    \)
    \nonumber\\
    &= 
    \Os(\ell_\text{max} \sqrt{T\ln(|\Hs|/\delta)}),
    \nonumber
\end{align}
noting that $\sum_{t=1}^T1/\sqrt{t} \le 2\sqrt{T}$.

\paragraph{Expected regret}
The expected regret can be obtained via integrating the deviations in (\ref{eq:exsulix:known:regretbound}), \ie
\begin{equation*}
    \Exp [W] \le \int_0^1 \frac{1}{\delta}\Prob \( W > \ln\frac{1}{\delta} \) d\delta.
\end{equation*}
Here, if we take $W = \frac{1}{\ell_\text{max}\(1+\sqrt{T/\ln|\Hs|}\)}\(\Reg_T-4\ell_\text{max}\sqrt{T\ln|\Hs|}\)  - \ln2 > \ln \frac{1}{\delta}$, which holds with probability at most $\delta$,
then,
\begin{equation*}
    \Exp [W] \le 1,
\end{equation*}
which suggests that
\begin{equation}
    \Exp[\Reg_T]
    \le 
     \ell_\text{max} 
    \(4\sqrt{T\ln|\Hs|}
    +   (1+\ln2) \(1+\sqrt{\frac{T}{\ln|\Hs|}}\)
    \)
    = \Os(\ell_\text{max} \sqrt{T\ln|\Hs|}).
    \label{eq:exsulix:exp:regretbound}
\end{equation}

\section{Discussion}
\label{sec:discussion}

\subsection{Inefficiency}
{\color{red}\label{sec:discussion:ineff}}
{\color{\hl}
The SR bound in (\ref{eq:fdrboundfromlemma}) converges to zero in $T$ unless $\Ine_T$ converges to $1$.
In other words, $\SR_T$ may fail to be bounded if
$\Ine_T \approx 1$ \eg $\sum_{t=1}^T e_t \approx T$,
due to the nature of the empirical metric with the variance induced by randomized bandit algorithms.
For instance, if the generator $G$ consistently produces incorrect outputs,
the learner will try to converge to an always-abstaining solution. However, due to randomization, if the learner mistakenly accepts even a single sample in the early stage of learning, then regardless of convergence to an always-abstaining solution, the empirical SR may still exceed the desired level $\alpha$.
Figures~\ref{fig:plot_over_alpha:nq:gpt-3.5:fdr} and~\ref{fig:nq:shift:gpt-3.5:fM2:fdr} show such a case caused by a poorly calibrated scoring function rather than the generator: for NQ with GPT-5.4, $f_{\texttt{con}}$ assigns the maximum score to $44.5\%$ of the samples, $14.8\%$ of which are incorrect, so \ours is driven toward always abstaining for smaller $\alpha$ and the empirical SR exceeds $\alpha$ (see the caption of Figure~\ref{fig:stochastic:main:plot_over_alpha}).
}

\subsection{Selecting $\lambda$}

We clarify that given the data sequence $\x_{1:T}$, the realized
selection inefficiency $\Ine_T$ is an outcome of the interaction between
the model capability of $G$, the calibration of the scoring function $f$,
and the choice of the trade-off parameter $\lambda$.
In particular, a larger $\lambda$ places more emphasis on the SR-related
loss term (\ref{eq:loss-sg}), which typically leads to more conservative behavior and faster
convergence of the SR constraint, at the cost of increased inefficiency.

Nevertheless, when the underlying model or the scoring function $f$ is
severely miscalibrated, one may heuristically increase $\lambda$ so that
the learner approaches the behavior of the most conservative
SR-feasible expert in hindsight, if such an expert exists
(see Section~\ref{sec:ineff:justification} for a discussion on selection efficiency).
In this sense, choosing $\lambda = \mathcal{O}(T^d)$ with $d \in \mathbb{R}^+$
provides a principled way to tune the trade-off between efficiency and robustness.

In our experiments, we set $\lambda = T^{1/2}$, matching the rate achieved by standard bandit regret bounds.
Empirical trends under different values of $\lambda$ are reported in
Figures~\ref{fig:fdroverlambda} and \ref{fig:ineffoverlambda}.

\subsection{Selection Efficiency}
\label{sec:ineff:justification}

In this section, we show that the always-abstaining policy is not optimal
in the aspect of full-feedback setting whenever there exists a selectively-abstaining
policy that satisfies the SR constraint, \ie $\argmin_\tau \sum_{t=1}^T[a_t(\tau) + \lambda d_t(\tau, \alpha)] \notin \Hs^\text{aa}$, if there exists some $\tau \in \Hs \setminus \Hs^\text{aa}$ that satisfies the SR guarantee,
where $\Hs^\text{aa} \coloneqq \{\tau \mid \sum_{t=1}^T a_t(\tau) = T\}$.

Let $e_t^\tau \coloneqq \Es_t(\Sh(;\tau))$, and $\tau^\text{aa} \in \Hs^\text{aa}$, which is always abstaining, then
for any $\tau^\text{sa} \in \Hs \setminus \Hs^\text{aa}$ that satisfies the SR guarantee, the following inequality holds,
\begin{gather}
    \frac{\sum_{t=1}^T \mathbbm{1}\left( \Sh(\x_t; \tau^\text{sa}) \neq \texttt{IDK}\right) \cdot e_t^{\tau^\text{sa}}}{\sum_{t=1}^T \mathbbm{1}\left( \Sh(\x_t; \tau^\text{sa}) \neq \texttt{IDK}\right)} \le \alpha \nonumber
    \\
    \Rightarrow \sum_{t=1}^T \mathbbm{1}\left( \Sh(\x_t; \tau^\text{sa}) \neq \texttt{IDK}\right)  \cdot e_t^{\tau^\text{sa}} \le \alpha \sum_{t=1}^T \mathbbm{1}\left( \Sh(\x_t; \tau^\text{sa}) \neq \texttt{IDK}\right).
    \label{discussion:not-triv:fdr}
\end{gather}
Let $L_T(\tau, \alpha) \coloneqq \sum_{t=1}^T \frac{a_t(\tau) + \lambda d_t(\tau, \alpha)}{1 + \lambda}$, which is the cumulative loss of $\tau$,
and consider the difference of cumulative loss between $\tau^\text{aa}$ and $\tau^\text{sa}$,
{
\begin{align*}
    {(1+\lambda) }&(L_T(\tau^\text{aa}) - L_T(\tau^\text{sa}))
    \\
    &= \sum_{t=1}^T[a_t(\tau^\text{aa}) + \lambda d_t(\tau^\text{aa}, \alpha)] - \sum_{t=1}^T[a_t(\tau^\text{sa}) + \lambda d_t(\tau^\text{sa}, \alpha)]
    \\
    &= \sum_{t=1}^T\bigg[\mathbbm{1}\left(\Sh(\x_t; \tau^\text{aa})= \texttt{IDK}\right)
    + \lambda \left(\mathbbm{1}( \Sh(\x_t; \tau^\text{aa}) \neq \texttt{IDK}) \cdot e_t^{\tau^\text{aa}} - \alpha\mathbbm{1}( \Sh(\x_t; \tau^\text{aa}) \neq \texttt{IDK}) + \alpha \right)\bigg] 
    \\
    & \qquad
    - \sum_{t=1}^T\bigg[\mathbbm{1}\left(\Sh(\x_t; \tau^\text{sa})= \texttt{IDK}\right)
    + \lambda \left(\mathbbm{1}( \Sh(\x_t; \tau^\text{sa}) \neq \texttt{IDK} ) \cdot e_t^{\tau^\text{sa}} - \alpha\mathbbm{1}( \Sh(\x_t; \tau^\text{sa}) \neq \texttt{IDK}) + \alpha \right)\bigg] 
    \\
    &= T +\lambda\alpha T - \Bigg[T - \sum_{t=1}^T \mathbbm{1}\left(\Sh(\x_t; \tau^\text{sa}) \neq \texttt{IDK} \right)
    + \lambda\bigg\{\alpha \bigg(T-\sum_{t=1}^T \mathbbm{1} \left(\Sh(\x_t; \tau^\text{sa}) \neq \texttt{IDK} \right)\bigg)
    \\
    & \qquad\qquad\qquad\qquad\qquad\qquad\qquad\qquad
    + \sum_{t=1}^T \mathbbm{1}\left( \Sh(\x_t; \tau^\text{sa}) \neq \texttt{IDK} \right) \cdot e_t^{\tau^\text{sa}}\bigg\}\Bigg]
    \\
    &= \sum_{t=1}^T \mathbbm{1}\left(\Sh(\x_t; \tau^\text{sa}) \neq \texttt{IDK} \right) + \lambda\alpha \sum_{t=1}^T \mathbbm{1}\left( \Sh(\x_t; \tau^\text{sa}) \neq \texttt{IDK} \right)
    -\lambda \sum_{t=1}^T \mathbbm{1}\left( \Sh(\x_t; \tau^\text{sa}) \neq \texttt{IDK} \right) \cdot e_t^{\tau^\text{sa}}.
\end{align*}
}
Since $\tau^\text{sa}$ satisfies the SR guarantee (\ref{discussion:not-triv:fdr}),
we get
\begin{equation}
    L_T(\tau^\text{aa}) - L_T(\tau^\text{sa})
    \ge \sum_{t=1}^T
    \frac{\mathbbm{1}\left(\Sh(\x_t; \tau^\text{sa}) \neq \texttt{IDK} \right)}{1+\lambda}
    = \frac{T - \sum_{t=1}^T a_t(\tau^\text{sa})}{1+\lambda} > 0, \label{proof:ineff:justification:lower}
\end{equation}
where $T - \sum_{t=1}^T a_t(\tau^\text{sa}) > 0$ holds due to $\tau^\text{sa} \notin \Hs^\text{aa}$.
This implies that, given sufficiently many data points, any $\tau^\text{sa}$ that satisfies the SR guarantee incurs at least $T - \sum_{t=1}^T a_t(\tau^\text{sa})$ less loss than $\tau^\text{aa}$.
{\color{\hl}This is empirically supported by Figure \ref{fig:stochastic:main:plot_over_alpha} and \ref{fig:exsul:shift:poa}.}

\clearpage
\section*{NeurIPS Paper Checklist}

\begin{enumerate}

\item {\bf Claims}
    \item[] Question: Do the main claims made in the abstract and introduction accurately reflect the paper's contributions and scope?
    \item[] Answer: \answerYes{}
    \item[] Justification: The main claims and contributions are stated in the abstract and Section \ref{sec:introduction}.

\item {\bf Limitations}
    \item[] Question: Does the paper discuss the limitations of the work performed by the authors?
    \item[] Answer: \answerYes{}
    \item[] Justification: We discuss the limitations of the proposed approach in Section~\ref{sec:conc}.

\item {\bf Theory assumptions and proofs}
    \item[] Question: For each theoretical result, does the paper provide the full set of assumptions and a complete (and correct) proof?
    \item[] Answer: \answerYes{}
    \item[] Justification: 

    \item {\bf Experimental result reproducibility}
    \item[] Question: Does the paper fully disclose all the information needed to reproduce the main experimental results of the paper to the extent that it affects the main claims and/or conclusions of the paper (regardless of whether the code and data are provided or not)?
    \item[] Answer: \answerYes{}
    \item[] Justification: Sections~\ref{sec:exp} and Appendix~\ref{sec:expsetup} provide the experimental setup, implementation details, hyperparameters, datasets, baselines, and evaluation protocols needed to reproduce the main results.

\item {\bf Open access to data and code}
    \item[] Question: Does the paper provide open access to the data and code, with sufficient instructions to faithfully reproduce the main experimental results, as described in supplemental material?
    \item[] Answer: \answerNo{}
    \item[] Justification: We do not provide open access to the code and data at submission time. However, Sections~\ref{sec:exp} and Appendix~\ref{sec:expsetup} provide the implementation details, hyperparameters, and evaluation protocols needed to reproduce the main results.

\item {\bf Experimental setting/details}
    \item[] Question: Does the paper specify all the training and test details (e.g., data splits, hyperparameters, how they were chosen, type of optimizer) necessary to understand the results?
    \item[] Answer: \answerYes{}
    \item[] Justification: Sections~\ref{sec:exp} and Appendix~\ref{sec:expsetup} specify the experimental setup, including data splits, hyperparameters, implementation details, and evaluation protocols needed to understand the results.

\item {\bf Experiment statistical significance}
    \item[] Question: Does the paper report error bars suitably and correctly defined or other appropriate information about the statistical significance of the experiments?
    \item[] Answer: \answerYes{}
    \item[] Justification: We report distributional information for the main experimental results using violin plots in Section \ref{sec:exp}, which show variability across repeated runs. We also provide theoretical guarantees in Section \ref{sec:advfeedback:algorithm} to support the reliability of the proposed algorithm.

\item {\bf Experiments compute resources}
    \item[] Question: For each experiment, does the paper provide sufficient information on the computer resources (type of compute workers, memory, time of execution) needed to reproduce the experiments?
    \item[] Answer: \answerYes{}
    \item[] Justification: We describe the compute resources used for our experiments in Appendix~\ref{sec:expsetup}.
    
\item {\bf Code of ethics}
    \item[] Question: Does the research conducted in the paper conform, in every respect, with the NeurIPS Code of Ethics \url{https://neurips.cc/public/EthicsGuidelines}?
    \item[] Answer: \answerYes{}
    \item[] Justification: The research was conducted in accordance with the NeurIPS Code of Ethics.

\item {\bf Broader impacts}
    \item[] Question: Does the paper discuss both potential positive societal impacts and negative societal impacts of the work performed?
    \item[] Answer: \answerNA{}
    \item[] Justification: The work is foundational algorithmic research and does not have a direct path to societal impact or deployment.
    
\item {\bf Safeguards}
    \item[] Question: Does the paper describe safeguards that have been put in place for responsible release of data or models that have a high risk for misuse (e.g., pre-trained language models, image generators, or scraped datasets)?
    \item[] Answer: \answerNA{}
    \item[] Justification: The paper does not release data or models with a high risk of misuse.

\item {\bf Licenses for existing assets}
    \item[] Question: Are the creators or original owners of assets (e.g., code, data, models), used in the paper, properly credited and are the license and terms of use explicitly mentioned and properly respected?
    \item[] Answer: \answerYes{}
    \item[] Justification: We properly cite the original dataset sources and respect their licenses and terms of use.

\item {\bf New assets}
    \item[] Question: Are new assets introduced in the paper well documented and is the documentation provided alongside the assets?
    \item[] Answer: \answerNA{}
    \item[] Justification: The paper does not release new assets such as datasets, code packages, or models.

\item {\bf Crowdsourcing and research with human subjects}
    \item[] Question: For crowdsourcing experiments and research with human subjects, does the paper include the full text of instructions given to participants and screenshots, if applicable, as well as details about compensation (if any)? 
    \item[] Answer: \answerNA{}
    \item[] Justification: The paper does not involve crowdsourcing or research with human subjects.

\item {\bf Institutional review board (IRB) approvals or equivalent for research with human subjects}
    \item[] Question: Does the paper describe potential risks incurred by study participants, whether such risks were disclosed to the subjects, and whether Institutional Review Board (IRB) approvals (or an equivalent approval/review based on the requirements of your country or institution) were obtained?
    \item[] Answer: \answerNA{}
    \item[] Justification: The paper does not involve crowdsourcing or research with human subjects.

\item {\bf Declaration of LLM usage}
    \item[] Question: Does the paper describe the usage of LLMs if it is an important, original, or non-standard component of the core methods in this research? Note that if the LLM is used only for writing, editing, or formatting purposes and does \emph{not} impact the core methodology, scientific rigor, or originality of the research, declaration is not required.
    \item[] Answer: \answerNA{}
    \item[] Justification: LLMs were not used as important, original, or non-standard components of the core method.

\end{enumerate}


\end{document}